\documentclass[10pt,journal,compsoc]{IEEEtran}
\usepackage{hyperref}
\usepackage{amsmath,amssymb,graphicx,adjustbox,multirow,hhline}
\usepackage{stackengine}
\usepackage{algorithm}
\usepackage{algpseudocode}
\usepackage{pifont}
\def\delequal{\mathrel{\stackon[1pt]{=}{$\scriptstyle\Delta$}}}

\newcommand*{\hermconj}{^{\mathsf{H}}}
\newcommand{\ykt}{\widetilde{\yv}_k}
\newcommand{\xkt}{\widetilde{\xv}_k}
\newcommand{\wkt}{\widetilde{\wv}_k}

\long\def\comment#1{}

\newcommand{\beq}{\begin{equation*}}
\newcommand{\eeq}{\end{equation*}}

\newfont{\bbb}{msbm10 scaled 700}

\newfont{\bb}{msbm10 scaled 1100}

\newcommand{\hv}{{\bf h}}

\newcommand{\wv}{{\bf w}}

\newcommand{\xv}{{\bf x}}
\newcommand{\yv}{{\bf y}}

\newcommand{\Am}{{\bf A}}

\newcommand{\Wm}{{\bf W}}

\newcommand{\Xm}{{\bf X}}
\newcommand{\Ym}{{\bf Y}}

\newcommand{\Kc}{{\cal K}}

\newcommand{\Nc}{{\cal N}}
\newcommand{\Oc}{{\cal O}}

\newcommand{\Sc}{{\cal S}}

\newcommand{\Vc}{{\cal V}}

\newcommand{\lambdav}{\hbox{\boldmath$\lambda$}}

\usepackage{caption}
\usepackage{subcaption}
\usepackage{ragged2e}

\ifCLASSINFOpdf
\else
\fi

\begin{document}
\title{Image Denoising via Collaborative Dual-Domain Patch Filtering}
\author{Muzammil~Behzad,~\IEEEmembership{Student Member,~IEEE}
}

\markboth{}%
{Shell \MakeLowercase{\textit{et al.}}: Bare Demo of IEEEtran.cls for Computer Society Journals}

\IEEEtitleabstractindextext{%
\begin{abstract} In this paper, we propose a novel image denoising algorithm exploiting features from both spatial as well as transformed domain. We implement intensity-invariance based improved grouping for collaborative support-agnostic sparse reconstruction. For collaboration firstly, we stack similar-structured patches via intensity-invariant correlation measure. The grouped patches collaborate to yield desirable sparse estimates for noise filtering. This is because similar patches share the same support in the transformed domain, such similar supports can be used as probabilities of active taps to refine the sparse estimates. This ultimately produces a very useful patch estimate thus increasing the quality of recovered image by discarding the noise-causing components. A region growing based spatially developed post-processor is then applied to further enhance the smooth regions by extracting the spatial domain features. We also extend our proposed method for denoising of color images. Comparison results with the state-of-the-art algorithms in terms of peak signal-to-noise ratio (PNSR) and structural similarity (SSIM) index from extensive experimentations via a broad range of scenarios demonstrate the superiority of our proposed algorithm.
\end{abstract}

}

\maketitle

\IEEEdisplaynontitleabstractindextext

%
\IEEEpeerreviewmaketitle

\IEEEraisesectionheading{\section{Introduction}
	\label{sec:introduction}}
\IEEEPARstart{I}{mage} denoising has been explored extensively over the past decades because of its tremendous importance toward the applications of statistical and seismic signal processing, computer and embedded vision, Internet-of-Things (IoT), robotics, and artificial intelligence. However, the noise contamination in images raises questions on the reliability of image analysis and is of a major interest during the image acquisition phase~\cite{6112764}. In such scenarios, the objective is to find an estimate $\widehat{\Xm}$ of a clean image $\Xm$ which is corrupted by the signal-independent additive white Gaussian noise (AWGN) as $\Ym = \Xm + \Wm$, where the noise $\Wm \sim \mathcal{N} (\textbf{0},\sigma_w^2\textbf{I})$.

Since image denoising is an ill-posed problem, it is theoretically complex to accurately restore a denoised version of the image from noise\cite{7059213}. However, a number of intelligent algorithms have been designed recently to tackle this problem. These algorithms span from pixel based filtering techniques (such as, Gaussian filtering, bilateral filtering and total variation regularization) to recently developed patch based methods. The patch based methods have shown to produce amazingly better results and have outperformed many traditional pixel based methods \cite{7321034}. In this connection, a significant number of proposed patch based approaches cover, for example, denoising \cite{4160959, 5482584, 6054053, 7418309, 7448946}, deblurring \cite{6804676, 7327182}, segmentation \cite{6475946, 6727538}, detection and tracking \cite{6420842,7472026}, and bio-medical applications~\cite{6399478, 7418225}.

In case of image denoising algorithms, the existing techniques are further classified into spatial-domain and transformed-domain based methods. Among the spatial-domain based denoising approaches, total variation minimization \cite{ipol.2012.g-tvd}, and recently developed methods exploiting the spatial self-similarity have showed promising results (e.g., see \cite{1703579,7298595}). The non-local means (NL-means) \cite{1467423} is one of the pioneer methods to benefit from spatial self-similarity. This algorithm replaces a reference pixel by the weighted average of other neighboring pixels. Here, the similarity in the neighborhood of each pixel with that of a reference pixel is taken as a self-similarity measure.

As a strong competitor, the transformed-domain based techniques rely on an underlying image regularity assumption and process the image by computing a transform representation, such as, wavelet transform. Generally, the idea of thresholding of coefficients in the transformed-domain is utilized \cite{7457928, 1014998,6226423,7350835,7298646}. In this regard, since sparsity is an abundant property of many natural-existing or artificially-synthesized signals, many algorithms advocated the use of sparse representations of images or its patches. For this purpose, compressed sensing (CS) algorithms are applied to find the sparse coefficients of a given patch. One of the important works in this category include the K-SVD algorithm for image denoising via dictionary learning \cite{elad2006image}. K-SVD computes a highly overcomplete dictionary using preliminary training. A similar work is being reported in \cite{4392496,doi:10.1137/070697653, ipol.2012.llm-ksvd, 5653853, 7067415} that significantly contributed toward this domain of research. However, a strong limitation of these algorithms is that they impose very high computational burden thereby compromising its~practicality.

The recently developed image denoising algorithms take into account the critical information from both spatial as well as the transformed-domain, and use that as an important piece of information to yield amazingly better results (e.g., see \cite{6820766, 6678291, 7351639}). One of such popular algorithms is the block-matching 3D filtering (BM3D) \cite{4271520}, which is considered as the state-of-the-art algorithm in image denoising. This algorithm operates at the patch level and first collects the similar blocks in the spatial domain followed by stacking them in 3D arrays. Afterwards, a hard thresholding is applied in the transformed-domain producing a prior image estimate. This estimate is further refined by a specially developed Wiener filter.

The currently trending state-of-the-art algorithms like NL-means, K-SVD and BM3D produce near to optimal results when tested on natural images as shown in \cite{5995309} and \cite{5339210}. However, denoising still remains a challenging problem that has vast room for improvements in many directions especially in the case of large noise, varying images sizes, and whether the image is natural or man-made. For example, while majority of these algorithms manage to separate the noise from image to a good extent, the performance of these algorithms is a function of image size. Furthermore, one of the major drawbacks is that they tend to blur out the recovered patches thus removing sharp image details. With such an undesired trade-off at hand, we present the following fundamental challenges that a handy image denoising algorithm should possess:
\begin{enumerate}
	\item The perceptually smooth areas of an image should be kept as flat as possible, and noise should be removed completely from such regions.
	\item The boundaries in an image should be well preserved since the generally used averaging approach results in blurring out those details.
	\item The texture details of an image should not be removed as they are of extreme importance and have key information. This is one of the most challenging tasks where majority of the algorithms cannot perform well resulting in loosing texture~details.
	\item Artifacts should not appear in the denoised image specifically at very high noise levels.
\end{enumerate}

\textit{\textbf{Contribution:}}
We propose a denoising algorithm to combat AWGN in images meanwhile preserving the details of an image. Our algorithm works by first computing the overlapping patches and then finding structurally similar patches for all reference patches. This is then followed by computing sparse estimates where a collaborative step is performed in sparse domain to compute \textit{apriori} information of the likelihood of the active locations. This information helps us getting a refined estimate of patches via refined estimation step. Finally, a post-processor is applied exploiting the spatial features to further refine smooth regions of the image. This motivate us to name our algorithm \emph{collaborative dual-domain patch filtering} (C2DF). The key features of our algorithm are as follows:
\begin{enumerate}
	\item A Bayesian approach is used to recover the sparse estimates of patches with significant improvements by incorporating any available \textit{apriori} information.
	\item Instead of going for similar intensity patches, we use an intensity-invariant approach. We hunt similar structured patches based on correlation coefficient that helps us perform better collaboration in the transformed-domain.
	\item  Our proposed method is agnostic support distribution and lends itself a computationally-desired implementation.
\end{enumerate}

\subsection{Notations}
\label{notations}
We represent all the vectors used in our work with small case and bold face letters (e.g., $\yv$), all the scalars with small case normal font letters (e.g., $y$). We reserve upper case and bold face letters (e.g., $\Ym$) for matrices. For sets, we use calligraphic notation (e.g., $\Nc$). We use $\yv_i$, $y(j)$ and $\Nc_k$ to denote $i$th column of matrix $\Ym$, $j$th element of vector $\yv$, and a subset of $\Nc$, respectively.


The remainder of this paper is organized as follows. In Section \ref{Problem_Formulation}, we formulate the denoising problem, while our proposed denoising algorithm is described in Section~\ref{ProposedAlgorithm}. Section \ref{SPARSE_RECOVERY_ALGORITHM_SELECTION} discusses the factors that we are considering for the choice of sparse recovery algorithm. This is followed by Section \ref{Computational_Complexity} where we compare the computational overhead of our algorithm. The experimental setup and extensive simulation results are provided in Section \ref{SIMULATION_RESULTS_AND_DISCUSSIONS}, and finally, Section \ref{Conclusion} concludes the paper.

\section{Problem Formulation}
\label{Problem_Formulation}
\subsection{System Model}
\label{System_Model}
We assume the globally adopted linear and spatially
invariant system model \cite{campisi2007blind}. Let $\Xm \in \mathbb{R}^{R\times C}$ represent the matrix version of a clean noiseless image, where $R$ and $C$ denotes the number of rows and columns, respectively. We aim to estimate the latent image $\Xm$ from its noise contaminated observations $\Ym$, modeled as,
\begin{align}\label{eq:Y_X_W}
\Ym &= \Xm + \Wm,
\end{align}
where, $\Wm$ is matrix version of the noise having independent and identically distributed (i.i.d.) entries taken from a Gaussian distribution with zero mean and variance $\sigma_w^2$, i.e., $\Wm \sim \mathcal{N} (\textbf{0},\sigma_w^2\textbf{I})$. Additionally, the estimated or denoised image is denoted by $\widehat{\Xm}$.

\subsection{Sparse Reconstruction}
\label{Sparse_Reconstruction}
\begin{figure}[b]
	\centering
	\includegraphics[width = 0.9\linewidth]{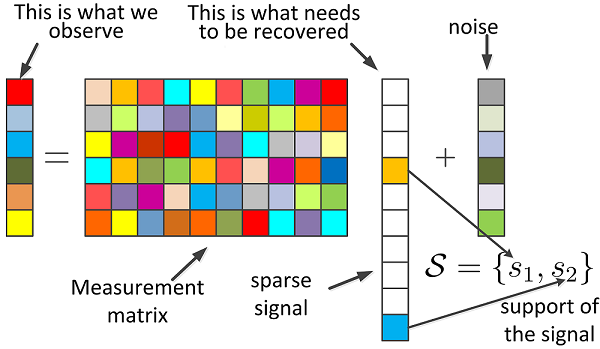}
	\caption{Sparse Recovery Model}
	\label{fig:fig1}
\end{figure}
Since many signals are generally sparse by birth, a number of CS algorithms can be applied on them to recover the meaningful information which gets destroyed during corruption via noise. For this purpose, the following under-determined model of linear equations is used
\begin{align}\label{eq:y_Ah_w}
\yv &= \Am \hv + \wv
\end{align}
where $\yv$, $\Am$, $\hv$ and $\wv$ represents noisy signal, an overcomplete dictionary, sparse representation of the signal and the noisy vector, respectively, as shown in Fig. \ref{fig:fig1}. The objective is to restore the degraded image, by some restoration processes, as close as possible to its original form, i.e., $\widehat{\Xm} \approx \Xm$.

The number of unknown elements in such a scenario is usually much larger than the number of observations. With CS, a true signal can be reconstructed by linear projections of the sparse signal using $\ell_1$-optimization with high probability \cite{6581876}. Many algorithms are proposed in the literature that compete to provide a better estimate of the sparse vector. We are specifically interested in utilizing a Bayesian scheme that enjoys low computational burden and outperforms currently existing sparse reconstruction techniques. In particular, a good estimate of the sparse vector can be provided even when there is no knowledge about signal support prior. More importantly, such sparse recovery algorithms are agnostic to support distribution, and hence, there is no need to estimate distribution parameters.


\section{The Proposed Collaborative Dual-Domain Patch Filtering (C2DF)}
\label{ProposedAlgorithm}
In this section, we present a detailed description of the methodology that we have used for designing our image denoising algorithm. Our proposed method has two major working blocks: 1) the denoiser, and 2) the post-processor, as discussed separately in the following sub-sections. In the denoiser block, an observed image is decomposed into several overlapping patches firstly, and similar-structured patches are computed. Afterwards, a sparse-domain based collaborative approach is used on the similarly-grouped patches to refine the patch estimates. Finally, the patches are placed to their original position. For further improvements in the results, we ultimately pass the denoised image to a post-processor that takes care of the smooth regions and produces significantly promising results. A summary of this has been presented as a block diagram in Fig. \ref{fig:fig31}, each stage of which we will discuss in the following sub-sections. We now explain in detail how the denoiser block works.

\subsection{Image Decomposition}
\label{Image_Decomposition}
For a given noisy image, we first of all form $N \times N$ size squared patches around every pixel where $N$ is always an odd number for proper processing\footnote{Our proposed image denoising algorithm is able to process the general case where patches can be non-squared, i.e., rectangular or even linear. However, for convenience and simplicity, we present the special case of squared patches in this article.}. Such decomposition results in overlapping patches that are useful to mitigate the effect of noise. Moreover, to facilitate the border pixels without introducing any artifacts, we firstly pad borders of the noisy image with $\lfloor \frac{N}{2} \rfloor$ pixels, before processing. This step yields a total number of $K = RC$ patches, where $R$ and $C$ denotes the number of rows and columns of the image, respectively. The patches are thus represented as
\begin{align}\label{eq:Ypatch}
\Ym_k = \Xm_k + \Wm_k, \quad \forall k \in \Kc.
\end{align}
where $\Kc = \{1,2,3,\dots, K-1, K\}$. In order to have computational simplicity and convenience, we denote the patches in (\ref{eq:Ypatch}) as vectors/1D signals and will be utilizing this notation in the coming sections of the paper. The vectorized representation of the patches is given as
\begin{align}\label{eq:Ypatch_vec}
\ykt = \xkt + \wkt, \quad \forall k \in \Kc.
\end{align}
where $\ykt, \xkt$ and $\wkt, \forall k$ are vectors of length $N^2$.

\subsection{Grouping Similar Patches}
\label{Patches_Grouping}
Once the overlapping patches are formed, the next step is to find a certain number of similar patches that would be used for collaboration. The grouping of patches using a similarity measure has led to a number significant improvements in a wide range of application like bio-medical signal processing, computer vision, machine intelligence, etc. (see e.g., \cite{6853394, 7051524, 7351075, 7457822, 7445872}). Some of the similarity based grouping techniques include self-organizing maps \cite{VanHulle2012}, vector quantization \cite{1056457}, fuzzy clustering \cite{hoppner1999fuzzy}, and an extensive review on this in \cite{jain1999data}. The recently developed denoising algorithms mainly use a distance based measure where similarity between different signals are realized in terms of the inverse of the point-wise distance between them. Therefore, a smaller distance between the signals would imply a higher similarity and vice versa. The generally used distance based similarity measure is the Euclidean distance as used by the state-of-the-art denoising algorithms like NL-means and BM3D.

However, despite being an effective way of finding similarity, Euclidean distance based similar-intensity grouping has a limitation; it limits the search for number of similar patches. For instance, even though natural images have some similarity in its structure, the number of similar patches vary. Consequently, in an image having smaller number of similar patches, the collaboration is not that effective disturbing the performance of denoising, severely in case of high noise. For this, novel methods are being proposed to find accurate similar patches. For example, the authors in \cite{7005524} search the similar patches by using not only a patch itself but the noise too by proposing noise similarity. Similarly, the authors in \cite{7444121} propose sequence-to-sequence similarity (SSS) which is essential for preserving edges.

To tackle this case more efficiently, we propose intensity-invariant grouping. The idea is to stack all the patches that have a similar inherent structure without relying on the intensity values as shown in the stage 1 of Fig. \ref{fig:fig31}. The correlation coefficient serves as the best tool to be utilized for the said purpose. Therefore, for two signals $\yv_k$ and $\yv_i$, the correlation coefficient is given as,
\begin{align}\label{eq:corr_coeff}
r(\yv_k,\yv_i) = \frac{ cov(\yv_k,\yv_i) }{\sigma_{\yv_k}\sigma_{\yv_i}},
\end{align}
where $-1 \ge r(\yv_k,\yv_i) \le 1$. A value close to $1$ or $-1$ means larger positive and negative correlation, respectively, while a value close to $0$ means smaller correlation.

Importantly, since the underlying structure would be reflected in the sparse domain, we not only group the patches that have positive correlation (near $+1$) but we also place those patches in the similarity group that have a negative correlation (near $-1$). This is because we search for the patches that have similar structure independent of the sign of correlation coefficient since the signs would be absorbed in the sparse domain. Most importantly, to make such method work more efficiently, the normalization of patches is needed\footnote{Consider the example of a completely black and a completely white patch which are totally different distance-wise but are similar in terms of structure when normalized. This normalization allows us to gather higher number of similar patches for a better collaboration}. Therefore, we normalize the patches in (\ref{eq:Ypatch_vec}) to get the following equivalent representation,
\begin{align}
\yv_k = \eta(\ykt) =
\begin{cases}
\frac{\ykt}{\|\ykt\|}, & \quad \|\ykt\| \ne 0\\
\ykt, & \quad \text{otherwise}
\end{cases}, \quad \forall k \in \Kc,
\end{align}
where $\eta(\cdot)$ denotes the operator for normalization, and $\yv_k$ is a normalized representation of the patch $\ykt$. Consequently, we have the following relationship
\begin{align}\label{eq:Ypatch_vec_normalized}
\yv_k = \xv_k + \wv_k, \quad \forall k \in \Kc
\end{align}
\begin{figure}[t]
	\begin{subfigure}[b]{4.2cm}
		\includegraphics[width=5cm, height=7.5cm]{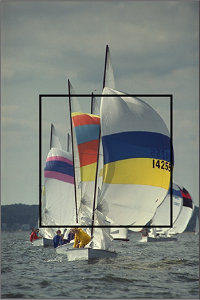}
		\caption{Observed Image}\label{fig:21a}
	\end{subfigure}
	\hspace{1cm}
	\begin{tabular}[b]{@{}c@{}}
		\begin{subfigure}[b]{4.2cm}
			\includegraphics[width=3.4cm, height=3.4cm]{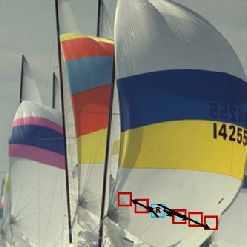}
			\caption{Eucl. Dist.}\label{fig:22b}
		\end{subfigure}\\
		\begin{subfigure}[b]{4.2cm}
			\includegraphics[width=3.4cm, height=3.4cm]{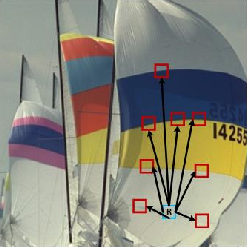}
			\caption{Corr. Coeff.}\label{fig:23c}
		\end{subfigure}
	\end{tabular}
	\caption{Sail boats (Source: Kodak gallery)}\label{fig:main2abc}
	\label{fig:fig2}
\end{figure}

Hence, a patch $\yv_k$, and those among all other patches that have an absolute correlation coefficient greater than $\yv_k$, say $\epsilon$, are placed together in a group. We name these as the neighbors of the $k$th patch, i.e., $\yv_k$. Therefore,
\begin{align}\label{eq:neighbours_set_indices}
\Nc_k = \{i:r(\yv_k,\yv_i) \ge \epsilon\}, \quad \forall k \in \Kc
\end{align}
represents a set of indices for all neighbors of patch $\yv_k$ and the index $k$ itself.

\begin{figure}[t]
	\begin{subfigure}[b]{4.2cm}
		\includegraphics[width=5cm, height=7.5cm]{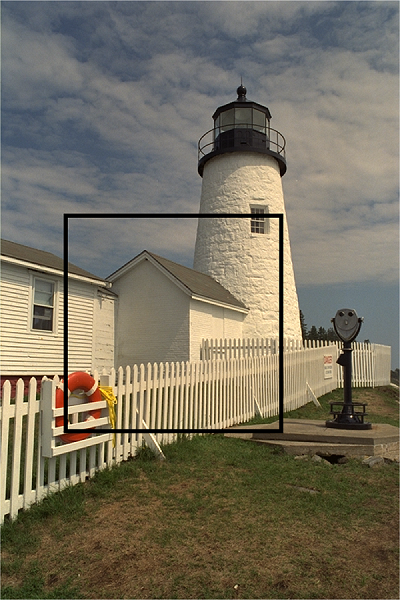}
		\caption{Observed Image}\label{fig:11a}
	\end{subfigure}
	\hspace{1cm}
	\begin{tabular}[b]{@{}c@{}}
		\begin{subfigure}[b]{4.2cm}
			\includegraphics[width=3.4cm, height=3.4cm]{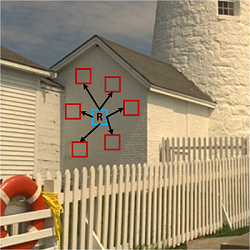}
			\caption{Eucl. Dist.}\label{fig:12b}
		\end{subfigure}\\
		\begin{subfigure}[b]{4.2cm}
			\includegraphics[width=3.4cm, height=3.4cm]{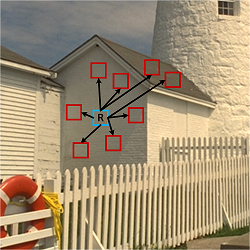}
			\caption{Corr. Coeff.}\label{fig:13c}
		\end{subfigure}
	\end{tabular}
	\caption{Light home (Source: Kodak gallery)}\label{fig:main1abc}
	\label{fig:fig3}
\end{figure}

To understand this completely, consider the images shown in Fig. \ref{fig:fig2} and \ref{fig:fig3} from Kodak gallery\footnote{\href{http://r0k.us/graphics/kodak/}{http://r0k.us/graphics/kodak/.}}. Here, we focus the limitations of distance based similarity, shown in \ref{fig:22b}, which in turn limit the performance of algorithms by producing smaller number of similar patches. A similar scenario is presented in Fig \ref{fig:fig3} where we take a portion of the image, shown in \ref{fig:12b}, and show that distance based measures suffer limitations. Such cases becomes severe when processing images that are contaminated by high noise.

On the contrary, using the intensity-invariant based measure results in a higher number of patches for collaboration. This is because not only such measure will give us the same intensity patches\footnote{By intensity-invariance, we mean that we do not specifically look for similar intensity patches rather we aim to find the patches that have similar underlying structure. This will of course include the patches that are similar intensity-wise but will also group those patches that are similar structure-wise.}, but it will also consider the patches as similar that have similar underlying structure. This is shown in Fig. \ref{fig:23c} and Fig. \ref{fig:13c} where a reference patch has access to similar structure patches throughout the image, and is significantly less-constrained.

Furthermore, It's worth noting that by virtue of the above presented definition in (\ref{eq:neighbours_set_indices}), the set of neighbors $\Nc_k$ are not restricted to be spatial neighbors and the set of neighbors $\Nc_k, \forall k\in \Kc$ are not disjoint. The upshot of this novel patches grouping is that it produces a higher number of neighbors for each patch which is beneficial for our sparse-domain based collaborative approach.

\subsection{Estimation Via Collaborative Denoising}
\label{Estimation_Via_Collaborative_Denoising}
Since images are inherently sparse in the wavelet domain, we aim to transform the patches in (\ref{eq:Ypatch_vec_normalized}) for computing the equivalent sparse representations. For this purpose, we use
\begin{align}\label{eq:sparse_equation}
\yv_k & = \Am \hv_k + \wv_k, \quad \forall k \in \Kc
\end{align}
where $\Am \in \mathbb{R}^{N^2 \times M}, M \gg N^2,$ is an overcomplete dictionary having wavelet basis. Furthermore, $\hv_k \in \mathbb{R}^{M}$  is the equivalent sparse representation of the spatial-domain patch~$\xv_k$, i.e., $\xv_k = \Am \hv_k$. 

For the sparse reconstruction, we let $\widehat{\hv}_k$ denote a recovered estimate of the sparse vector $\hv_k$ obtained via a sparse recovery algorithm, and let $\Sc_k$ represent the set of active indices in the sparse vector, i.e., its support set. In order to isolate noise, we process each patch individually. For that, MMSE estimate of sparse vector $\hv_k$ of all patches are computed, without any a priori information, given as
\begin{align}\label{eq:MMSE1}
\widehat{\hv}_k  \delequal \mathop{\mathbb{E}}[\hv_k|\xv_k] =\sum_{\Sc_k} p(\Sc_k|\xv_k)\mathop{\mathbb{E}} [\hv_k|\xv_k,\Sc_k].
\end{align}
The following explains how the sum, the posterior $p(\Sc_k|\xv_k)$, and the expectation $\mathop{\mathbb{E}} [\hv_k|\xv_k,\Sc_k]$ in (\ref{eq:MMSE1}) are evaluated. Given the support $\Sc_k$, (\ref{eq:sparse_equation}) becomes
\begin{align}\label{eq:MMSE2}
\xv_k = \Am_{\Sc_k}\hv_{\Sc_k} + \wv_k
\end{align}
where $\Am_{\Sc_k}$ is the matrix containing $\Sc_k$ indexed columns of $\Am$. Likewise, $\hv_{\Sc_k}$ is formed by $\Sc_k$ indexed entries of $\hv_k$. Since the distribution of $\hv_k$ is unknown making $\mathop{\mathbb{E}} [\hv_k|\xv_k,\Sc_k]$ very difficult to compute, we use the best linear unbiased estimate (BLUE) as
\begin{align}\label{eq:MMSE3}
\mathop{\mathbb{E}} [\hv_k|\xv_k,\Sc_k]	\gets	(\Am_{\Sc_k}^{\hermconj}\Am_{\Sc_k})^{-1}\Am_{\Sc_k}^{\hermconj}\xv_k.
\end{align}
Using Bayes rule, we can write the posterior as
\begin{align}\label{eq:MMSE4}
p(\Sc_k|\xv_k)=\frac{p(\xv_k|\Sc_k)p(\Sc_k)}{p(\xv_k)}.
\end{align}
We can ignore $p(\xv_k)$ in (\ref{eq:MMSE4}) as it's a common factor to all posterior probabilities. Since entries of $\hv_k$ are activated with Bernoulli distribution having $p$ as success probability, then
\begin{align}\label{eq:MMSE5}
p(\Sc_k)=p^{|\Sc_k|}(1-p)^{Q-|\Sc_k|}.
\end{align}
For $p(\xv_k|\Sc_k)$, if $\hv_{\Sc_k}$ is Gaussian, then $p(\xv_k|\Sc_k)$ would also be Gaussian which is easy to compute. On the contrary, evaluating $p(\xv_k|\Sc_k)$ is difficult for unknown or non-Gaussian $\hv_{\Sc_k}$ distribution. To tackle this, it can be noted that $\xv_k$ is formed by a vector in the subspace that is spanned by $\Am_{\Sc_k}$ columns with additive Gaussian noise $\wv_k$. By projecting $\xv_k$ on orthogonal complement space of $\Am_{\Sc_k}$, the non-Gaussian component is eliminated. This is achieved using the projection matrix $\textbf{P}^{\perp}_{\Sc_k}=\textbf{I}-\textbf{P}_{\Sc_k}=\textbf{I}-\Am_{\Sc_k}(\Am_{\Sc_k}^{\hermconj}\Am_{\Sc_k})^{-1}\Am_{\Sc_k}^{\hermconj}$ giving $\textbf{P}^{\perp}_{\Sc_k}\xv_k=\textbf{P}^{\perp}_{\Sc_k}\wv_k$ that is Gaussian with zero mean and covariance $\textbf{K}_k=\sigma_{\wv_k}^2\textbf{P}_{\Sc_k}^{\perp}$. Dropping some exponential terms, and simplifying gives us
\begin{align}\label{eq:MMSE6}
p(\xv_k|\Sc_k)	\simeq \exp(-\frac{1}{2\sigma_{\wv_k}^2}\|\textbf{P}^{\perp}_{\Sc_k}\xv_k\|^{2}).
\end{align}
In this way, we can evaluate the sum in (\ref{eq:MMSE1}). However, evaluating this sum for larger $Q$ demands more computational complexity that can be controlled by summing over less support sets $\Sc_k \in \Sc_k^{d}$ having significant posteriors. A greedy algorithm can be used to find $\Sc_k^{d}$ that we'll discuss further in the coming sections.

It is worth noting that in an ideal situation, $\Sc_k = \Sc_i, \forall i\in \Nc_k$, should hold true for all $k \in \Kc$, i.e., the support sets of similar patches should be similar in general. This motivates us to utilize the transformed-domain representation of the patches to devise a sparse-domain based collaborative denoising algorithm. However, this may not be the exact case in reality, and the support sets of similar patches may have some differences since $\Nc_k$ is a function of a non-zero $\epsilon$ as well as $\wv_k$. Based on this, the value of the threshold $\epsilon$ serves as a key parameter in the collaborative approach that's why it should be selected very properly. This is one of the main reasons why we are using the correlation coefficient based similarity approach so that higher similarity between the groups can be guaranteed.

Even though the correlation coefficient based approach would result in identifying the similar structured patches, the disturbances caused by the noise introduction would still persist yielding a disagreement among the support sets of similarly grouped patches. However, we would like to bring this into notice tthat this disagreement is actually a blessing in disguise. Provided adequately large $\epsilon$, majority of the outliers $\Vc_k = \bigcup_{i\in\Nc_k} \Sc_i \backslash \bigcap_{i\in\Nc_k} \Sc_i$ in the support sets are there, with high probability, due to noise. Such information ultimately guides us to find and take care of the noise-producing locations in the recovered estimate $\widehat{\hv}_k$. One naive tactic is to diminish the contribution of the non-zero components of $\widehat{\hv}_k$ located at $\Vc_k$ by eliminating it, and use this resulting sparse vector estimate to compute an estimate of the denoised spatial-domain patch using $\widehat{\xv}_k = \Am \widehat{\hv}_k$, as used in the majority of the sparse domain based denoising algorithms in the literature. However, this would basically result in discarding significantly critical information specifically in the high noise regime since some legitimate non-zero components may be mistaken for noise-producing components. In view of the stated problem, we resort to a much moderate method to tackle this case efficiently.

\begin{figure*}[t]
	\centering
	\includegraphics[width=\linewidth]{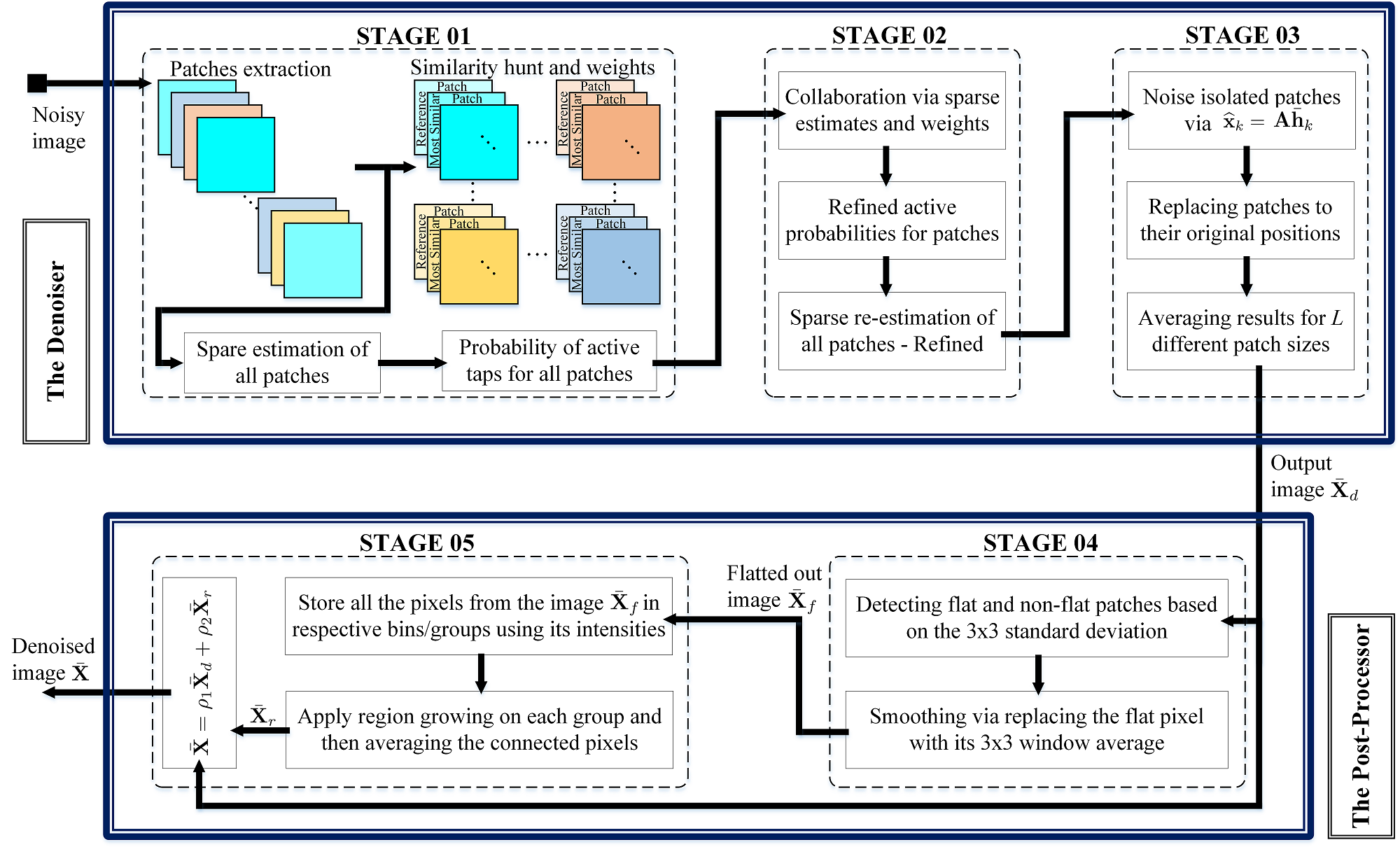}
	\caption{Flowchart of proposed C2DF denoising algorithm}
	\label{fig:fig31}
\end{figure*}
In this method, we take advantage by using the active taps probabilities of $\hv_k$, i.e., the non-zero locations of the sparse vector. The approach is that since the similar patches are grouped together, and their support would be similar, so the legitimate non-zero active locations of their corresponding sparse vectors will have high probabilities of being active as described briefly in stage 2 of Fig. \ref{fig:fig31}. Consequently, we propose this novel sparse-domain based collaboration among similar patches via active probabilities. In particular, for the $k$th patch, we let $\lambdav_k\in\mathbb{R}^{M}$ denote the active probabilities vector for the estimate $\widehat{\hv}_k$. Driven by this, we find the weighted average as follows
\begin{align}\label{eq:est_active_prob}
\lambdav_k^\prime &=
\frac{1}{\Nc_k}\sum_{j\in \Nc_k} \alpha_{j,k} \lambdav_j, \quad \forall k \in \Kc.
\end{align}
This serves as a refined estimate of the vector corresponding to the active probability of clean $\hv_k$. Here, the weighting factor is proportional to the similarity between the patches, and hence, the probability vectors
\begin{align}
\alpha_{j,k} \propto r(\yv_j, \yv_k), \quad j\ne k.
\end{align}

This simple yet effective approach significantly reduces the contribution of solitary active locations meanwhile keeping the information on the active taps common to majority of the similar patches in $\Nc_k$. Additionally, by virtue of the law of large numbers, (\ref{eq:est_active_prob}) will produce a better refined estimate specifically since $|\Nc_k|$ is large due to the intensity-invariant grouping method.

The obtained clean $\lambdav_k^\prime$ is a valuable piece of information that serves as a guide for the a priori information about the active taps locations of clean sparse-domain representation of the $k$th patch $\xv_k$. This a priori information is supplied to a sparse recovery algorithm to compute a refined estimate of true $\hv_k$, let's denote it by $\bar{\hv}_k$, and thus an estimate of true as well as denoised $k$th patch that we represent as $\widehat{\xv}_k$. To get the denoised patch, we de-normalize the noise-removed patch as follows
\begin{align}
\widehat{\xv}_k =
\begin{cases}
\eta^{-1}(\Am \bar{\hv}_k) = \Am \bar{\hv}_k \|\ykt\| \quad & \|\ykt\| \ne 0\\
\eta^{-1}(\Am \bar{\hv}_k) = \Am \bar{\hv}_k \quad & \text{otherwise}
\end{cases}, \quad \forall k \in \Kc.
\end{align}

\subsubsection{Dictionary De-correlation}
\label{Dictionary_De-correlation}
As we presented in (\ref{eq:sparse_equation}), each patch can be written as linear combination of basis elements from the dictionary. The columns of this dictionary are derived from wavelet basis, and are normalized to have unit norms. Prior finding support sets of $\widehat{\hv}_k$ via sparse estimation of patches, we reduce the correlation between dictionary columns for a robust computational and performance ability. Consequently, we remove weak supports by rejecting highly correlated columns as the information they encode could easily be encoded by other columns which correlate with them.
\begin{align}\label{eq:dictionary_decorrelation}
\Am = \Gamma_{\beta}(\Am')
\end{align}
where $\Gamma_{\beta}(.)$ is the de-correlation operator that removes all the columns of $\Am'$ with correlation greater than $\beta$.

\subsection{Reconstructing Denoised Image}
\label{Formation_of_Denoised_Image}
As explained in Section \ref{Image_Decomposition}, we decompose an image to form overlapping patches for a much better performance. Consequently, every pixel in the image is present in $N^2$ due to the overlapping approach, and therefore, each pixel has the same number of estimated denoised intensity values. To place the patches back to their original positions, and reconstruct the denoised image $\widehat{\Xm^a}$, we average the results of $N^2$ pixels for each pixel which in turn provides another level of denoising impurities. To increase the performance further, we implement a weighted average approach for the denoising results based on different $L$ patch sizes. As a result we get the further purified restored image from the denoiser, as shown in stage 3 of Fig. \ref{fig:fig31}, block as follows
\begin{align}\label{eq:diff_patch_size}
\bar{\Xm}_d = \sum\limits_i \gamma_i \widehat{\Xm^a}_i, \quad i = 3, 5, 7 \dots
\end{align}
where $\gamma_i$ and $\widehat{\Xm^a}_i$ are the weights and image from the denoiser block based on $i$th odd patch size, respectively. The aforementioned steps of the denoiser block have also been presented in the Algorithm \ref{Algo:The_Denoiser}.
\begin{algorithm}[b!]
	\caption{The Denoiser}
	\label{Algo:The_Denoiser}
	\begin{algorithmic}[1]
		\Procedure{Collaborative Sparse Recovery Based Denoising}{}	 		
		\For{$i$th odd patch size, $i = 3, 5, 7 \text{ and } 9$}
		\State Extract patches $\ykt, \forall k \in \Kc$
		\For{each $k$th patch $\yv_k$}
		\State Compute $\Nc_k = \{i:r(\yv_k,\yv_i) \ge \epsilon\}$
		\State Find weighting matrix containing $\alpha_{j,k}$ entries
		\State Find active probabilities $\lambdav_k$ via $\widehat{\hv}_k$
		\State Compute $\lambdav_k^\prime =\frac{1}{\Nc_k}\sum_{j\in \Nc_k} \alpha_{j,k} \lambdav_j$
		\State Find $\bar{\hv}_k$ via SABMP using $\lambdav_k^\prime$
		\State Denoise the patch using $\widehat{\xv}_k = \eta^{-1}(\Am \bar{\hv}_k)$
		\EndFor
		\State Restore all the denoised patches to from $\widehat{\Xm^a}_i$
		\EndFor
		\State Average the results $\bar{\Xm}_d = \sum\limits_i \gamma_i \widehat{\Xm^a}_i$
		\EndProcedure
	\end{algorithmic}
\end{algorithm}
\subsection{The Post-Processor}
\label{Post_Processor}
Once we have the denoised image from the denoiser block $\bar{\Xm}_d$, we pass it to the post-processor block to further take care of the smooth regions, and remove the remaining noise-components, if any. The post-processor block has two major steps to tackle the remaining noise elements: 1) detecting and processing flat/smooth patches, and 2) applying region growing to enhance the smoother regions.

\subsubsection{Detecting Flatness}
\label{Detecting_Flatness}
\begin{figure}[b]
	\centering
	\begin{subfigure}{4.2cm}
		\centering
		\includegraphics[width=4cm, height=4cm]{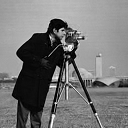}
		\caption{$512 \times 512$ Cameraman}
		\label{fig:std_filt_a}
	\end{subfigure}%
	\begin{subfigure}{4.2cm}
		\centering
		\includegraphics[width=4cm, height=4cm]{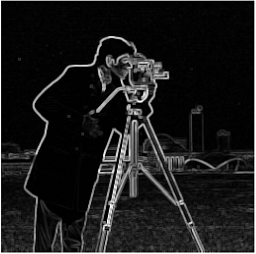}
		\caption{Standard Deviation}
		\label{fig:std_filt_b}
	\end{subfigure}
	\caption{An example of detecting flat regions of \textit{Cameraman} image using standard deviation of a $3 \times 3$ neighborhood as the detection measure}
	\label{fig:std_filt}
\end{figure}
To detect flat regions, we use standard deviation based detection method, an example of which has been shown in Fig. \ref{fig:std_filt}. In Fig. \ref{fig:std_filt_a}, we present a clean $512 \times 512$ \textit{Cameraman} image, while in Fig. \ref{fig:std_filt_b}, we show its standard deviation version. In this method, an image has to be padded first to accommodate the border pixels. Then, a $3 \times 3$ window based neighborhood approach is applied to find out the flatness of the patch centered at the reference pixel. The standard deviation of this $3 \times 3$ patch decides whether to label it as a flat or non-flat region, i.e, a smaller standard deviation value would correspond to a flat region, while a larger value would mean that it's a non-flat region and has edge details.

As can be seen in Fig. \ref{fig:std_filt_b}, the edges have been separated properly, and the flat regions have been segmented out. Now, this of course yield quite promising detection for clean images but might miss out some flat patches in the noisy image. For this reason, we denoise our image first to have a properly-recovered image, and then pass it to the post-processor to perform the stated operations. The threshold $\zeta$ to decide whether a patch should be labeled as smooth or non-smooth has been set based on observing a number of natural as well as synthetic images.

Once the patches are detected and labeled as flat, we then process to smooth these out for discarding the remaining noisy components hidden in the flat regions, as summarized in stage 4 of Fig. \ref{fig:fig31}. For a flat patch, we replace the center pixel by the average of pixels in its $3 \times 3$ neighborhood. As a result, we get the refined flat pixels\footnote{By flat pixel, we mean a pixel whose neighborhood has been detected as flat. In our case, if the standard deviation of the $3 \times 3$ neighborhood of a reference pixel is smaller than $\zeta$, then it's called as a flat pixel.} that we put back to their original positions to reconstruct the image as follows
\begin{align}
\bar{\Xm}_{f}(z) =
\begin{cases}
\bar{\xv}^f_k(z) \hspace{0.17cm}= \frac{1}{|\bar{\xv}^d_k|}\sum_{j=1}^{|\bar{\xv}^d_k|} \bar{\xv}^d_k(j) \hspace{0cm} & \sigma(\bar{\xv}^d_k) \le \zeta \vspace{0.15cm} \\
\bar{\xv}^{nf}_k(z) = \bar{\xv}^d_k(z) \hspace{0cm} & \text{otherwise}
\end{cases}, \hspace{0cm} \forall k \in \Kc
\end{align}
where $\bar{\xv}^d_k$ is the $k$th patch extracted from the output image $\bar{\Xm}_d$ of the denoiser block, and $z$ represent the index location of each pixel. The superscripts $f$ and $nf$ in the terms $\bar{\xv}^f_k(z)$ and $\bar{\xv}^{nf}_k(z)$ correspond to the processed flat and non-flat pixel, respectively, and $\bar{\Xm}_{f}(z)$ represents the $z$th pixel of the smoothed-out image $\bar{\Xm}_{f}$ using the aforementioned process.
\begin{algorithm}[h!]
	\caption{The Post-Processor}
	\label{Algo:Post_Processor}
	\begin{algorithmic}[1]
		\Procedure{Region Growing Based Averaging}{}
		\State Extract patches $\bar{\xv}^d_k, \forall k \in \Kc$
		\For{each $k$th patch $\bar{\xv}^d_k$} 
		\If{$\sigma(\bar{\xv}^d_k) \le \zeta$} \State $\bar{\xv}^f_k(z) = \frac{1}{|\bar{\xv}^d_k|}\sum_{j=1}^{|\bar{\xv}^d_k|} \bar{\xv}^d_k(j)$		
		\EndIf
		\EndFor
		\State Restore the patches to form $\bar{\Xm}_{f}$
		\State Form pixel intensity based bin groups
		\State Apply region growing to find connected pixels
		\State Refine connected pixels
		\State Construct post-processed image $\bar{\Xm}_{r}$		 
		\EndProcedure
	\end{algorithmic}
\end{algorithm}

\begin{figure*}[t]
	\centering
	\begin{subfigure}[b]{\textwidth}
		\centering
		\includegraphics[width=0.95\linewidth]{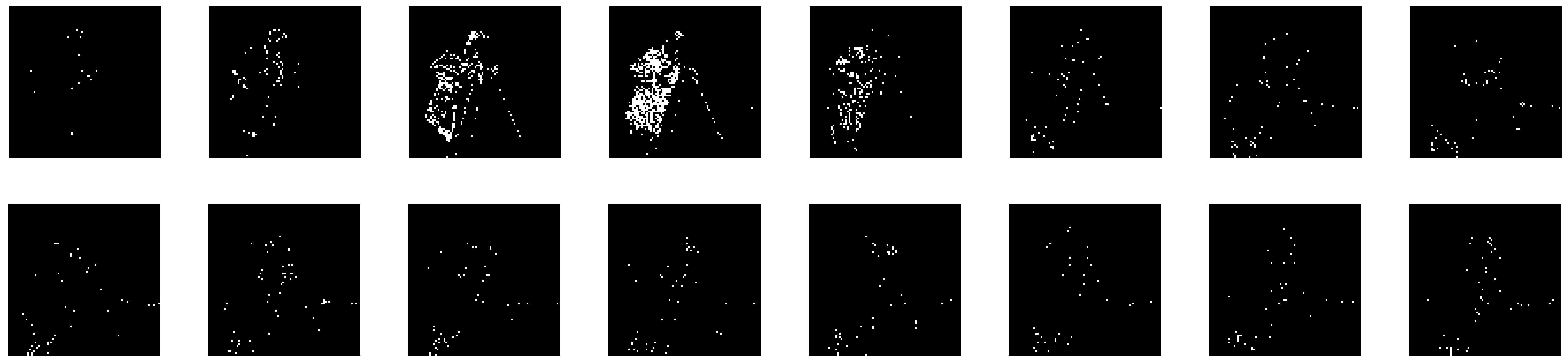}
		\label{fig:RG1}
	\end{subfigure}\\
	\vspace{0.45cm}
	\begin{subfigure}[b]{\textwidth}
		\centering
		\includegraphics[width=0.95\linewidth]{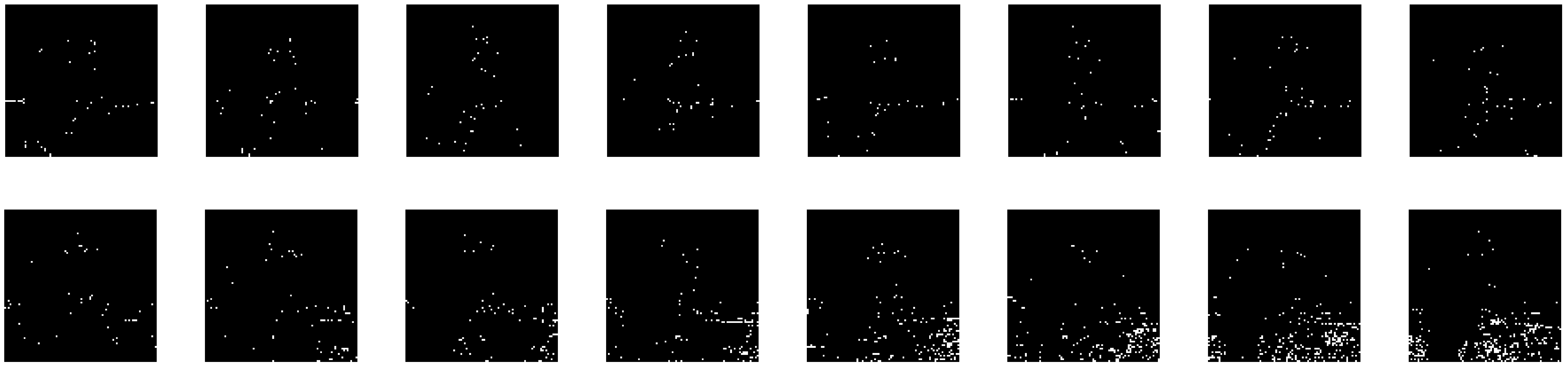}
		\label{fig:RG2}
	\end{subfigure}\\
	\vspace{0.45cm}
	\begin{subfigure}[b]{\textwidth}
		\centering
		\includegraphics[width=0.95\linewidth]{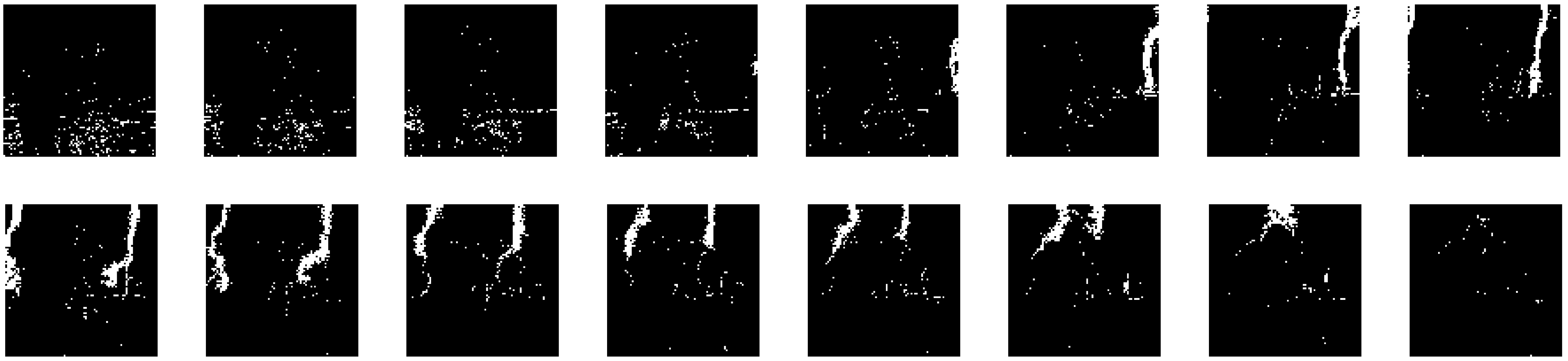}
		\label{fig:RG3}
	\end{subfigure}\\
	\vspace{0.45cm}
	\begin{subfigure}[b]{\textwidth}
		\centering
		\includegraphics[width=0.95\linewidth]{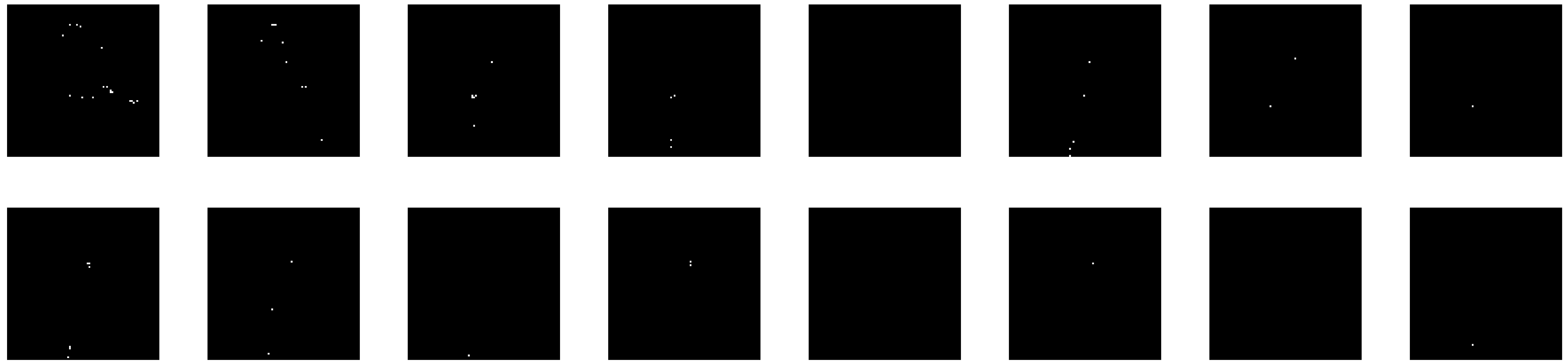}
		\label{fig:RG4}
	\end{subfigure}
	\caption{An example of dividing the \textit{Cameraman} image into 64 different groups/bins (left to right): first row; group 1-8, second row; group 2-16, third row; group 17-24, fourth row; group 25-32, fifth row; group 33-40, sixth row; group 41-48, seventh row; group 49-56, 8th row; group 57-64}
	\label{fig:Reg_Growing}
\end{figure*}
\subsubsection{Region Growing}
\label{Region_Growing}
As a final step, we perform region growing on $\bar{\Xm}_{f}$. For this, we store the pixels in different number of bins based on their intensity levels. For instance, we assign group 1 to the pixels that have, let's say, intensity range from 0-3, group 2 to intensities from 4-7, and so on. We do this for all the pixels thereby creating different bins having pixel intensities and its locations stored. We show an example of applying such intensity-leveling on \textit{Cameraman} image in Fig. \ref{fig:Reg_Growing}. In this figure, we display all bins or intensity groups as binary images where white pixels correspond to the relevant group.

For each bin, we apply region growing to find the connected pixels. This means that the similar intensity pixels are identified locally first. Afterwards, if the number of connected pixel is significant, we replace those pixels by its average. Similarly, we repeat this process for all the bins that ultimately provide us with the region growing processed image that we denote by $\bar{\Xm}_{r}$. Finally, we get our final denoised image $\bar{\Xm}$, shown in stage 5 of Fig. \ref{fig:fig31}, using the weighted average of the image $\bar{\Xm}_{d}$ from denoiser and the region growing processed image $\bar{\Xm}_{r}$ as follows
\begin{align}\label{eq:post_processed}
\bar{\Xm} =  \rho_1 \bar{\Xm}_{d} + \rho_2 \bar{\Xm}_{r}
\end{align}
where $\rho_1$ and $\rho_2$ are the weights which are function of the noise variance. The steps of our proposed post-processor are presented briefly in Algorithm \ref{Algo:Post_Processor}.

\section{Sparse Recovery Algorithm Selection}
\label{SPARSE_RECOVERY_ALGORITHM_SELECTION}
Our C2DF image denoising algorithm involves estimation of the sparse vectors $\widehat{\hv}_k$ and $\bar{\hv}_k$ as discussed previously. Even though there exist a number of algorithms for sparse recovery that provide the estimate efficiently, we have to be cautious in our selection for such sparse recovery algorithm. In particular, the nature of tackling our denoising problem dictates that such algorithm should:
\begin{itemize}
	\item not pose strict conditions on the dictionary matrix $\Am$,
	\item be able to estimate parameters such as sparsity and variance of unknown vectors if not provided,
	\item be invariant to the distribution of unknown, and 
	\item be capable of utilizing any available a priori info.
\end{itemize}
Several number of sparse algorithms are proposed in the literature that provides some of the aforementioned attributes. However, very few have all of the mentioned attributes. Among such sparse recovery methods, we are interested specifically in SABMP \cite{6581876} due to its nature of fulfilling our requirements. This algorithm is capable of MMSE estimation even in the case when the probability distribution of the unknown vector is unavailable. Additionally, it also provides active taps probabilities along with the estimated sparse vector which is one of the main advantages from which our proposed algorithm is benefiting.
\begin{figure*}[t]
	\begin{subfigure}{\textwidth}
		\centering
		\begin{subfigure}{.16\textwidth}
			\centering
			\includegraphics[width=2.9cm, height=2.8cm]{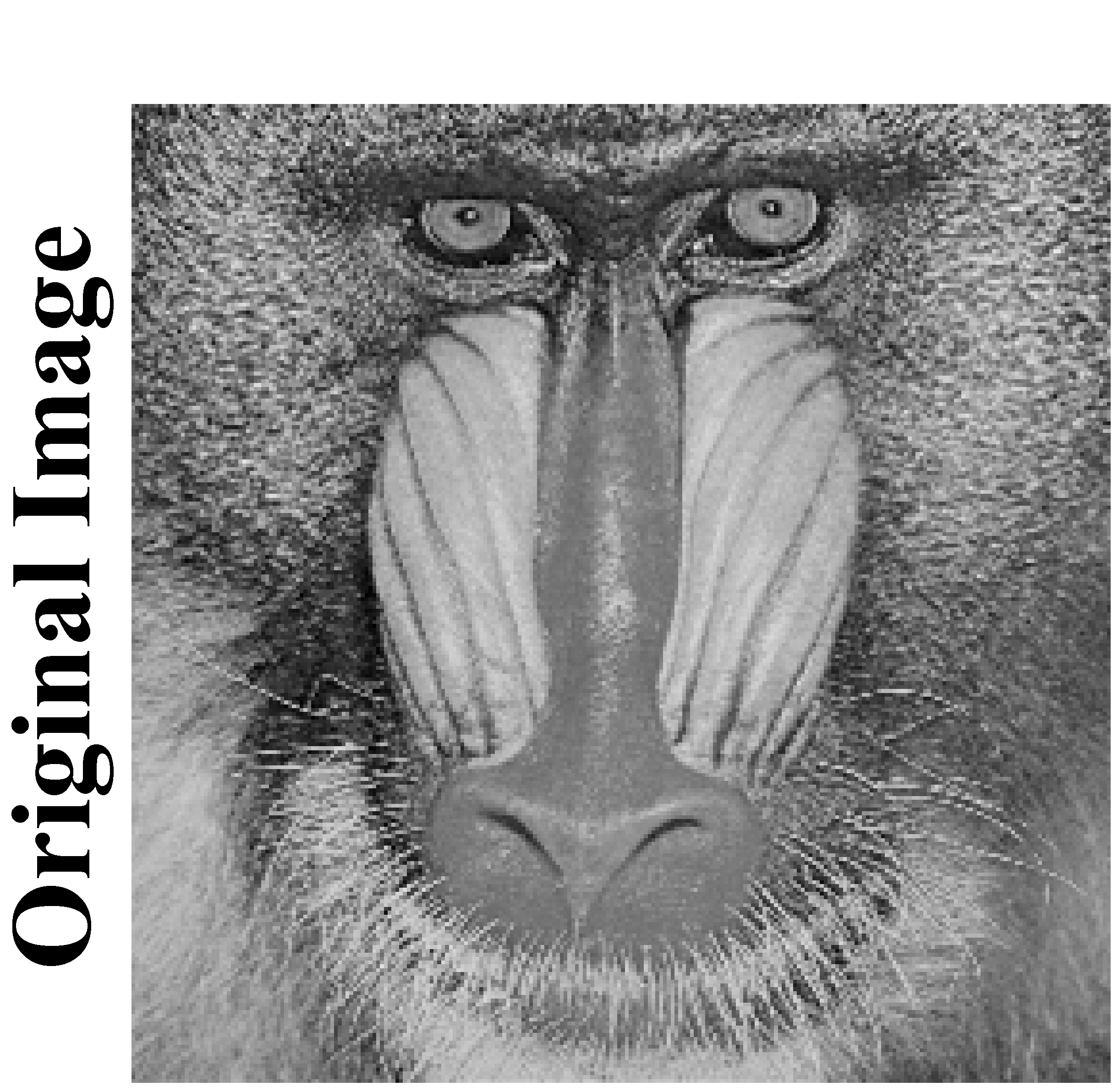}
		\end{subfigure}%
		\begin{subfigure}{.16\textwidth}
			\centering
			\includegraphics[width=2.9cm, height=2.8cm]{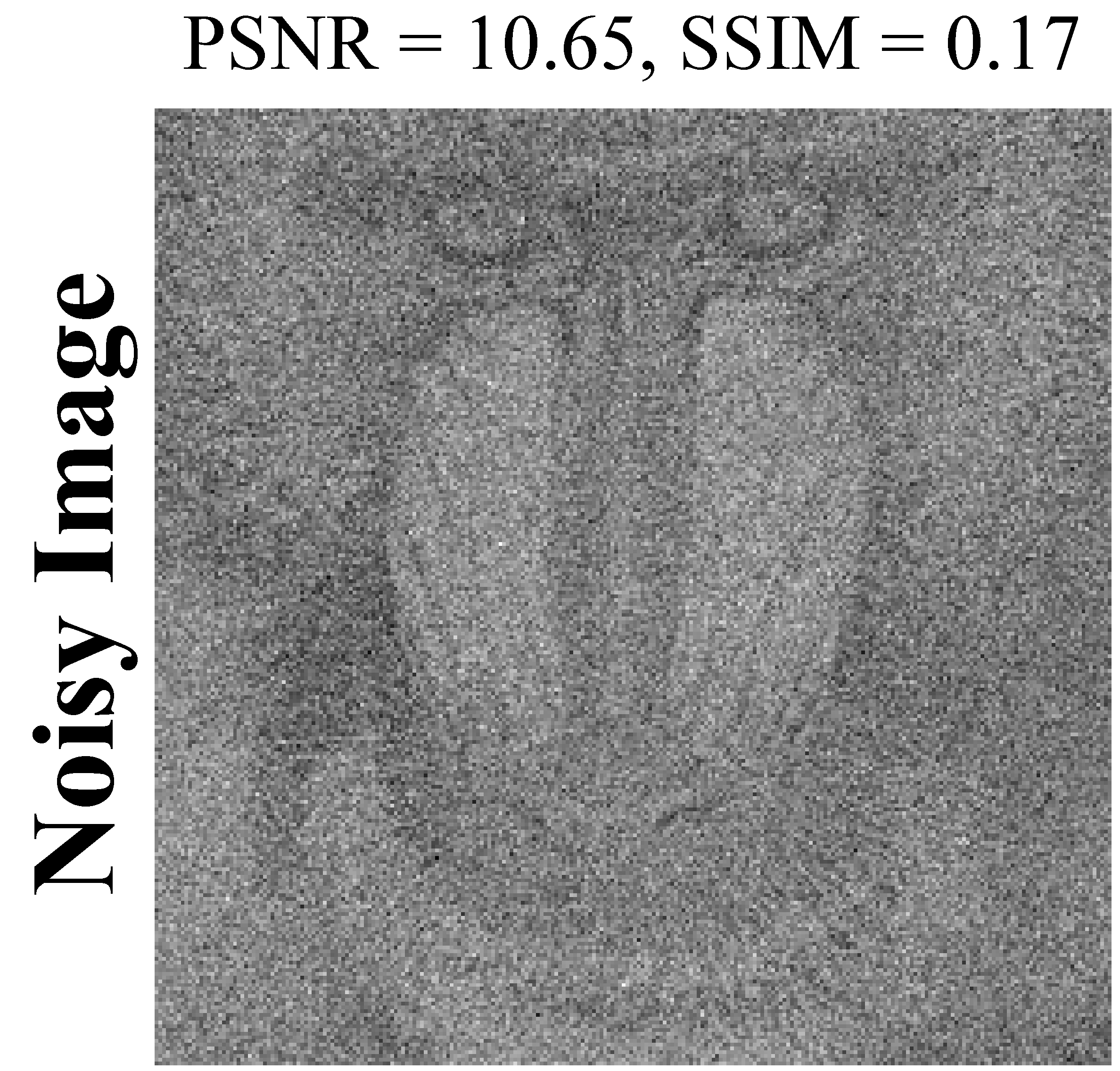}
		\end{subfigure}
		\begin{subfigure}{.16\textwidth}
			\centering
			\includegraphics[width=2.9cm, height=2.8cm]{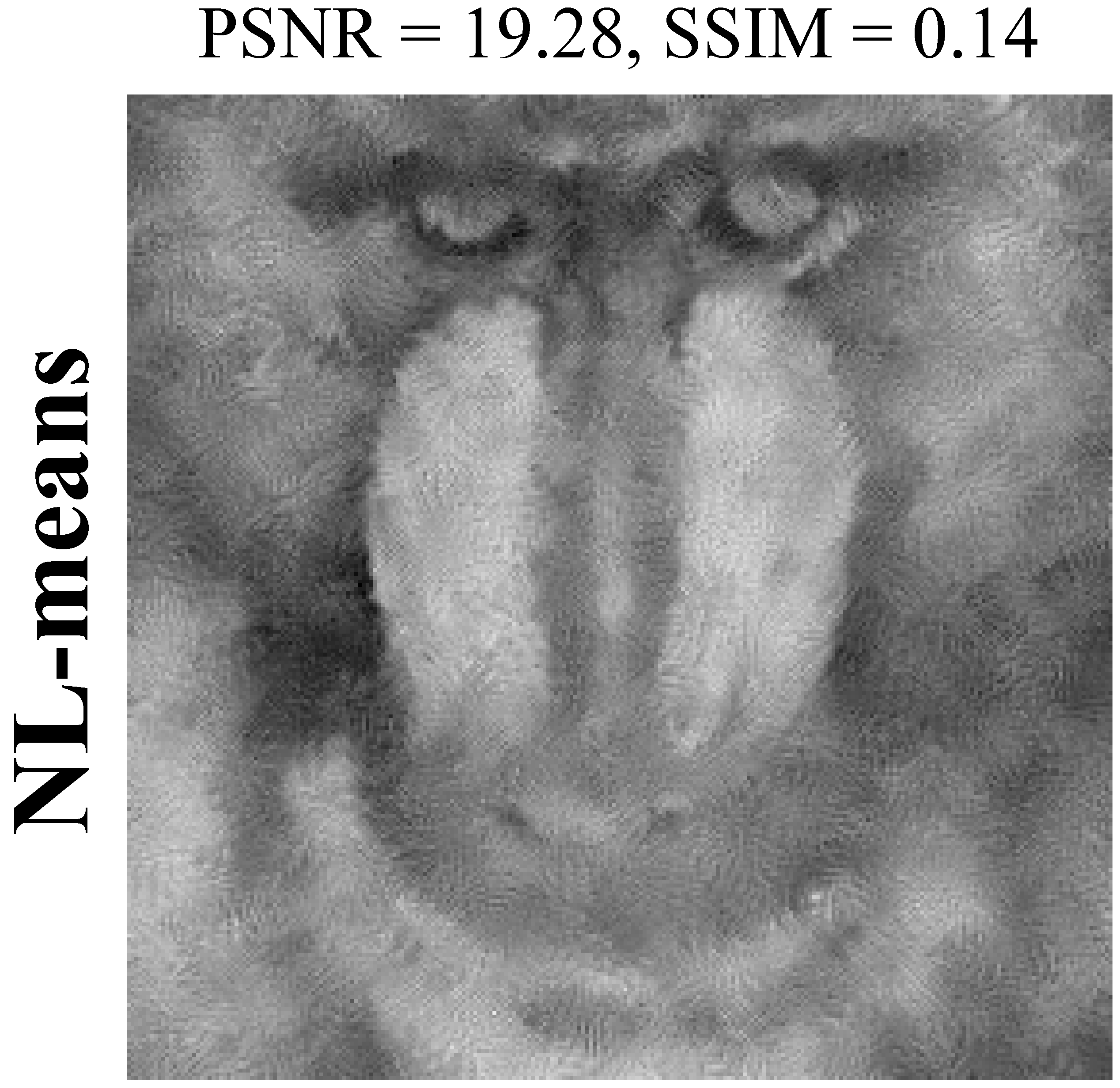}
		\end{subfigure}
		\begin{subfigure}{.16\textwidth}
			\centering
			\includegraphics[width=2.9cm, height=2.8cm]{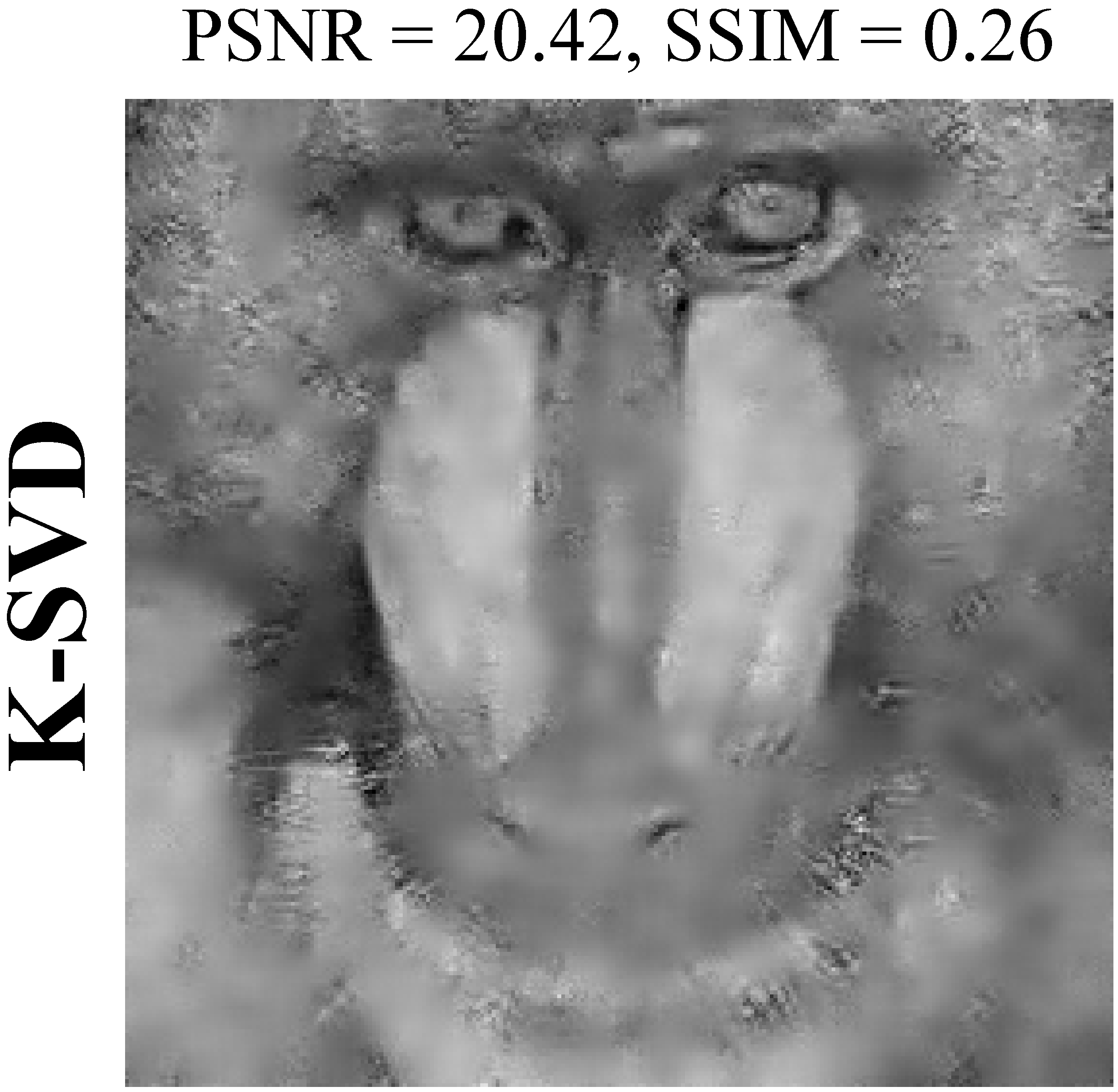}
		\end{subfigure}
		\begin{subfigure}{.16\textwidth}
			\centering
			\includegraphics[width=2.9cm, height=2.8cm]{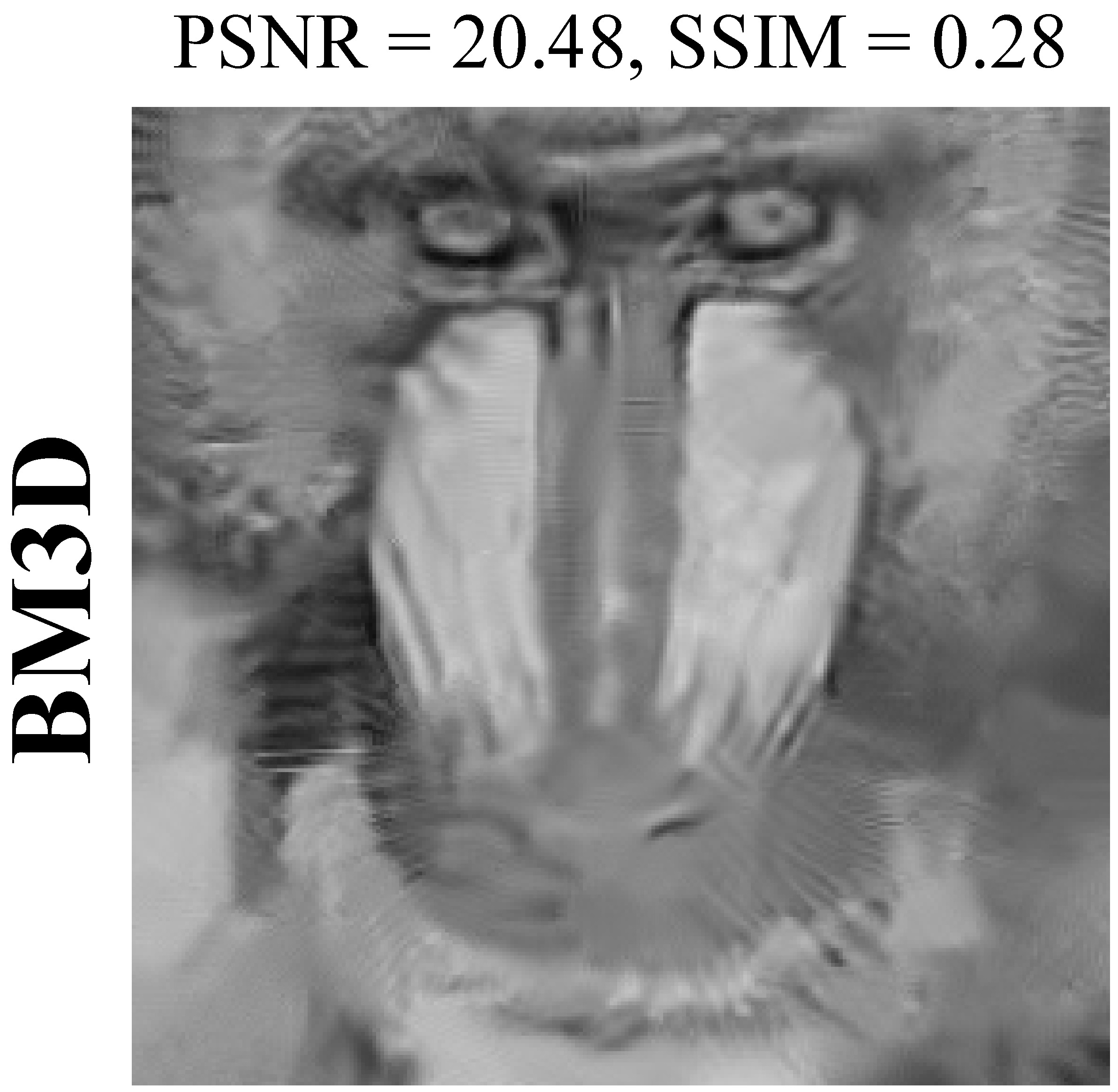}
		\end{subfigure}
		\begin{subfigure}{.16\textwidth}
			\centering
			\includegraphics[width=2.9cm, height=2.8cm]{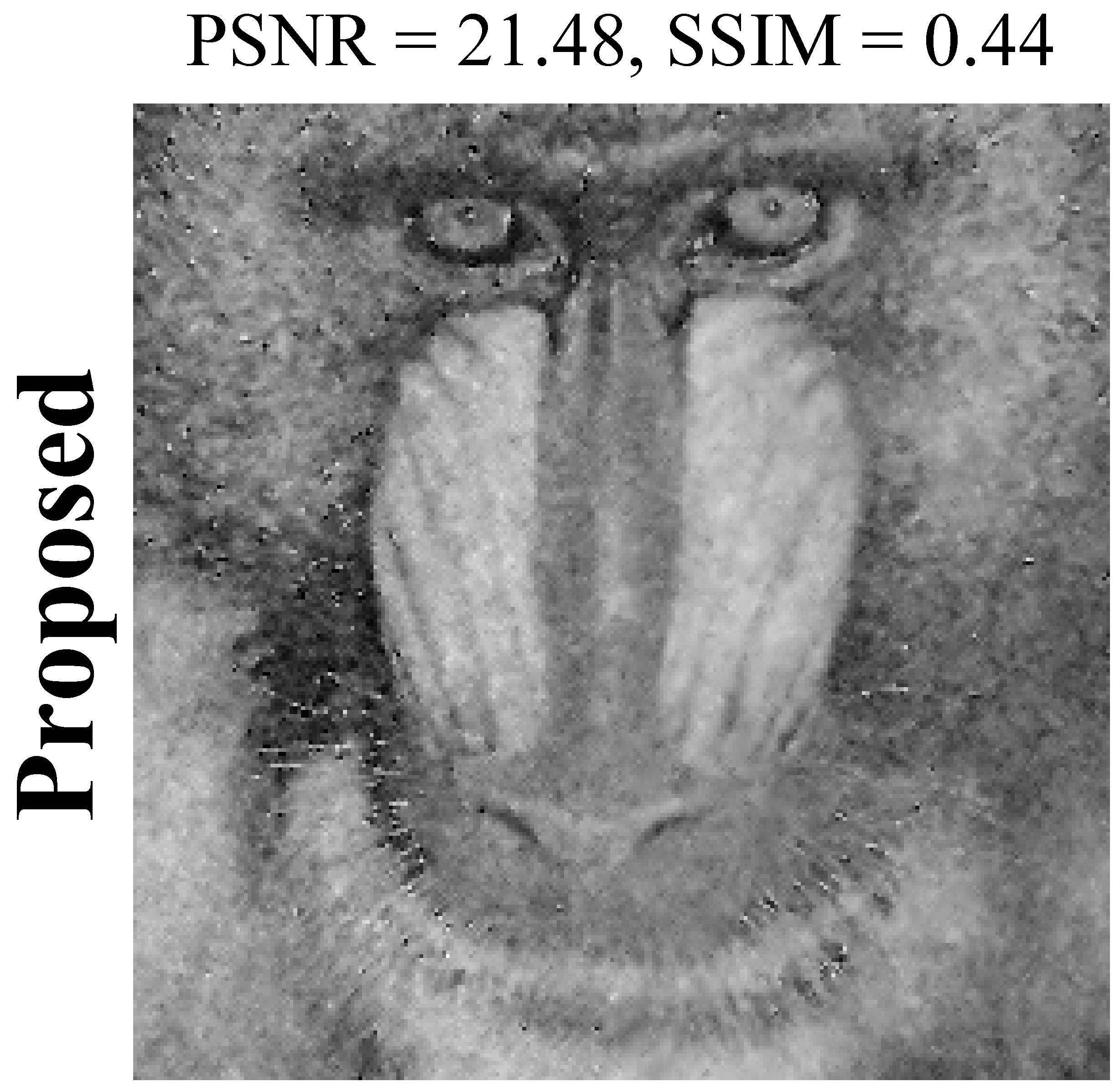}
		\end{subfigure}
	\end{subfigure} \\
	\begin{subfigure}{\textwidth}
		\centering
		\begin{subfigure}{.16\textwidth}
			\centering
			\includegraphics[width=2.9cm, height=2.8cm]{./Figures/Simu_Res_1/mandril_original.png}
		\end{subfigure}%
		\begin{subfigure}{.16\textwidth}
			\centering
			\includegraphics[width=2.9cm, height=2.8cm]{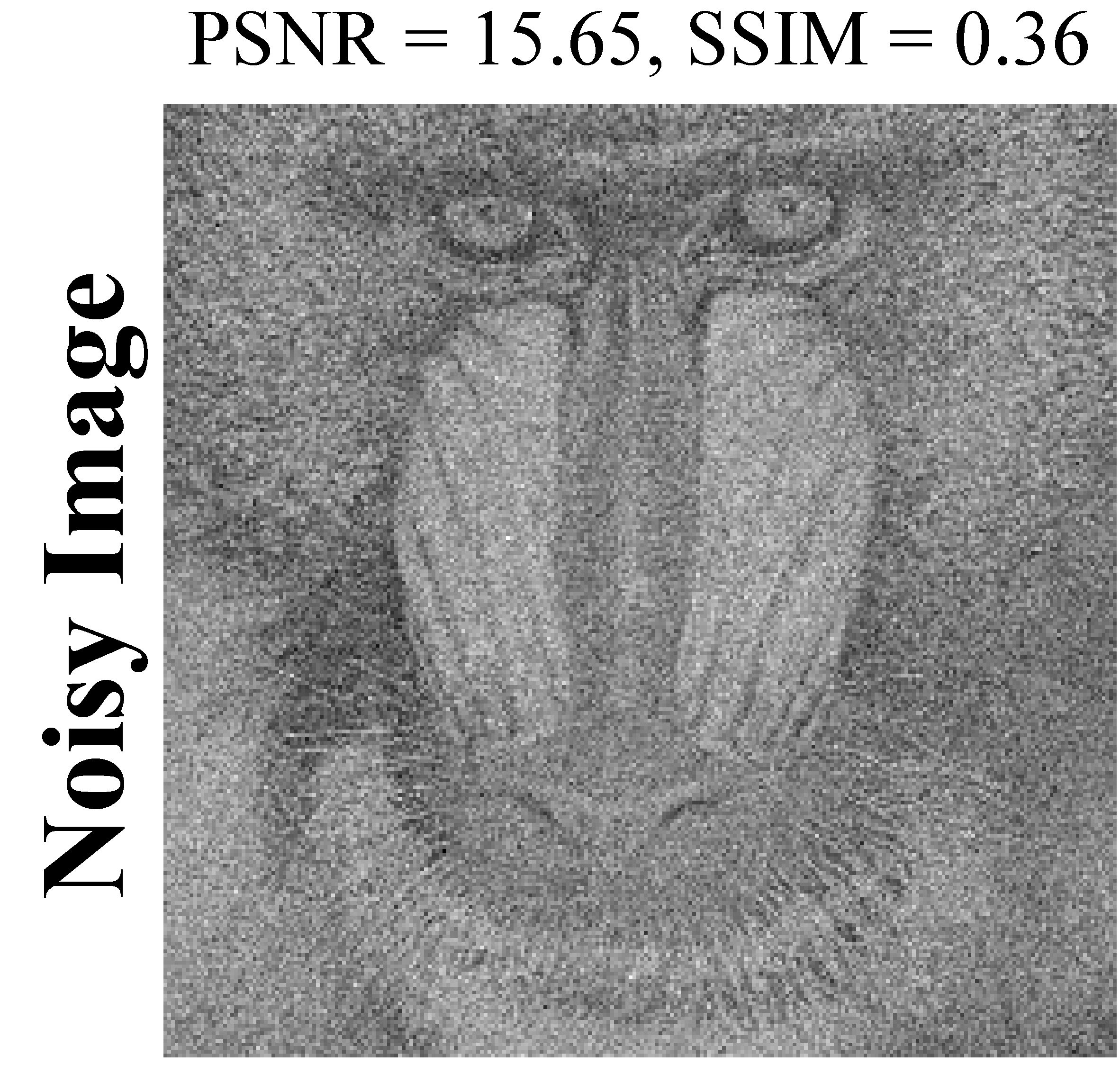}
		\end{subfigure}
		\begin{subfigure}{.16\textwidth}
			\centering
			\includegraphics[width=2.9cm, height=2.8cm]{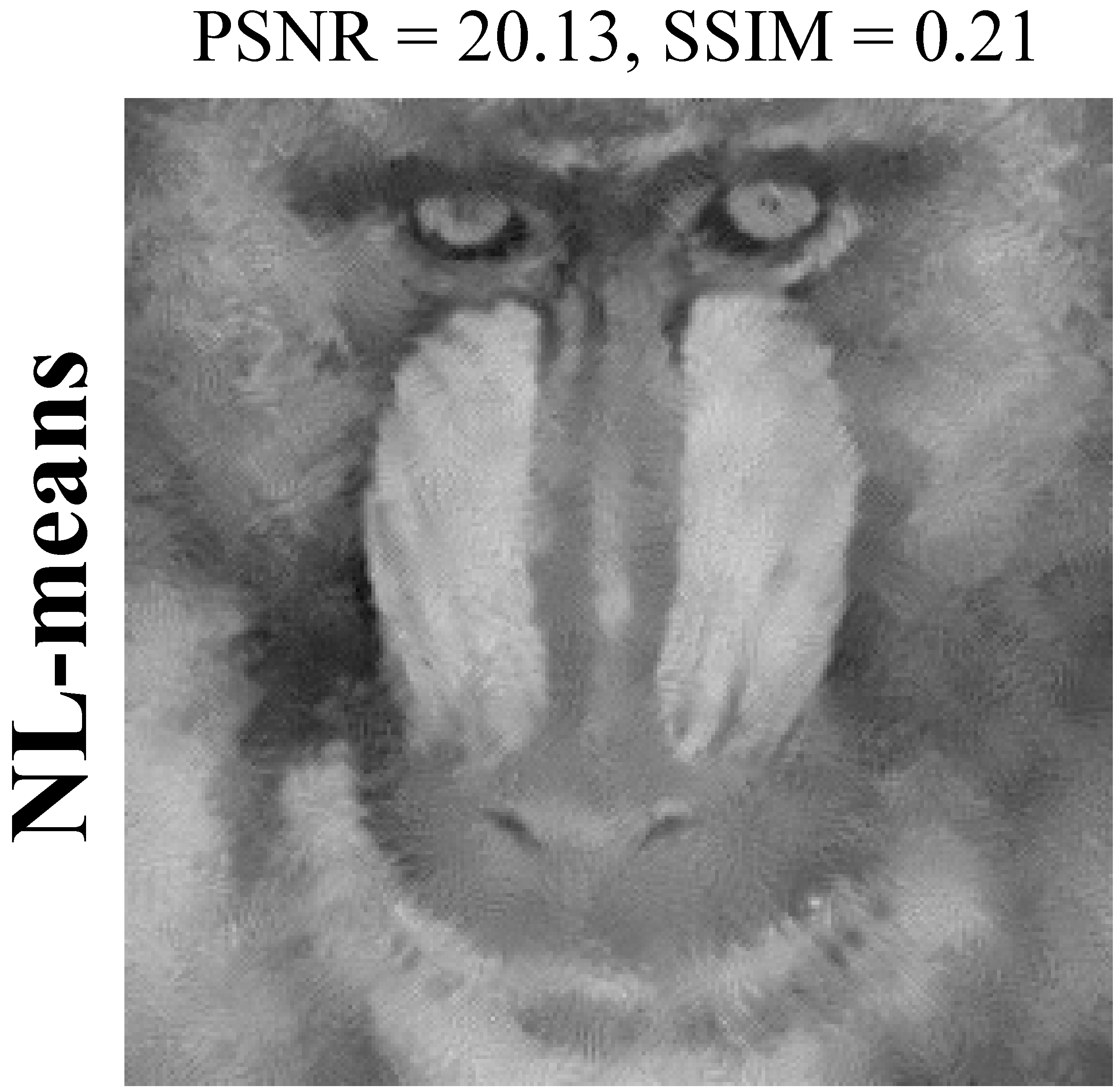}
		\end{subfigure}
		\begin{subfigure}{.16\textwidth}
			\centering
			\includegraphics[width=2.9cm, height=2.8cm]{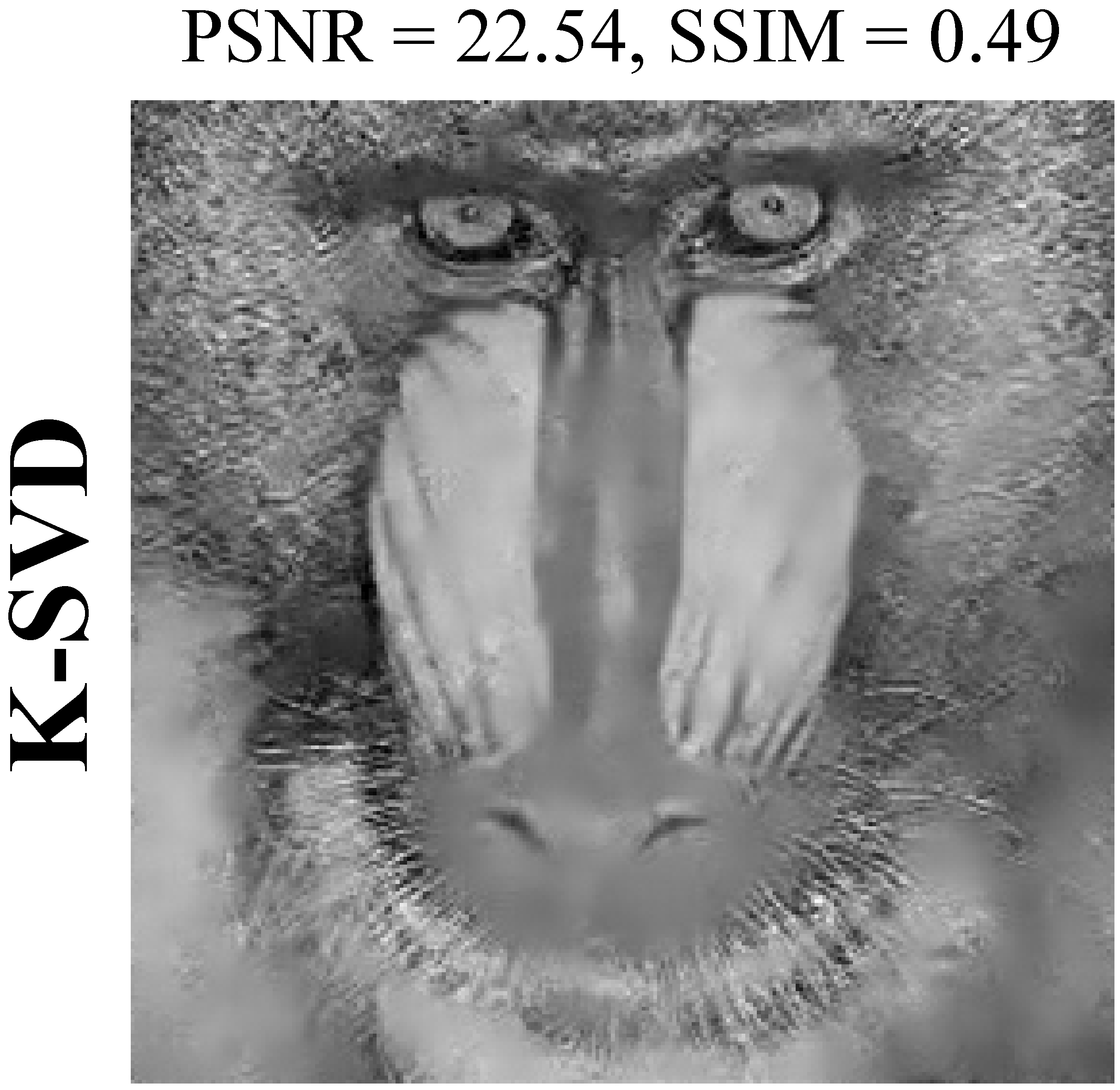}
		\end{subfigure}
		\begin{subfigure}{.16\textwidth}
			\centering
			\includegraphics[width=2.9cm, height=2.8cm]{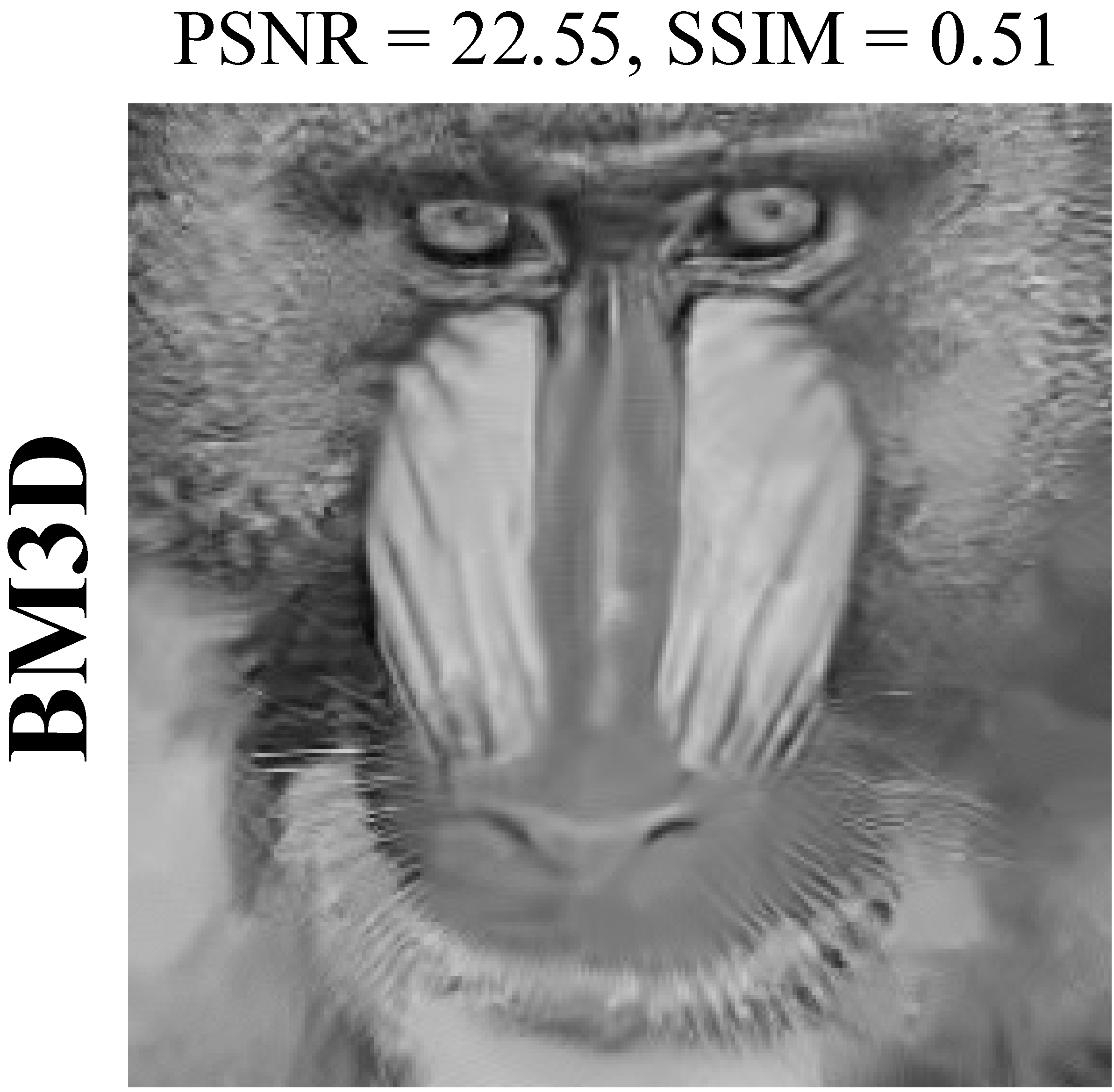}
		\end{subfigure}
		\begin{subfigure}{.16\textwidth}
			\centering
			\includegraphics[width=2.9cm, height=2.8cm]{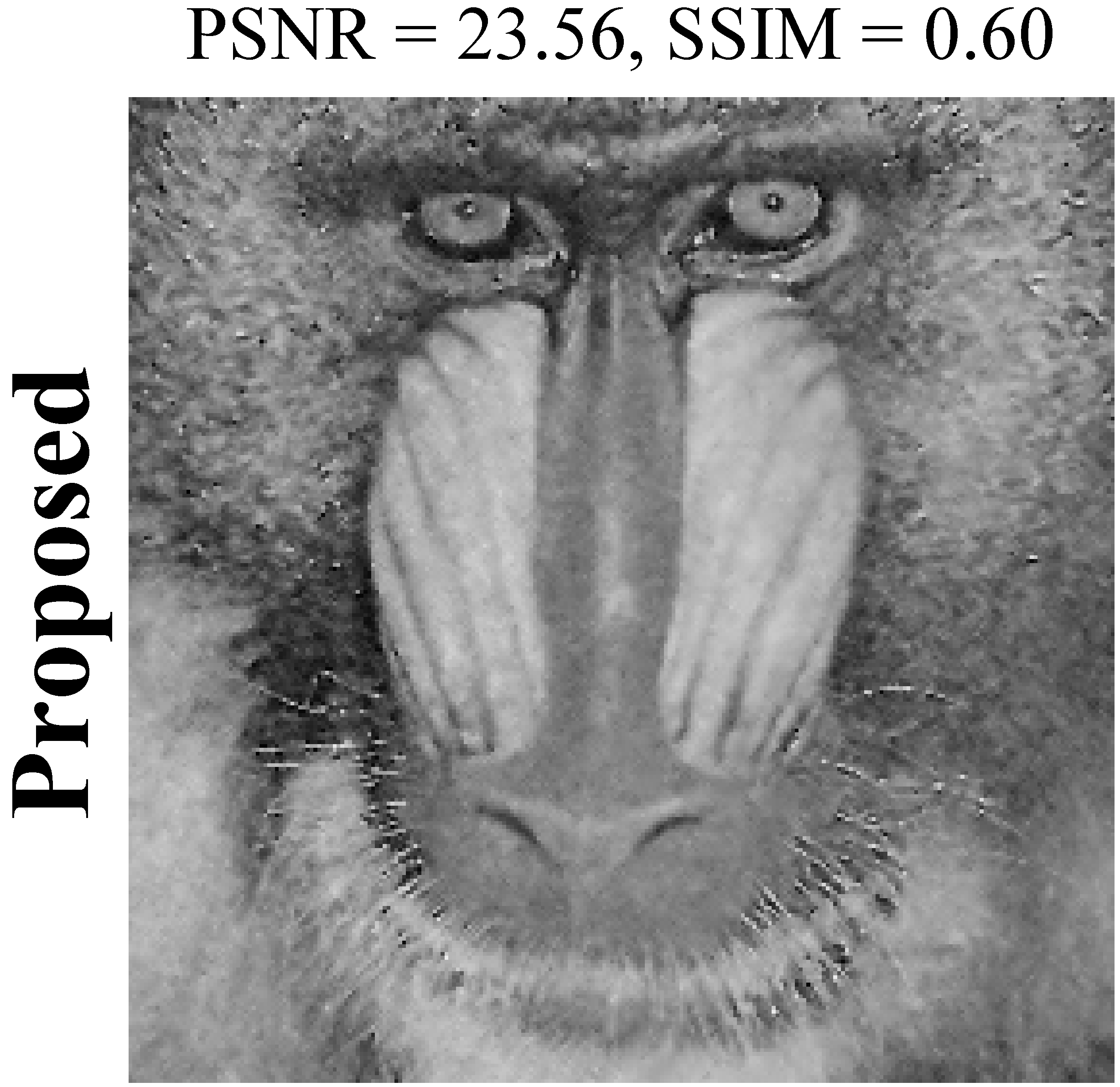}
		\end{subfigure}
	\end{subfigure}\\
	\begin{subfigure}{\textwidth}
		\centering
		\begin{subfigure}{.16\textwidth}
			\centering
			\includegraphics[width=2.9cm, height=2.8cm]{./Figures/Simu_Res_1/mandril_original.png}
		\end{subfigure}%
		\begin{subfigure}{.16\textwidth}
			\centering
			\includegraphics[width=2.9cm, height=2.8cm]{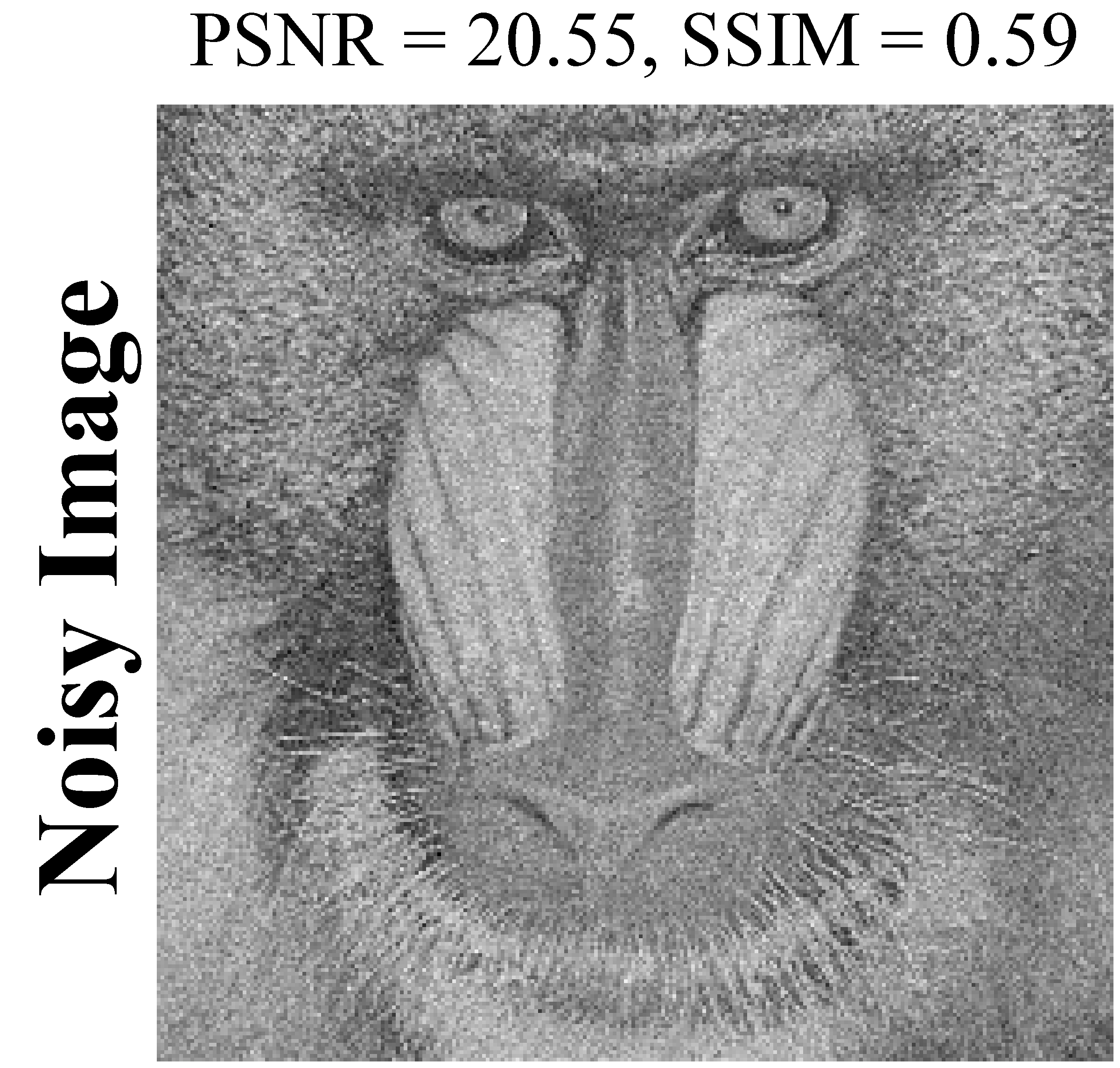}
		\end{subfigure}
		\begin{subfigure}{.16\textwidth}
			\centering
			\includegraphics[width=2.9cm, height=2.8cm]{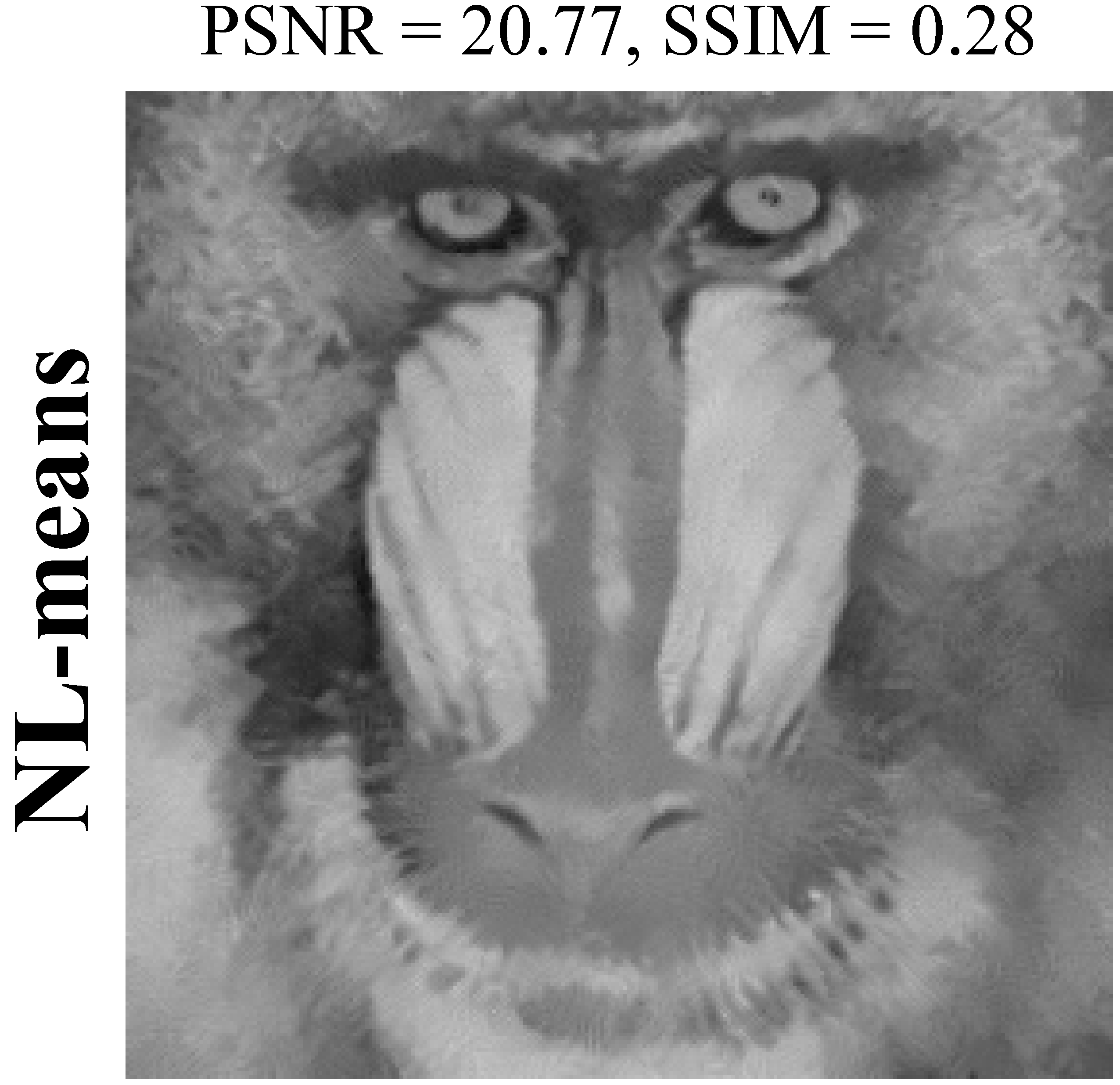}
		\end{subfigure}
		\begin{subfigure}{.16\textwidth}
			\centering
			\includegraphics[width=2.9cm, height=2.8cm]{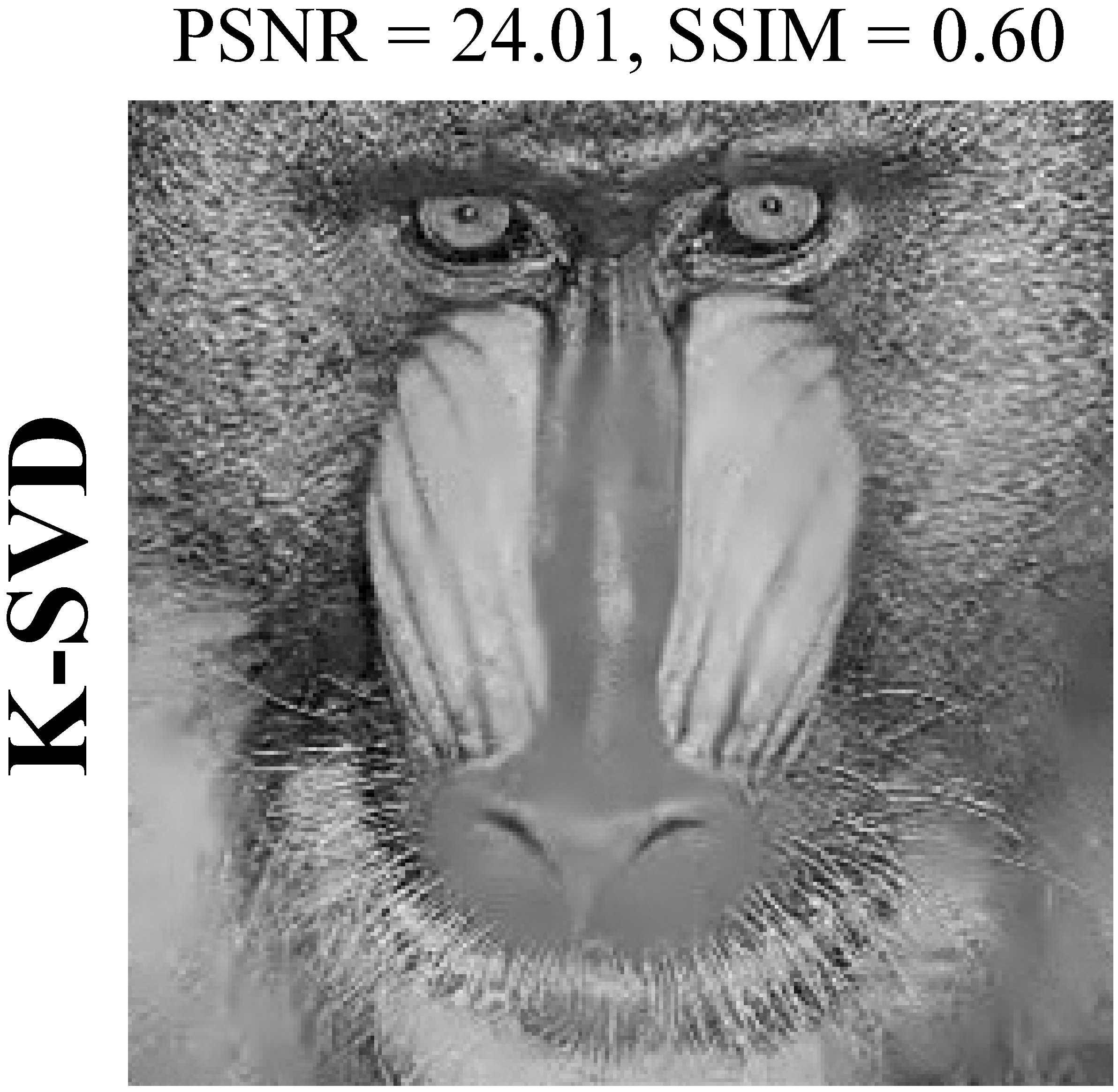}
		\end{subfigure}
		\begin{subfigure}{.16\textwidth}
			\centering
			\includegraphics[width=2.9cm, height=2.8cm]{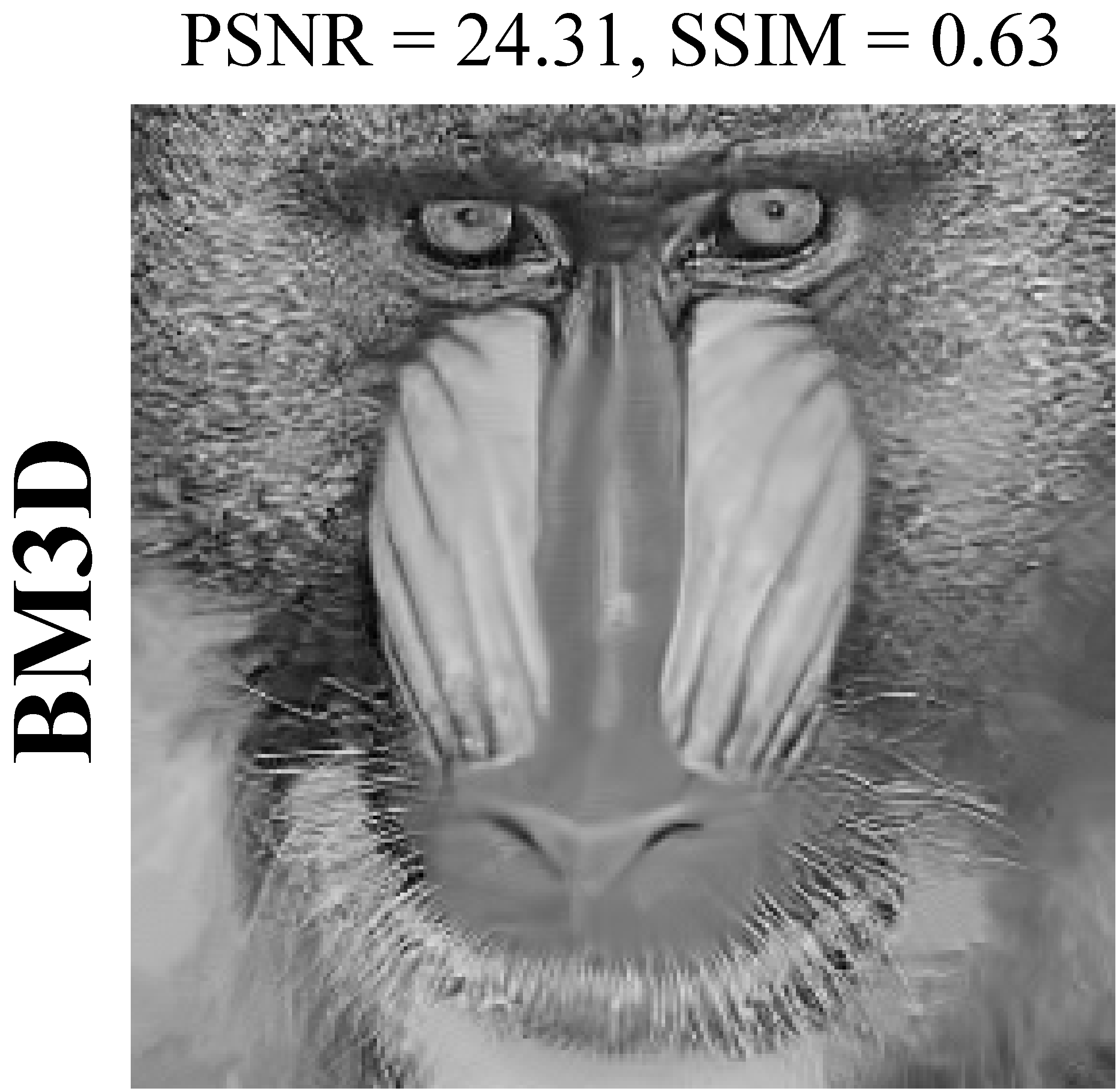}
		\end{subfigure}
		\begin{subfigure}{.16\textwidth}
			\centering
			\includegraphics[width=2.9cm, height=2.8cm]{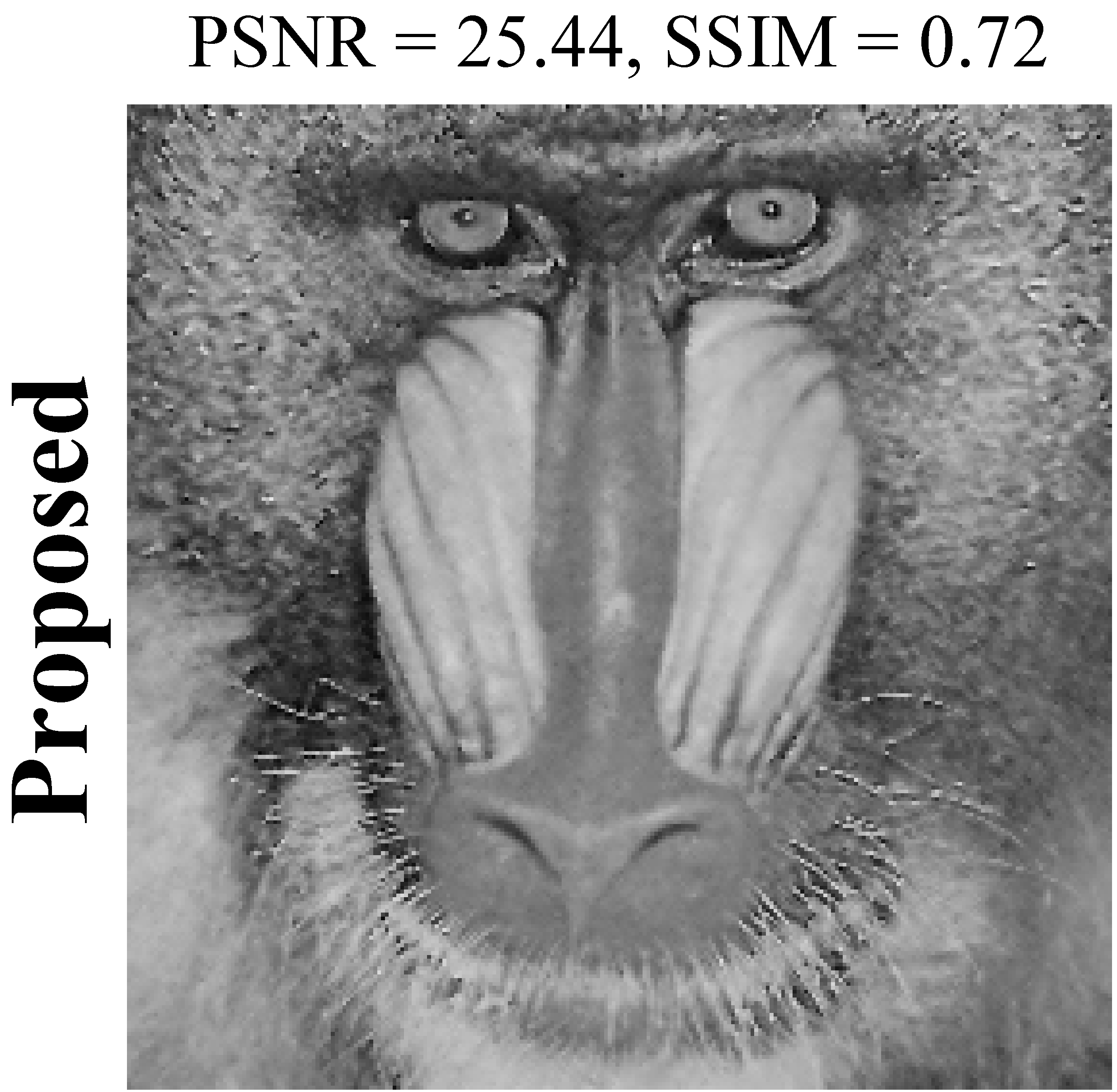}
		\end{subfigure}
	\end{subfigure}
	\caption{Denoising \textit{Mandrill}: 1st row at SNR$_{dB}$/$\sigma$ = -5/103, 2nd row at SNR$_{dB}$/$\sigma$ = 0/58, 3rd row at SNR$_{dB}$/$\sigma$ = 5/58}
	\label{fig:Sim_Res_1}
\end{figure*}

\section{Computational Complexity}
\label{Computational_Complexity}
The complexity of our proposed image denoising algorithm is dominated by that of the sparse recovery algorithm that we use, which fortunately has a low computational complexity when compared to other similar existing algorithms. With the dimensions of our problem at hand, the complexity for estimating one $\hv_k$ via SABMP is of the order $\Oc(MN^2P)$ where $P$ is the expected number of non-zeros that is generally a very small number. Lastly, to estimate all of the $K$ patches and for $L$ various iterations for different patch sizes, the computational complexity will gauge to an order of $\Oc(KLMN^2P)$.

\section{Simulation Results and Discussions}
\label{SIMULATION_RESULTS_AND_DISCUSSIONS}
In this section, we present the experimental results of our proposed C2DF image denoising algorithm, and compare the results with three state-of-the-art denoising methods namely NL-means \cite{1467423}, K-SVD \cite{elad2006image} and BM3D \cite{4271520}. The comparison takes place over a number of different sceneries to show the performance gain of our algorithm using standard test images. We also present the results of other natural images from the database available online at SIPI\footnote{\href{http://sipi.usc.edu/database/}{http://sipi.usc.edu/database/.}} to validate the applicability and efficiency of our method irrespective of any specific scenario.

We use a range of noise levels covering low noise to extremely high noise regimes. In particular, these high noise levels make the competition much more challenging by confusing signal components with those of noise components showcasing the limitations of existing methods where these fail to perform well specifically in preserving structures and details. The entries of dictionary in our case comprises of wavelet as well as DCT basis. We use SNR ranging from -5 dB to 25 dB. Further, we use patch size of 3, 5, 7 and 9, i.e, $L = 4$, and the denoising results based on these are averaged in the end for significant improvements.
\begin{figure*}[t]
	\centering
	\begin{subfigure}{.33\textwidth}
		\centering
		\includegraphics[width=5.75cm, height=5.75cm]{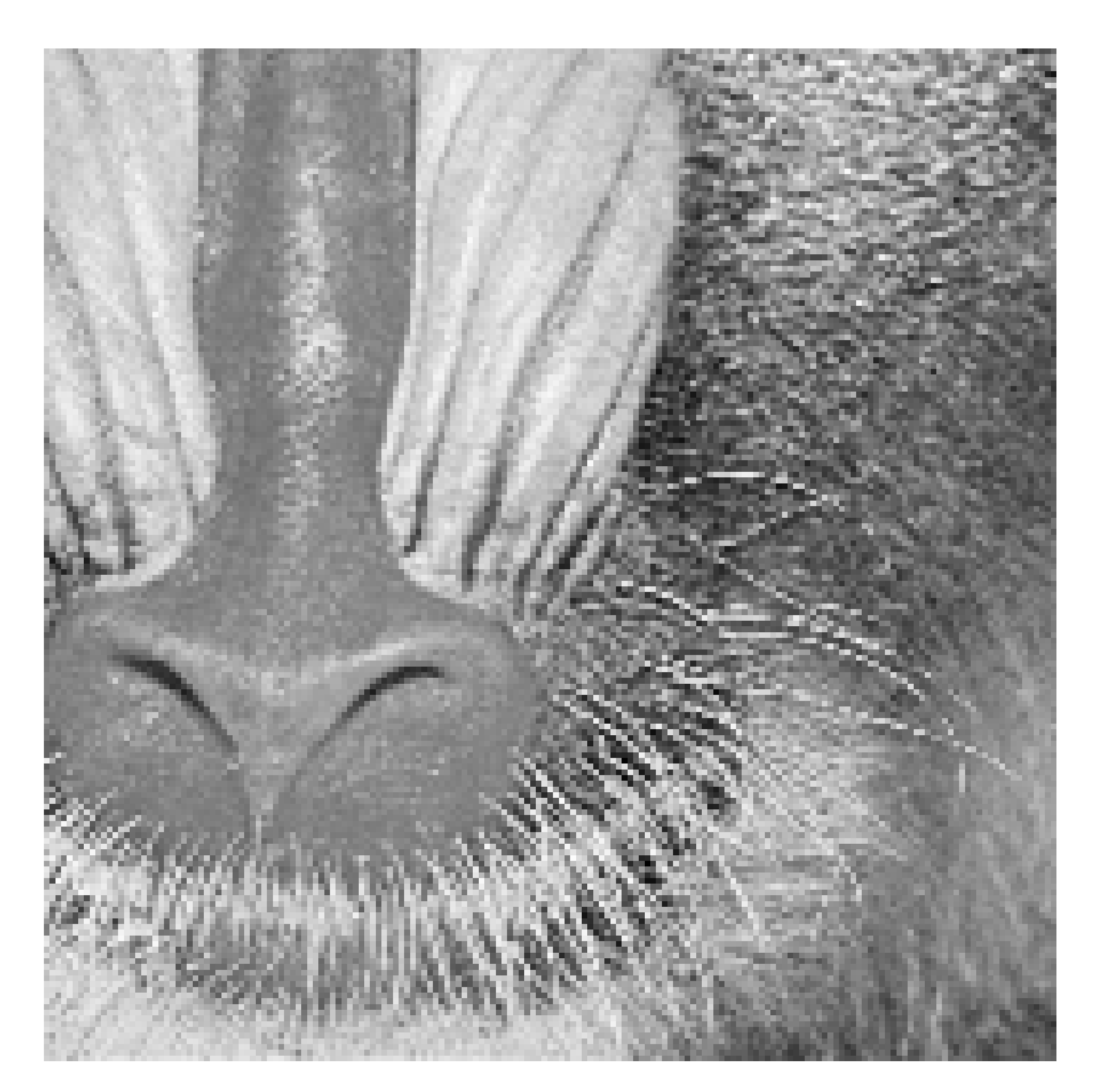}
		\caption{}
	\end{subfigure}%
	\begin{subfigure}{.33\textwidth}
		\centering
		\includegraphics[width=5.75cm, height=5.75cm]{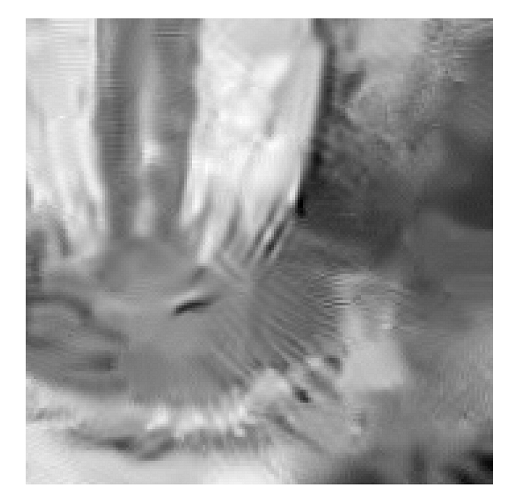}
		\caption{}
	\end{subfigure}
	\begin{subfigure}{.33\textwidth}
		\centering
		\includegraphics[width=5.75cm, height=5.75cm]{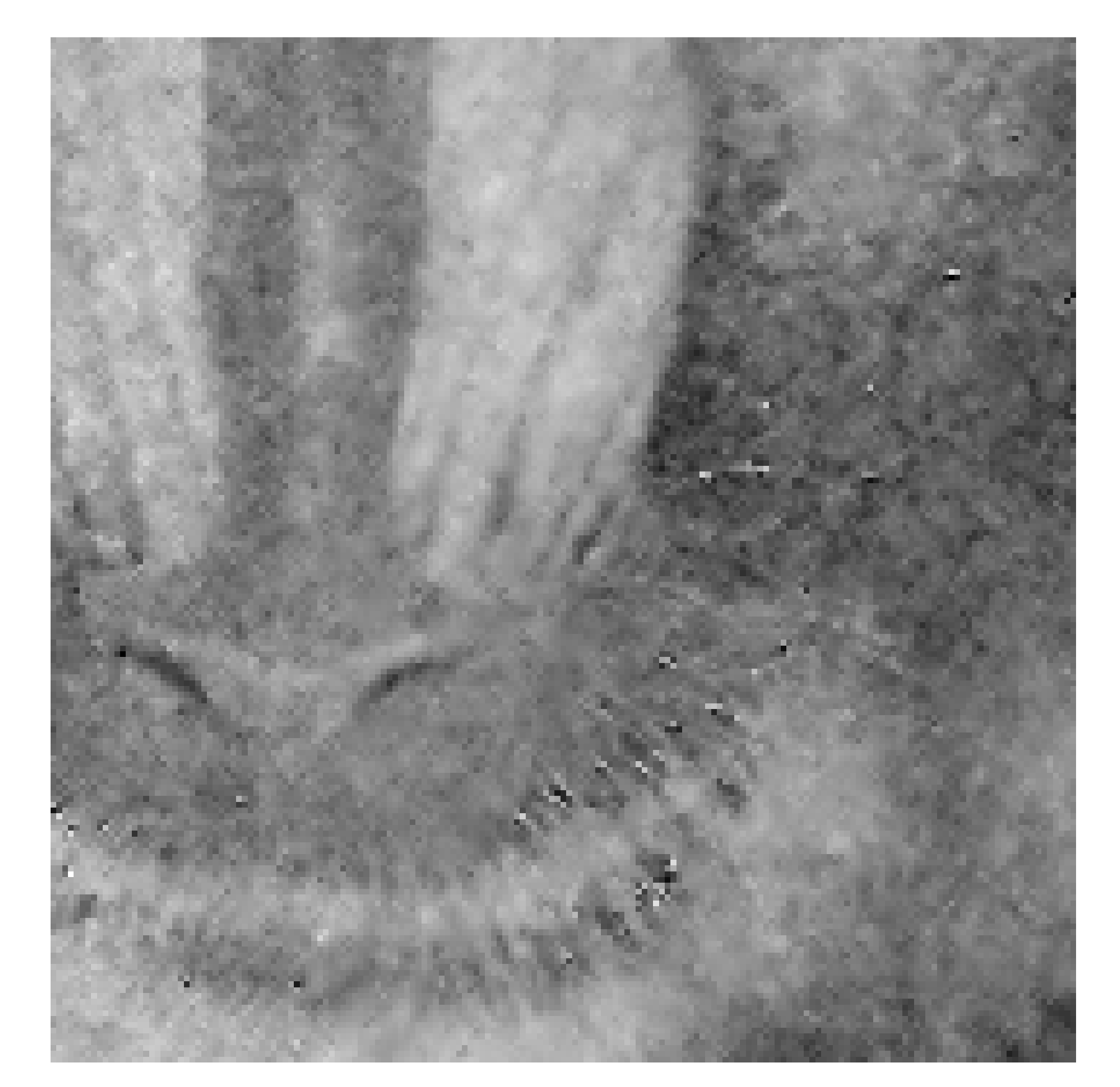}
		\caption{}
	\end{subfigure}
	\caption{Zoomed versions of the (a) \textit{Mandrill} image, denoised by (b) BM3D, and (c) C2DF at SNR$_{dB}$/$\sigma$ = -5/103}
	\label{fig:zoomed}
\end{figure*}
\begin{figure*}[h!]
	\begin{subfigure}{\textwidth}
		\centering
		\begin{subfigure}{.16\textwidth}
			\centering
			\includegraphics[width=2.9cm, height=2.8cm]{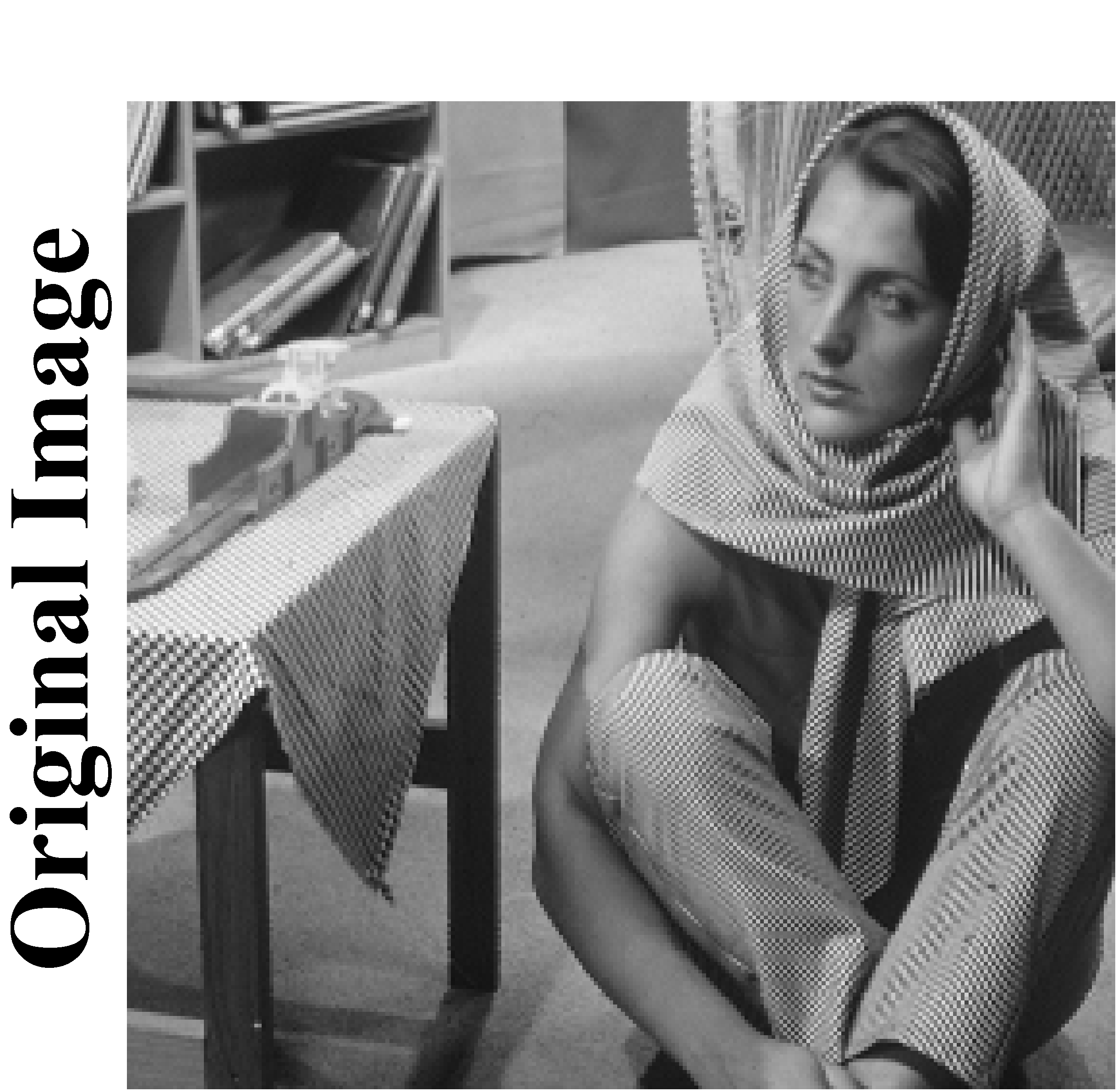}
		\end{subfigure}%
		\begin{subfigure}{.16\textwidth}
			\centering
			\includegraphics[width=2.9cm, height=2.8cm]{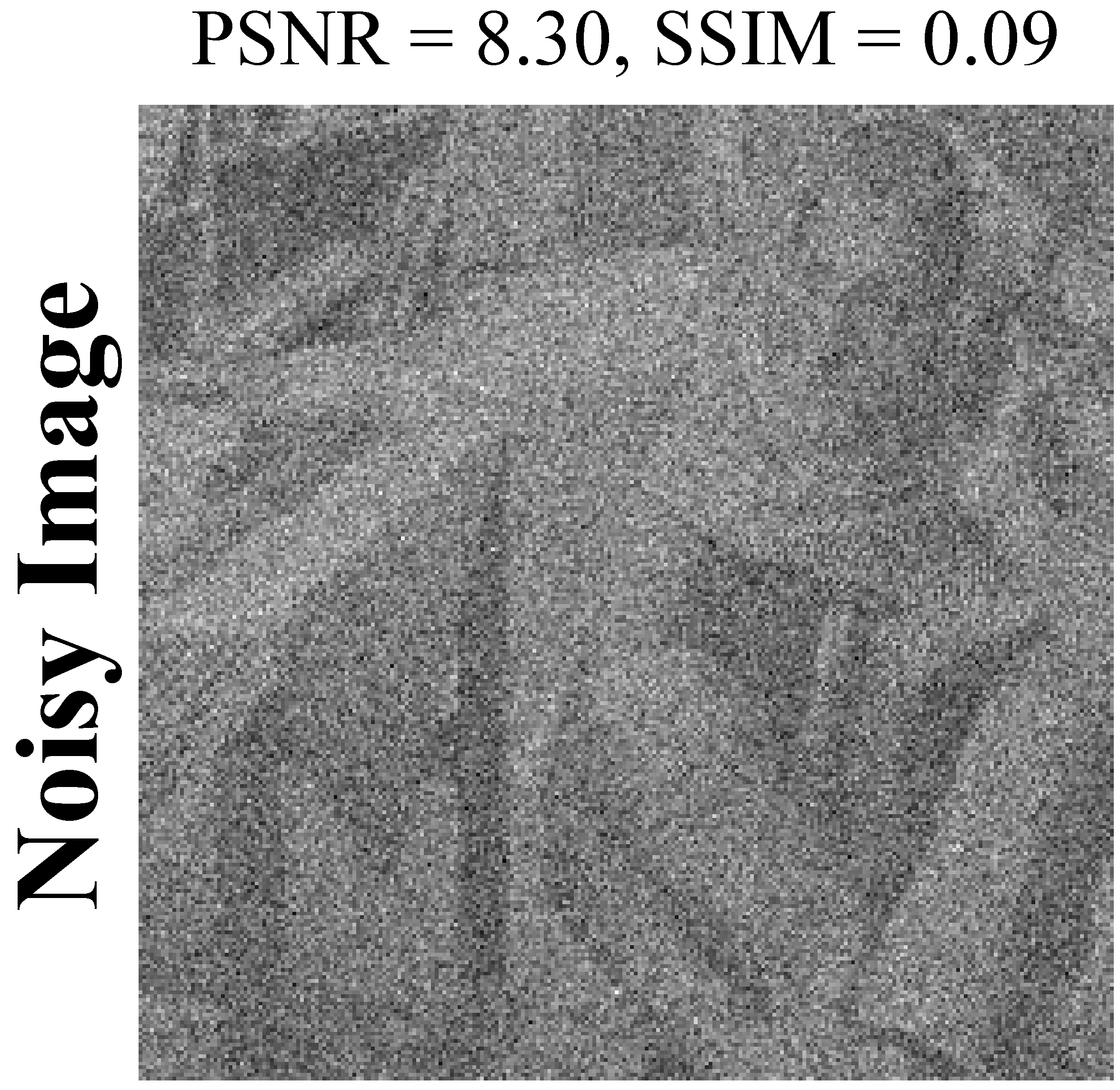}
		\end{subfigure}
		\begin{subfigure}{.16\textwidth}
			\centering
			\includegraphics[width=2.9cm, height=2.8cm]{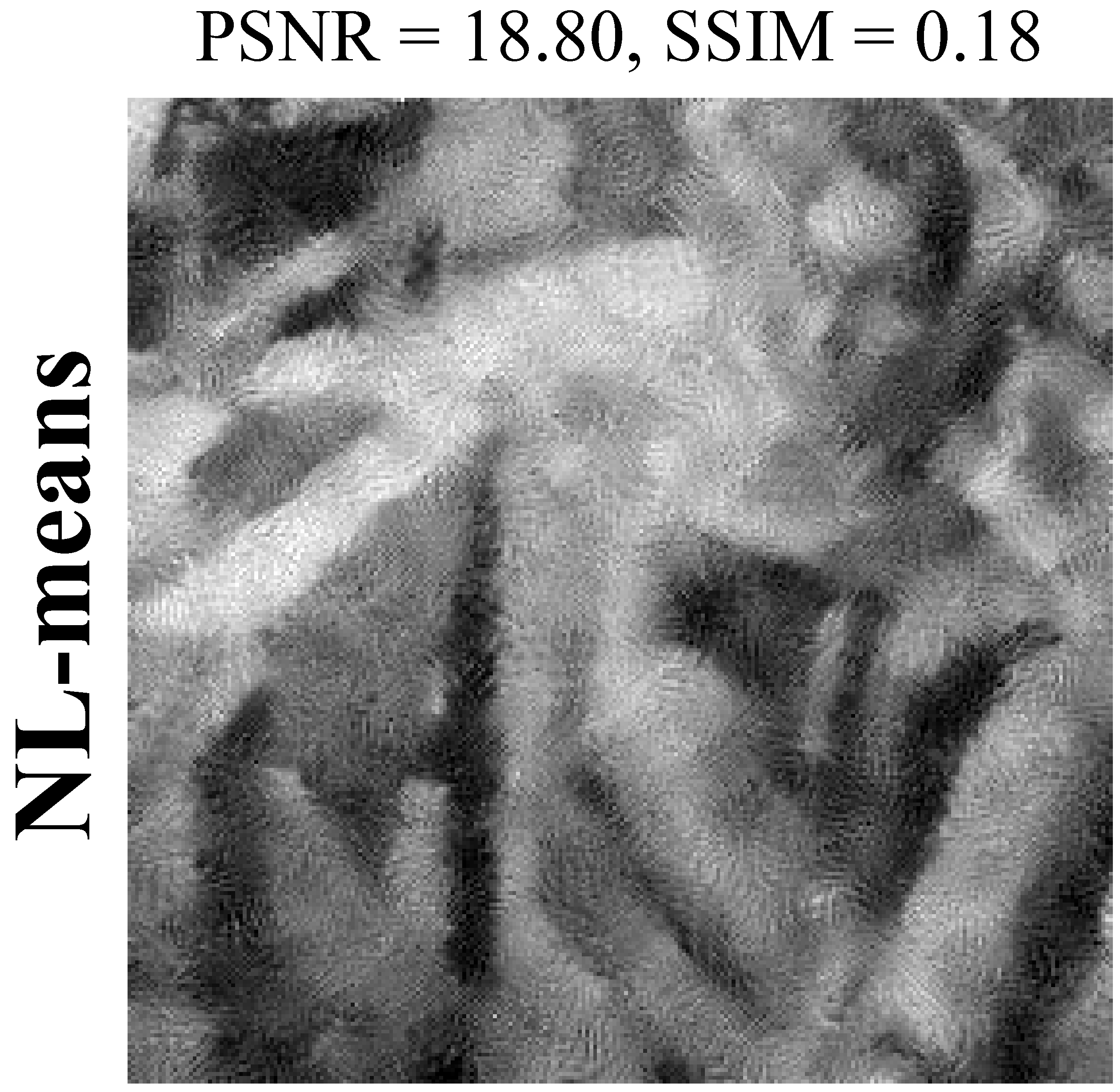}
		\end{subfigure}
		\begin{subfigure}{.16\textwidth}
			\centering
			\includegraphics[width=2.9cm, height=2.8cm]{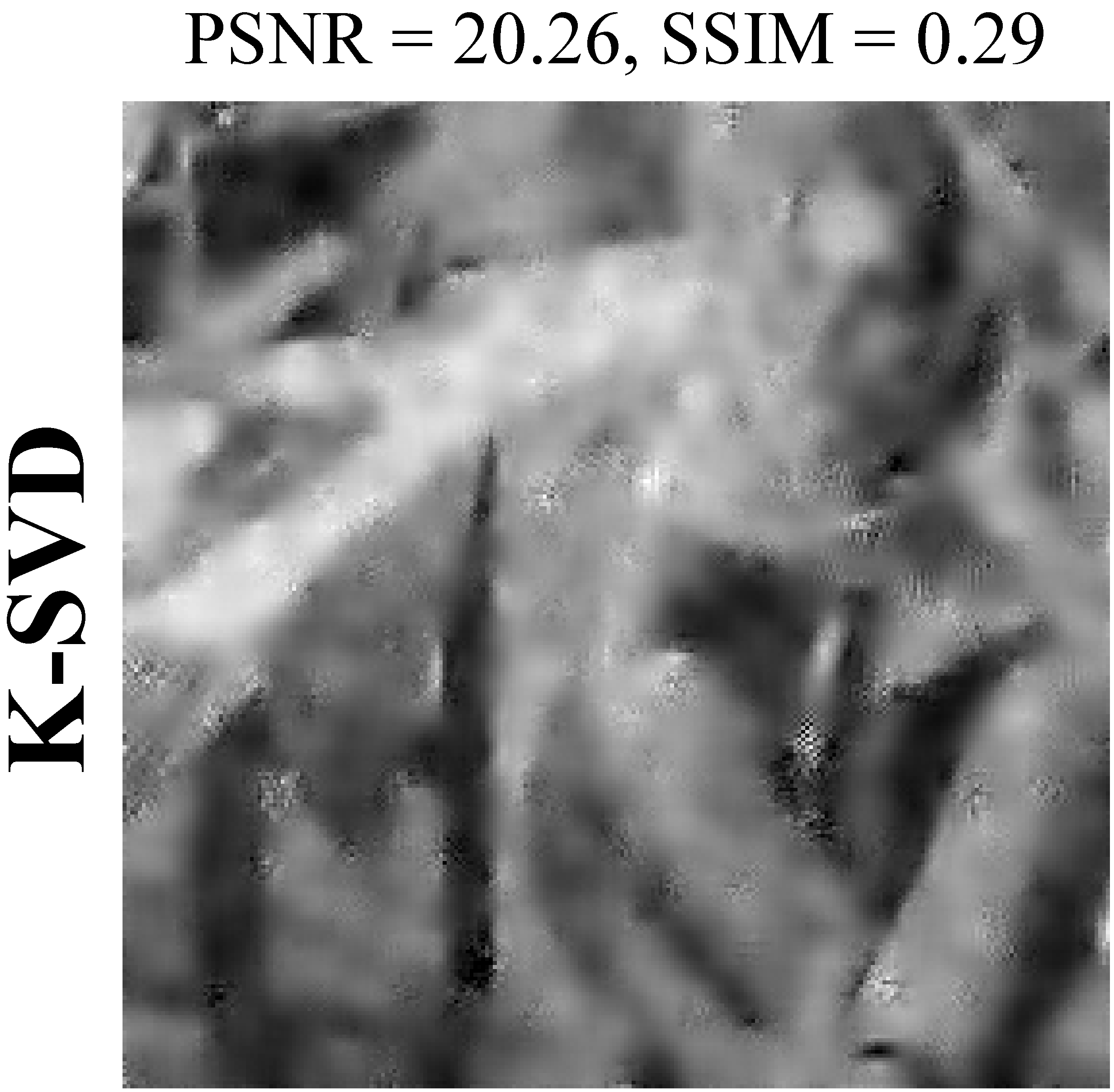}
		\end{subfigure}
		\begin{subfigure}{.16\textwidth}
			\centering
			\includegraphics[width=2.9cm, height=2.8cm]{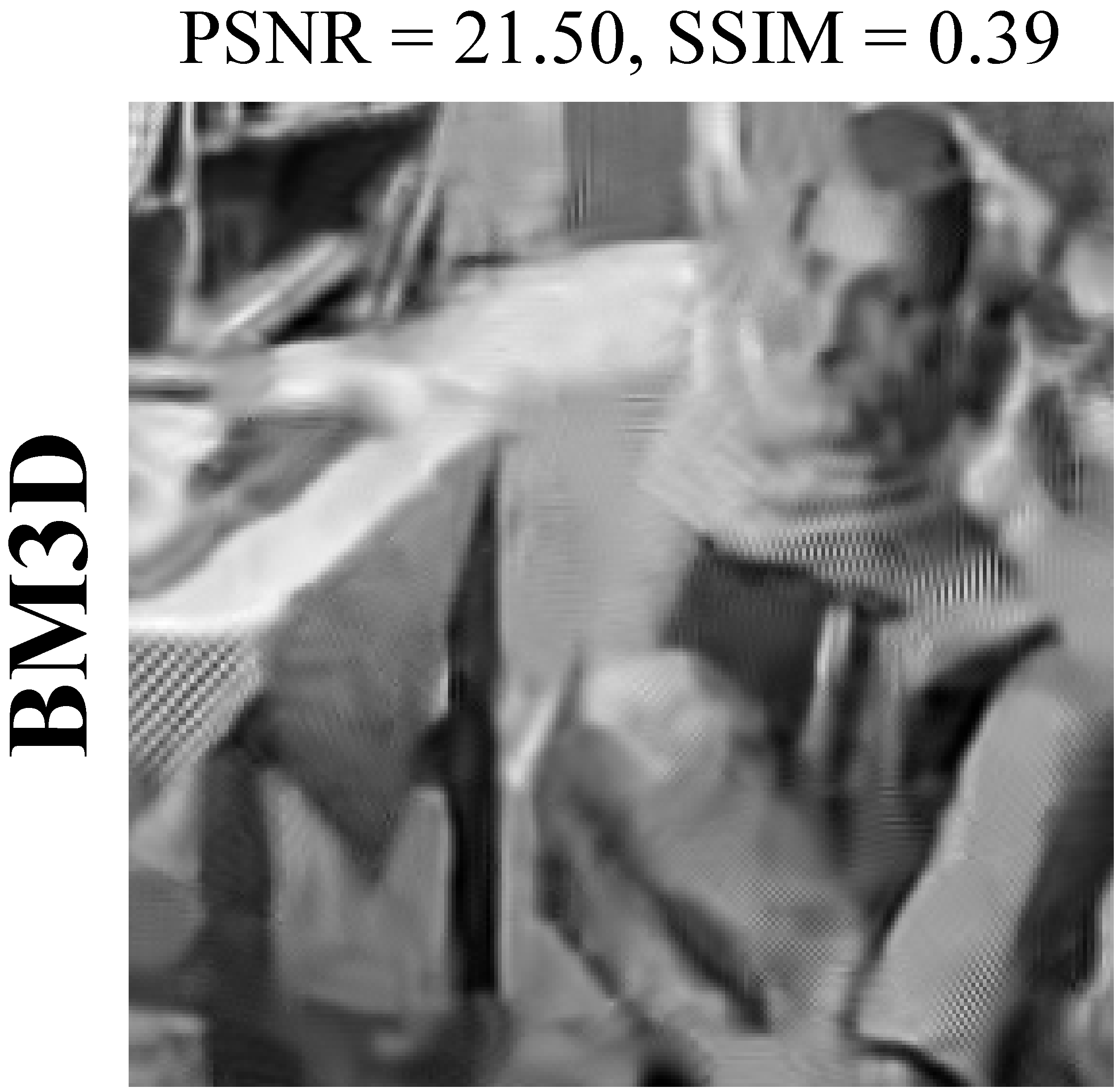}
		\end{subfigure}
		\begin{subfigure}{.16\textwidth}
			\centering
			\includegraphics[width=2.9cm, height=2.8cm]{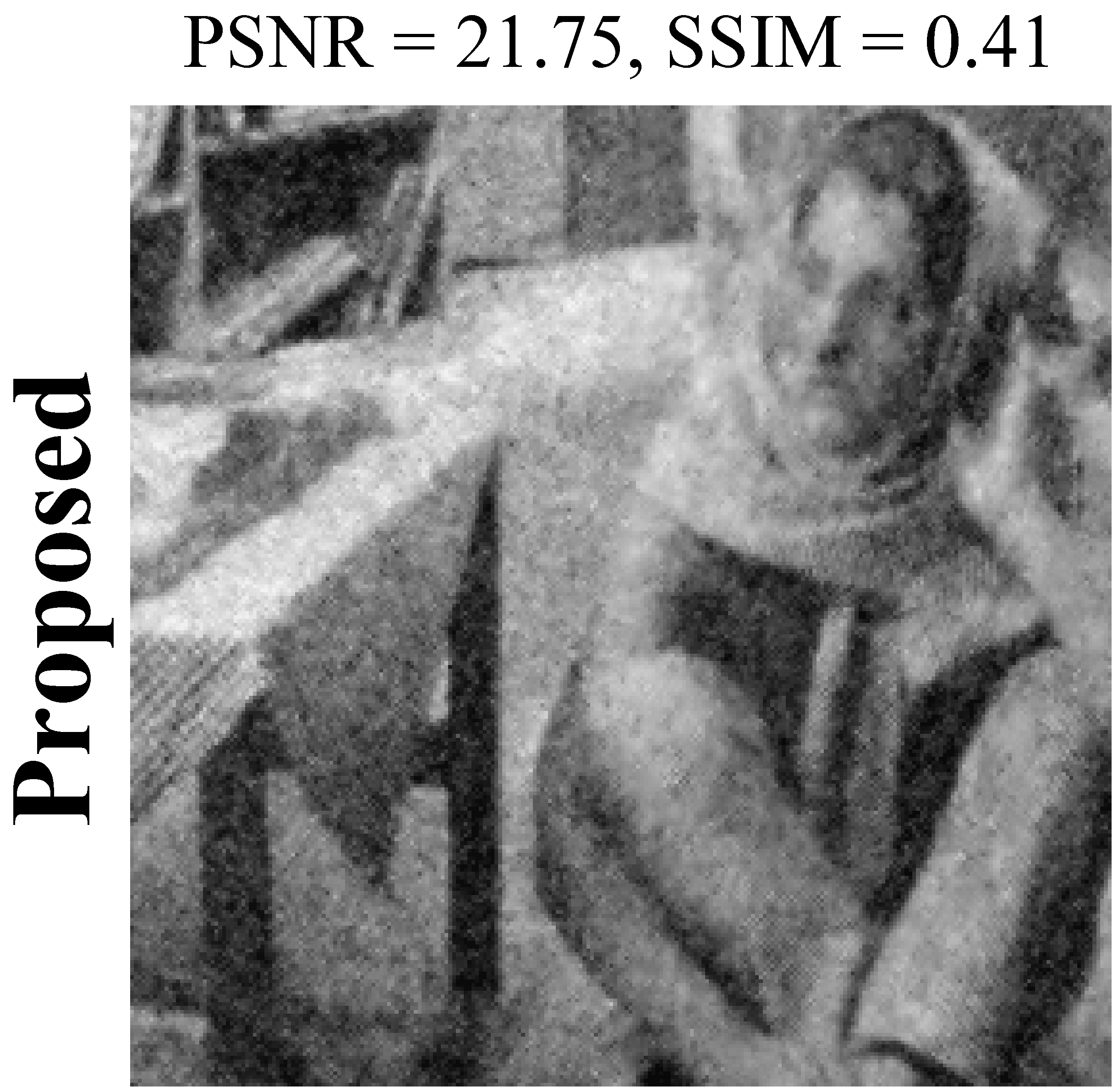}
		\end{subfigure}
	\end{subfigure} \\
	\begin{subfigure}{\textwidth}
		\centering
		\begin{subfigure}{.16\textwidth}
			\centering
			\includegraphics[width=2.9cm, height=2.8cm]{./Figures/Simu_Res_2/barbara_original.png}
		\end{subfigure}%
		\begin{subfigure}{.16\textwidth}
			\centering
			\includegraphics[width=2.9cm, height=2.8cm]{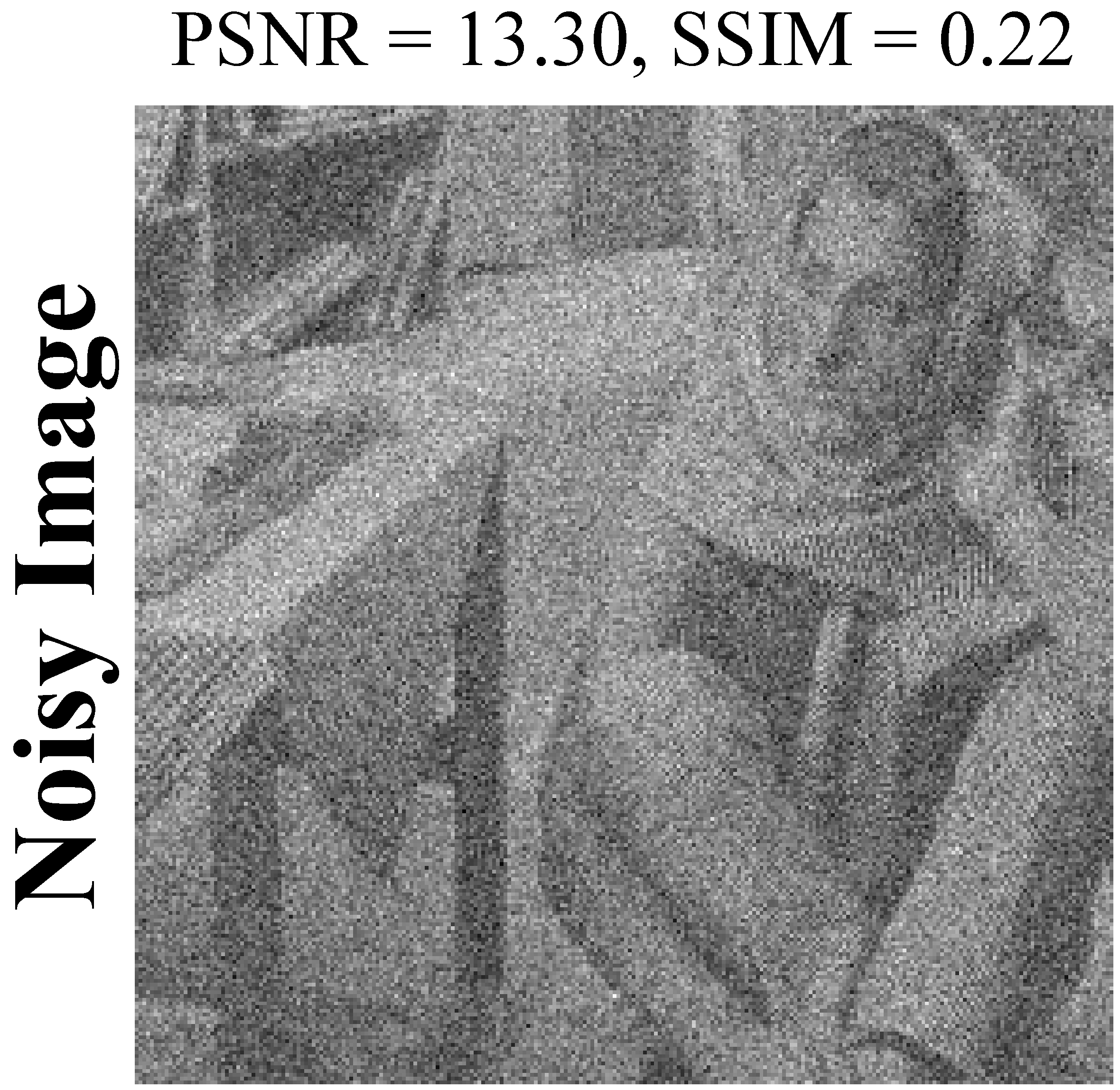}
		\end{subfigure}
		\begin{subfigure}{.16\textwidth}
			\centering
			\includegraphics[width=2.9cm, height=2.8cm]{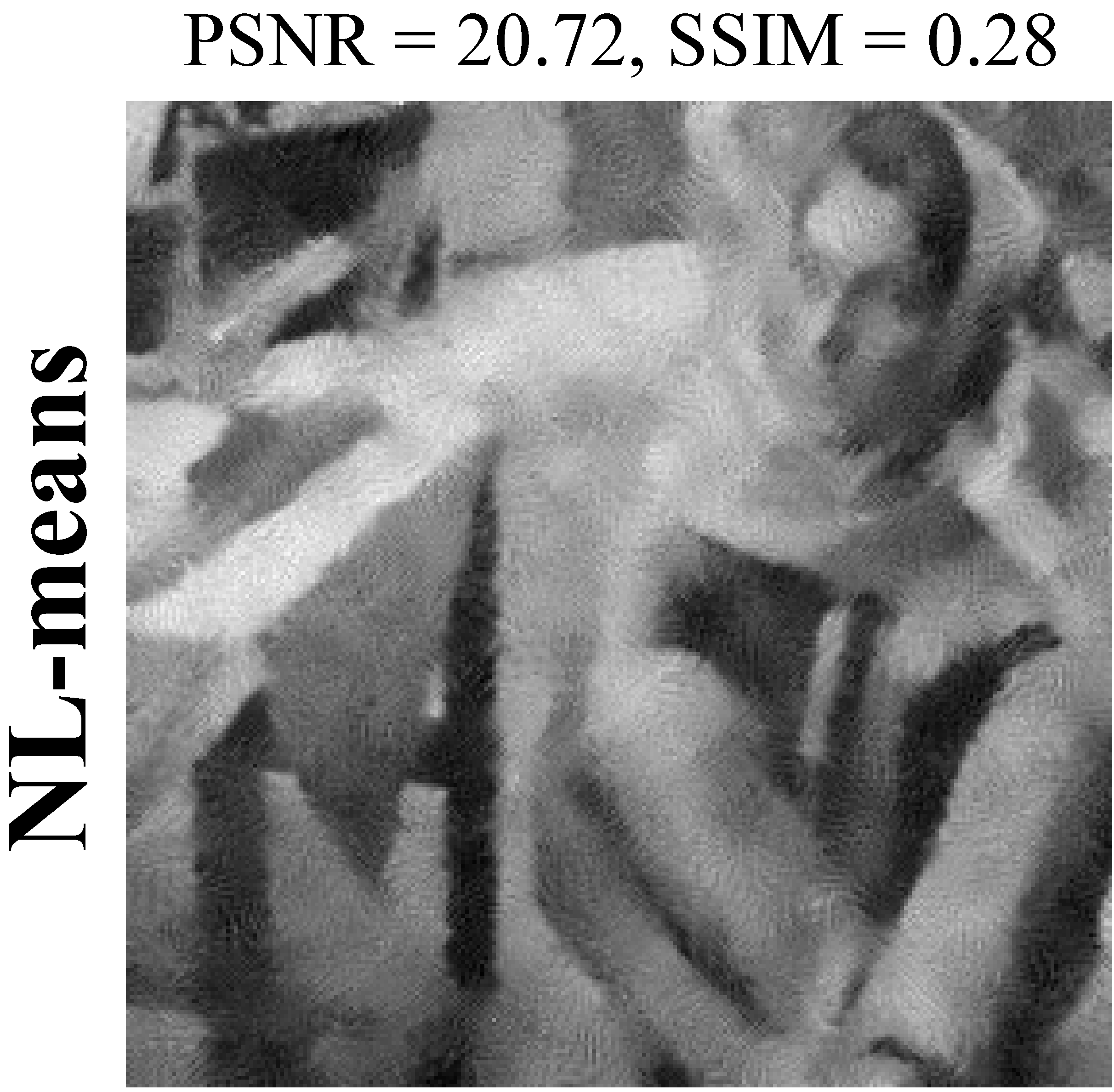}
		\end{subfigure}
		\begin{subfigure}{.16\textwidth}
			\centering
			\includegraphics[width=2.9cm, height=2.8cm]{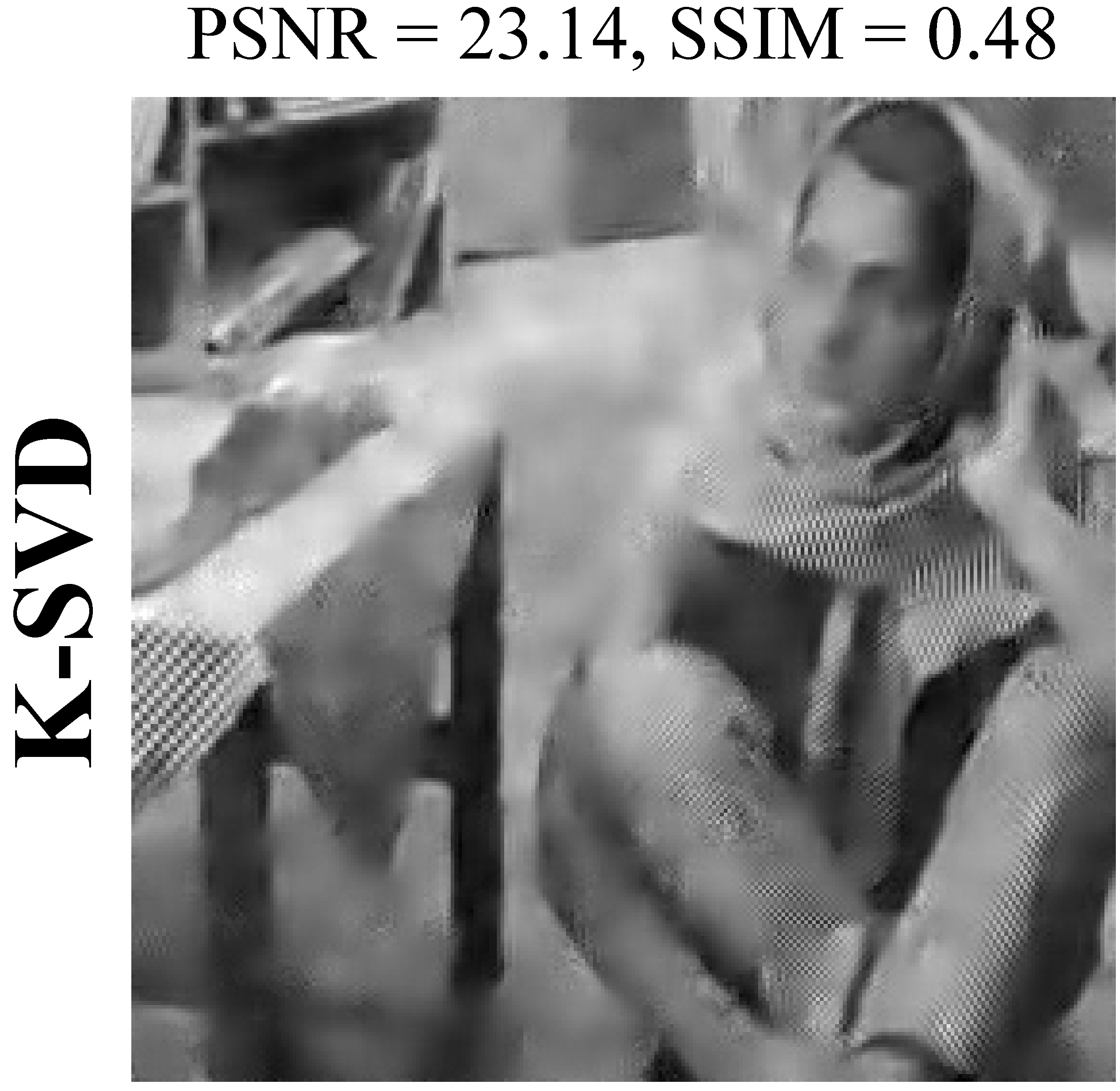}
		\end{subfigure}
		\begin{subfigure}{.16\textwidth}
			\centering
			\includegraphics[width=2.9cm, height=2.8cm]{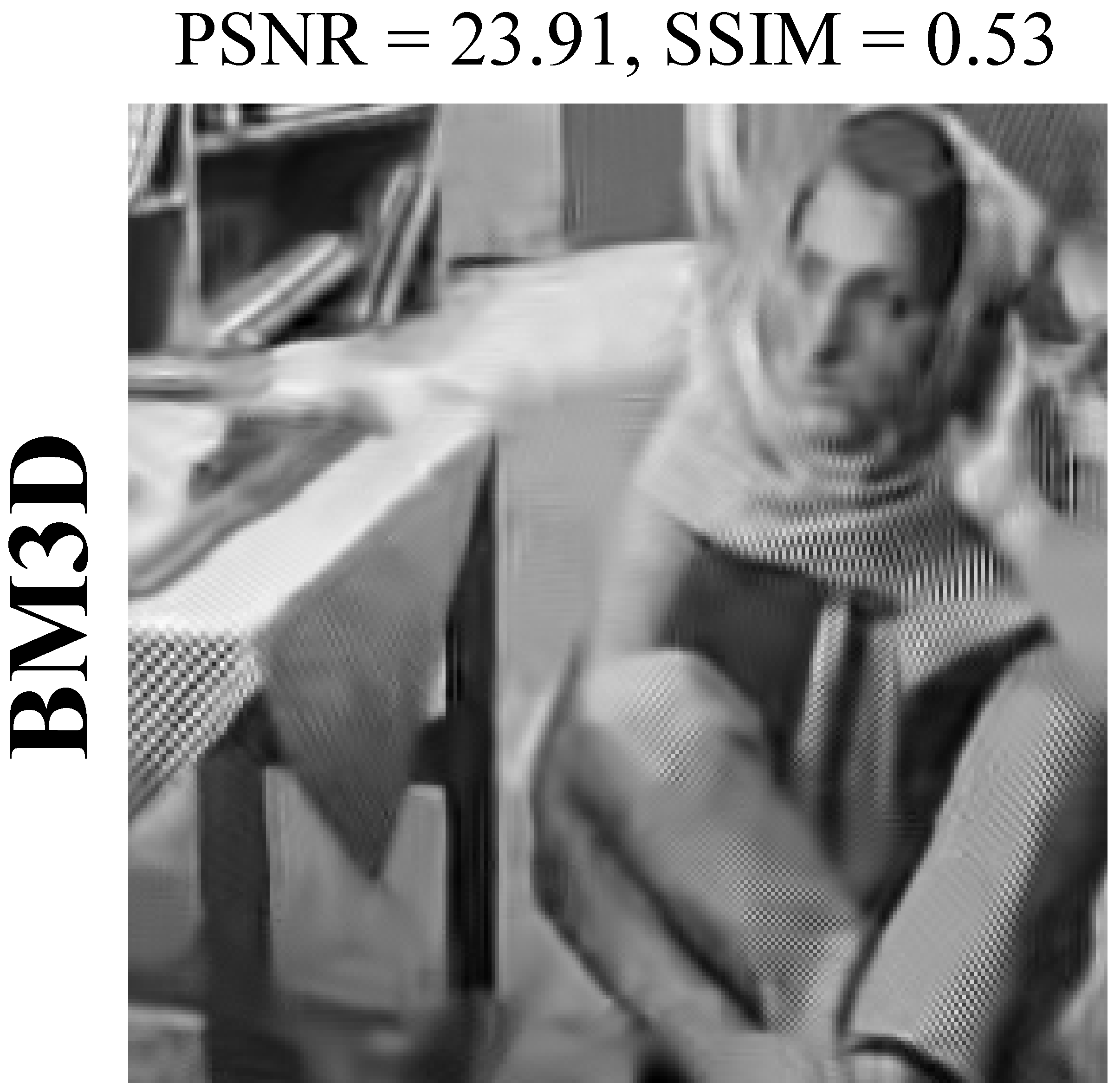}
		\end{subfigure}
		\begin{subfigure}{.16\textwidth}
			\centering
			\includegraphics[width=2.9cm, height=2.8cm]{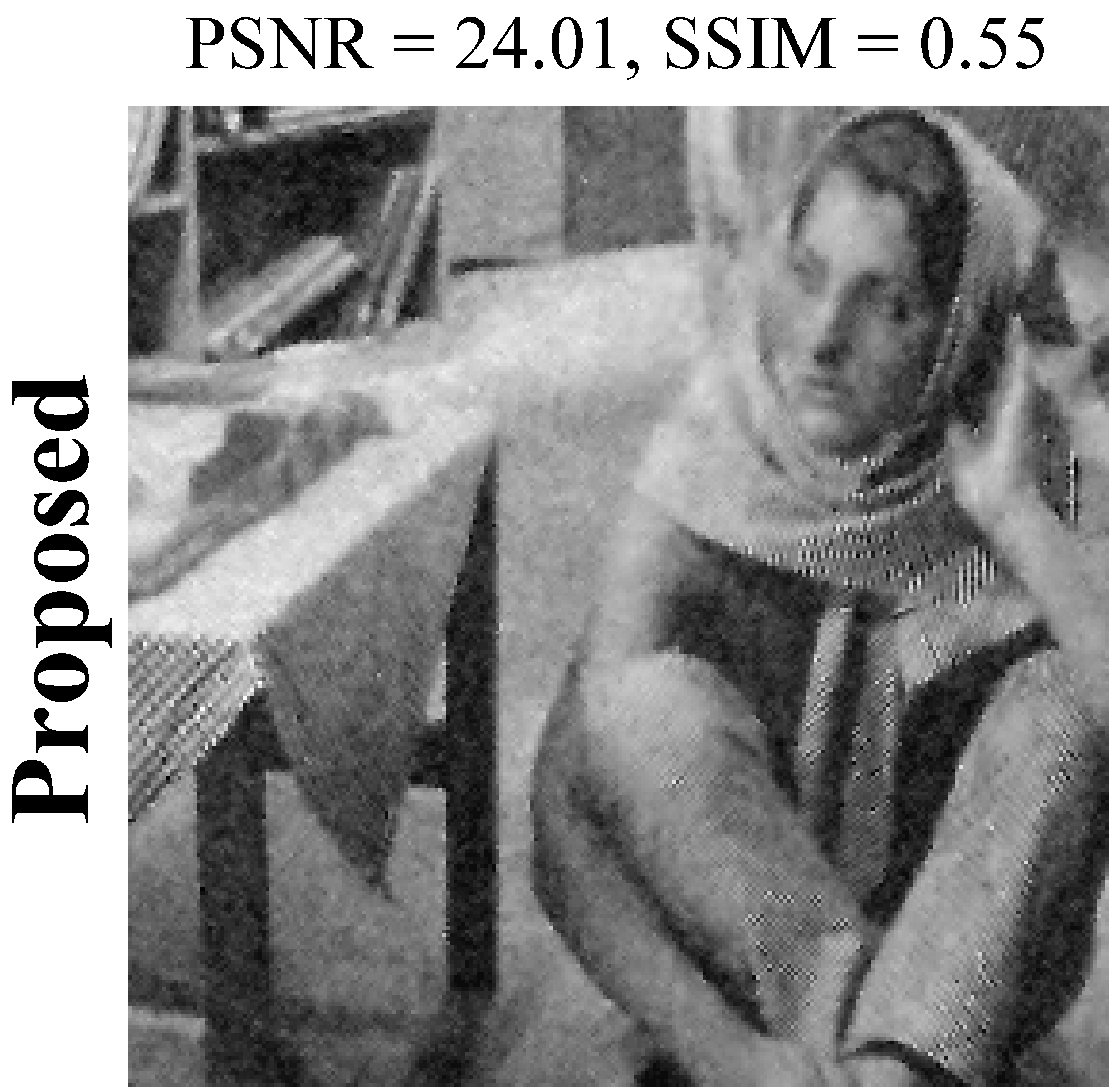}
		\end{subfigure}
	\end{subfigure}\\
	\begin{subfigure}{\textwidth}
		\centering
		\begin{subfigure}{.16\textwidth}
			\centering
			\includegraphics[width=2.9cm, height=2.8cm]{./Figures/Simu_Res_2/barbara_original.png}
		\end{subfigure}%
		\begin{subfigure}{.16\textwidth}
			\centering
			\includegraphics[width=2.9cm, height=2.8cm]{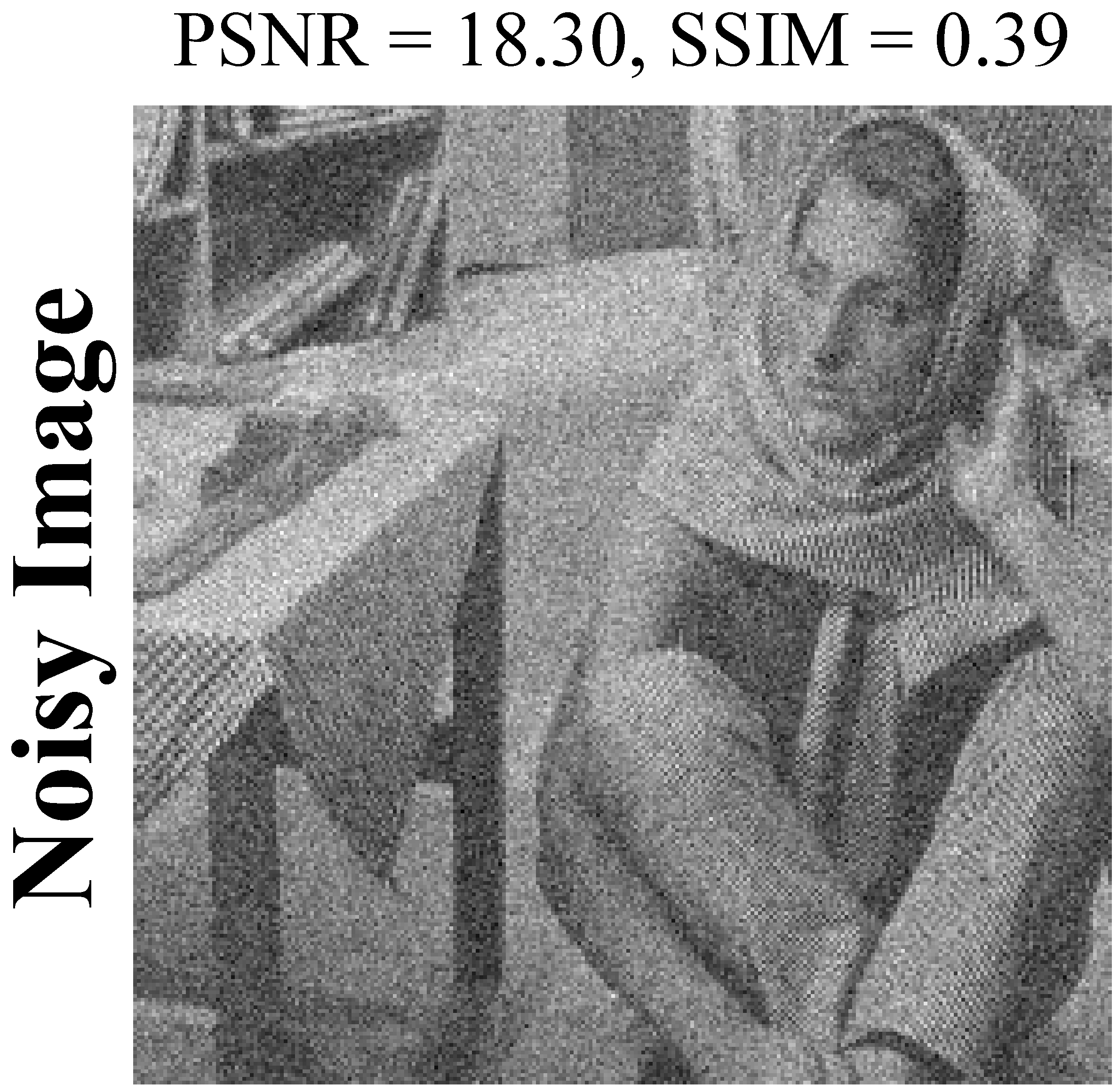}
		\end{subfigure}
		\begin{subfigure}{.16\textwidth}
			\centering
			\includegraphics[width=2.9cm, height=2.8cm]{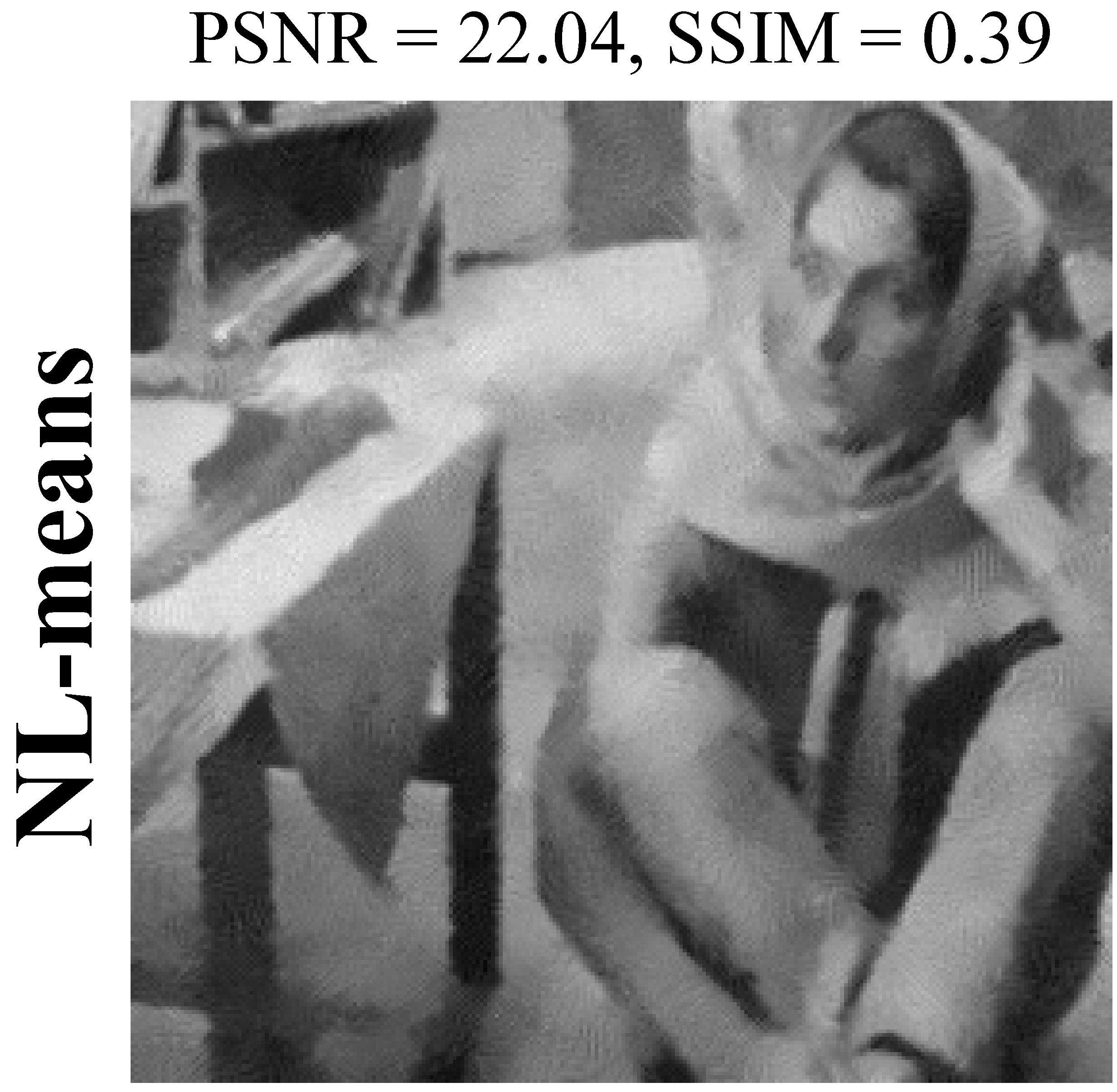}
		\end{subfigure}
		\begin{subfigure}{.16\textwidth}
			\centering
			\includegraphics[width=2.9cm, height=2.8cm]{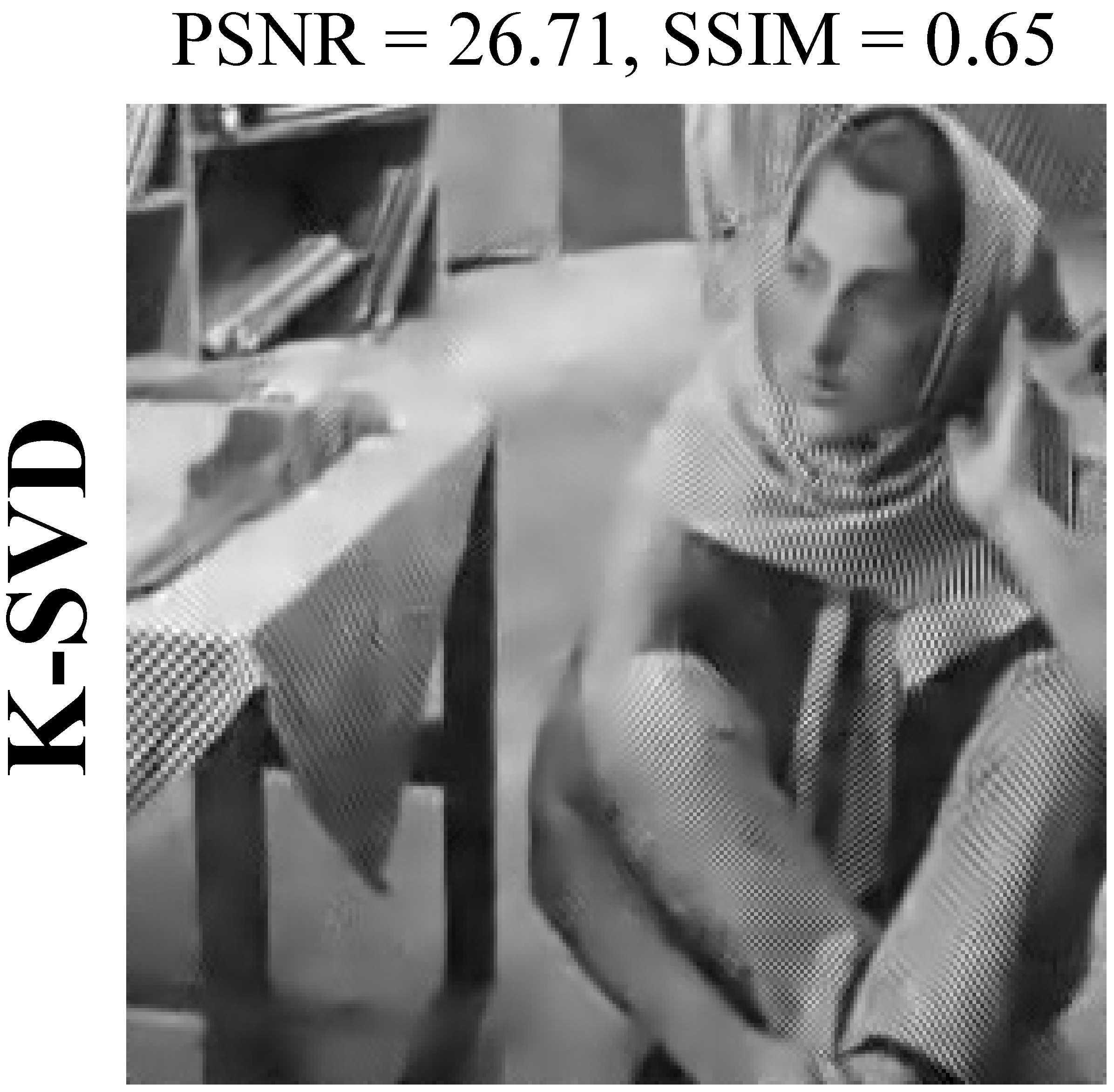}
		\end{subfigure}
		\begin{subfigure}{.16\textwidth}
			\centering
			\includegraphics[width=2.9cm, height=2.8cm]{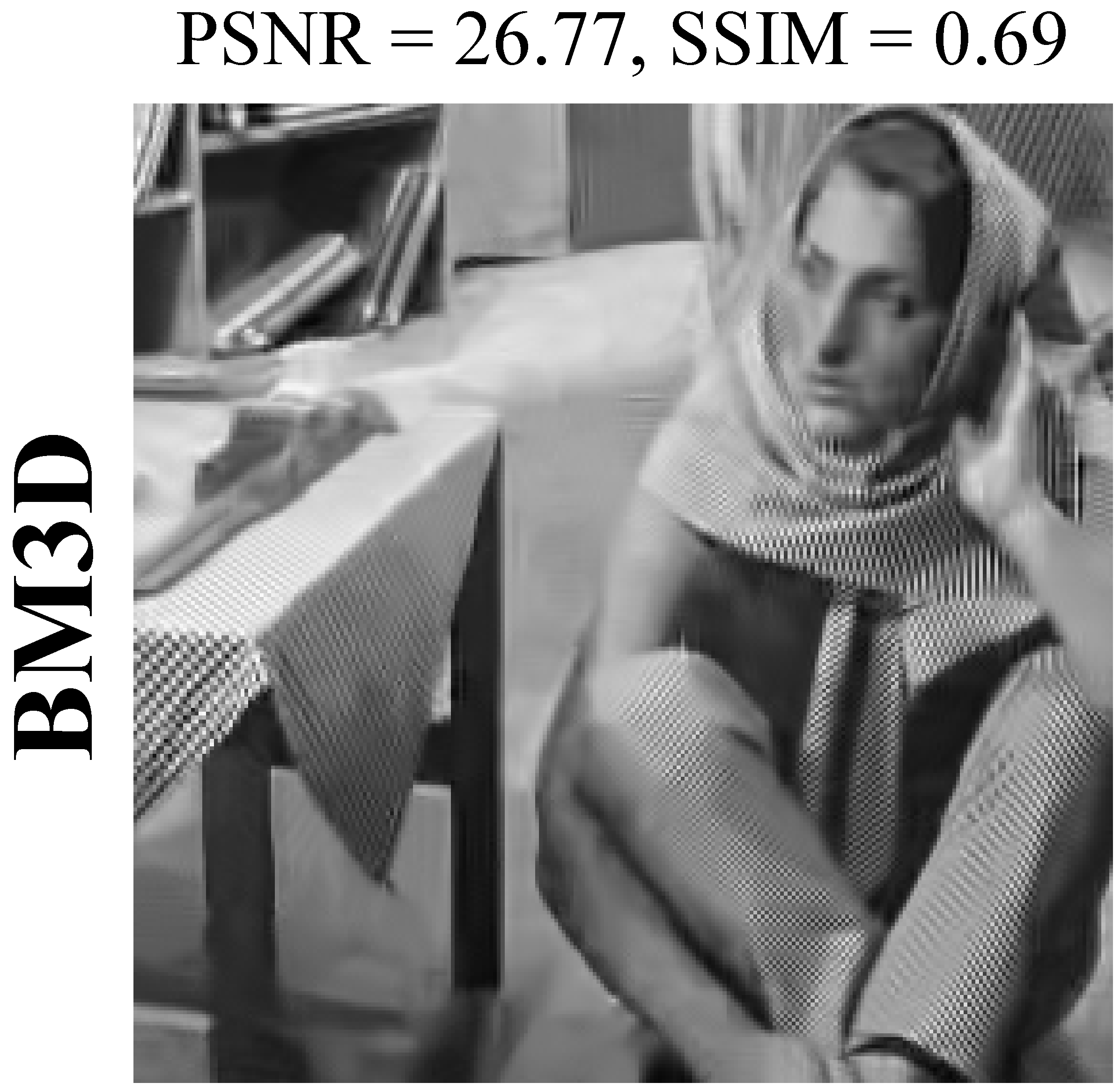}
		\end{subfigure}
		\begin{subfigure}{.16\textwidth}
			\centering
			\includegraphics[width=2.9cm, height=2.8cm]{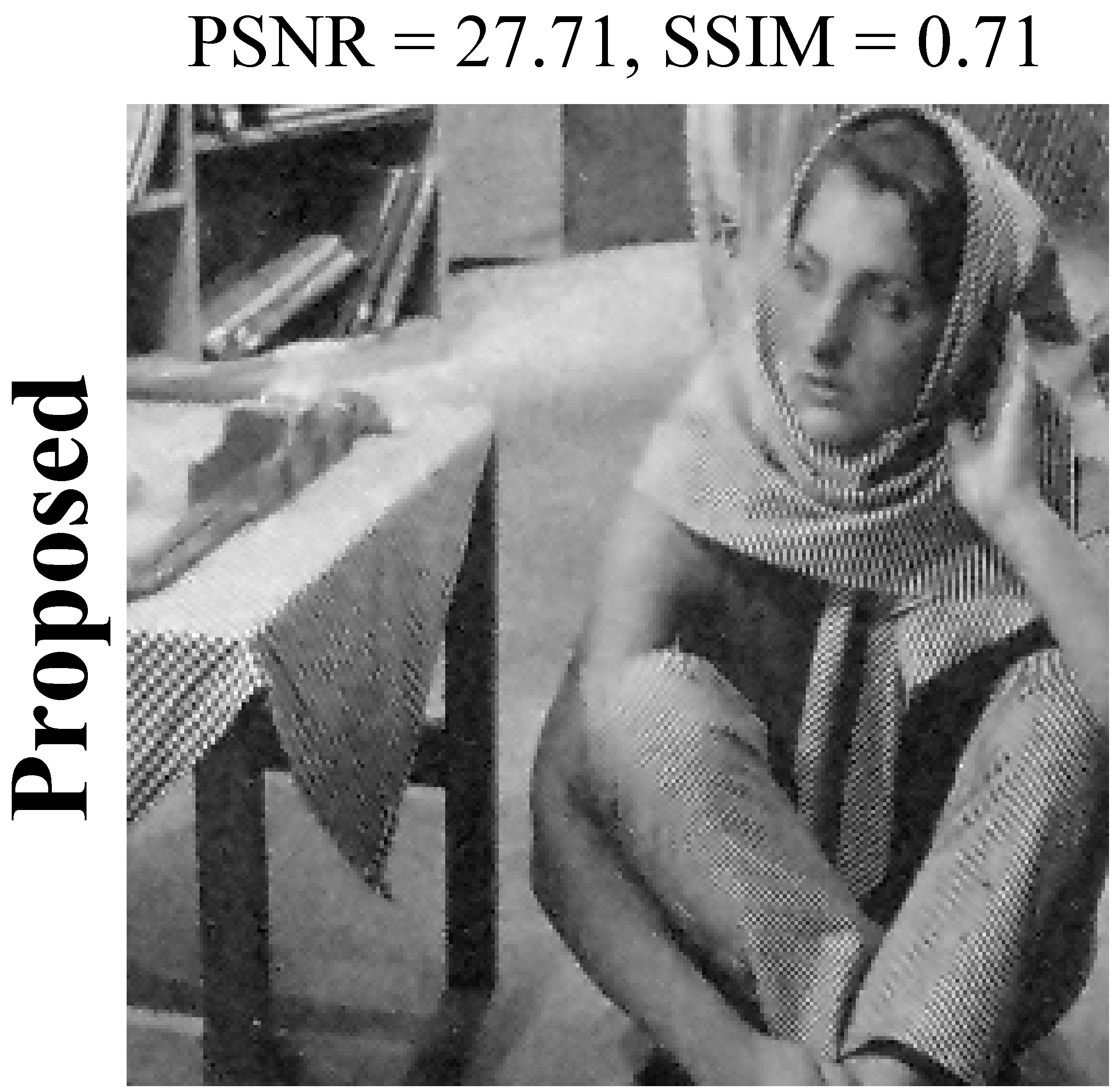}
		\end{subfigure}
	\end{subfigure}
	\caption{Denoising \textit{Barbara}: 1st row at SNR$_{dB}$/$\sigma$ = -5/103, 2nd row at SNR$_{dB}$/$\sigma$ = 0/58, 3rd row at SNR$_{dB}$/$\sigma$ = 5/58}
	\label{fig:Sim_Res_2}
\end{figure*}
\begin{figure*}[h!]
	\begin{subfigure}{\textwidth}
		\centering
		\begin{subfigure}{.13\textwidth}
			\centering
			\includegraphics[width=2.5cm, height=2.4cm]{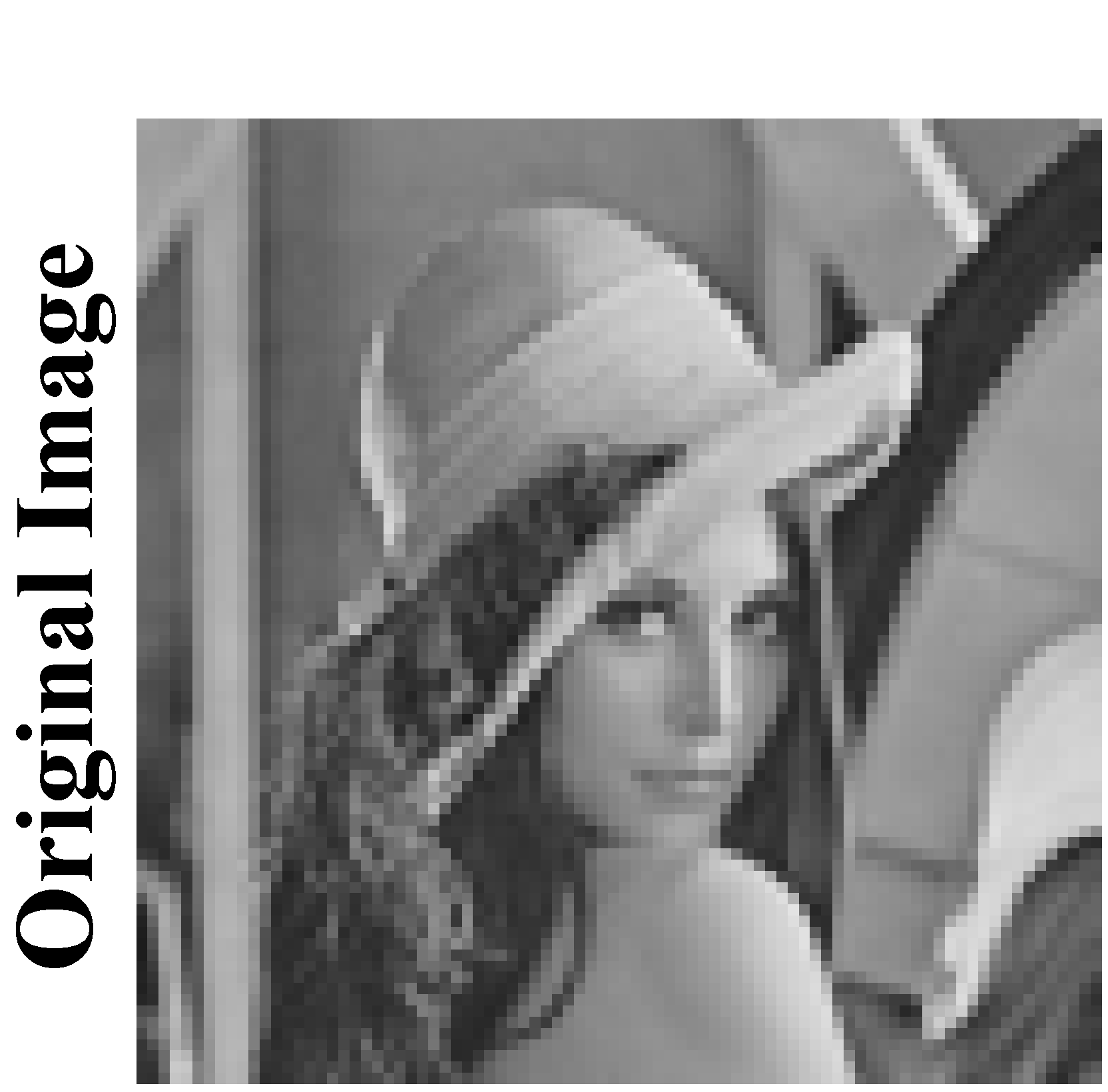}
		\end{subfigure}%
		\begin{subfigure}{.13\textwidth}
			\centering
			\includegraphics[width=2.5cm, height=2.4cm]{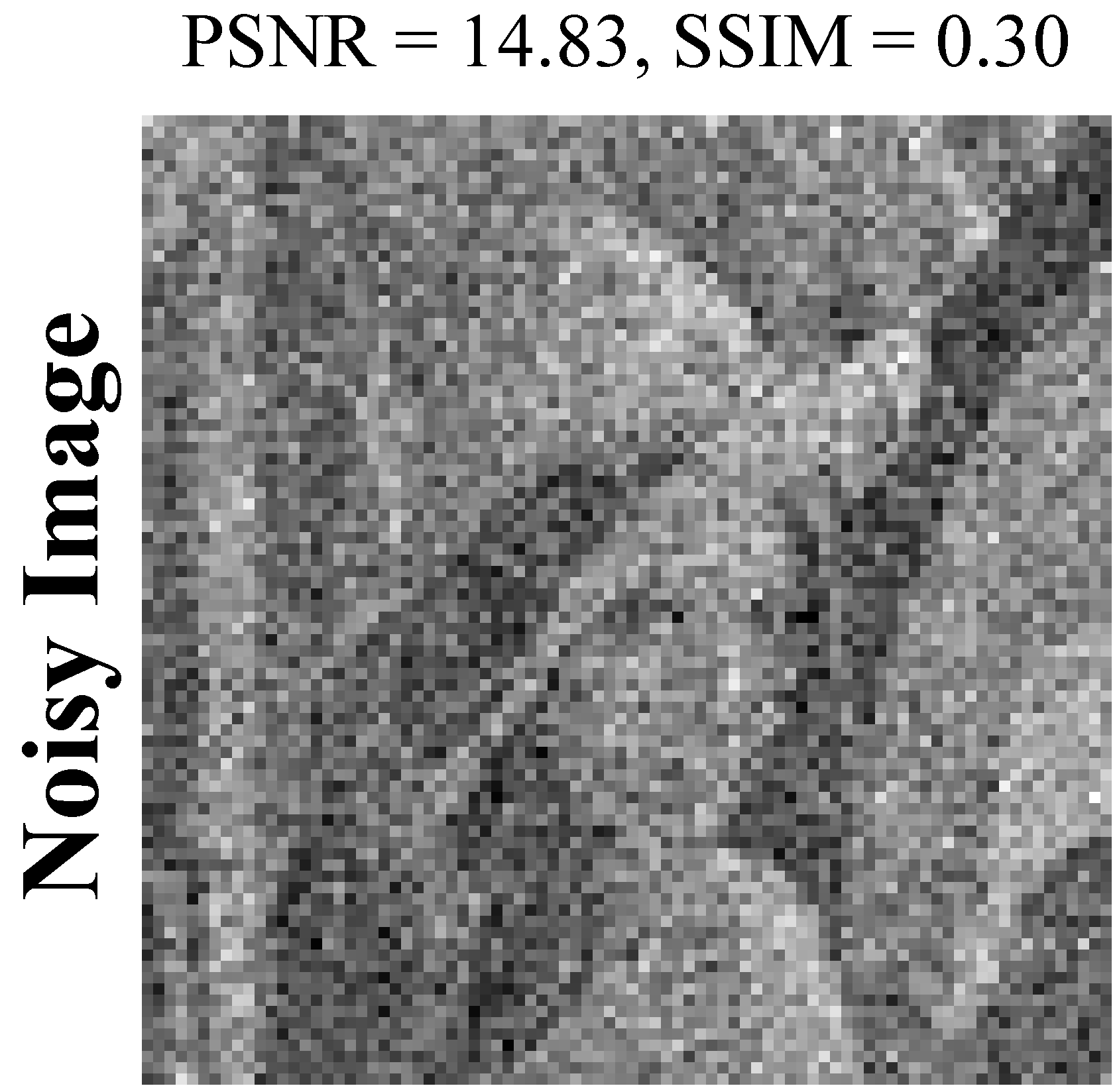}
		\end{subfigure}
		\begin{subfigure}{.13\textwidth}
			\centering
			\includegraphics[width=2.5cm, height=2.4cm]{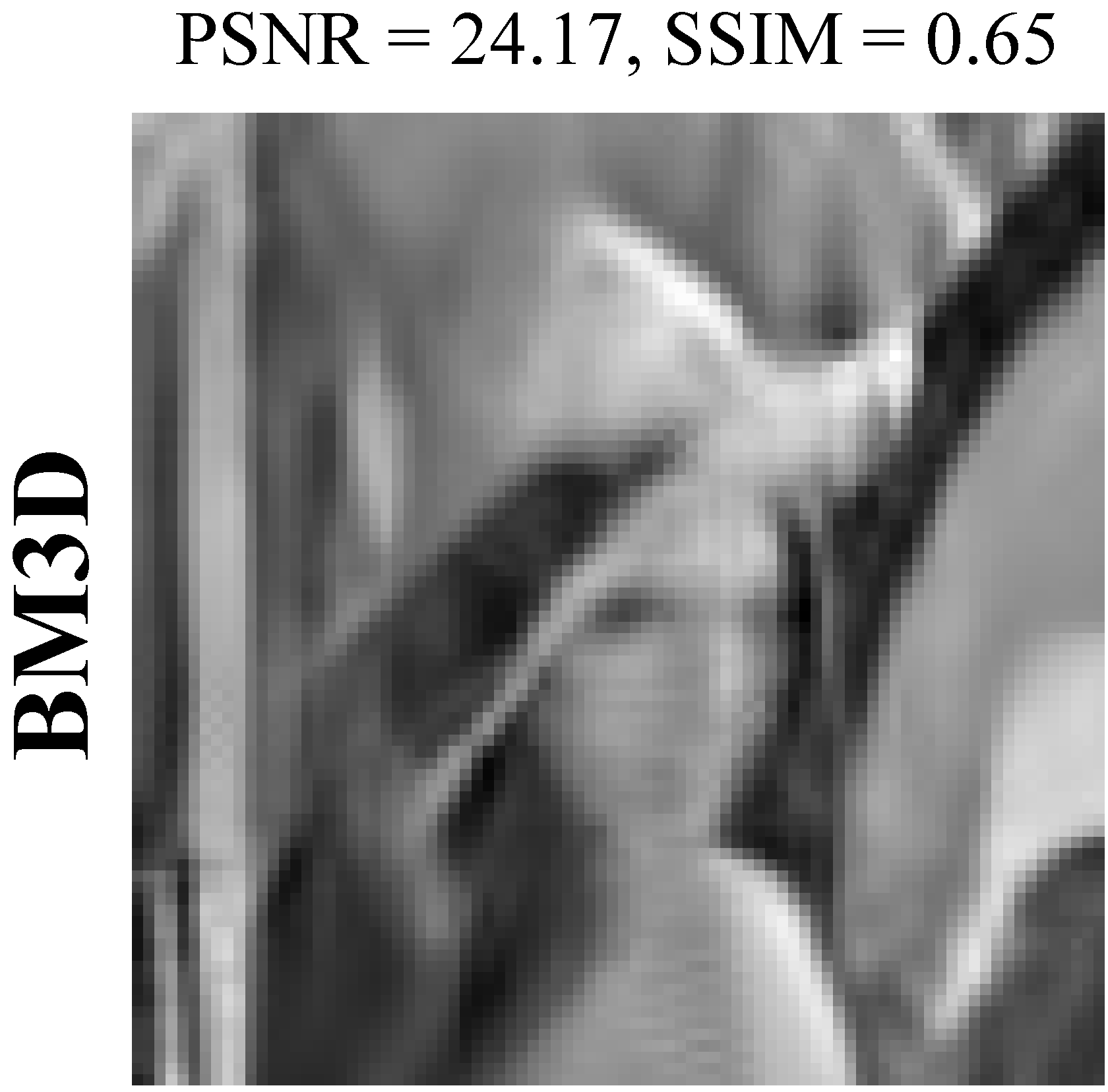}
		\end{subfigure}
		\begin{subfigure}{.13\textwidth}
			\centering
			\includegraphics[width=2.5cm, height=2.4cm]{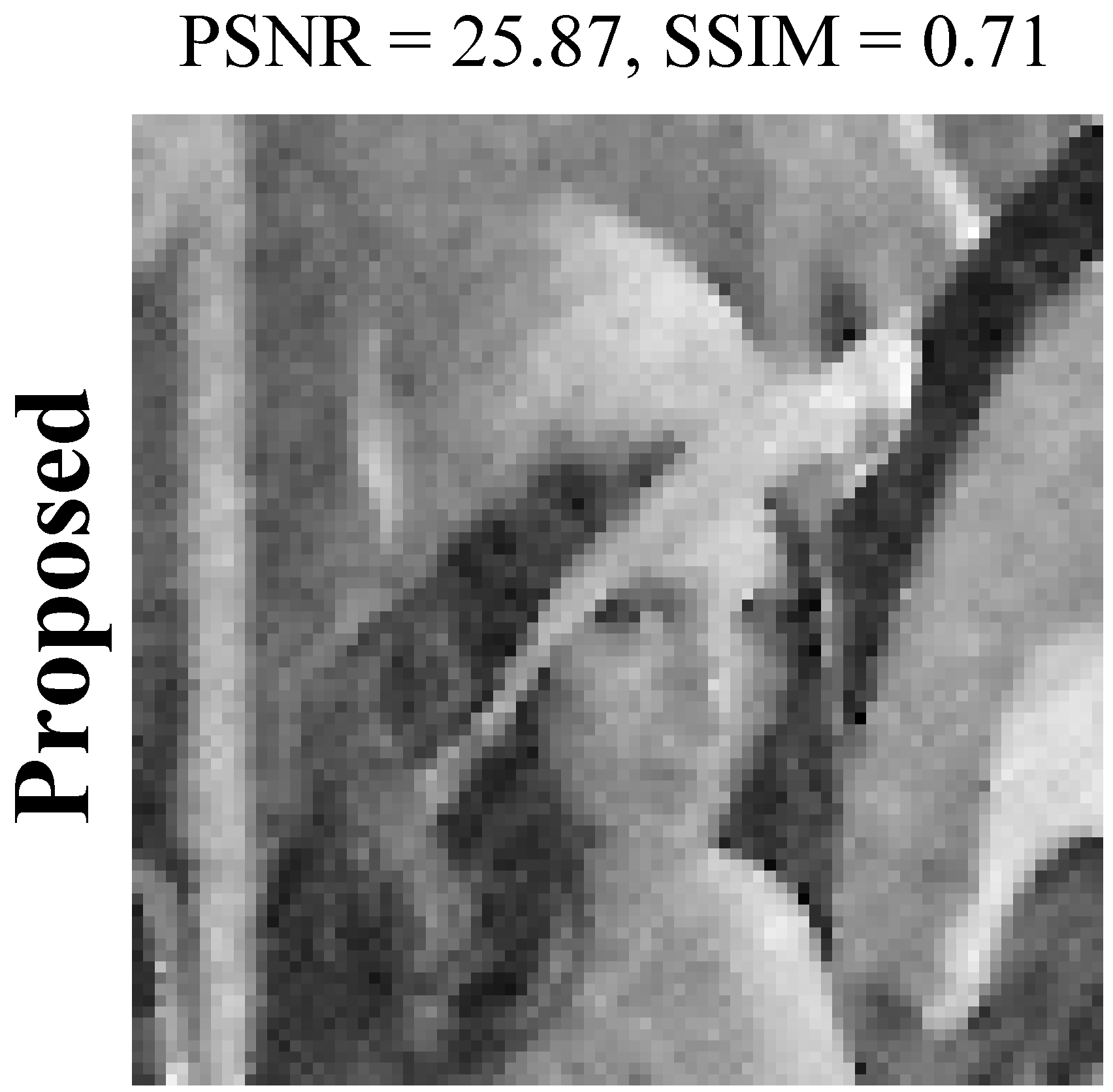}
		\end{subfigure}
		\begin{subfigure}{.13\textwidth}
			\centering
			\includegraphics[width=2.5cm, height=2.4cm]{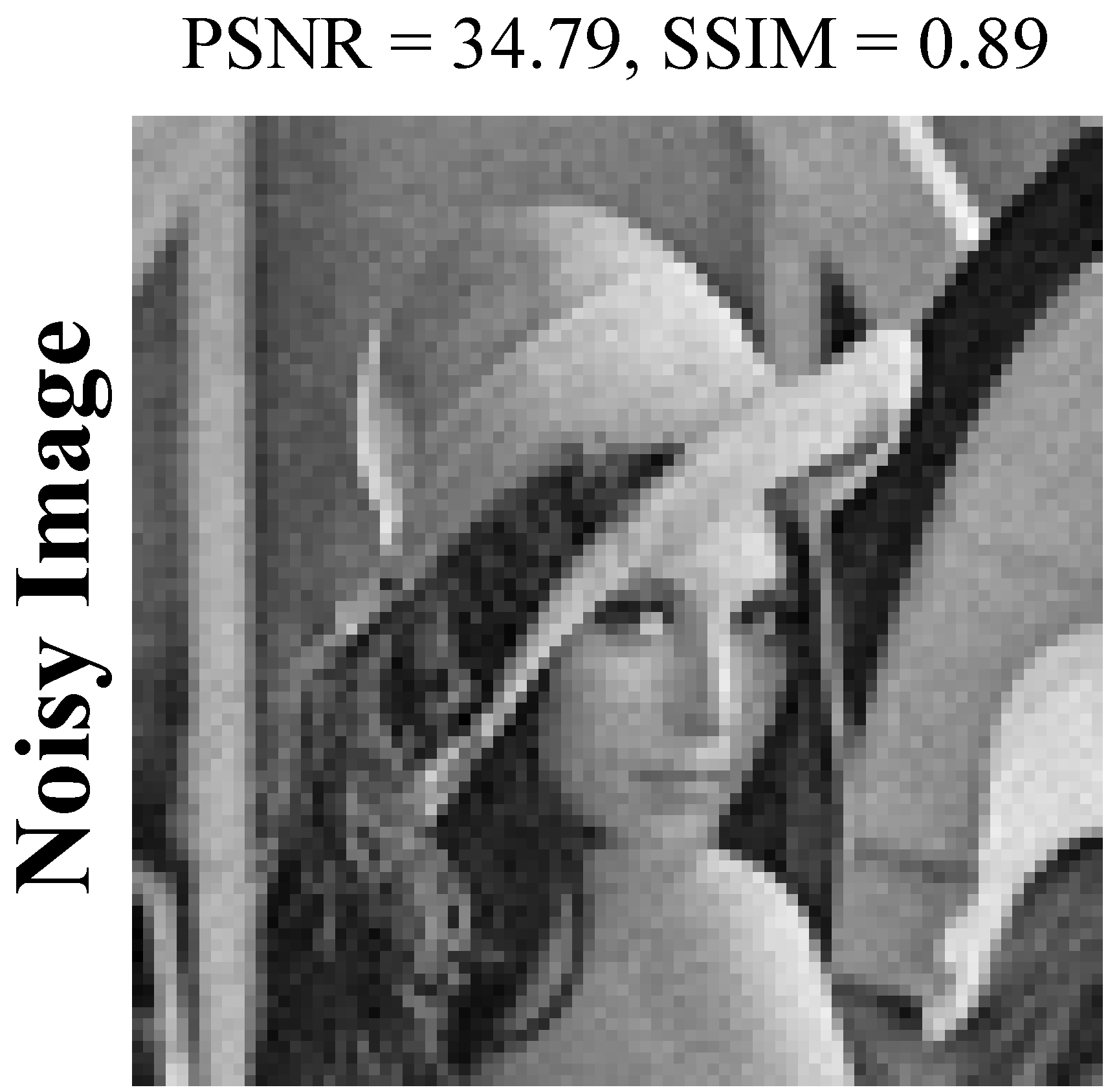}
		\end{subfigure}
		\begin{subfigure}{.13\textwidth}
			\centering
			\includegraphics[width=2.5cm, height=2.4cm]{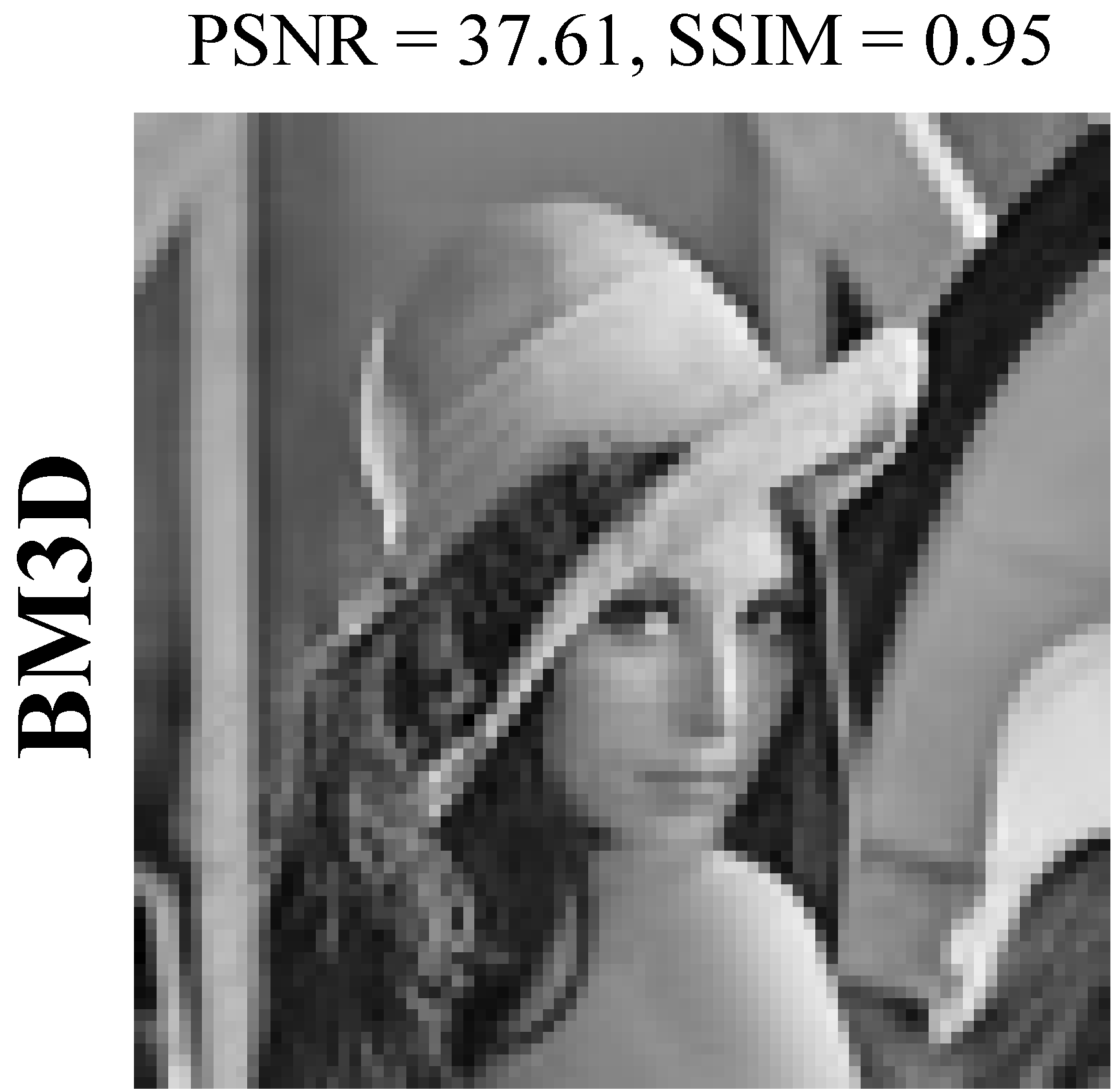}
		\end{subfigure}
		\begin{subfigure}{.13\textwidth}
			\centering
			\includegraphics[width=2.5cm, height=2.4cm]{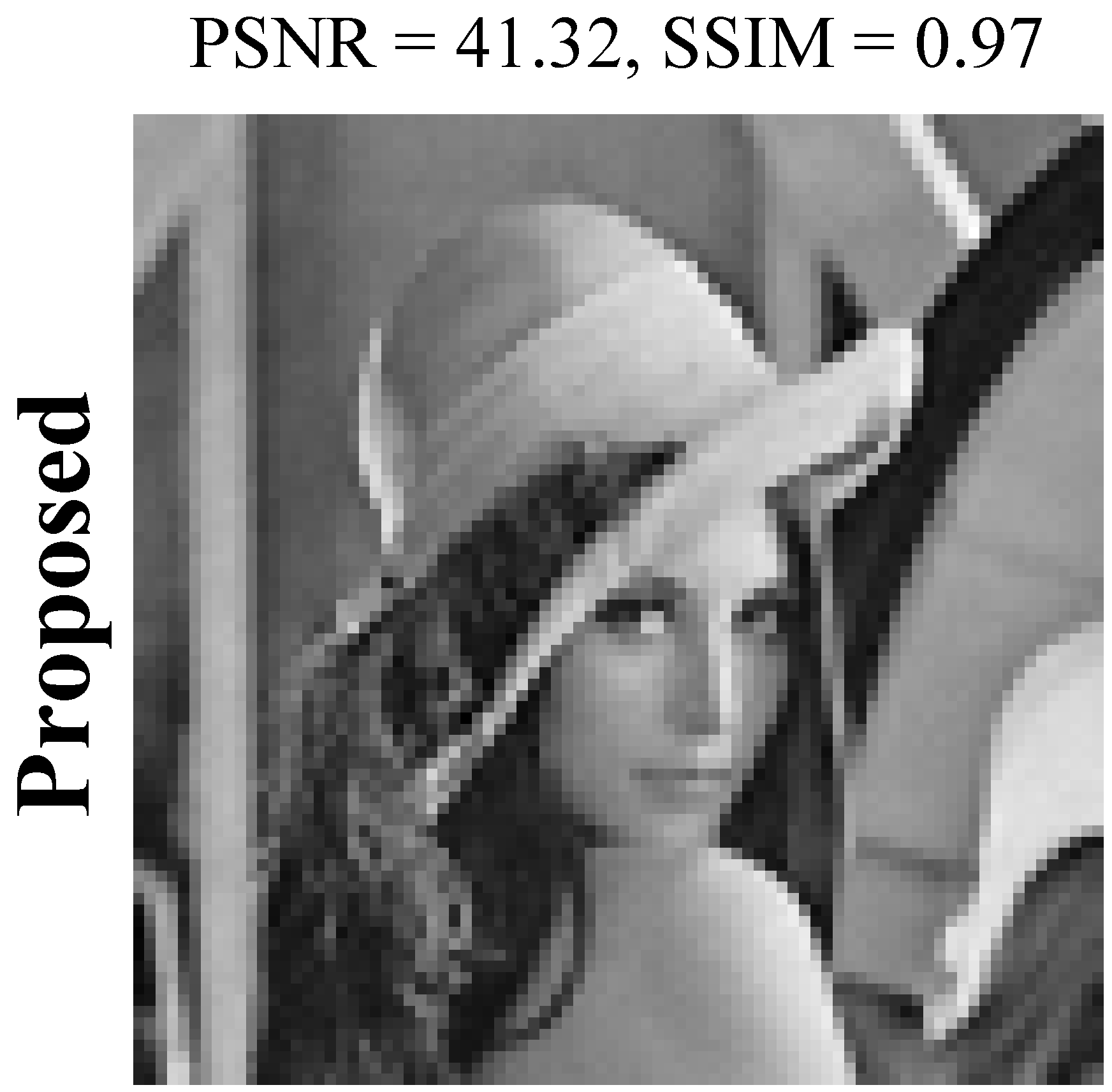}
		\end{subfigure}
	\end{subfigure}\\	
	\begin{subfigure}{\textwidth}
		\centering
		\begin{subfigure}{.13\textwidth}
			\centering
			\includegraphics[width=2.5cm, height=2.4cm]{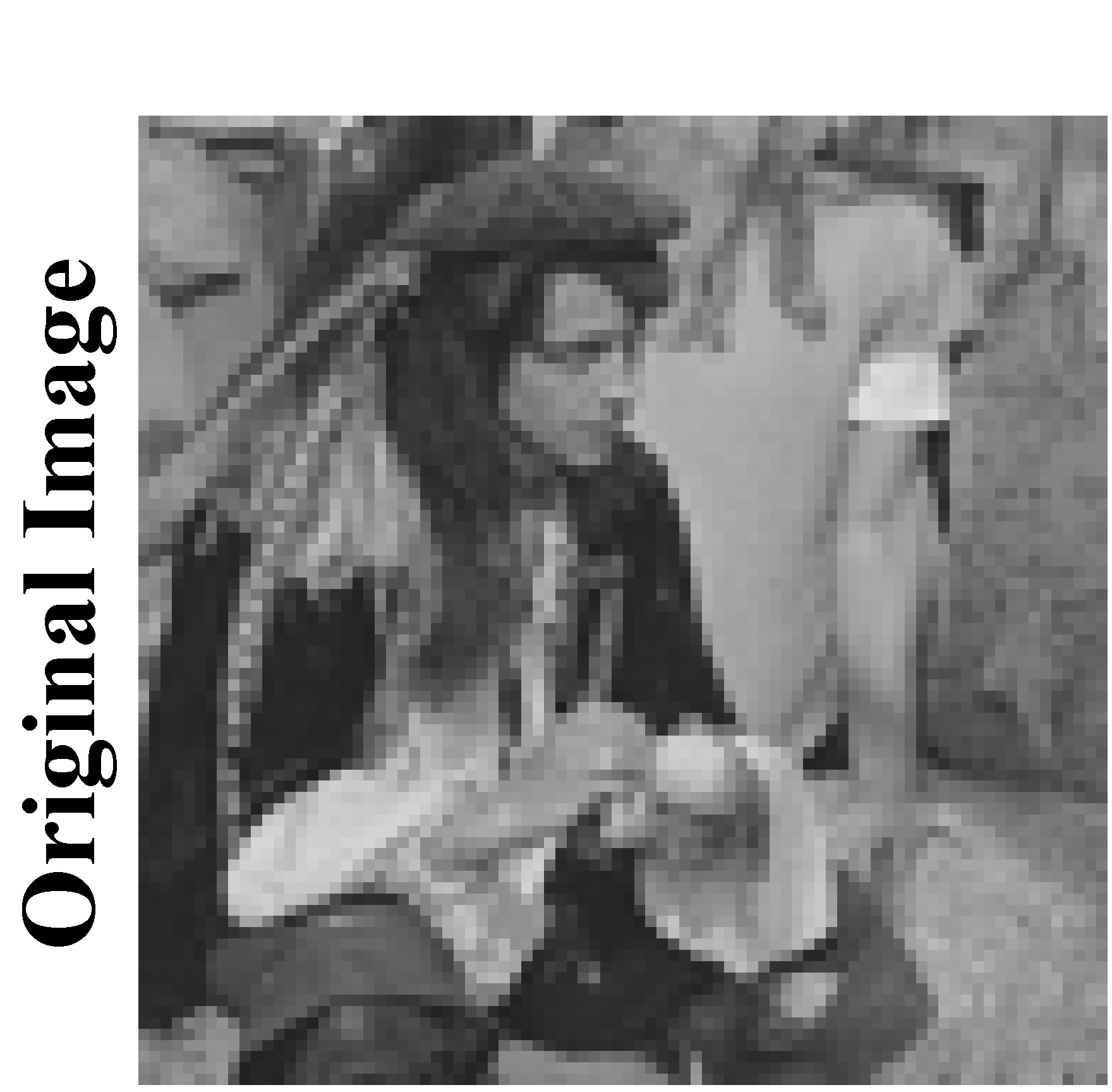}
		\end{subfigure}%
		\begin{subfigure}{.13\textwidth}
			\centering
			\includegraphics[width=2.5cm, height=2.4cm]{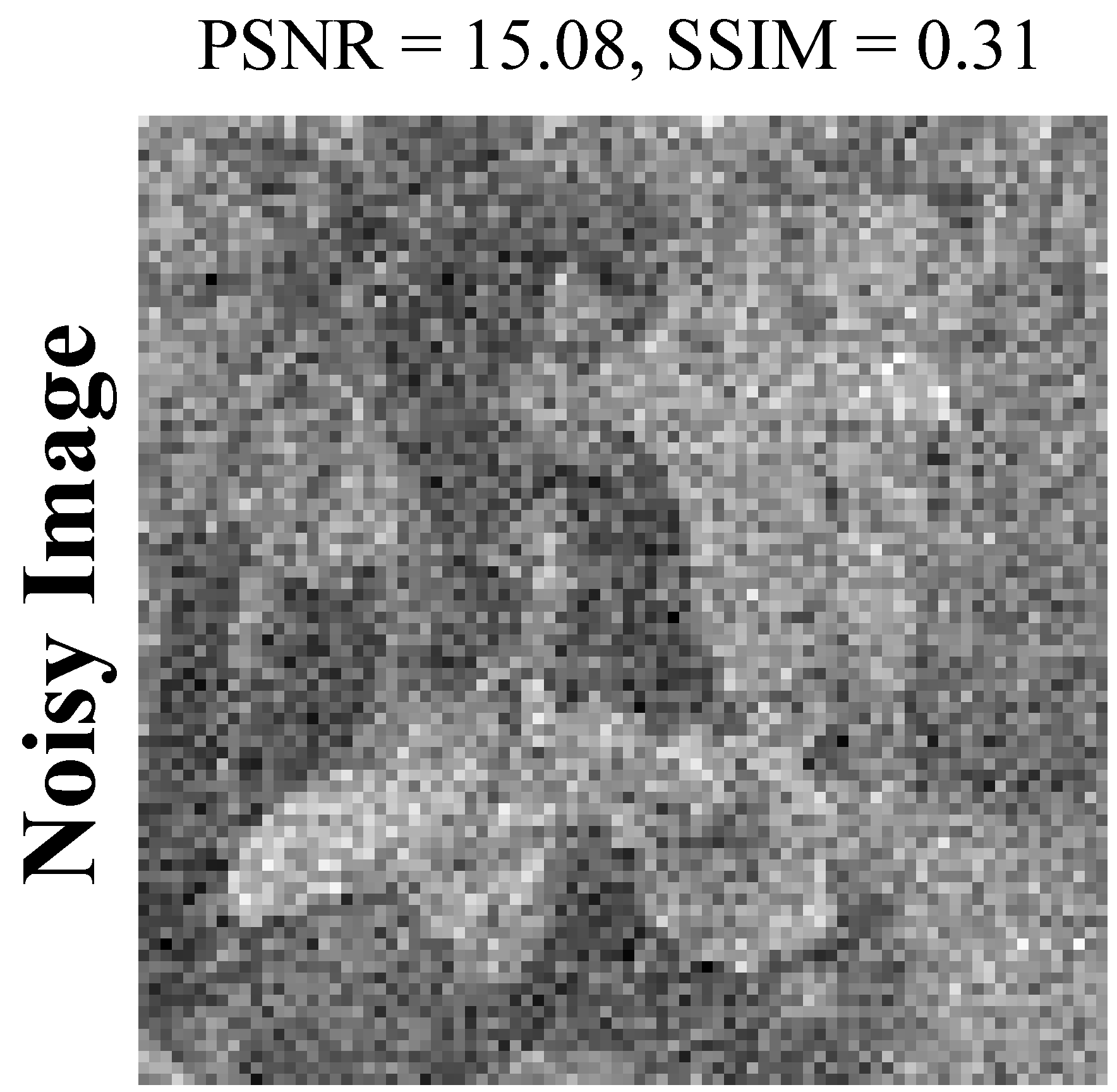}
		\end{subfigure}
		\begin{subfigure}{.13\textwidth}
			\centering
			\includegraphics[width=2.5cm, height=2.4cm]{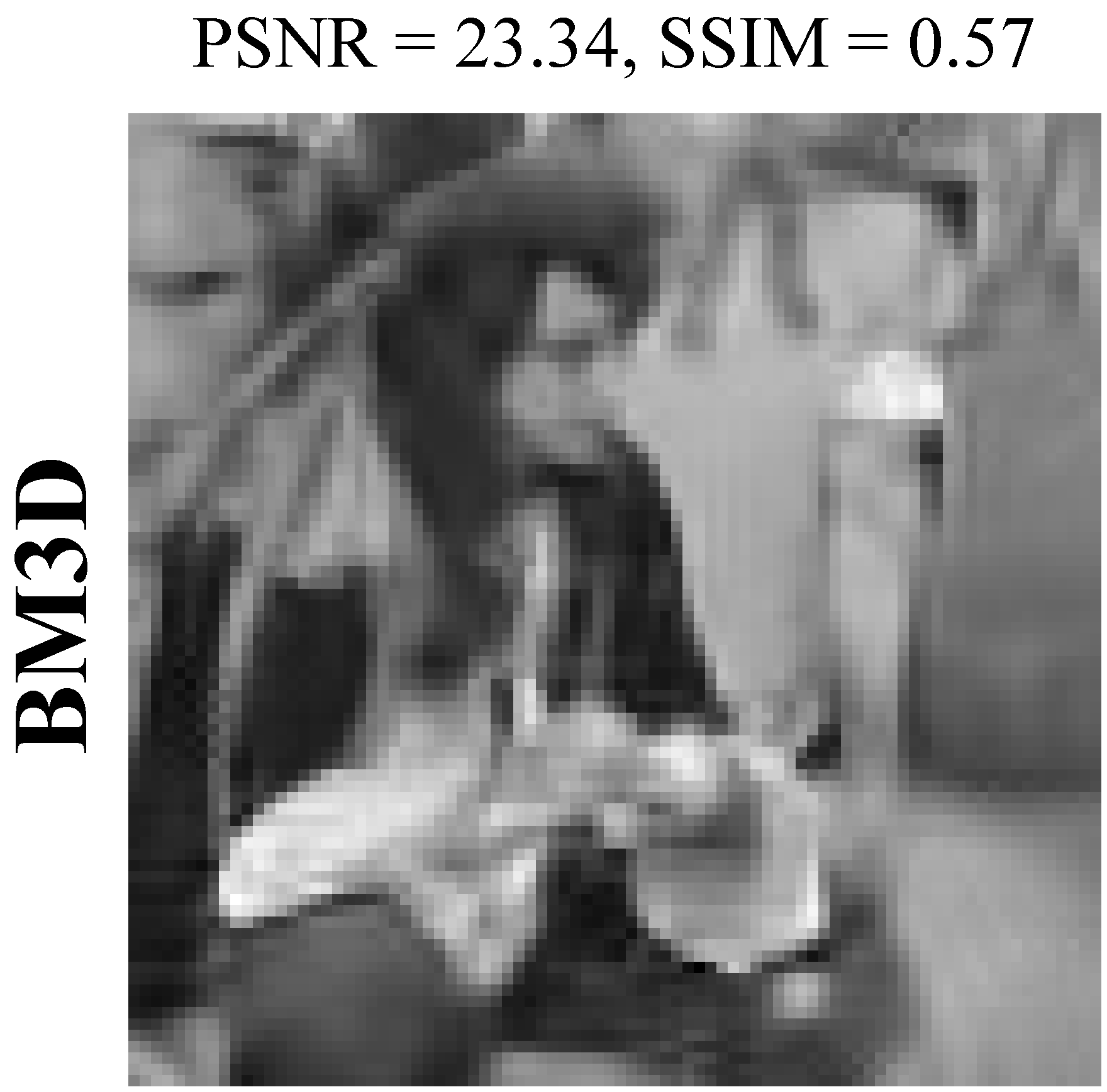}
		\end{subfigure}
		\begin{subfigure}{.13\textwidth}
			\centering
			\includegraphics[width=2.5cm, height=2.4cm]{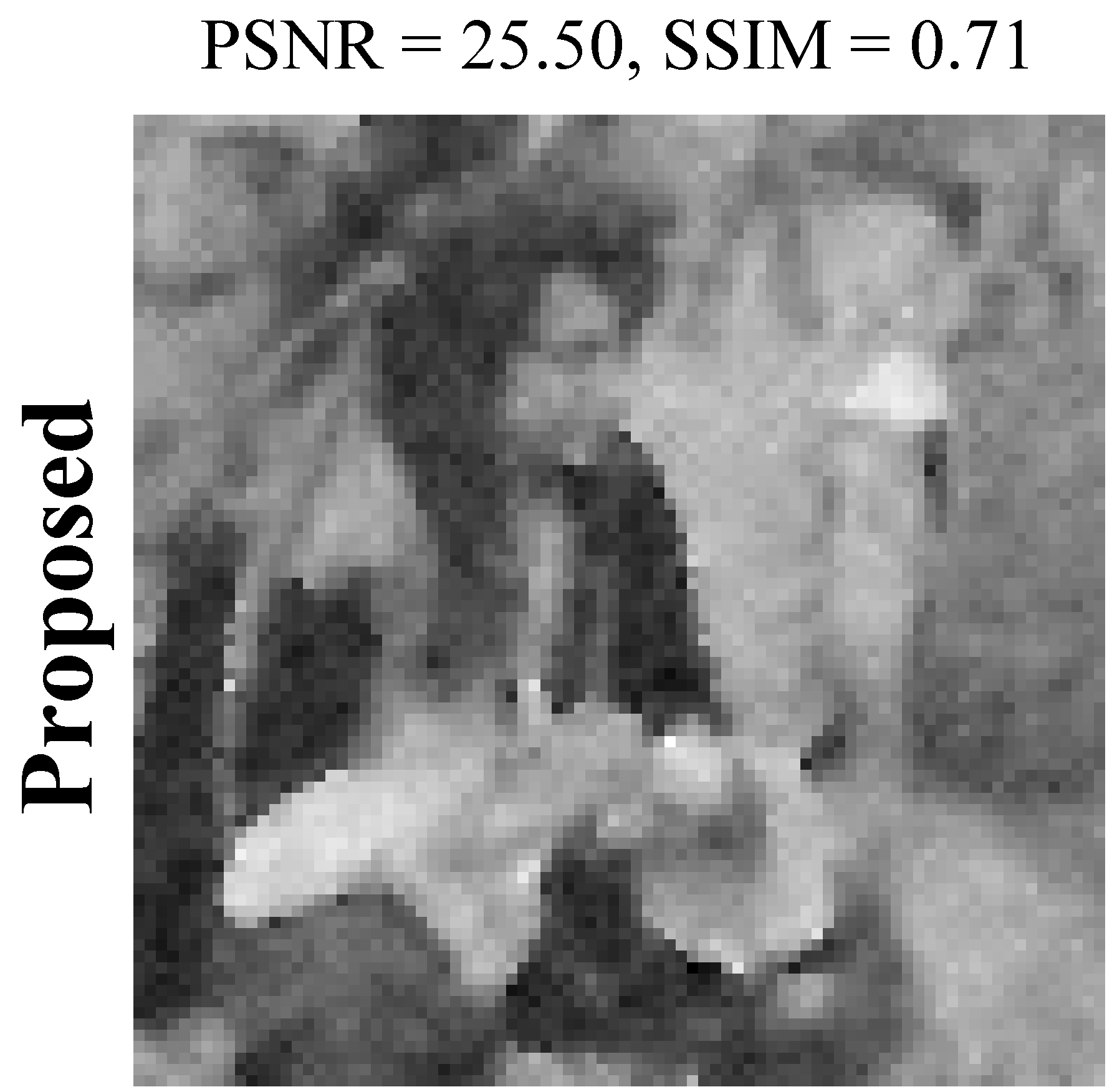}
		\end{subfigure}
		\begin{subfigure}{.13\textwidth}
			\centering
			\includegraphics[width=2.5cm, height=2.4cm]{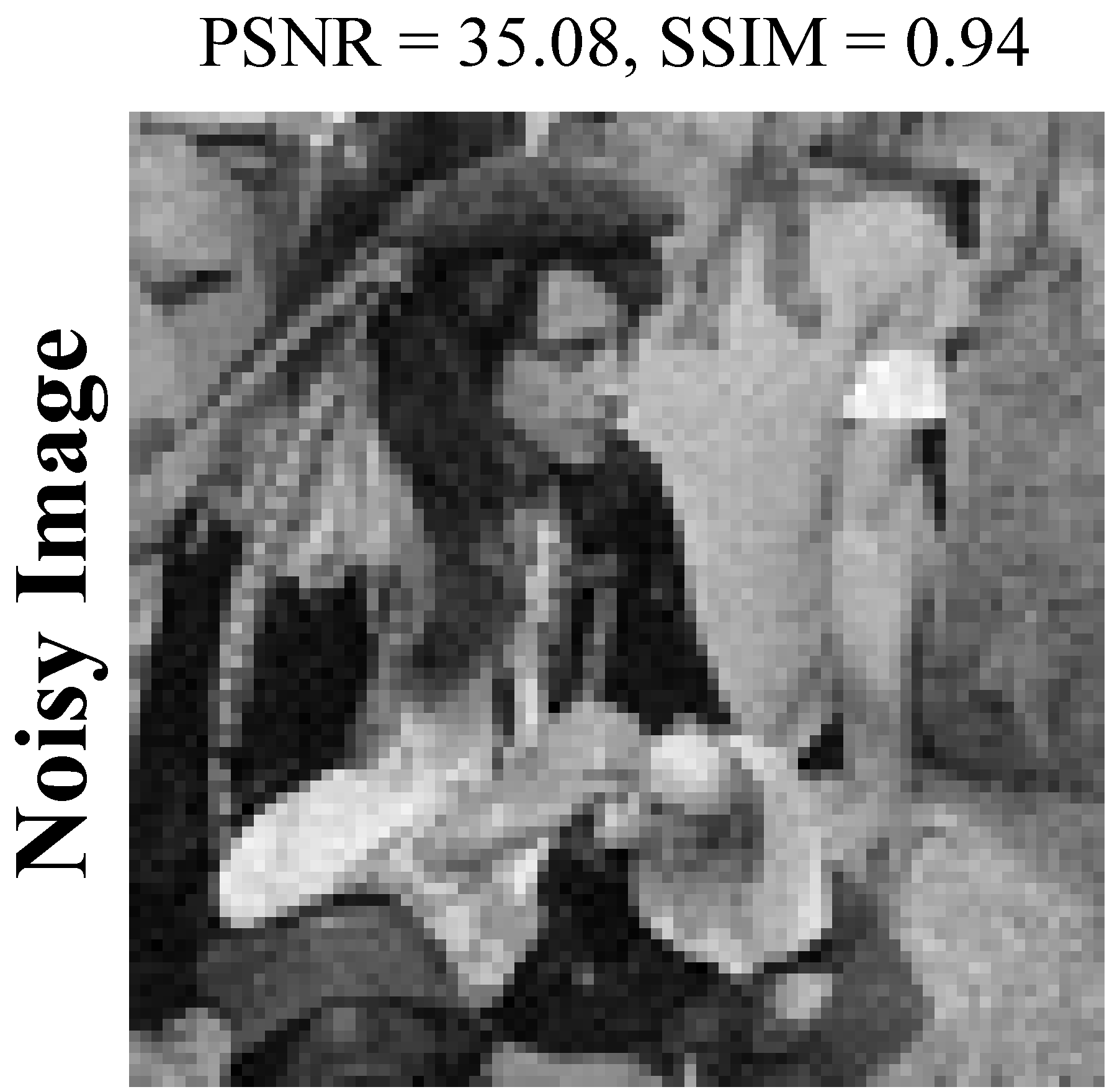}
		\end{subfigure}
		\begin{subfigure}{.13\textwidth}
			\centering
			\includegraphics[width=2.5cm, height=2.4cm]{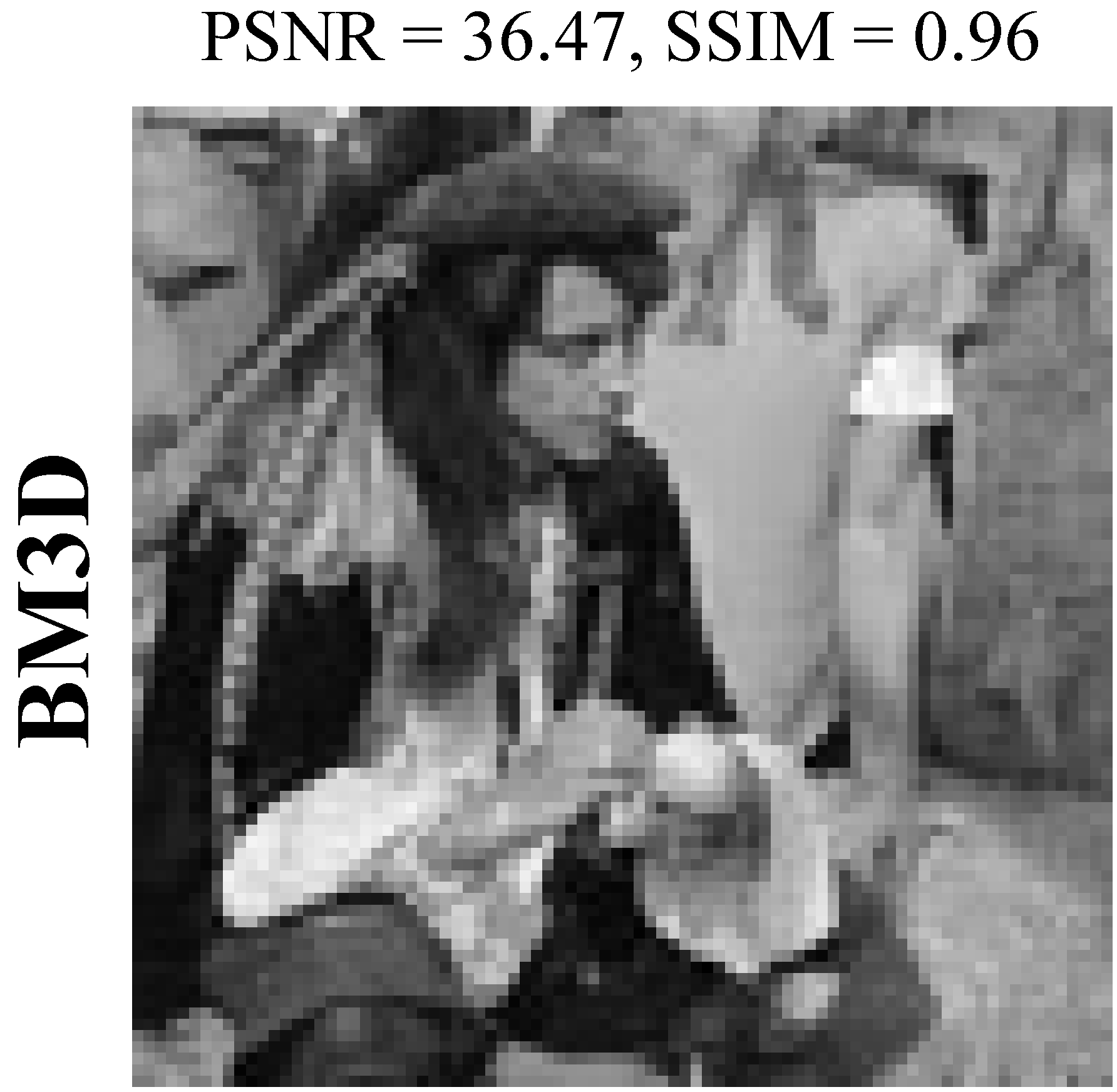}
		\end{subfigure}
		\begin{subfigure}{.13\textwidth}
			\centering
			\includegraphics[width=2.5cm, height=2.4cm]{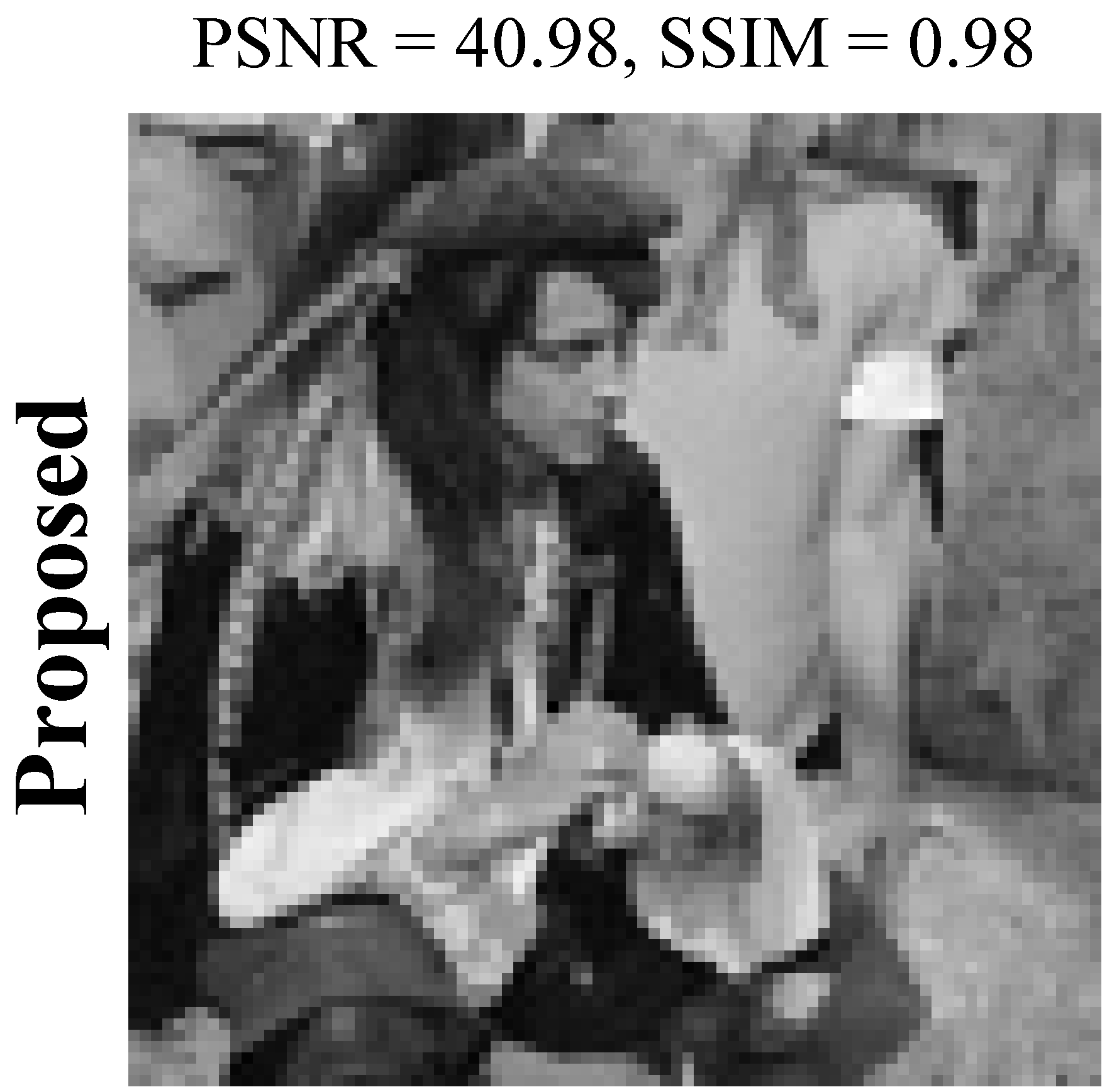}
		\end{subfigure}
	\end{subfigure} 
	\caption{1st row left to right: original \textit{Lena} image, noisy and denoised by BM3D and C2DF at SNR = 0 dB, and noisy and denoised by BM3D and C2DF at SNR = 20 dB. 2nd row left to right: original \textit{Man} image, noisy and denoised by BM3D and C2DF at SNR = 0 dB, and noisy and denoised by BM3D and C2DF at~SNR~=~20 dB} \vspace{1cm}
	\label{fig:Sim_Res_5}
\end{figure*}
\begin{figure*}[t]
	\centering
	\begin{subfigure}{.24\textwidth}
		\centering
		\includegraphics[width=4.2cm, height=5cm]{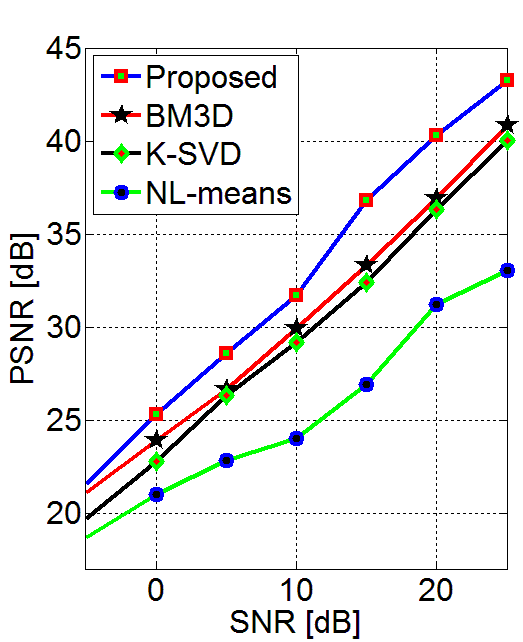}
		\caption{}
	\end{subfigure}%
	\begin{subfigure}{.24\textwidth}
		\centering
		\includegraphics[width=4.2cm, height=5cm]{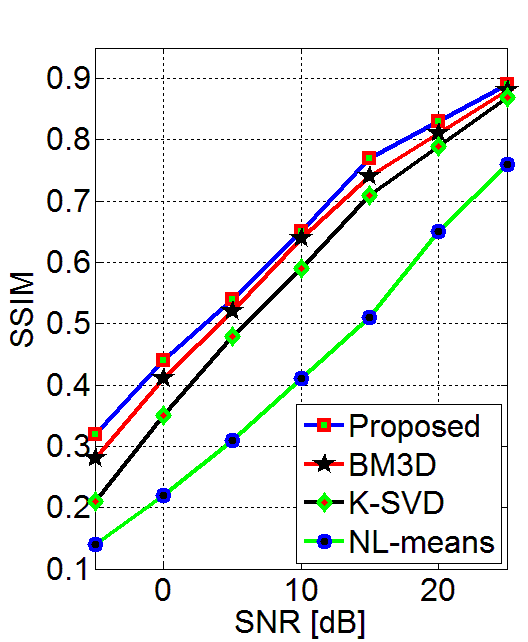}
		\caption{}
	\end{subfigure}
	\begin{subfigure}{.24\textwidth}
		\centering
		\includegraphics[width=4.2cm, height=5cm]{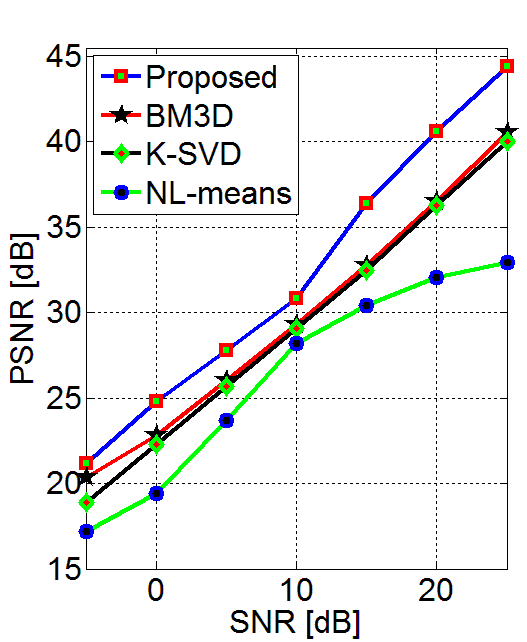}
		\caption{}
	\end{subfigure}
	\begin{subfigure}{.24\textwidth}
		\centering
		\includegraphics[width=4.2cm, height=5cm]{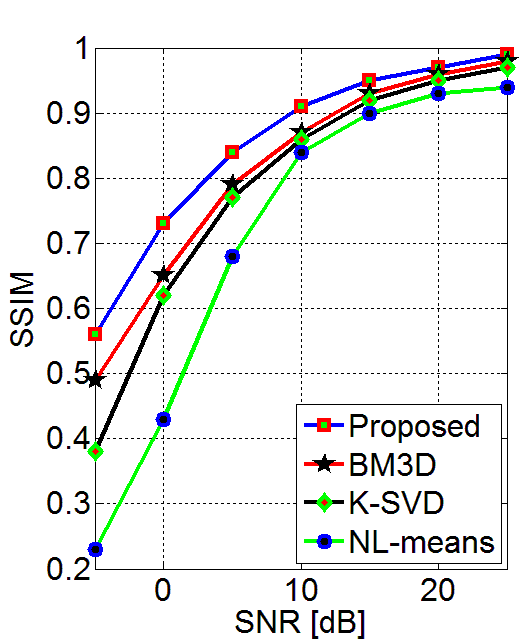}
		\caption{}
	\end{subfigure}
	\caption{Denoising comparison: (a) \textit{Cameraman} PSNR, (b) \textit{Cameraman} SSIM, (c) \textit{Peppers} PSNR, and (d) \textit{Peppers} SSIM}
	\label{fig:Sim_Res_3}
\end{figure*}
\begin{table*}[h!]
	\centering
	\begin{tabular}{|c|c|c|c|c|c|c|c|c|c|c|c|}\hline\hline
		\multicolumn{2}{|c|}{SNR [dB] / $\sigma$} & Living Room & Lena & Barbara & House & Man & Mandrill & Boat & Peppers \\ \cline{1-10}
		\hline\hline
		\multirow{2}{*}{-5/103}
		& BM3D & 22.15/0.42 & 21.53/0.50 & 21.50/0.39 & 23.26/0.34 & 21.21/0.43 & 20.48/0.28 & 22.17/0.35 & 20.33/0.49 \\ \cline{2-10}
		& \textbf{C2DF} & \textbf{23.27/0.54} & \textbf{22.47/0.55} & \textbf{21.75/0.41} & \textbf{23.61/0.36} & \textbf{22.45/0.55} & \textbf{21.48/0.44} & \textbf{23.13/0.48} & \textbf{21.19/0.56} \\ \hline
		
		\multirow{2}{*}{0/58}
		& BM3D & 24.16/0.57 & 24/17/0.65 & 23.91/0.53 & 26.15/0.45 & 23.34/0.57 & 22.55/0.51 & 24.16/0.50 & 22.82/0.65 \\  \cline{2-10}
		& \textbf{C2DF} & \textbf{26.12/0.69} & \textbf{25.87/0.71} & \textbf{24.01/0.55} & \textbf{27.00/0.47} & \textbf{25.50/0.71} & \textbf{23.56/0.60} & \textbf{26.03/0.64} & \textbf{24.78/0.73} \\ \hline
		
		\multirow{2}{*}{5/33}
		& BM3D & 26.98/0.74 & 27.48/0.78 & 26.77/0.69 & 29.83/0.55 & 26.20/0.73 & 24.31/0.63 & 26.90/0.67 & 26.02/0.79 \\  \cline{2-10}
		& \textbf{C2DF} & \textbf{28.31/0.74} & \textbf{28.77/0.81} & \textbf{27.71/0.71} & \textbf{29.84/0.56} & \textbf{28.04/0.81} & \textbf{25.44/0.72} & \textbf{28.38/0.75} & \textbf{27.81/0.84} \\ \hline
		
		\multirow{2}{*}{10/18}
		& BM3D & 30.12/0.86 & 30.72/0.87 & 29.80/0.79 & 33.17/0.62 & 29.19/0.84 & 27.83/0.80 & 29.94/0.81 & 29.30/0.87 \\  \cline{2-10}
		& \textbf{C2DF} & \textbf{33.26/0.91} & \textbf{33.07/0.88} & \textbf{31.60/0.82} & \textbf{34.14/0.63} & \textbf{32.56/0.91} & \textbf{30.19/0.89} & \textbf{32.91/0.87} & \textbf{30.81/0.91} \\ \hline
		
		\multirow{2}{*}{15/10}
		& BM3D & 33.75/0.93 & 34.02/0.92 & 32.92/0.85 & 36.57/0.67 & 32.59/0.92 & 31.81/0.90 & 33.27/0.90 & 32.79/0.93 \\  \cline{2-10}
		& \textbf{C2DF} & \textbf{37.57/0.96} & \textbf{37.39/0.94} & \textbf{35.74/0.89} & \textbf{38.59/0.71} & \textbf{36.93/0.96} & \textbf{36.59/0.96} & \textbf{36.84/0.94} & \textbf{36.41/0.95} \\ \hline
		
		\multirow{2}{*}{20/6}
		& BM3D & 37.67/0.97 & 37.71/0.95 & 36.24/0.90 & 39.90/0.74 & 36.47/0.96 & 36.20/0.96 & 36.78/0.95 & 36.53/0.96 \\  \cline{2-10}
		& \textbf{C2DF} & \textbf{41.40/0.98} & \textbf{41.32/0.97} & \textbf{39.81/0.94} & \textbf{42.37/0.78} & \textbf{40.08/0.98} & \textbf{40.66/0.98} & \textbf{40.11/0.97} & \textbf{40.61/0.97} \\ \hline
		
		\multirow{2}{*}{25/3}
		& BM3D & 41.90/0.98 & 41.51/0.97 & 39.96/0.94 & 43.14/0.80 & 40.77/0.98 & 40.86/0.98 & 40.28/0.97 & 40.53/0.98 \\  \cline{2-10}
		& \textbf{C2DF} & \textbf{44.42/0.99} & \textbf{44.81/0.98} & \textbf{43.44/0.97} & \textbf{45.52/0.85} & \textbf{44.37/0.99} & \textbf{43.92/0.99} & \textbf{42.38/0.98} & \textbf{44.41/0.99} \\ \hline\hline
	\end{tabular}
	\caption{Denoising comparison of grayscale images using BM3D and C2DF both in terms of PSNR and SSIM}
	\label{table1}
\end{table*}
\begin{table}[h!]
	\centering
	\begin{tabular}{|c|c|c|c|c|}\hline\hline
		\multicolumn{2}{|c|}{Image Name} & $\mathcal{N} (0,50)$ & $\mathcal{N} (0,40)$ & $\mathcal{N} (0,30)$  \\ \cline{1-4}
		\hline\hline
		\multirow{2}{*}{1.1.01}
		& BM3D & 20.00/0.24 & 20.48/0.35 & 21.65/0.52 \\ \cline{2-5}
		& \textbf{C2DF} & \textbf{21.92/0.55} & \textbf{22.58/0.61} & \textbf{23.56/0.68} \\ \hline
		
		\multirow{2}{*}{1.1.02}
		& BM3D & 20.08/0.46 & 20.91/0.59 & 22.38/0.71 \\  \cline{2-5}
		& \textbf{C2DF} & \textbf{21.80/0.62} & \textbf{22.62/0.69} & \textbf{23.68/0.76} \\ \hline
		
		\multirow{2}{*}{1.1.03}
		& BM3D & 23.10/0.16 & 23.48/0.22 & 24.12/0.32 \\  \cline{2-5}
		& \textbf{C2DF} & \textbf{24.30/0.42} & \textbf{24.89/0.46} & \textbf{25.64/0.53} \\ \hline
		
		\multirow{2}{*}{1.1.07}
		& BM3D & 25.31/0.10 & 25.55/0.13 & 25.77/0.16 \\  \cline{2-5}
		& \textbf{C2DF} & \textbf{25.91/0.35} & \textbf{26.51/0.39} & \textbf{27.22/0.44} \\ \hline
		
		\multirow{2}{*}{1.2.01}
		& BM3D & 18.18/0.39 & 19.03/0.56 & 20.68/0.72 \\  \cline{2-5}
		& \textbf{C2DF} & \textbf{20.38/0.65} & \textbf{21.25/0.72} & \textbf{22.33/0.78} \\ \hline
		
		\multirow{2}{*}{1.2.05}
		& BM3D & 18.28/0.49 & 19.10/0.62 & 20.84/0.76 \\  \cline{2-5}
		& \textbf{C2DF} & \textbf{20.23/0.68} & \textbf{21.11/0.74} & \textbf{22.19/0.80} \\ \hline
		
		\multirow{2}{*}{1.2.07}
		& BM3D & 17.80/0.29 & 18.67/0.48 & 20.38/0.68 \\  \cline{2-5}
		& \textbf{C2DF} & \textbf{20.02/0.61} & \textbf{20.88/0.68 }& \textbf{22.00/0.76} \\ \hline
		
		\multirow{2}{*}{1.2.13}
		& BM3D & 17.96/0.69 & 18.84/0.76 & 20.63/0.85 \\  \cline{2-5}
		& \textbf{C2DF} & \textbf{20.22/0.81} & \textbf{21.43/0.85} & \textbf{22.96/0.90} \\ \hline
		
		\multirow{2}{*}{1.3.01}
		& BM3D & 20.94/0.18 & 21.48/0.30 & 22.40/0.44 \\  \cline{2-5}
		& \textbf{C2DF} & \textbf{22.52/0.45} & \textbf{23.10/0.51} & \textbf{23.97/0.60} \\ \hline
		
		\multirow{2}{*}{1.3.04}
		& BM3D & 21.55/0.31 & 22.30/0.44 & 23.43/0.57 \\  \cline{2-5}
		& \textbf{C2DF} & \textbf{23.91/0.65} & \textbf{24.78/0.70} & \textbf{25.72/0.75} \\ \hline \hline
	\end{tabular}
	\caption{Denoising texture images from SIPI database using BM3D and C2DF under $\Wm \sim \mathcal{N} (\textbf{0},\sigma_w\textbf{I})$}
	\label{table2}
\end{table}
\begin{table}[h!]
\centering
\begin{tabular}{|c|c|c|c|c|}\hline\hline
	\multicolumn{2}{|c|}{Image Name} & $\mathcal{N} (0,50)$ & $\mathcal{N} (0,40)$ & $\mathcal{N} (0,30)$  \\ \cline{1-4}
	\hline\hline
	\multirow{2}{*}{2.1.01}
	& BM3D & 20.11/0.36 & 20.55/0.42 & 21.26/0.51 \\ \cline{2-5}
	& \textbf{C2DF} & \textbf{21.47/0.57} & \textbf{21.95/0.62} & \textbf{22.46/0.67} \\ \hline
	
	\multirow{2}{*}{2.1.02}
	& BM3D & 20.00/0.35 & 20.53/0.43 & 21.22/0.51 \\  \cline{2-5}
	& \textbf{C2DF} & \textbf{21.07/0.52} & \textbf{21.50/0.56} & \textbf{22.00/0.62} \\ \hline
	
	\multirow{2}{*}{2.1.05}
	& BM3D & 20.22/0.48 & 20.72/0.54 & 21.67/0.61 \\  \cline{2-5}
	& \textbf{C2DF} & \textbf{21.13/0.59} & \textbf{21.69/0.64} & \textbf{22.32/0.69} \\ \hline
	
	\multirow{2}{*}{2.1.06}
	& BM3D & 20.83/0.21 & 21.00/0.26 & 21.25/0.33 \\  \cline{2-5}
	& \textbf{C2DF} & \textbf{21.38/0.38} & \textbf{21.56/0.41} & \textbf{21.75/0.45} \\ \hline
	
	\multirow{2}{*}{2.1.12}
	& BM3D & 18.31/0.11 & 18.43/0.15 & 18.57/0.20 \\  \cline{2-5}
	& \textbf{C2DF} & \textbf{18.57/0.30} & \textbf{18.68/0.33} & \textbf{18.80/0.36} \\ \hline
	
	\multirow{2}{*}{2.2.07}
	& BM3D & 26.44/0.32 & 27.17/0.37 & 28.06/0.43 \\  \cline{2-5}
	& \textbf{C2DF} & \textbf{27.48/0.45} & \textbf{28.43/0.51} & \textbf{29.53/0.57} \\ \hline
	
	\multirow{2}{*}{2.2.11}
	& BM3D & 24.47/0.12 & 24.95/0.19 & 25.58/0.27 \\  \cline{2-5}
	& \textbf{C2DF} & \textbf{26.04/0.42} & \textbf{26.78/0.48} & \textbf{27.71/0.55} \\ \hline
	
	\multirow{2}{*}{2.2.13}
	& BM3D & 24.65/0.23 & 25.12/0.30 & 25.89/0.38 \\  \cline{2-5}
	& \textbf{C2DF} & \textbf{26.17/0.53} & \textbf{27.23/0.60} & \textbf{28.52/0.67} \\ \hline
	
	\multirow{2}{*}{2.2.14}
	& BM3D & 24.96/0.22 & 25.43/0.29 & 26.11/0.37 \\  \cline{2-5}
	& \textbf{C2DF} & \textbf{26.43/0.47} & \textbf{27.20/0.53} & \textbf{28.17/0.60} \\ \hline
	
	\multirow{2}{*}{2.2.17}
	& BM3D & 23.09/0.21 & 23.60/0.27 & 24.32/0.35 \\  \cline{2-5}
	& \textbf{C2DF} & \textbf{24.81/0.46} & \textbf{25.46/0.51} & \textbf{26.26/0.57} \\ \hline \hline
\end{tabular}
\caption{Denoising aerial images from SIPI database using BM3D and C2DF under $\Wm \sim \mathcal{N} (\textbf{0},\sigma_w\textbf{I})$}
\label{table3}
\end{table}

\subsection{General Comparisons}
\label{General_Comparisons}
Consider Fig. \ref{fig:Sim_Res_1}, where we compare the results of denoising $256 \times 256$ standard \textit{Mandrill} image using the state-of-the-art methods and our proposed method. Since this image is rich in details, it serves as one of the best images for comparisons. As shown, the competing algorithms blur out the details especially in case of high noise. On the contrary, our method accurately tackles the noise components in the detail-rich and complex parts of the image. For instance, compare the nostrils, hair and beard in the mandrill image, as shown in Fig. \ref{fig:zoomed}, which got blurred out by other competing methods but are recovered to a very good extent by C2DF. Avoiding such blurring is one of the key issues since it can have critical consequences in many applications, e.g., bio-medical application where, let's say, the task is to detect tumors. In that case, the detection may go wrong severely due to blurring, and hence can be life-threatening.

A similar comparison of denoising $256 \times 256$ \textit{Barbara} image has been shown in Fig. \ref{fig:Sim_Res_2} that clearly depicts successful recovery of image details discarding the noise components. Further in Fig. \ref{fig:Sim_Res_5}, we present the comparison of denoising the standard \textit{Lena} and \textit{Man} images at both high and low noise to showcase that our algorithm perform equally well at low noise tenure. Please note that in this figure, we have compare the results with BM3D only as it outperforms all the existing methods. These results clearly validates the efficiency of C2DF outperforming the existing~state-of-the-art methods.

In Fig. \ref{fig:Sim_Res_3}, we compare the denoising results of grayscale \textit{Cameraman} and \textit{Peppers} image in terms of PSNR and SSIM. The images were corrupted by a range of different noise levels from SNR = -5 dB to SNR = 25 dB, and the corresponding results are plotted. We have shown to outperform the existing methods over a range of SNRs. Specifically, since the \textit{Peppers} image has a lot of flat regions that ultimately favors the competing algorithms, as these tend to blur out the smooth and flat parts of the image, we have outperformed these algorithms even in such a scenario. Our PSNR and SSIM in both the images and at any noise level is better than that of the existing methods. We recommend the readers to zoom into the electronic files of this article for a much better comparison of the images shown.
\begin{figure*}[t]
	\begin{subfigure}{\textwidth}
		\centering
		\begin{subfigure}{.24\textwidth}
			\centering
			\includegraphics[width=4cm, height=2.4cm]{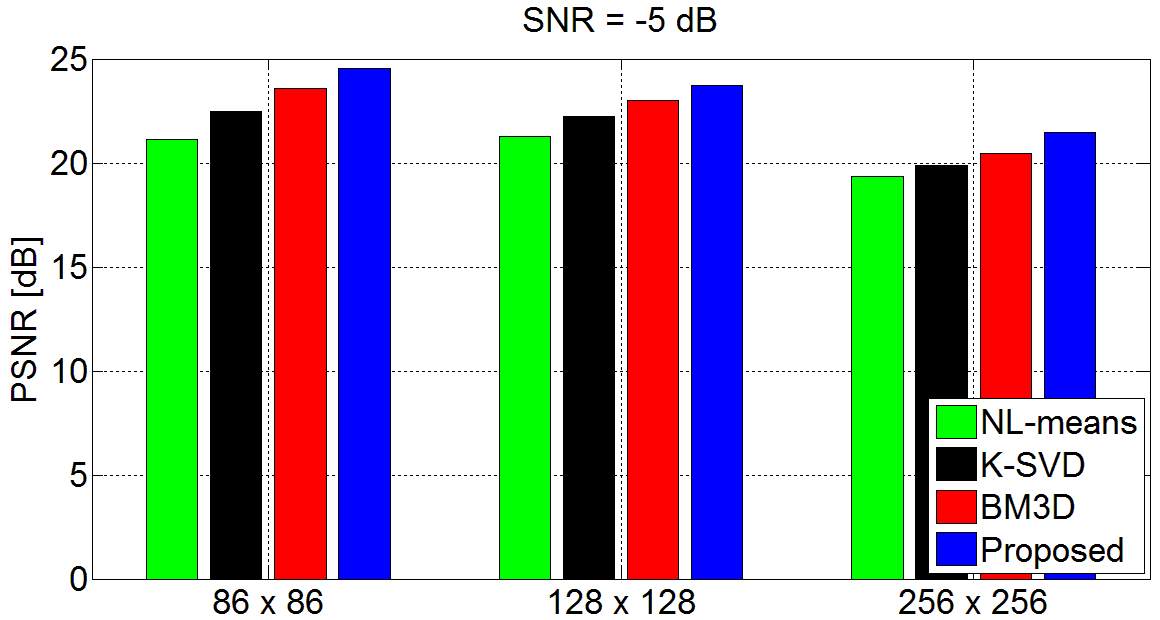}
		\end{subfigure}%
		\begin{subfigure}{.24\textwidth}
			\centering
			\includegraphics[width=4cm, height=2.4cm]{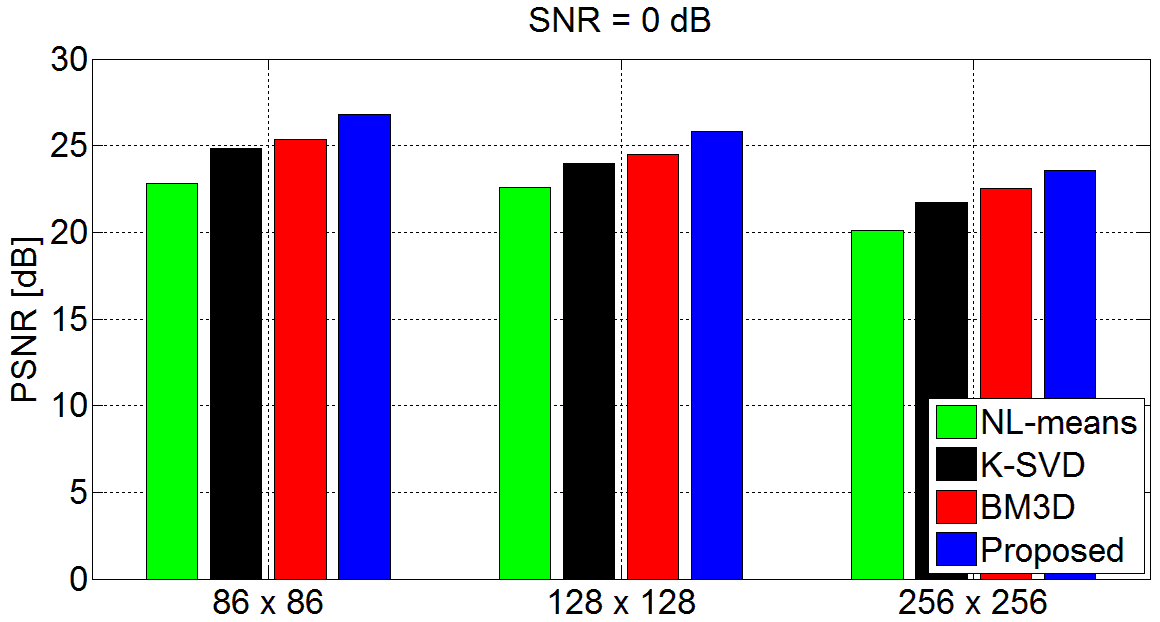}
		\end{subfigure}
		\begin{subfigure}{.24\textwidth}
			\centering
			\includegraphics[width=4cm, height=2.4cm]{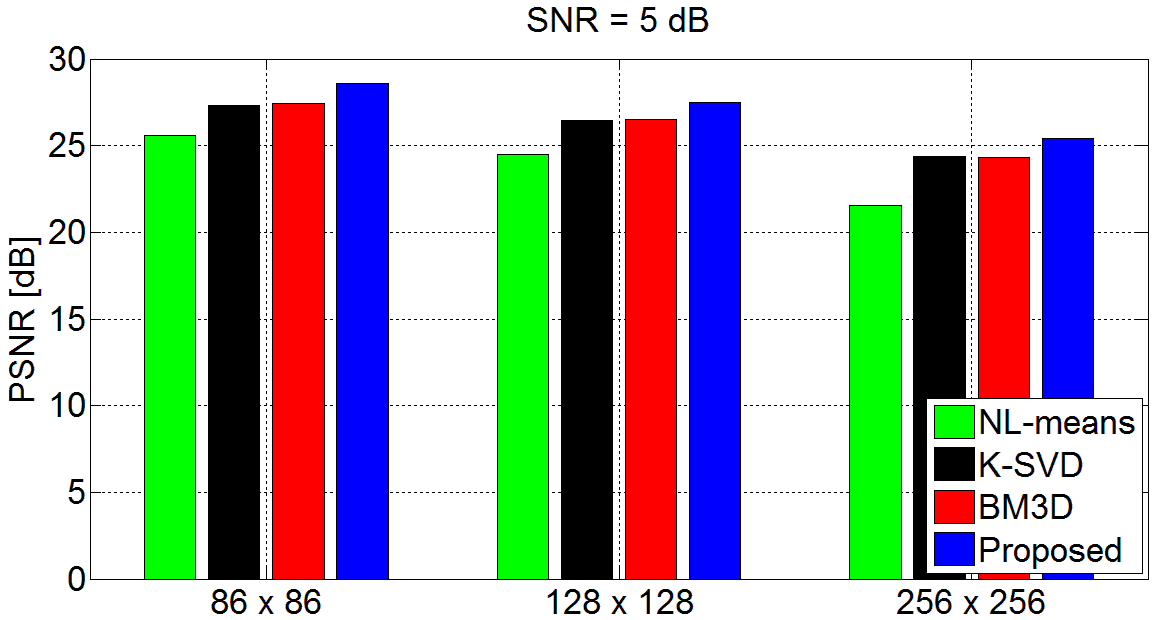}
		\end{subfigure}
		\begin{subfigure}{.24\textwidth}
			\centering
			\includegraphics[width=4cm, height=2.4cm]{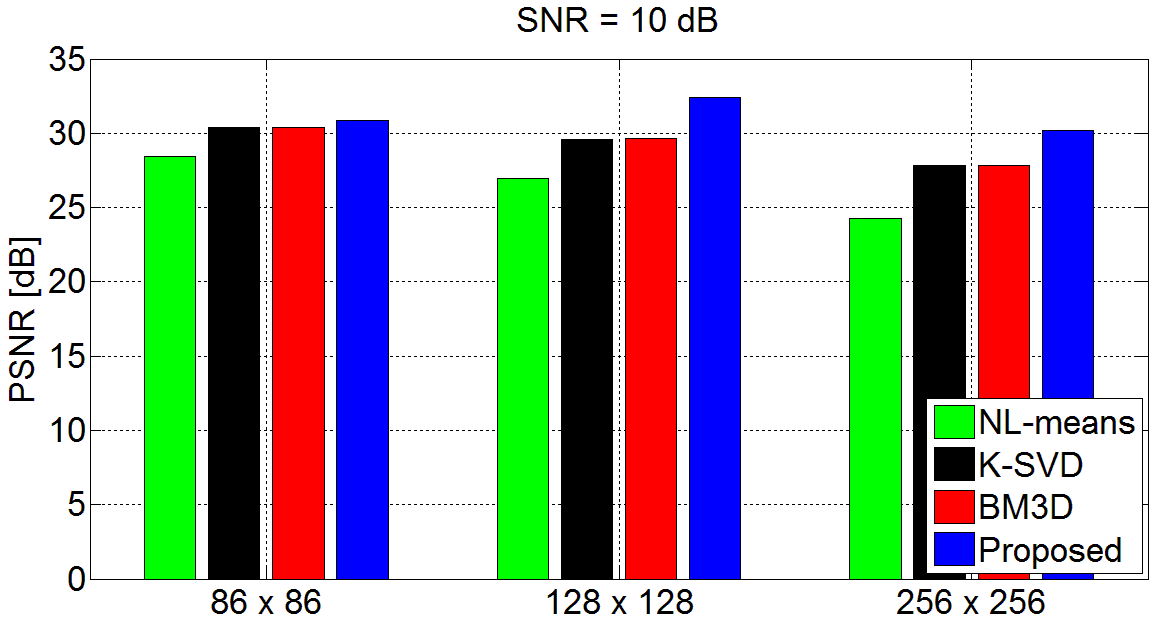}
		\end{subfigure}
	\end{subfigure}\\	
	\begin{subfigure}{\textwidth}
		\centering
		\begin{subfigure}{.24\textwidth}
			\centering
			\includegraphics[width=4cm, height=2.4cm]{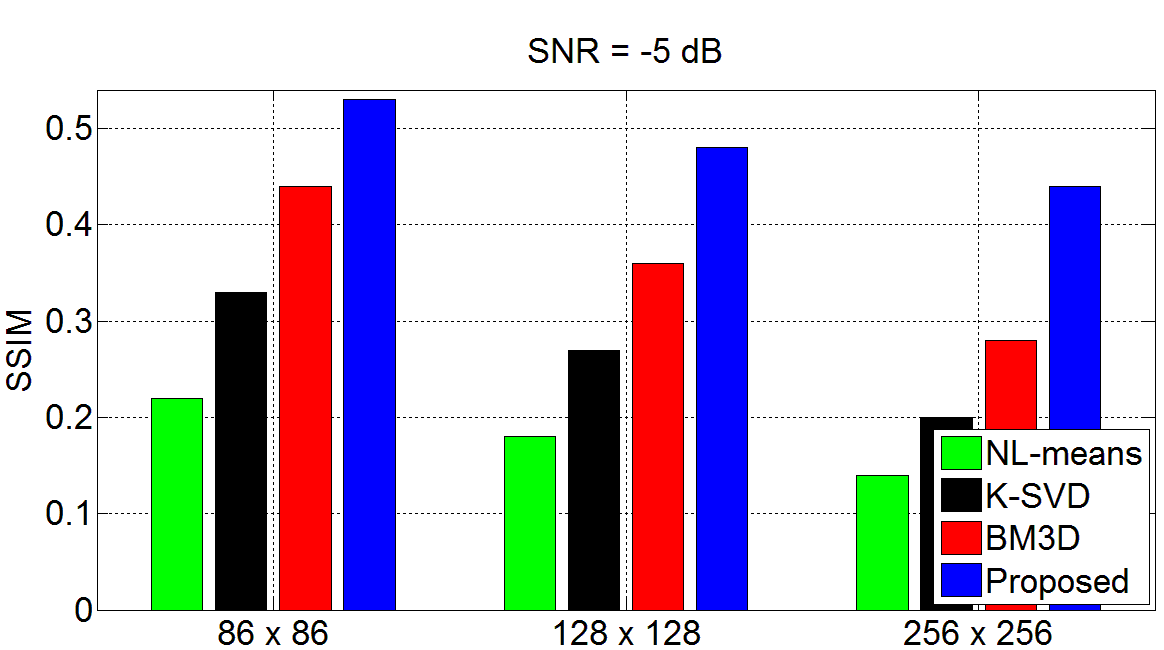}
		\end{subfigure}%
		\begin{subfigure}{.24\textwidth}
			\centering
			\includegraphics[width=4cm, height=2.4cm]{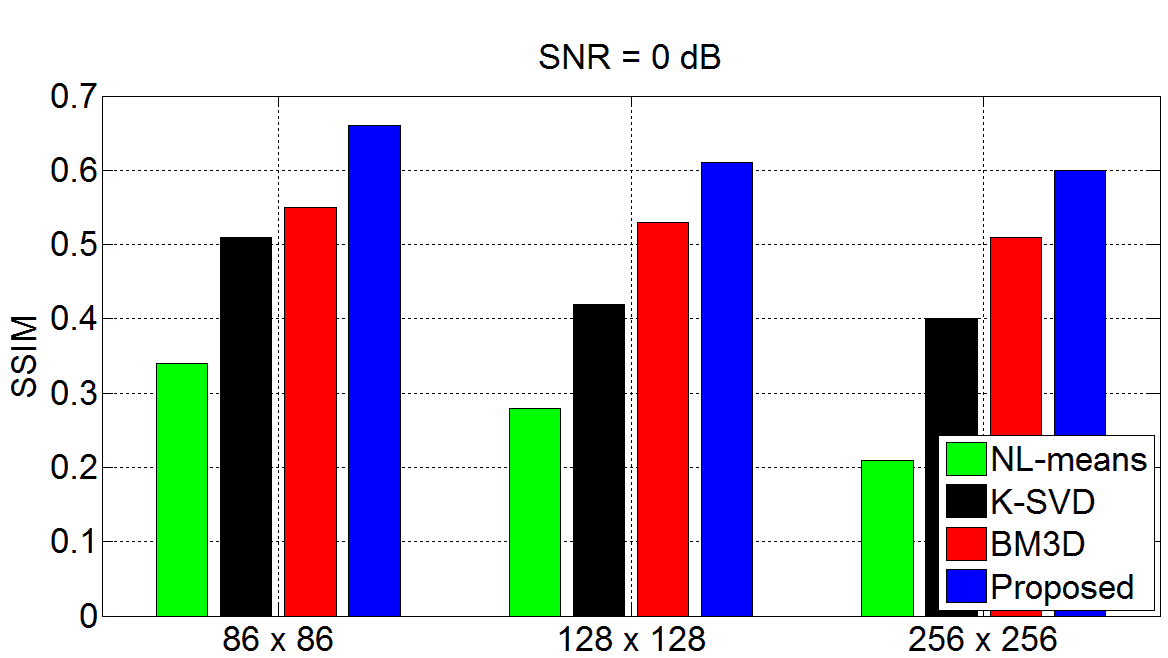}
		\end{subfigure}
		\begin{subfigure}{.24\textwidth}
			\centering
			\includegraphics[width=4cm, height=2.4cm]{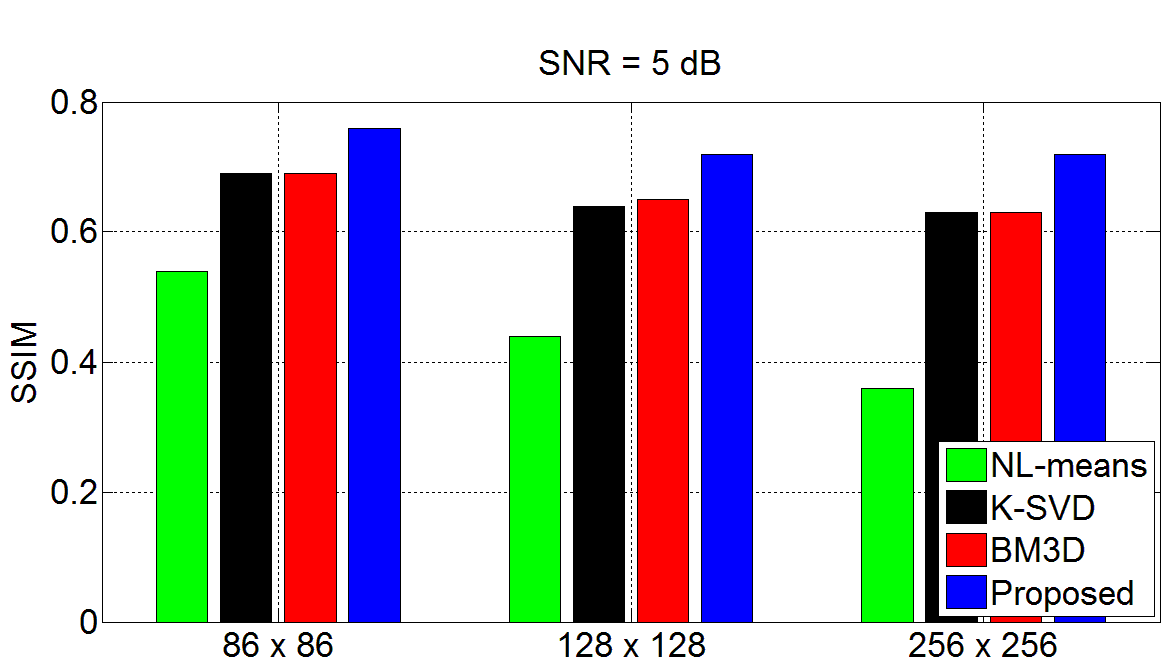}
		\end{subfigure}
		\begin{subfigure}{.24\textwidth}
			\centering
			\includegraphics[width=4cm, height=2.4cm]{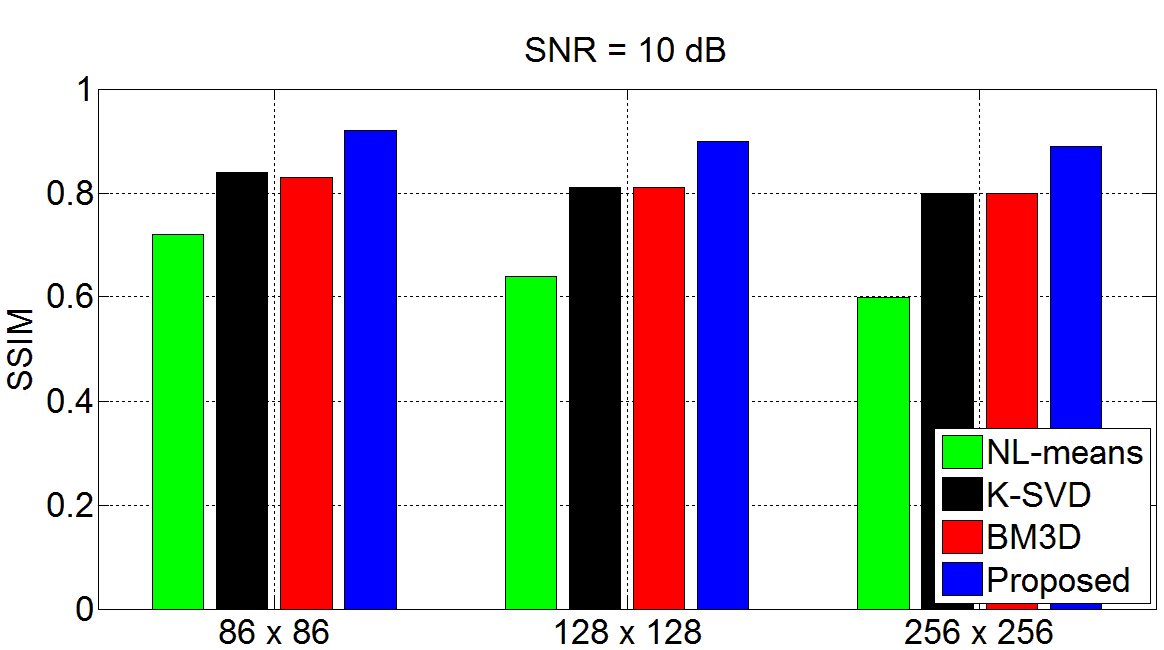}
		\end{subfigure}
	\end{subfigure} 
	\caption{Comparison of denoising $86 \times 86$, $128 \times 128$ and $256 \times 256$ size grayscale \textit{Mandrill} images in terms of PSNR (1st row) and SSIM (2nd row) using NL-means, K-SVD, BM3D and C2DF at SNR = -5, 0, 5 and 10 dB (left-to-right)}
	\label{fig:Sim_Res_6}
\end{figure*}
\begin{figure*}[h!]
	\begin{subfigure}{\textwidth}
		\centering
		\begin{subfigure}{.16\textwidth}
			\centering
			\includegraphics[width=2.9cm, height=2.8cm]{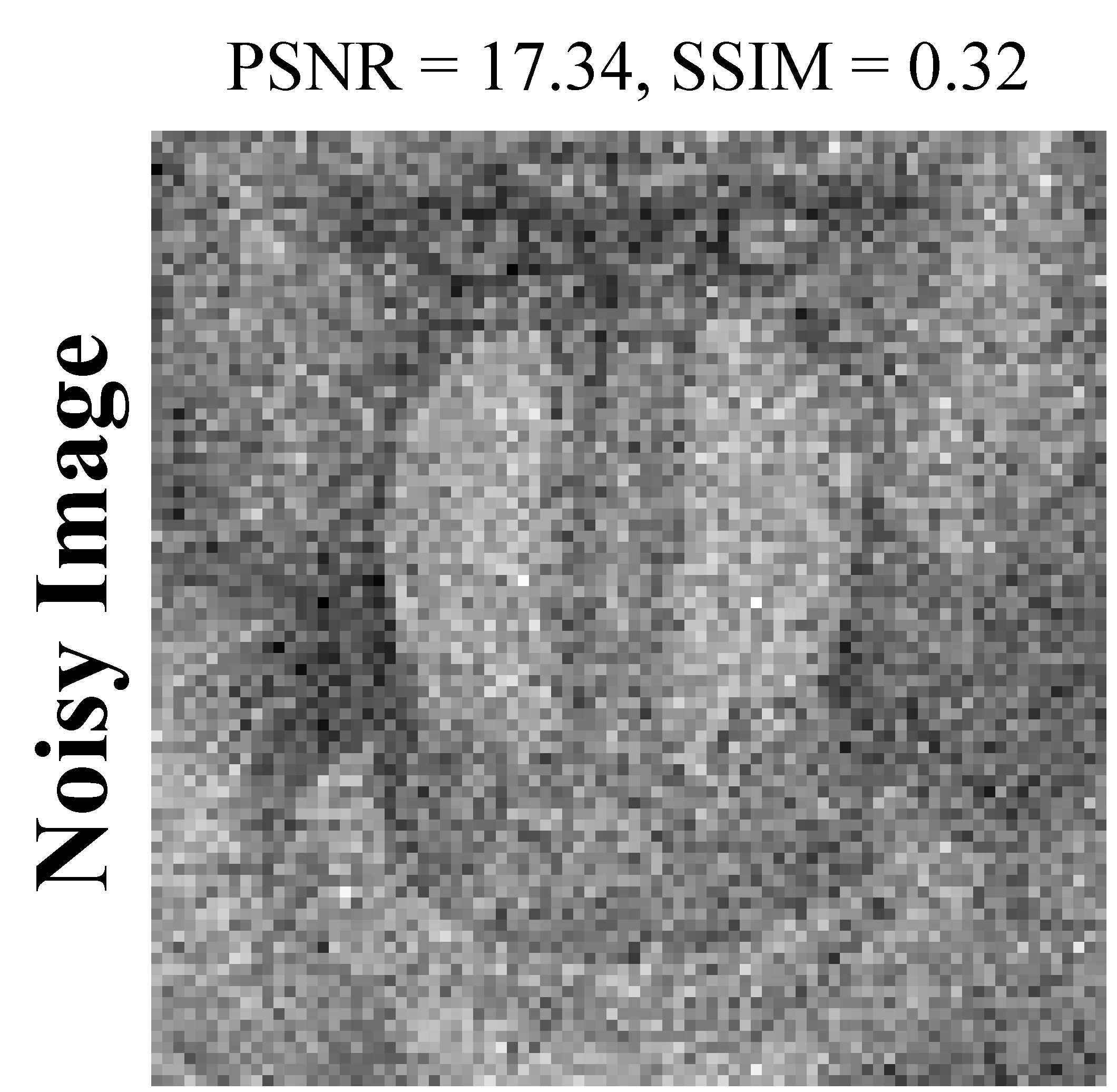}
		\end{subfigure}%
		\begin{subfigure}{.16\textwidth}
			\centering
			\includegraphics[width=2.9cm, height=2.8cm]{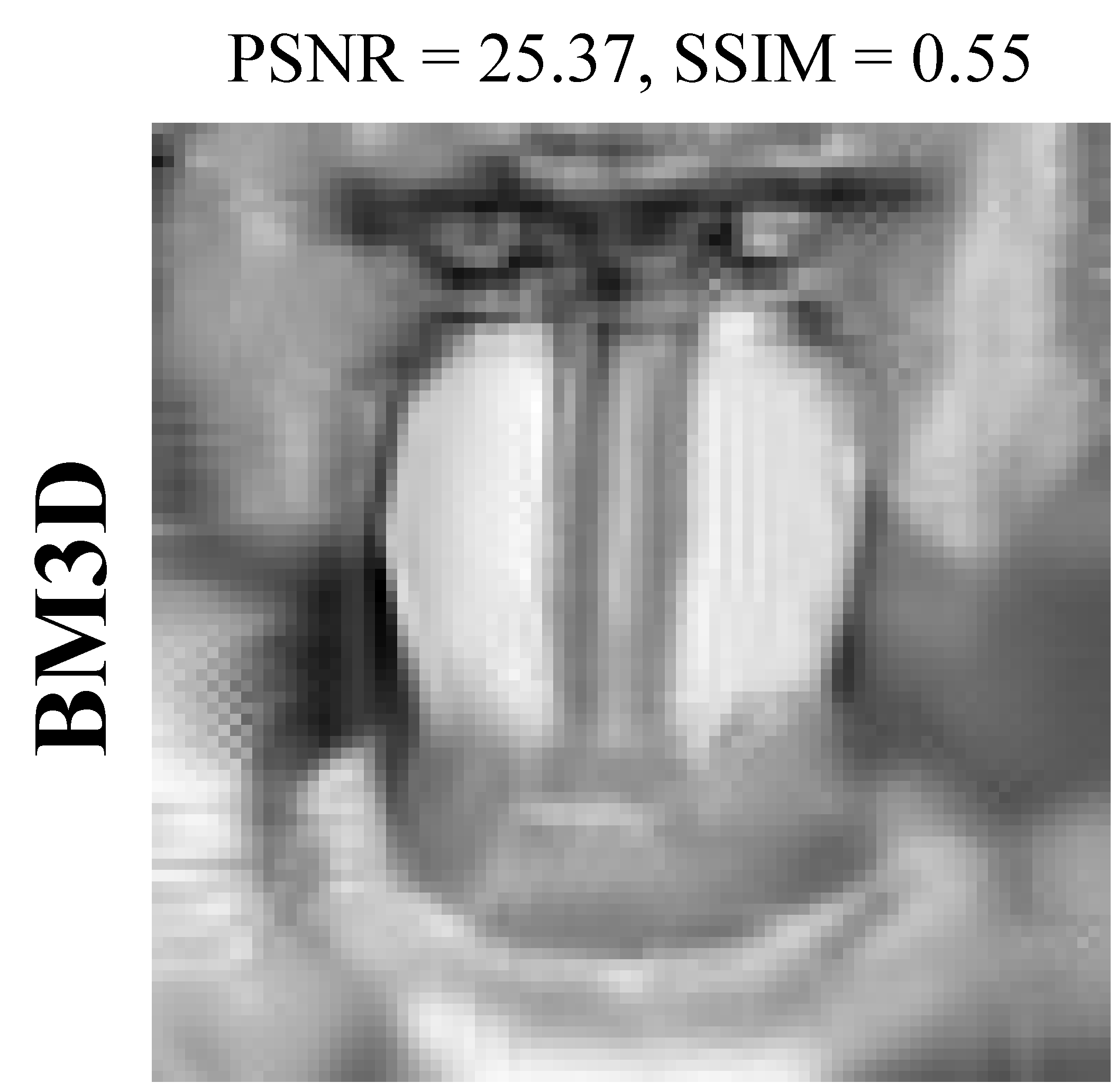}
		\end{subfigure}
		\begin{subfigure}{.16\textwidth}
			\centering
			\includegraphics[width=2.9cm, height=2.8cm]{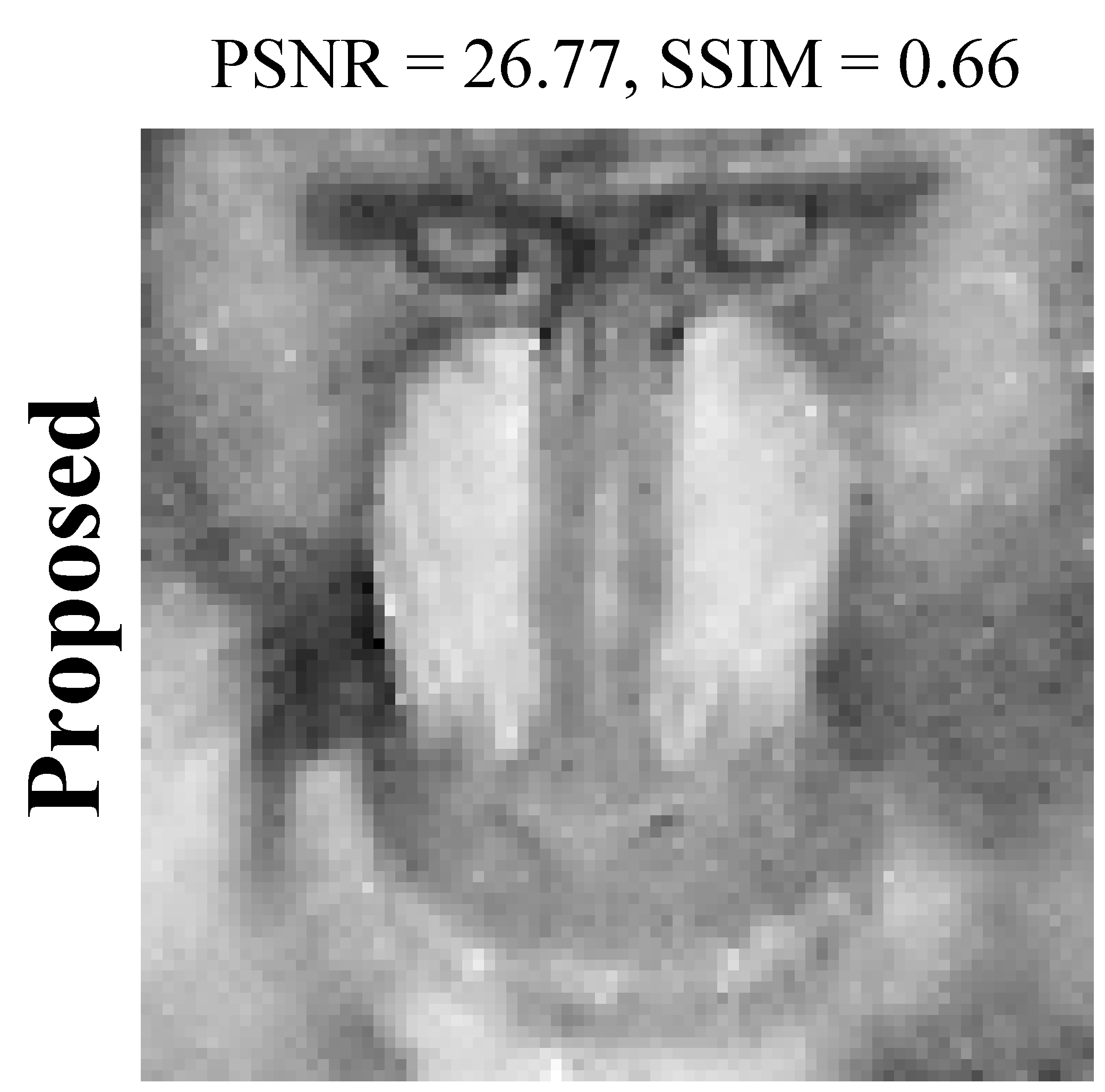}
		\end{subfigure}
		\begin{subfigure}{.16\textwidth}
			\centering
			\includegraphics[width=2.9cm, height=2.8cm]{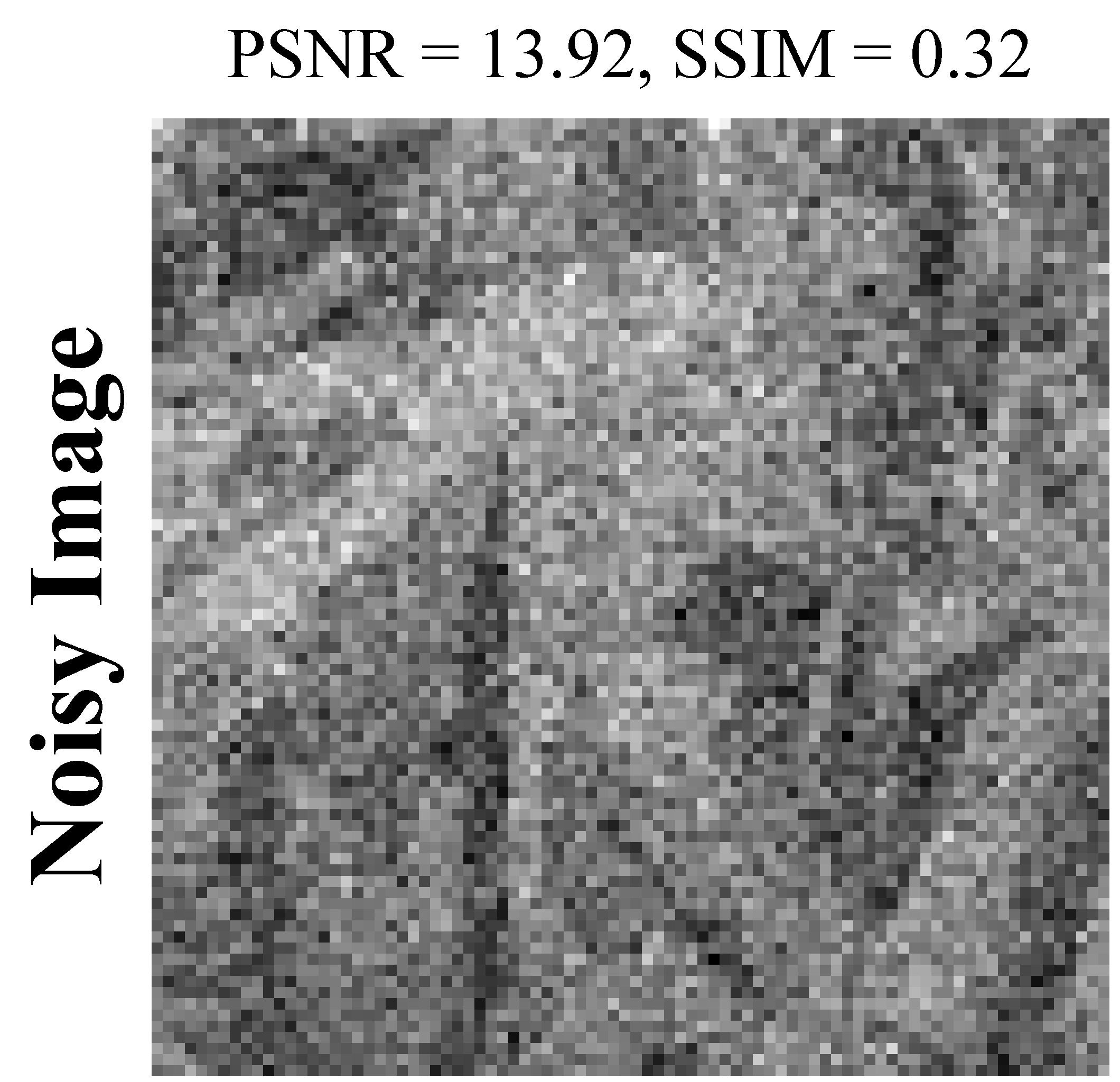}
		\end{subfigure}
		\begin{subfigure}{.16\textwidth}
			\centering
			\includegraphics[width=2.9cm, height=2.8cm]{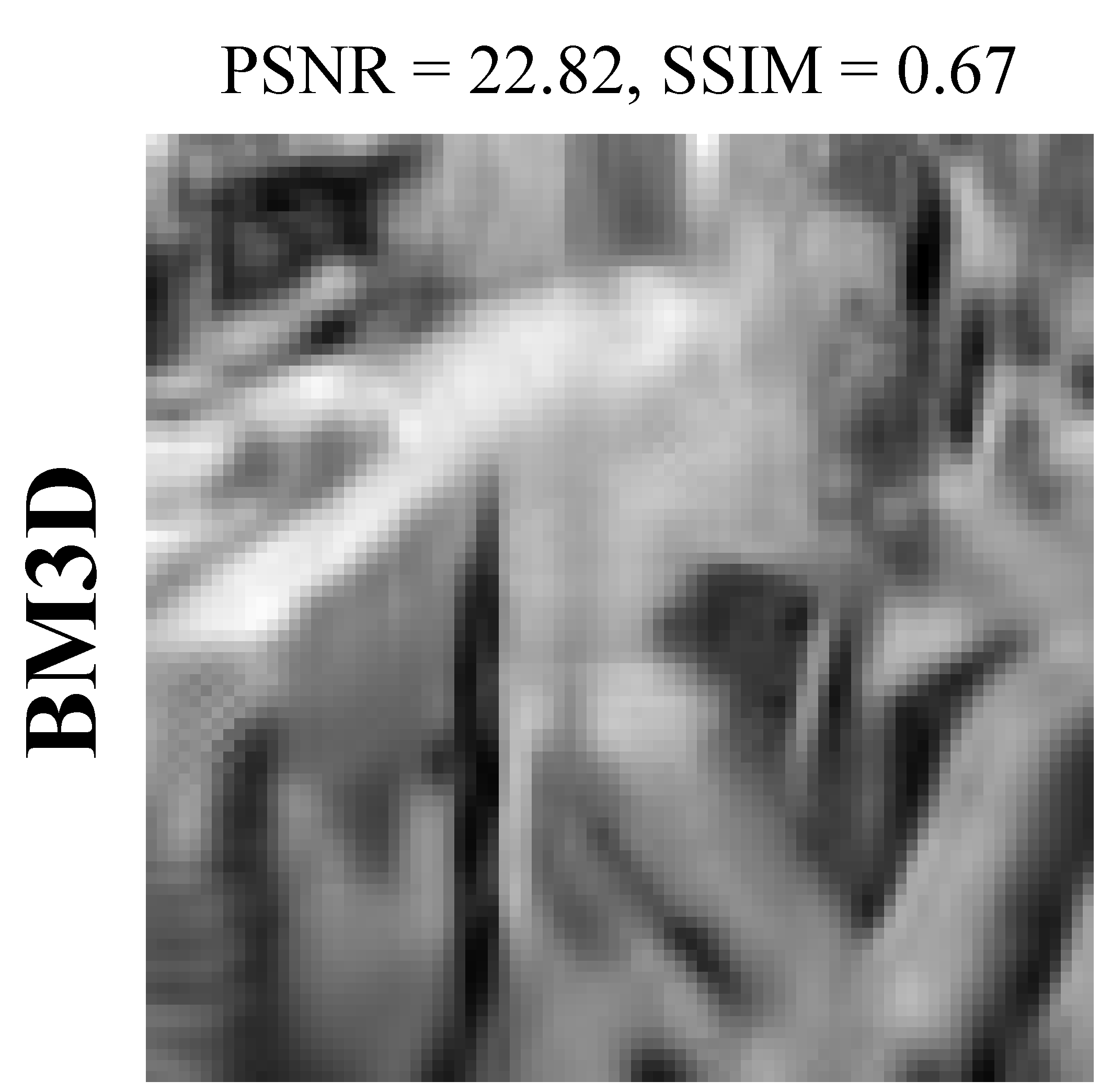}
		\end{subfigure}
		\begin{subfigure}{.16\textwidth}
			\centering
			\includegraphics[width=2.9cm, height=2.8cm]{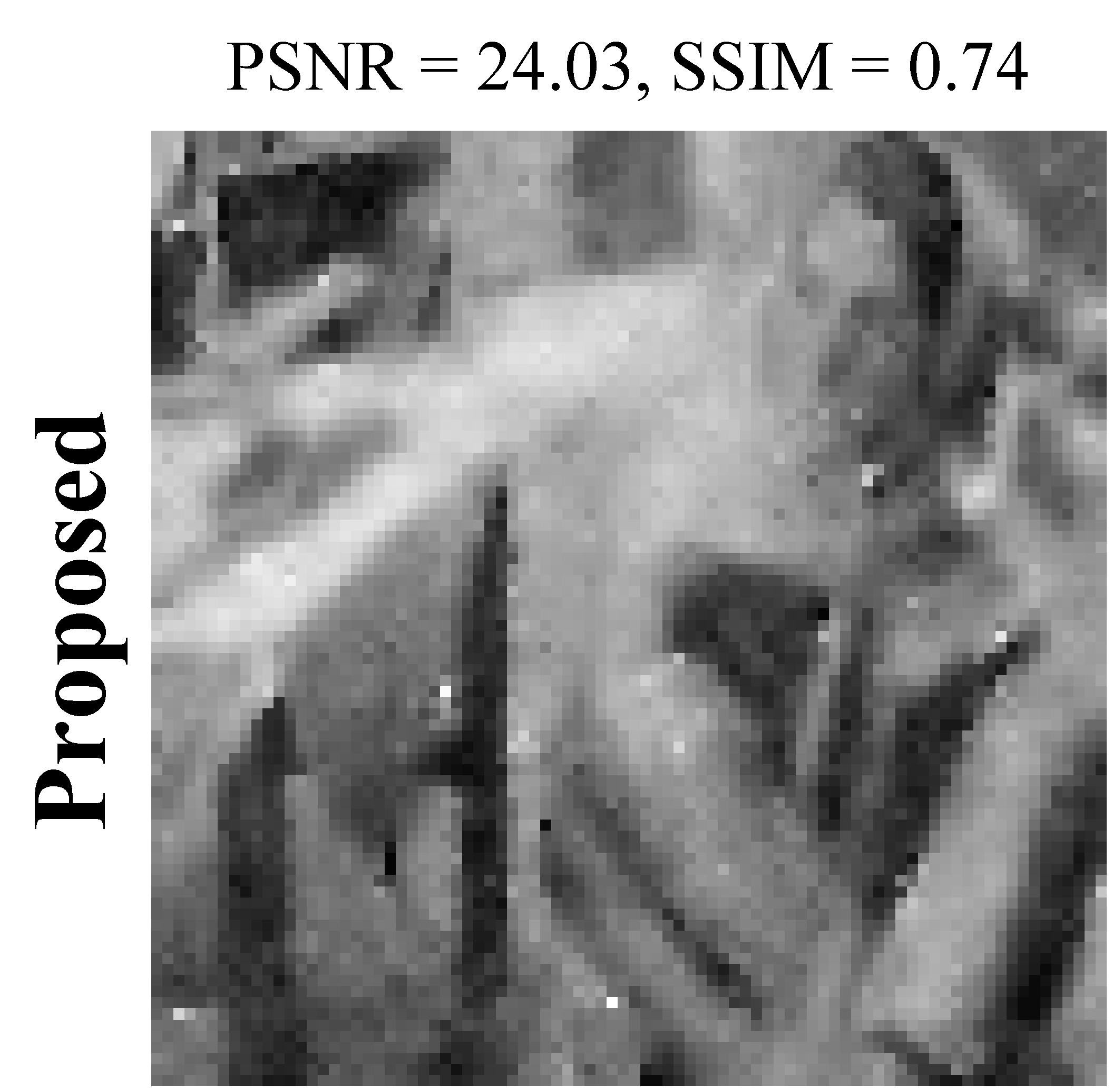}
		\end{subfigure}
	\end{subfigure} \\
	\begin{subfigure}{\textwidth}
		\centering
		\begin{subfigure}{.16\textwidth}
			\centering
			\includegraphics[width=2.9cm, height=2.8cm]{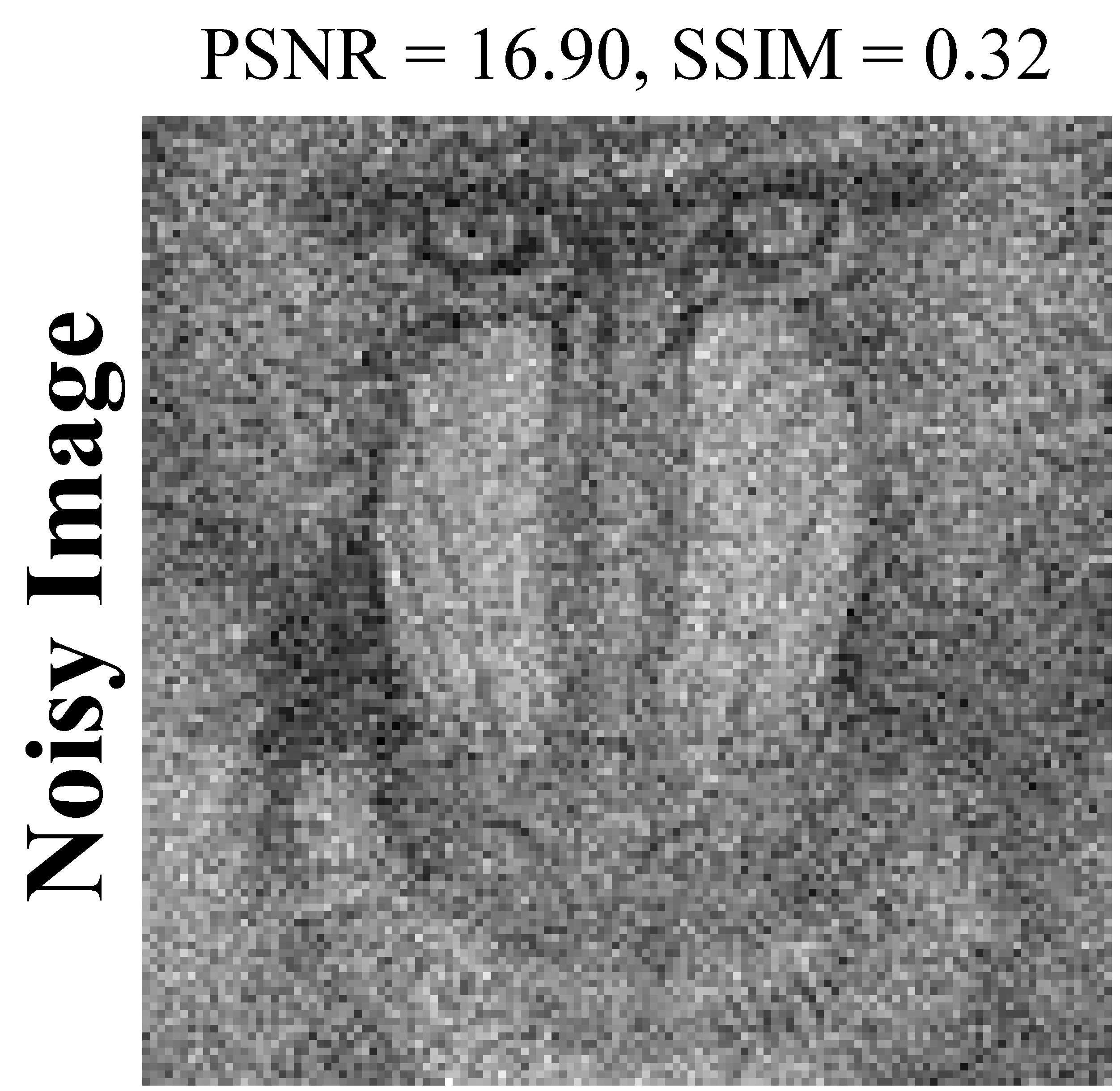}
		\end{subfigure}%
		\begin{subfigure}{.16\textwidth}
			\centering
			\includegraphics[width=2.9cm, height=2.8cm]{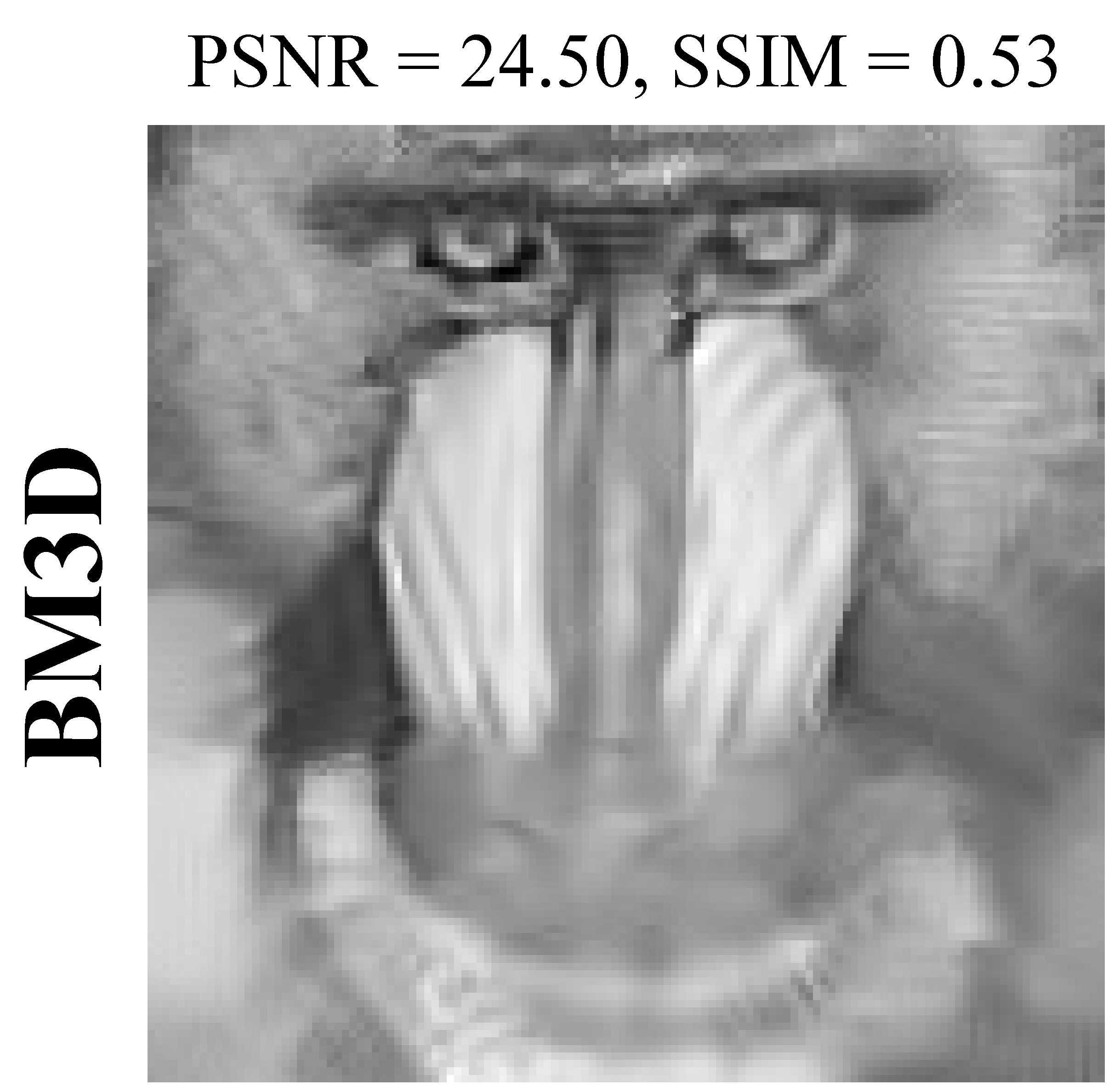}
		\end{subfigure}
		\begin{subfigure}{.16\textwidth}
			\centering
			\includegraphics[width=2.9cm, height=2.8cm]{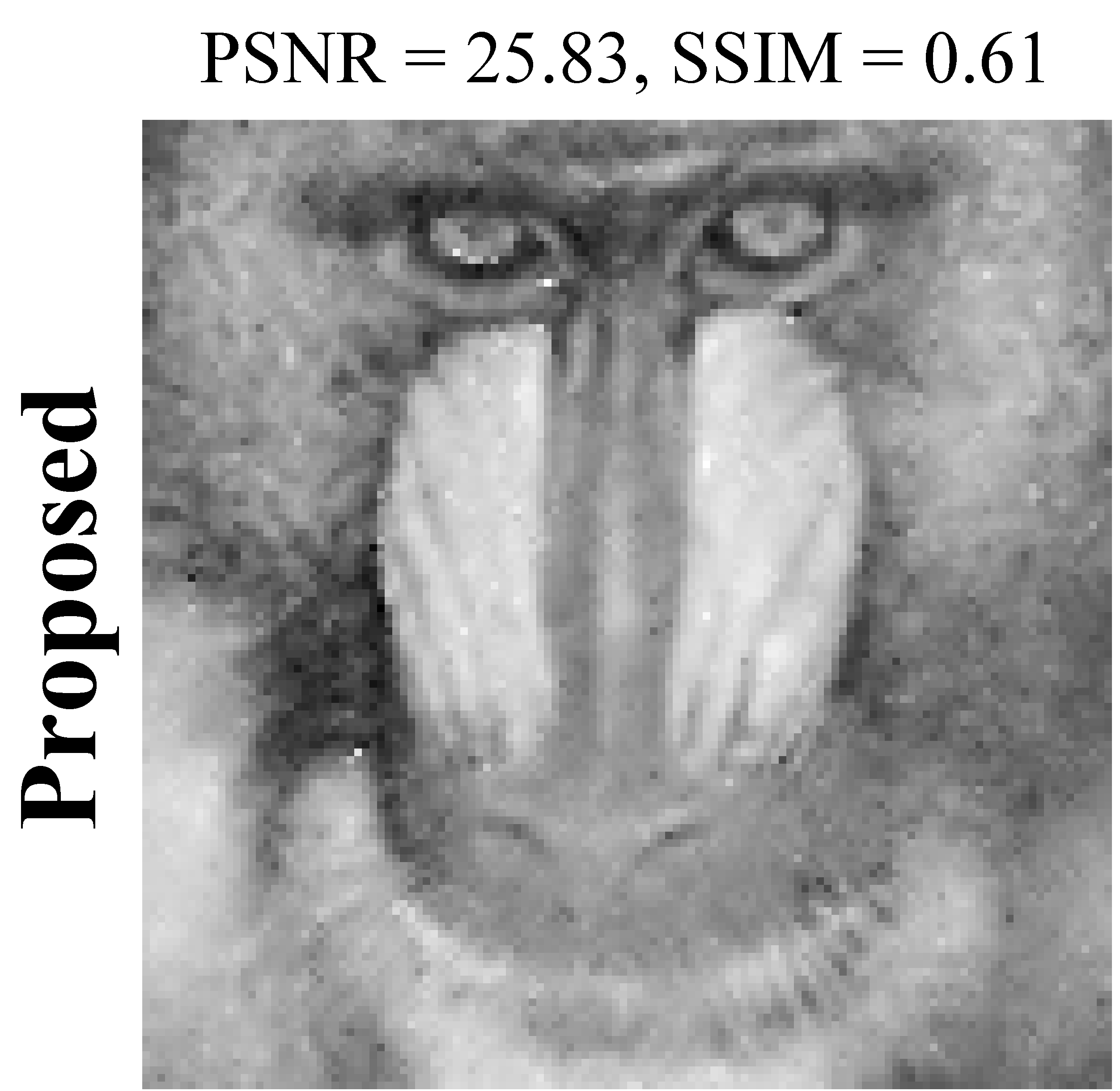}
		\end{subfigure}
		\begin{subfigure}{.16\textwidth}
			\centering
			\includegraphics[width=2.9cm, height=2.8cm]{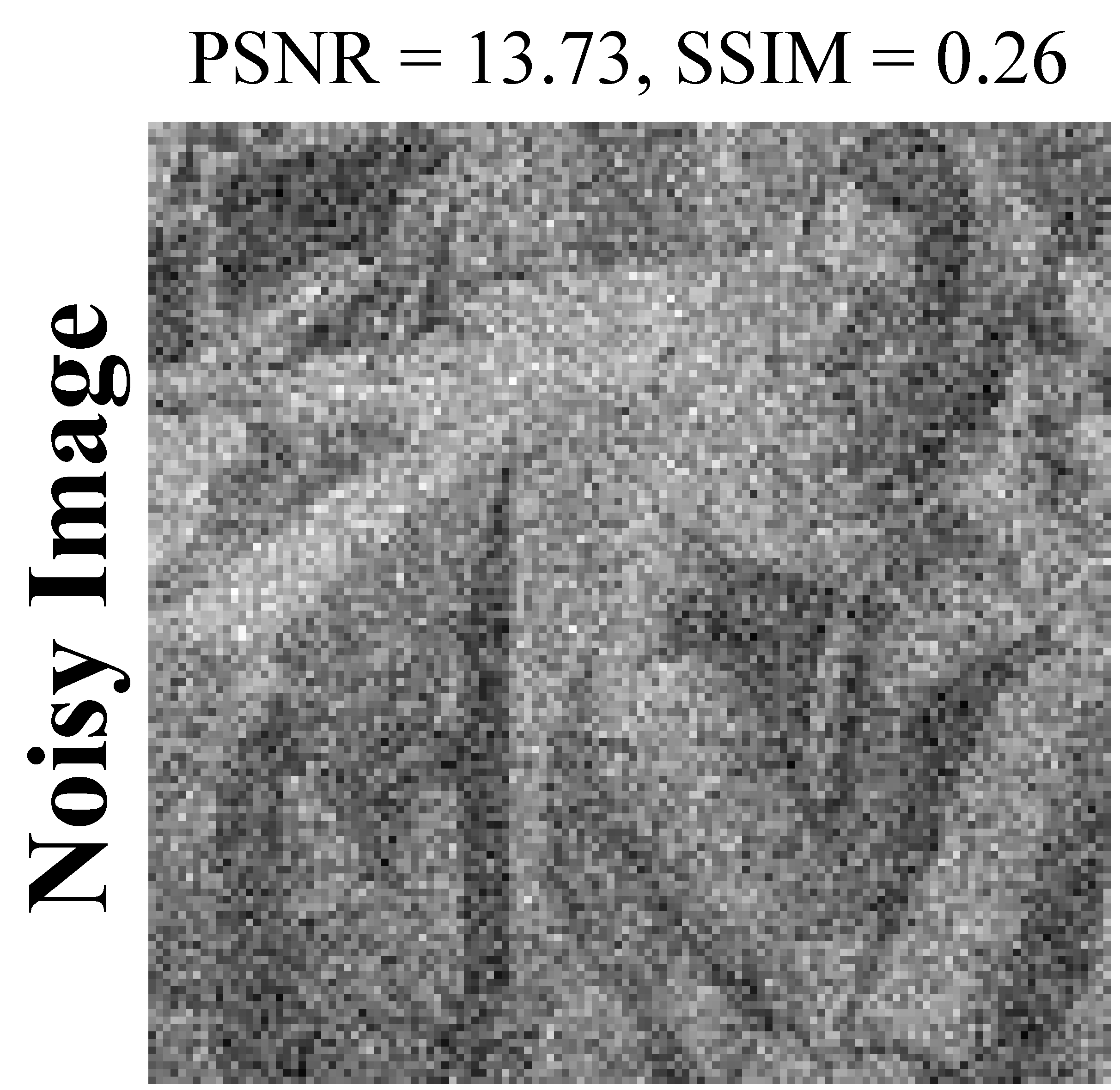}
		\end{subfigure}
		\begin{subfigure}{.16\textwidth}
			\centering
			\includegraphics[width=2.9cm, height=2.8cm]{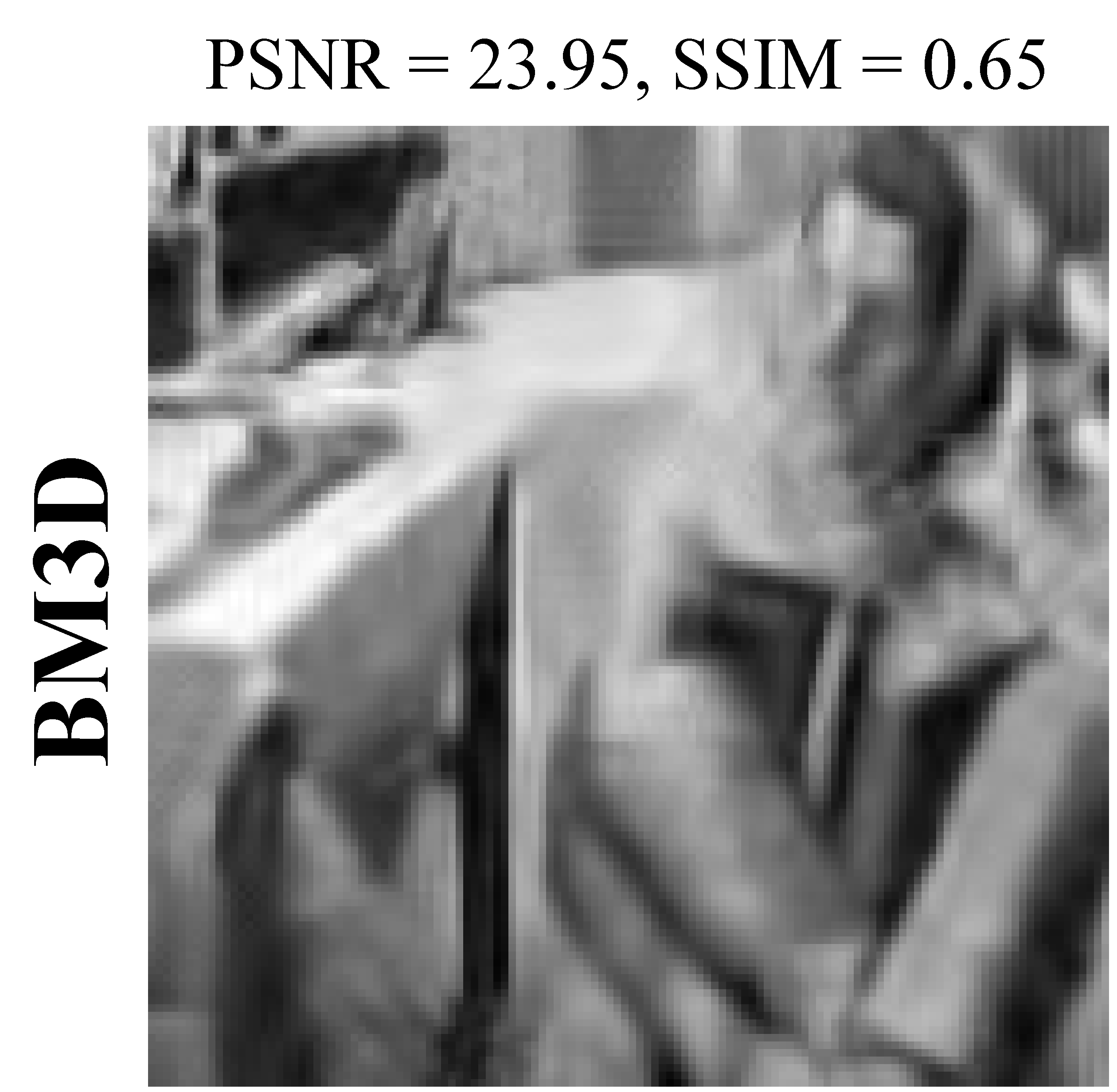}
		\end{subfigure}
		\begin{subfigure}{.16\textwidth}
			\centering
			\includegraphics[width=2.9cm, height=2.8cm]{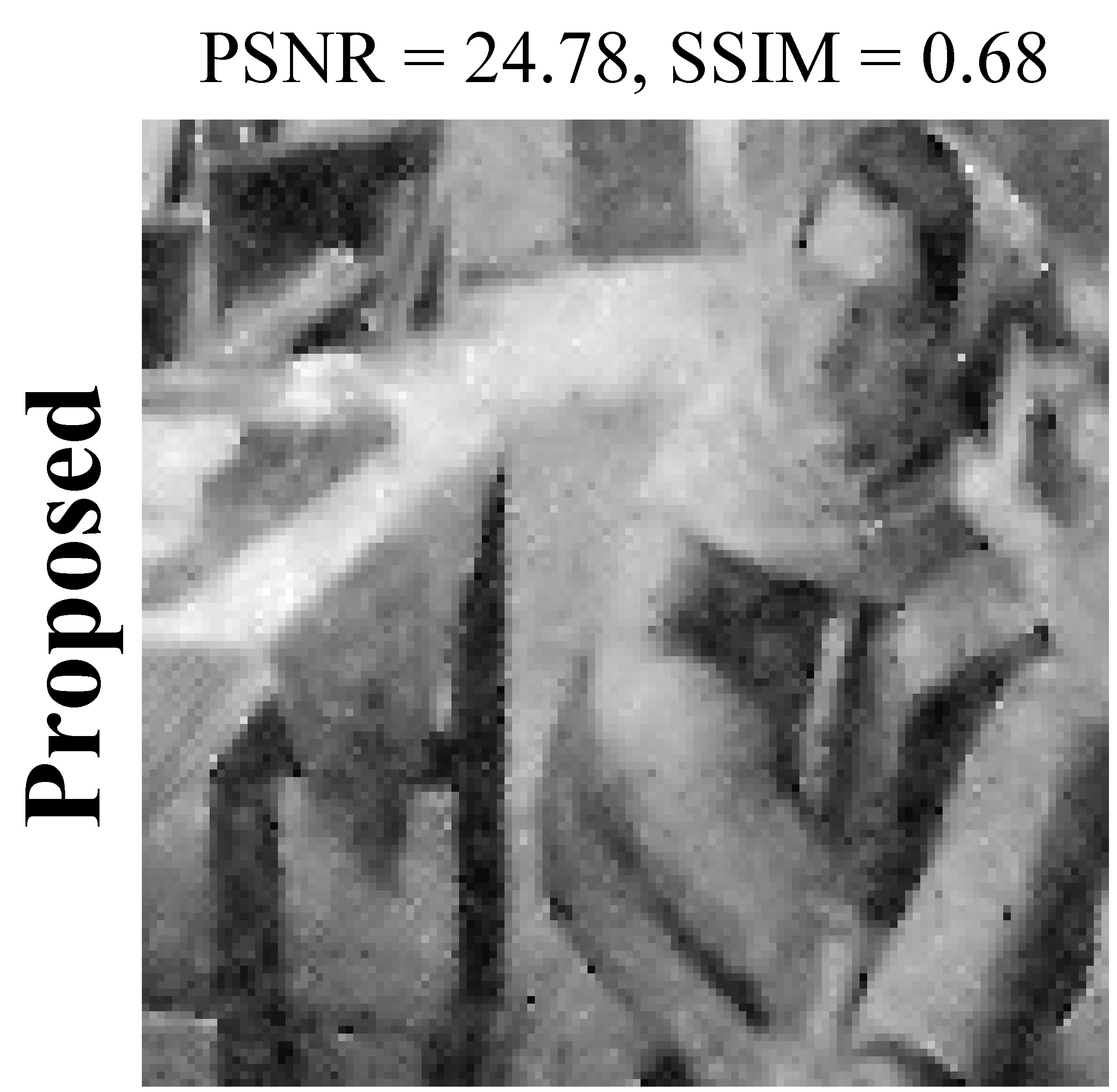}
		\end{subfigure}
	\end{subfigure}
	\caption{Denoising results of \textit{Barbara} and \textit{Mandrill} at SNR = 0 dB: 1st row $86 \times 86$, 2nd row $128 \times 128$ size images}
	\label{fig:Sim_Res_8}
\end{figure*}

For a detailed comparison of the denoising performance over various grayscale images widely used in image processing community, and over a range of noise levels, we summarize the PSNR ans SSIM results in Table \ref{table1}. Since BM3D outperforms both NL-means and K-SVD, therefore we only compare our results with that of BM3D in this table. Moreover in Table \ref{table2} and \ref{table3}, we also present the results of denoising grayscale texture and aerial images from SIPI database to prove that our algorithm can be applied globally to any image. From the provided tables, it is clear that our proposed image denoising algorithm outperforms the state-of-the-art algorithm in each scenario and has proven itself to be a much efficient algorithm for image denoising.

\subsection{Various Resolution Images}
\label{Various_Resolution_Images}
In this section, we present the denoising comparison over resolution of different sizes as opposed to the traditional comparison methods where only single image size is used throughout the experimentations. This is very important to validate the effectiveness of any algorithm, i.e., how better can any algorithm tackle noise at different image sizes. For this purpose, we take grayscale $86 \times 86$, $128 \times 128$ and $256 \times 256$ size \textit{Barabara} and \textit{Mandrill} images, and denoise them using BM3D and our method at different noise levels.

The results of denoising \textit{Mandrill} image at different resolutions and at SNR = -5, 0, 5 and 10 dB is shown in Fig.~\ref{fig:Sim_Res_6}. This figure shows that C2DF not only outperforms state-of-the-art denoising algorithms at large image size but is also capable of removing noise components effectively at small image sizes. The first row in this figure corresponds to the resultant PSNR achieved by denoising at SNR = -5, 0, 5 and 10 dB, while the second row shows the corresponding SSIM results. 

A pictorial representation of the denoised images has been shown in Fig. \ref{fig:Sim_Res_8} to depict the limitation of existing algorithms. As discussed already, even the state-of-the-art method like BM3D tend to blur out the images especially at low resolution images and high noise. For instance, observe the first row of this figure where the facial details of the \textit{Mandrill} image has been blurred out while the details are preserved in results by our method. A similar problem is detected for the \textit{Barbara} where, e.g., the hands, facial details and the table legs get disappeared but exist in our denoised results. The results for $256 \times 256$ has already been shown previously in Fig. \ref{fig:Sim_Res_1}. As can be clearly concluded from these figures, the proposed method shows an effective performance in all the sceneries and leads the PSNR and SSIM points table by quite a good margin.

\begin{figure*}[t]
	\begin{subfigure}{\textwidth}
		\centering
		\begin{subfigure}{.13\textwidth}
			\centering
			\includegraphics[width=2.5cm, height=2.4cm]{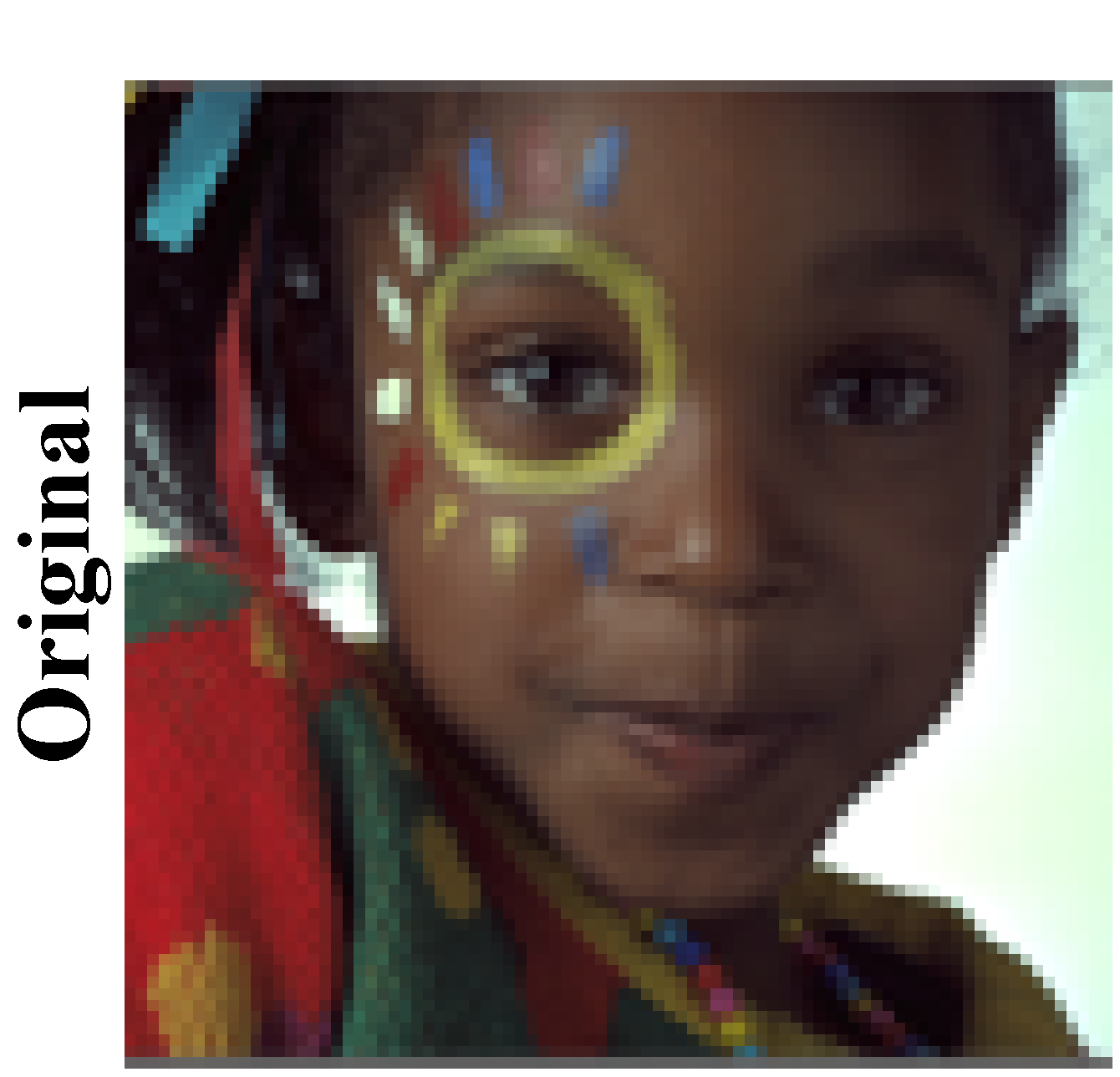}
		\end{subfigure}%
		\begin{subfigure}{.13\textwidth}
			\centering
			\includegraphics[width=2.5cm, height=2.4cm]{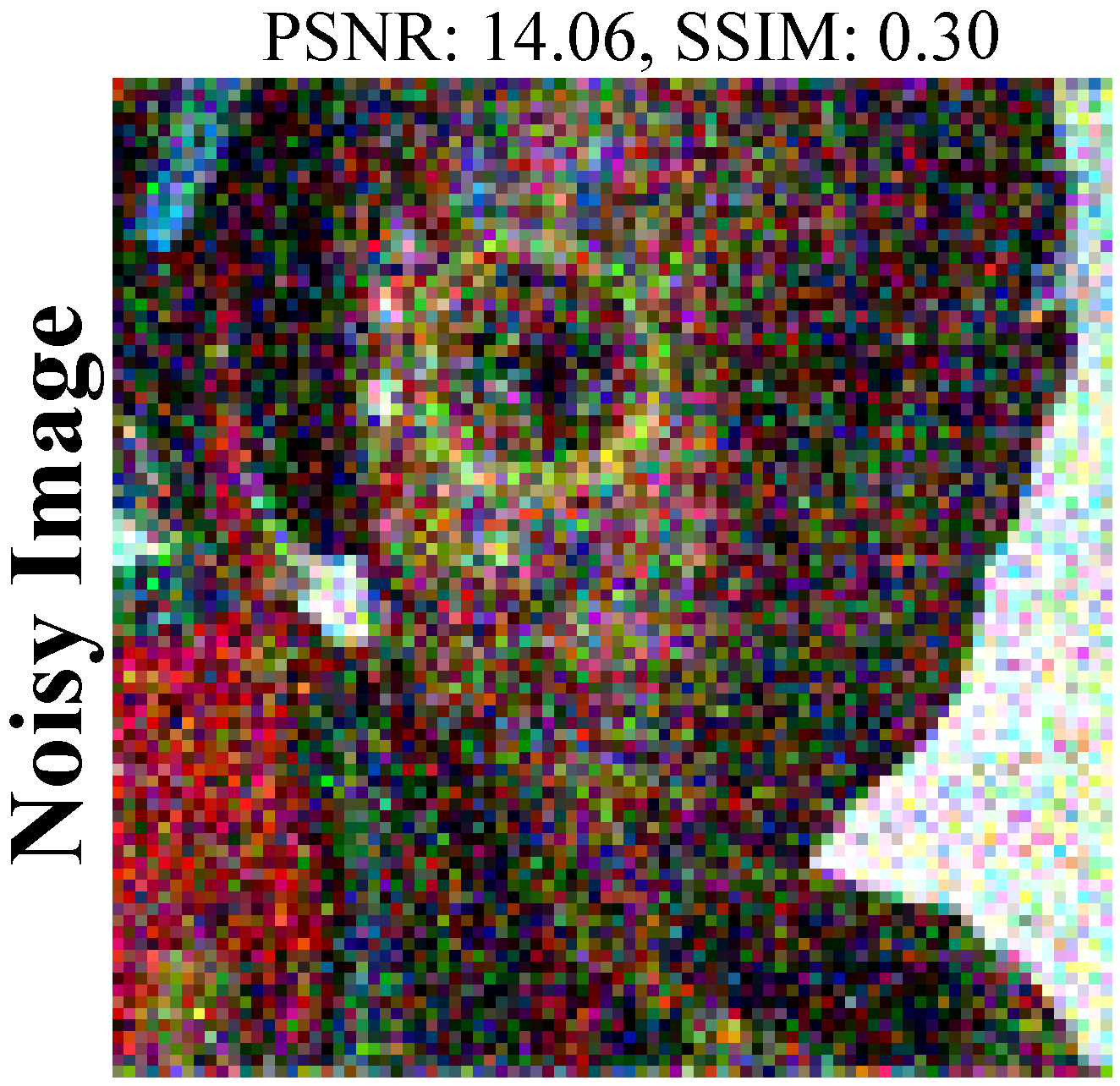}
		\end{subfigure}
		\begin{subfigure}{.13\textwidth}
			\centering
			\includegraphics[width=2.5cm, height=2.4cm]{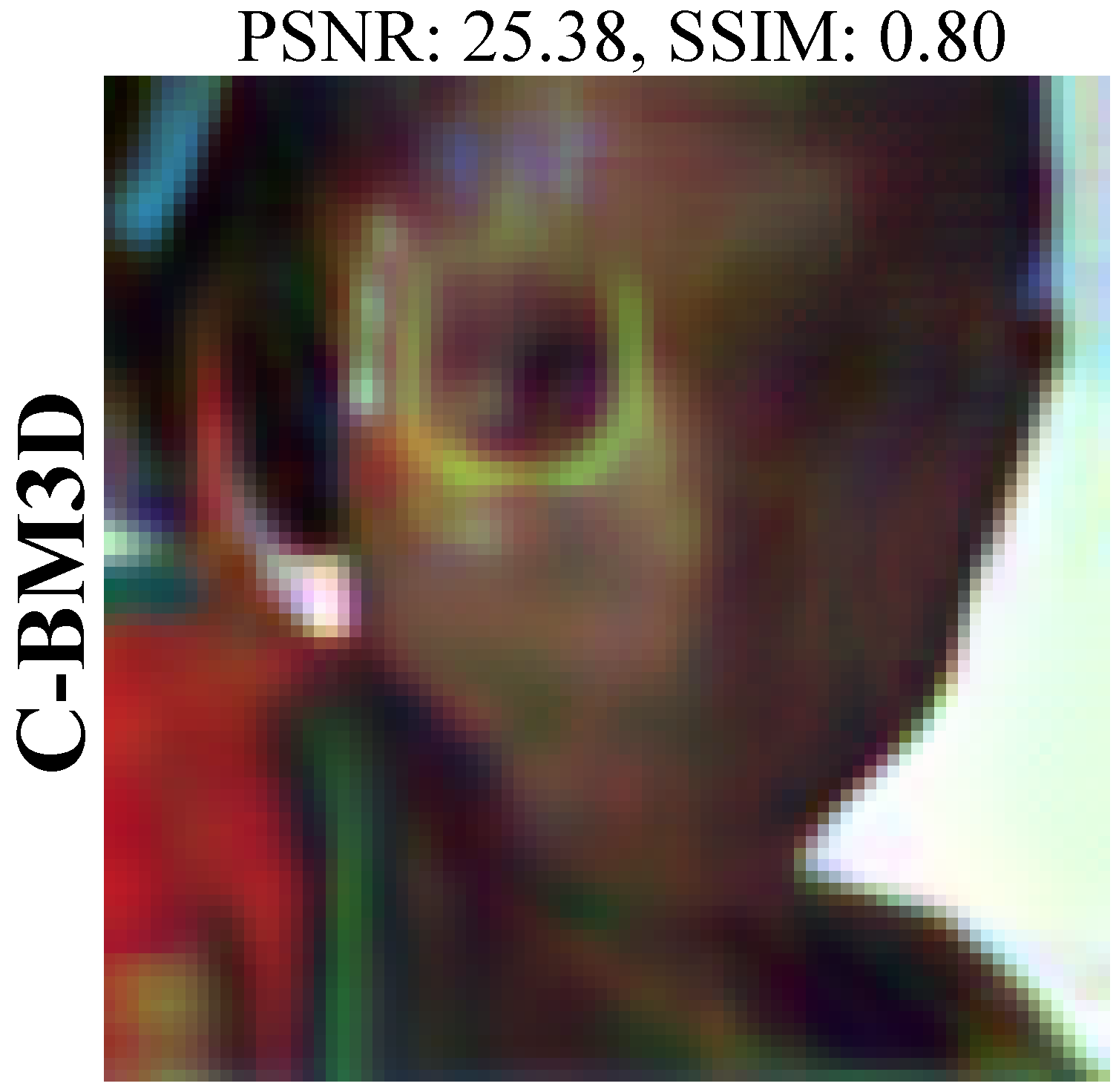}
		\end{subfigure}
		\begin{subfigure}{.13\textwidth}
			\centering
			\includegraphics[width=2.5cm, height=2.4cm]{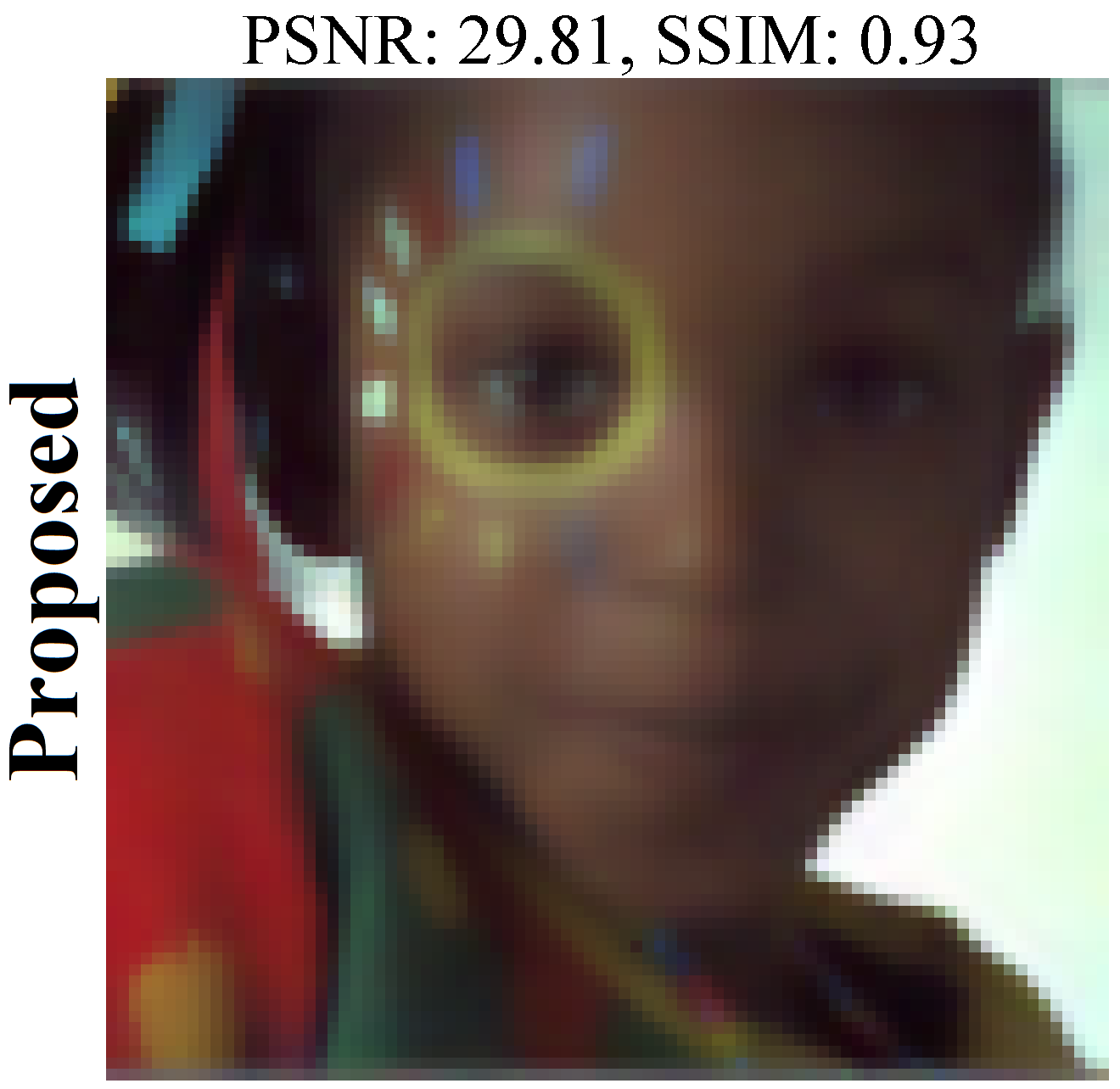}
		\end{subfigure}
		\begin{subfigure}{.13\textwidth}
			\centering
			\includegraphics[width=2.5cm, height=2.4cm]{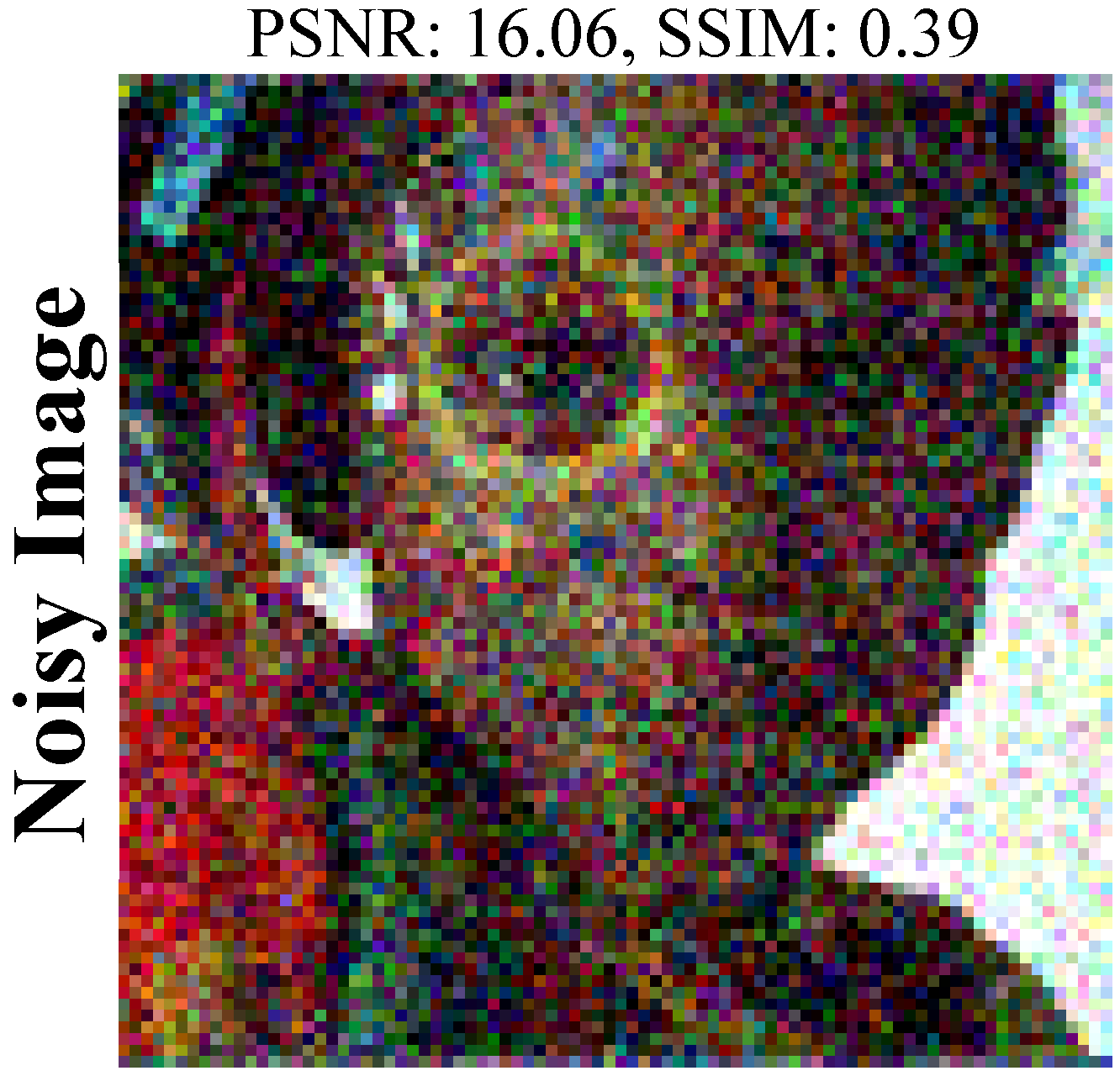}
		\end{subfigure}
		\begin{subfigure}{.13\textwidth}
			\centering
			\includegraphics[width=2.5cm, height=2.4cm]{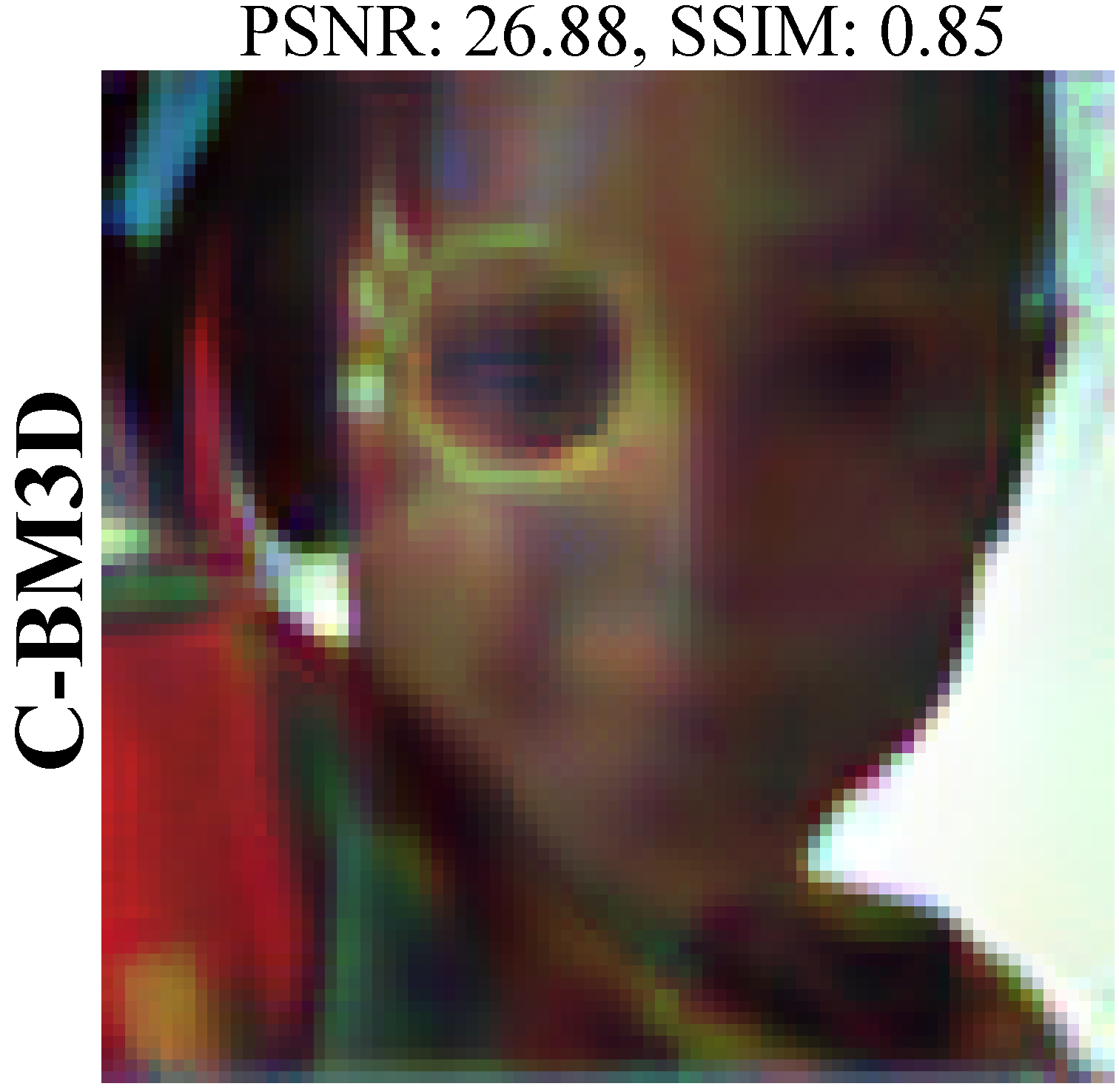}
		\end{subfigure}
		\begin{subfigure}{.13\textwidth}
			\centering
			\includegraphics[width=2.5cm, height=2.4cm]{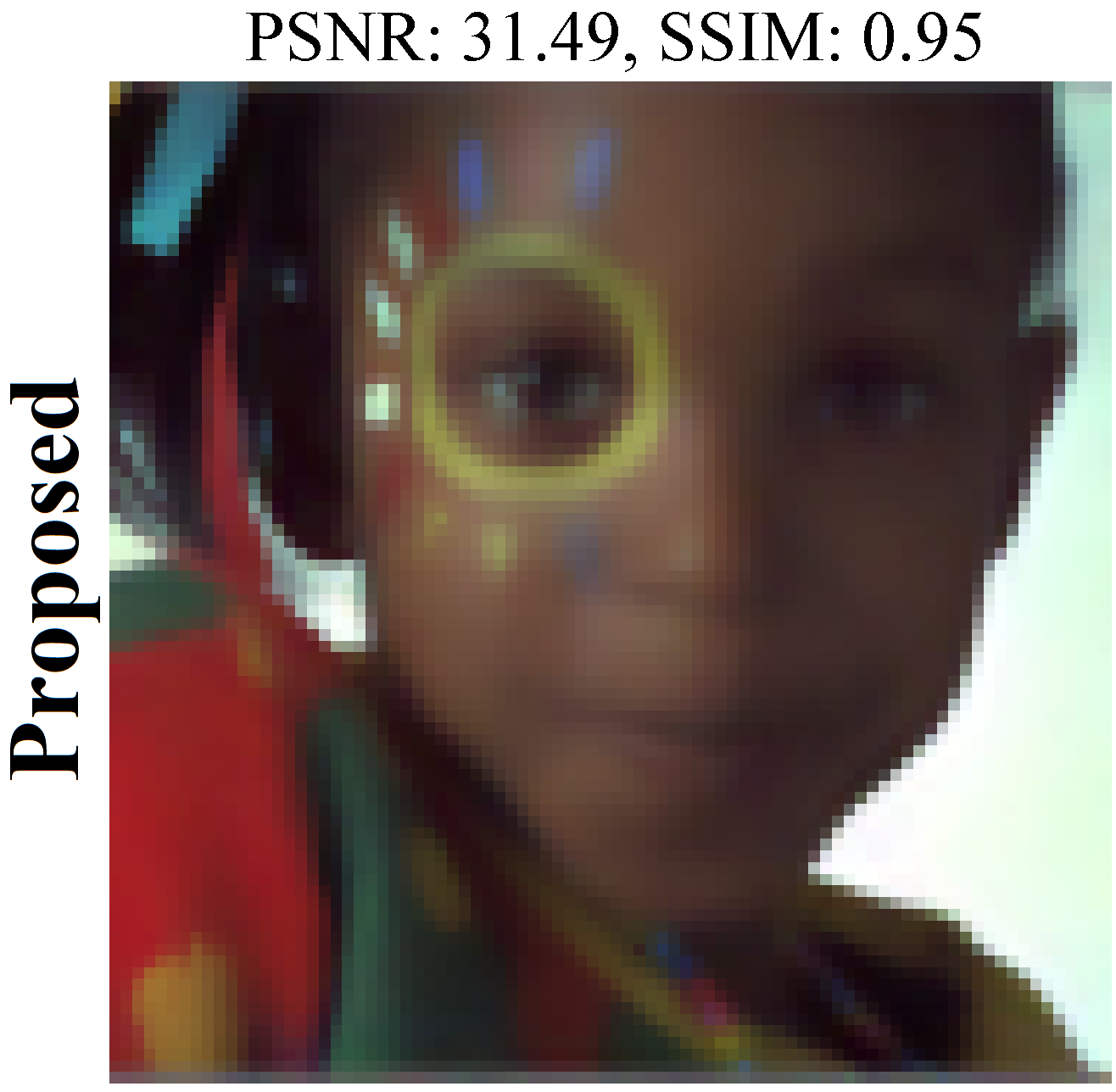}
		\end{subfigure}
	\end{subfigure}\\	
	\begin{subfigure}{\textwidth}
		\centering
		\begin{subfigure}{.13\textwidth}
			\centering
			\includegraphics[width=2.5cm, height=2.4cm]{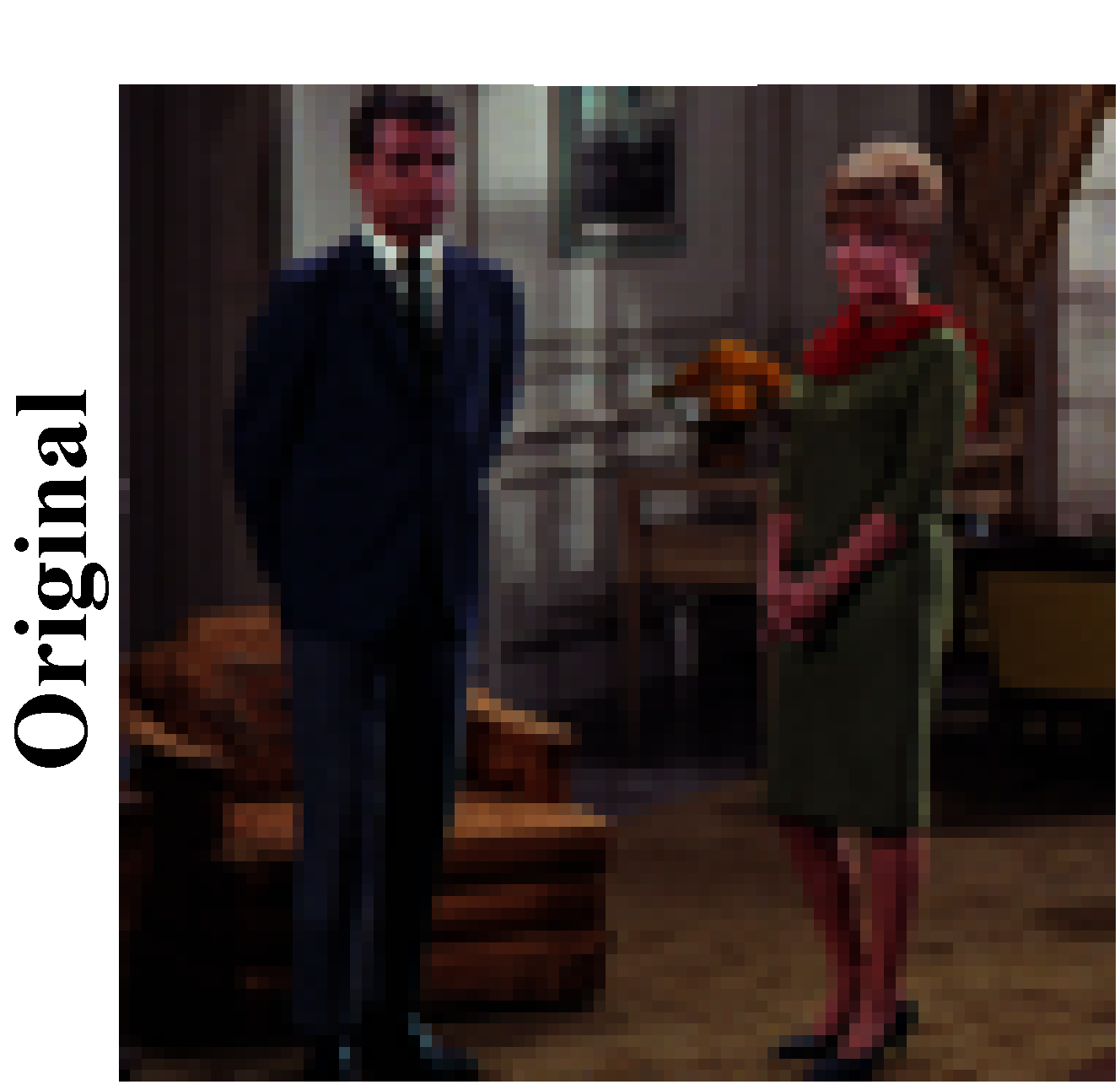}
		\end{subfigure}%
		\begin{subfigure}{.13\textwidth}
			\centering
			\includegraphics[width=2.5cm, height=2.4cm]{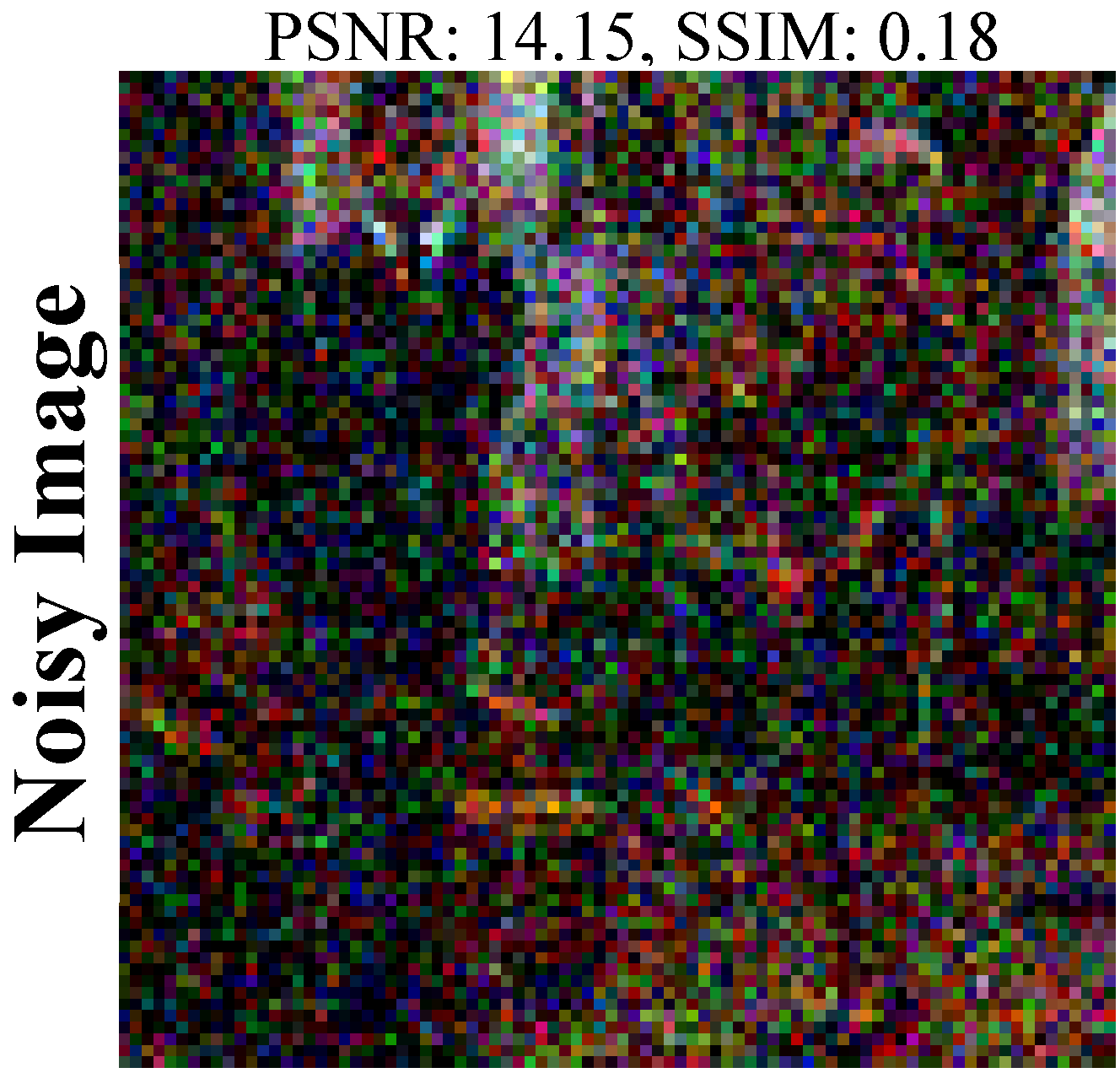}
		\end{subfigure}
		\begin{subfigure}{.13\textwidth}
			\centering
			\includegraphics[width=2.5cm, height=2.4cm]{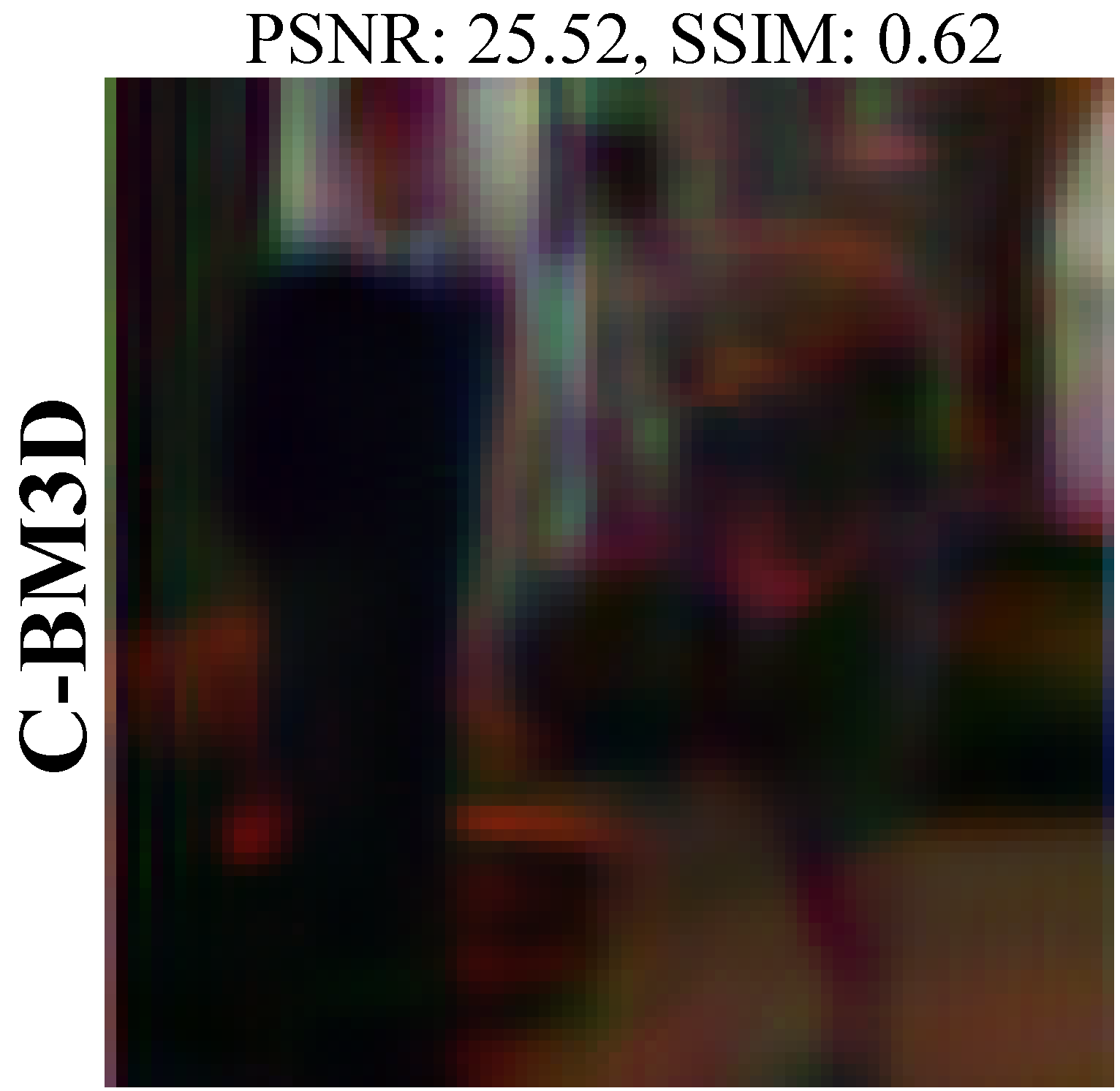}
		\end{subfigure}
		\begin{subfigure}{.13\textwidth}
			\centering
			\includegraphics[width=2.5cm, height=2.4cm]{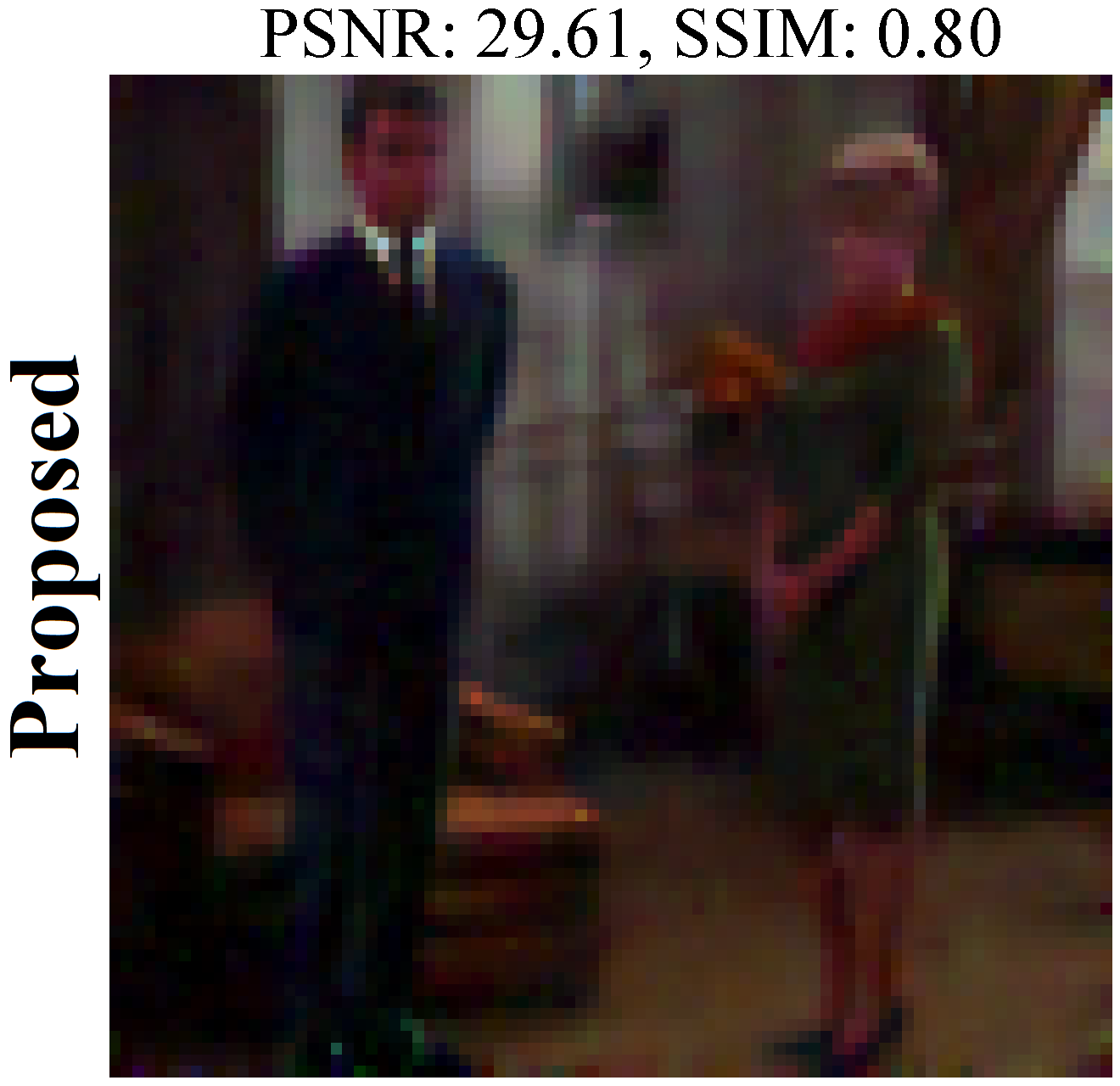}
		\end{subfigure}
		\begin{subfigure}{.13\textwidth}
			\centering
			\includegraphics[width=2.5cm, height=2.4cm]{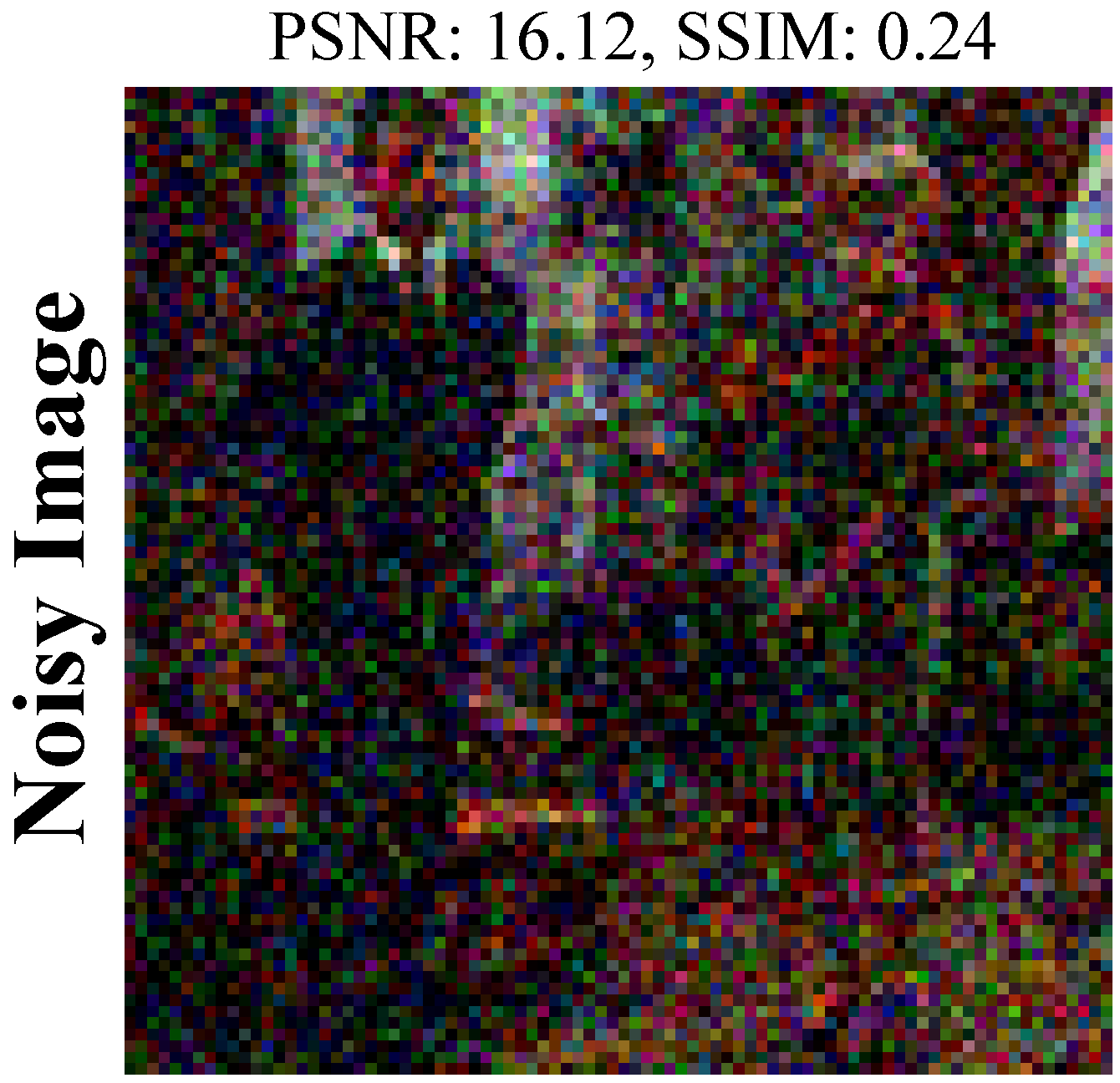}
		\end{subfigure}
		\begin{subfigure}{.13\textwidth}
			\centering
			\includegraphics[width=2.5cm, height=2.4cm]{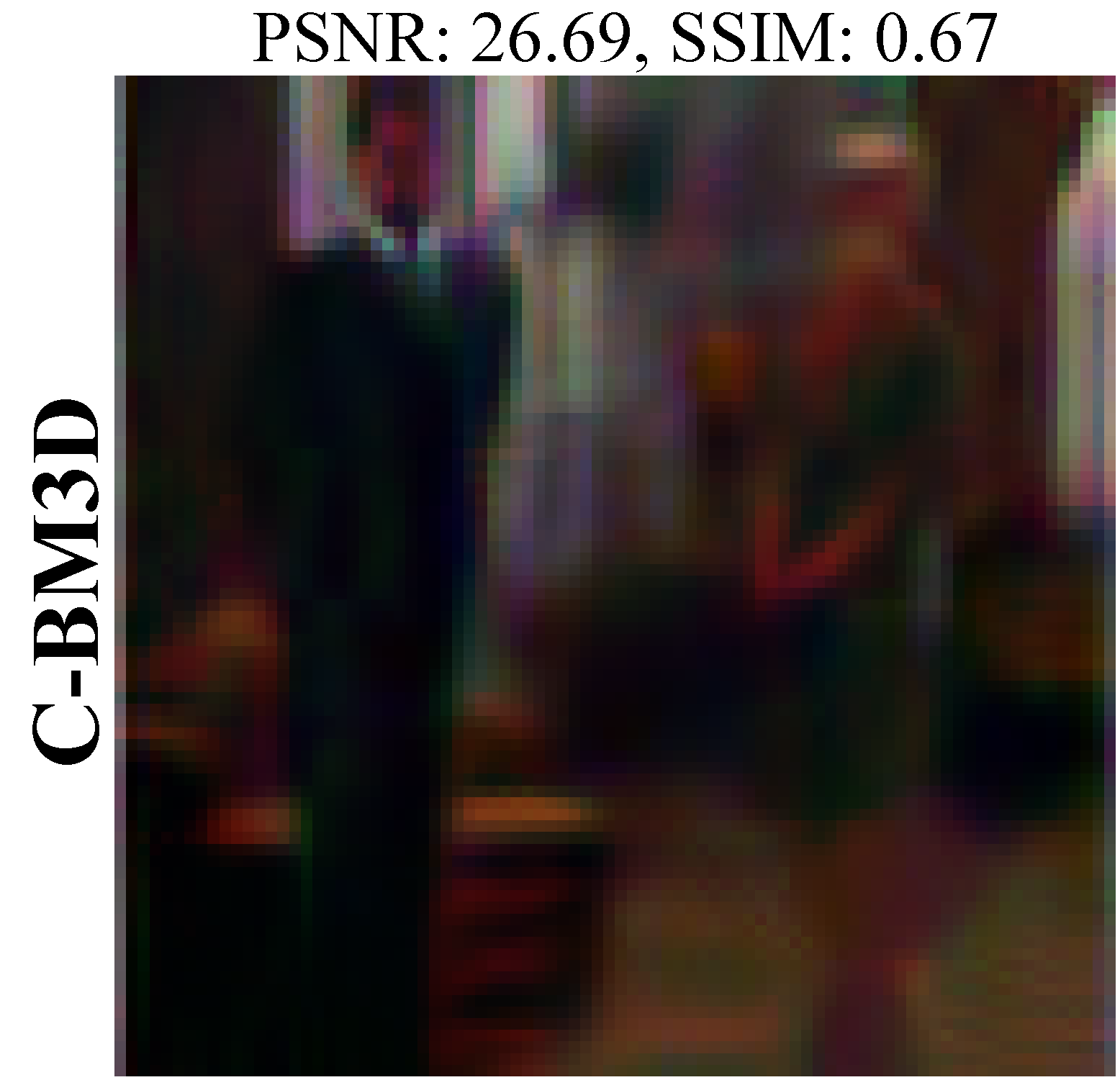}
		\end{subfigure}
		\begin{subfigure}{.13\textwidth}
			\centering
			\includegraphics[width=2.5cm, height=2.4cm]{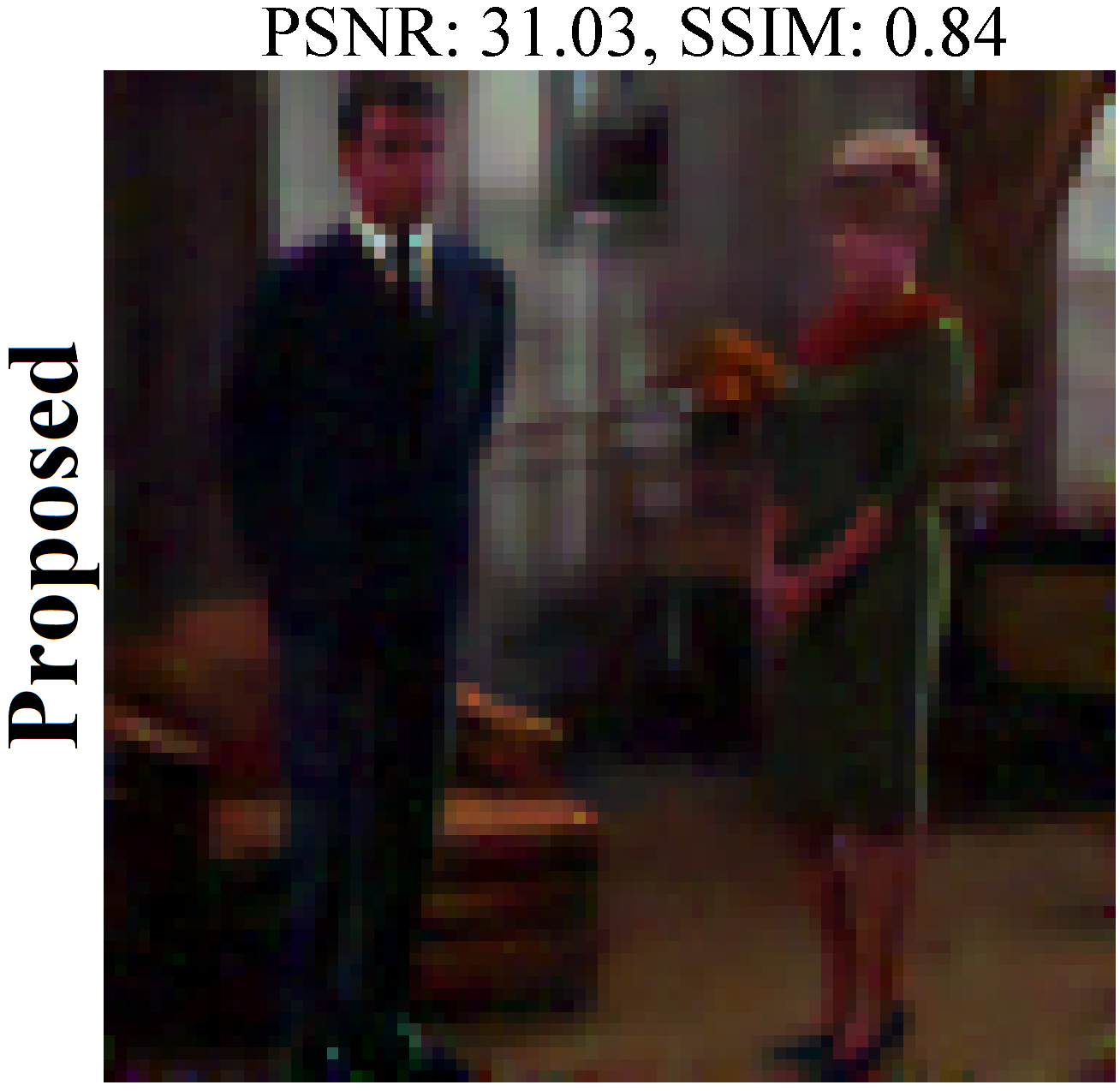}
		\end{subfigure}
	\end{subfigure}\\	
	\begin{subfigure}{\textwidth}
		\centering
		\begin{subfigure}{.13\textwidth}
			\centering
			\includegraphics[width=2.5cm, height=2.4cm]{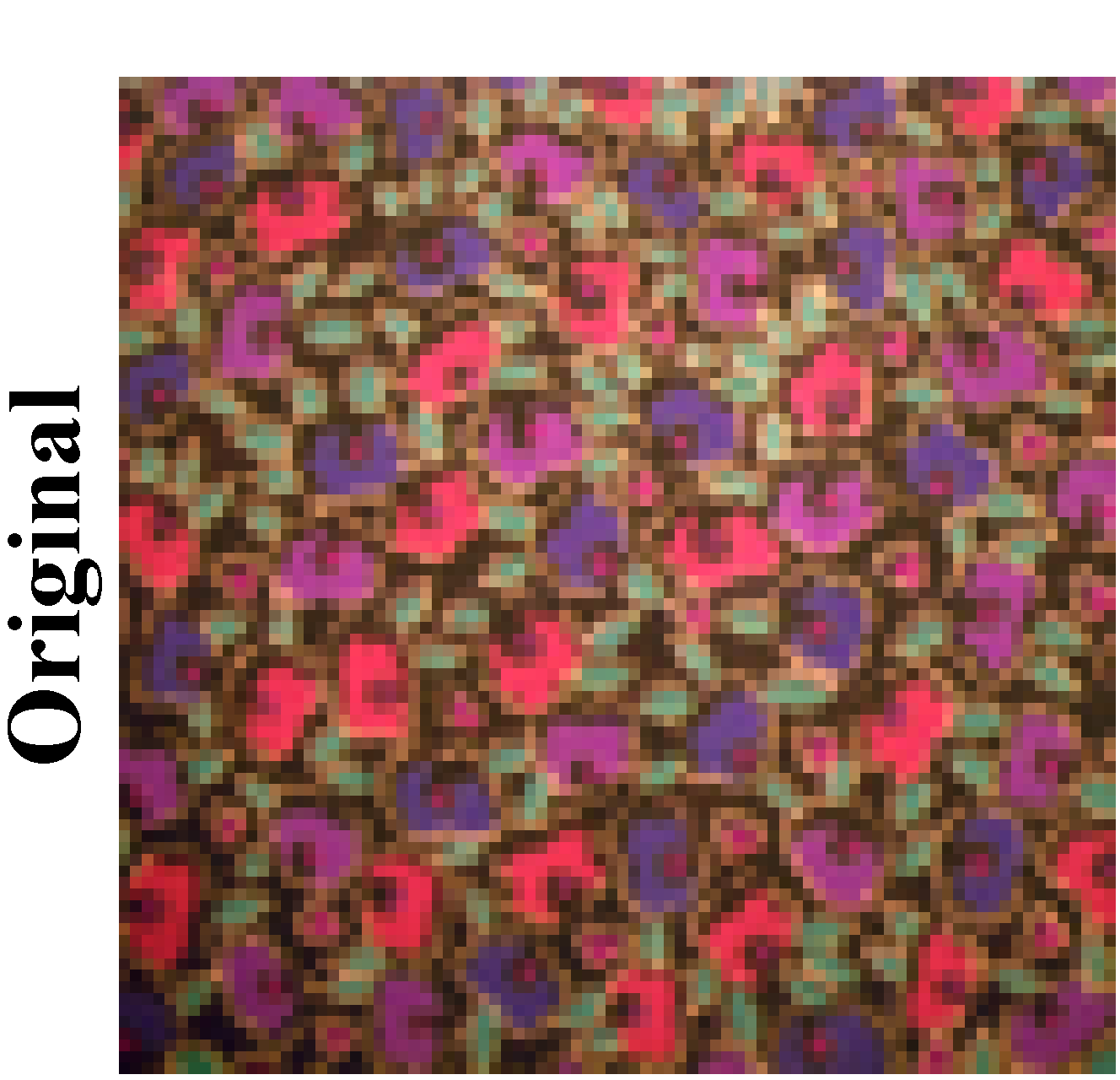}
		\end{subfigure}%
		\begin{subfigure}{.13\textwidth}
			\centering
			\includegraphics[width=2.5cm, height=2.4cm]{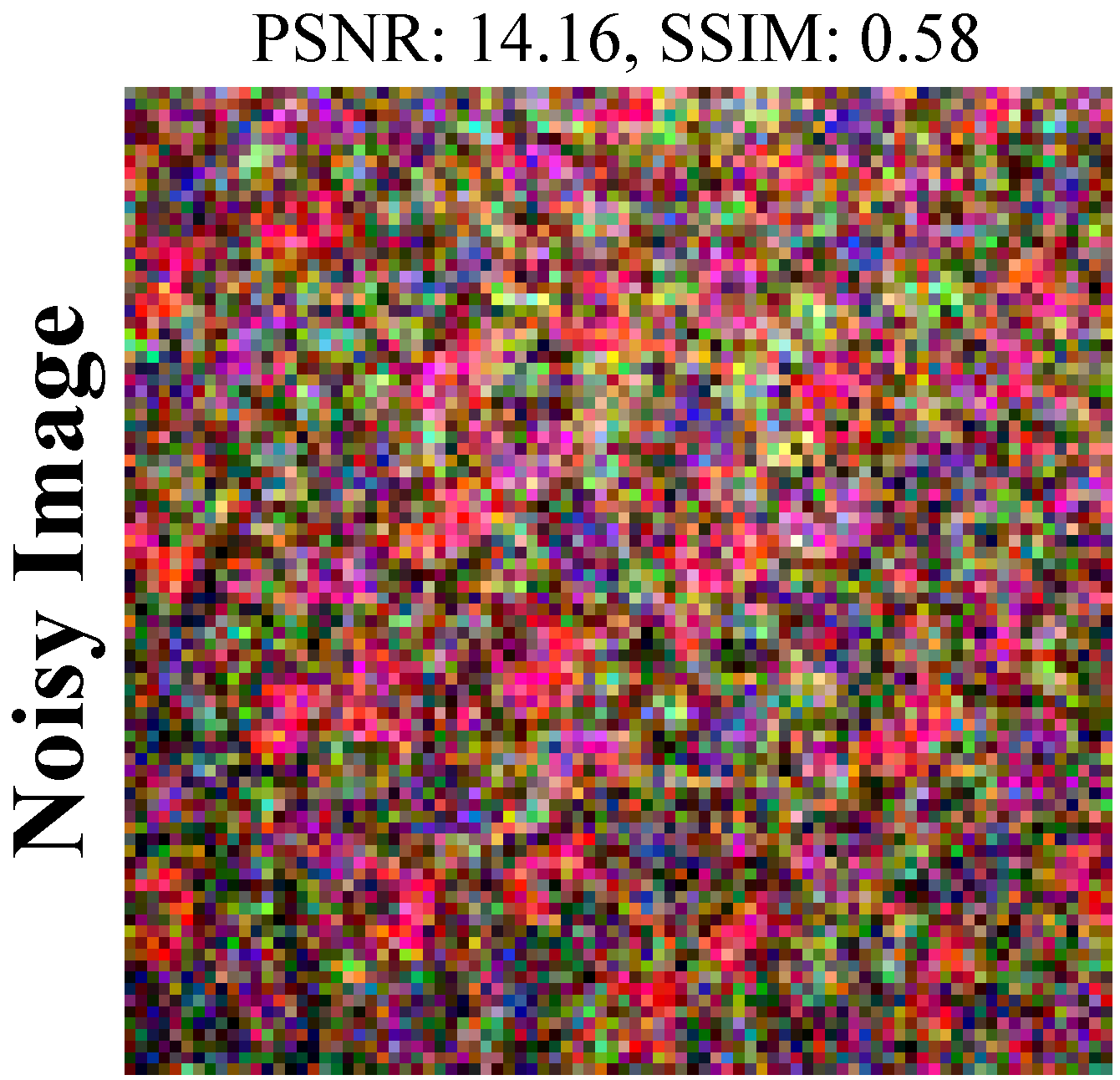}
		\end{subfigure}
		\begin{subfigure}{.13\textwidth}
			\centering
			\includegraphics[width=2.5cm, height=2.4cm]{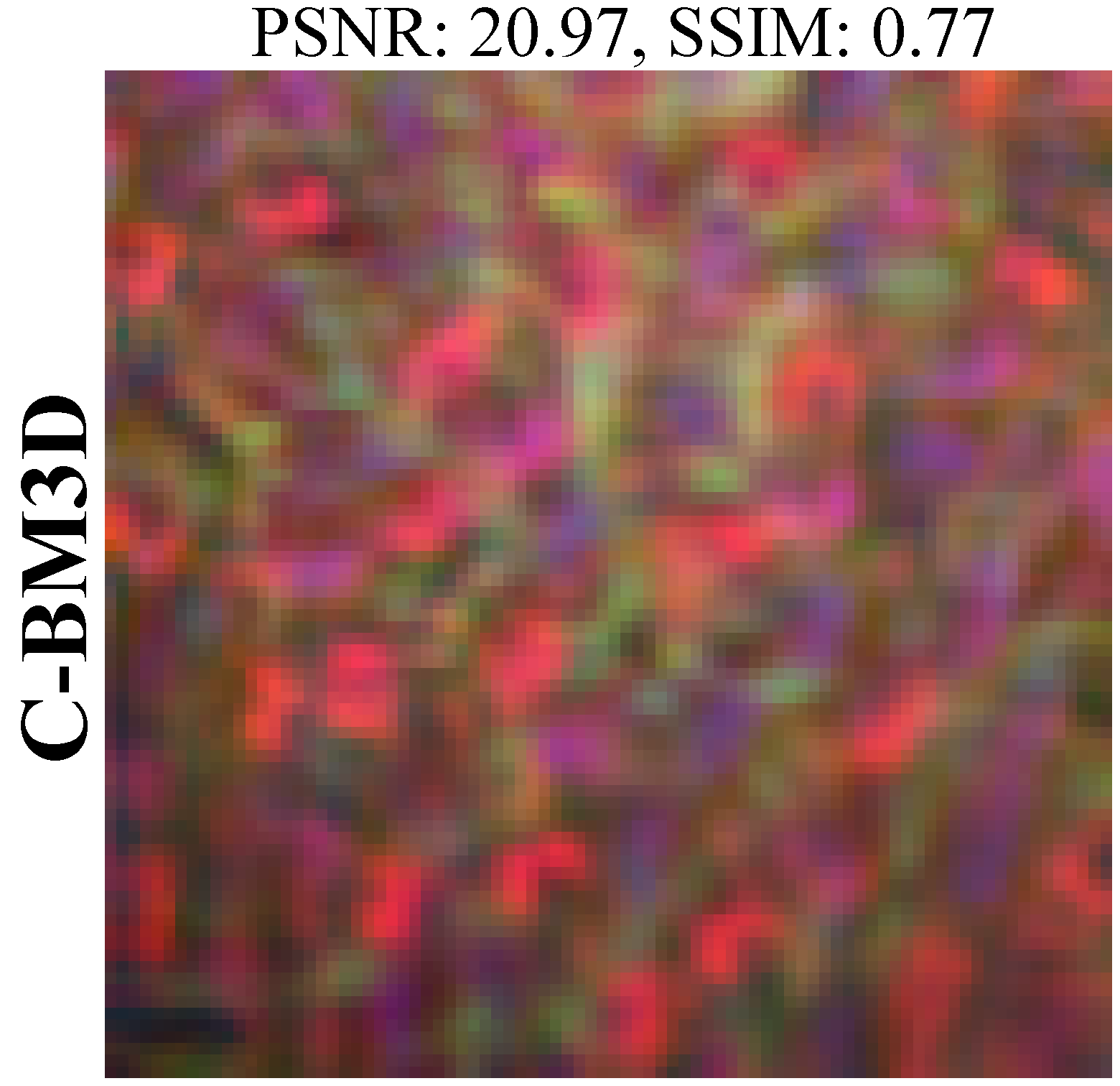}
		\end{subfigure}
		\begin{subfigure}{.13\textwidth}
			\centering
			\includegraphics[width=2.5cm, height=2.4cm]{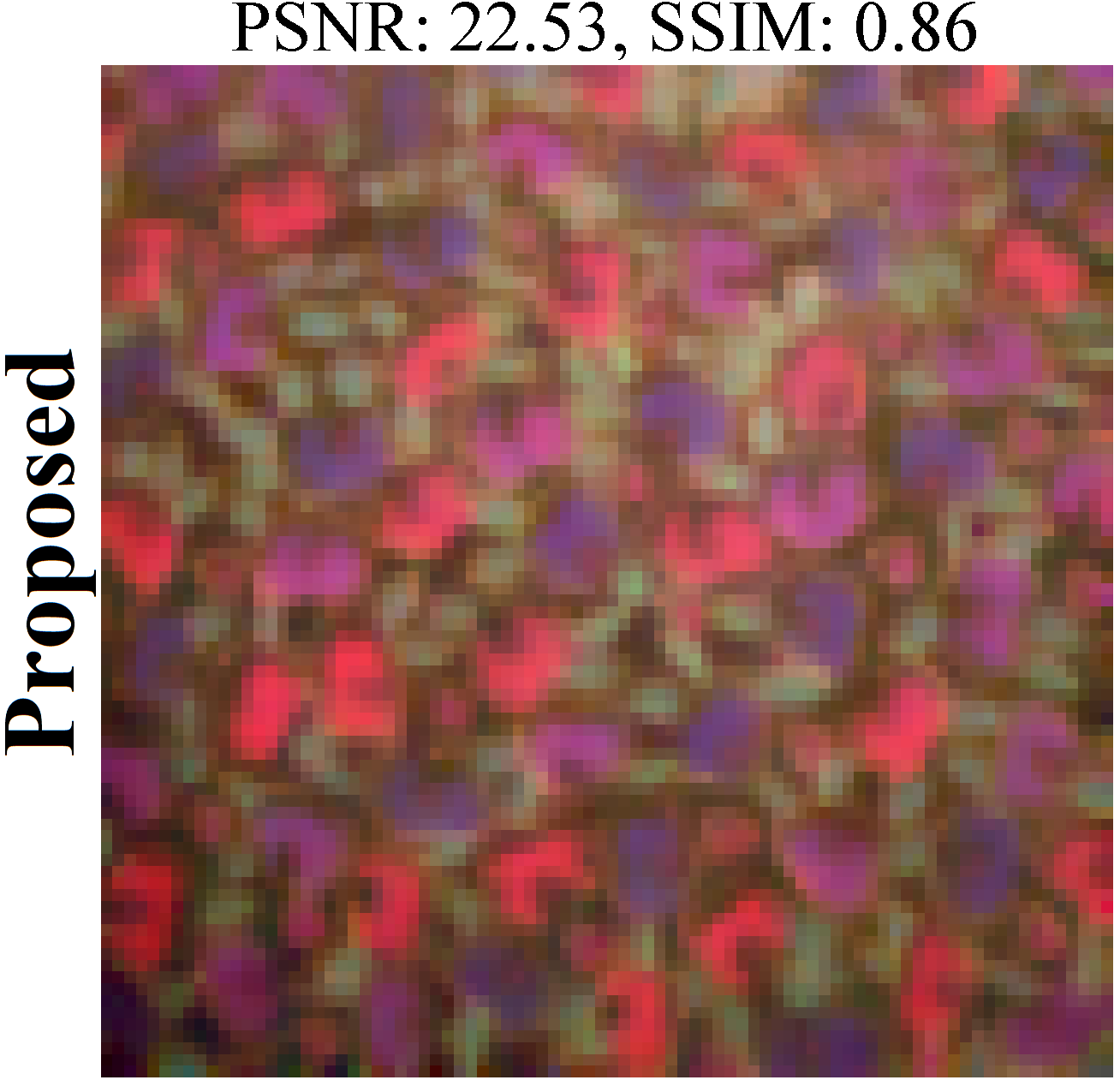}
		\end{subfigure}
		\begin{subfigure}{.13\textwidth}
			\centering
			\includegraphics[width=2.5cm, height=2.4cm]{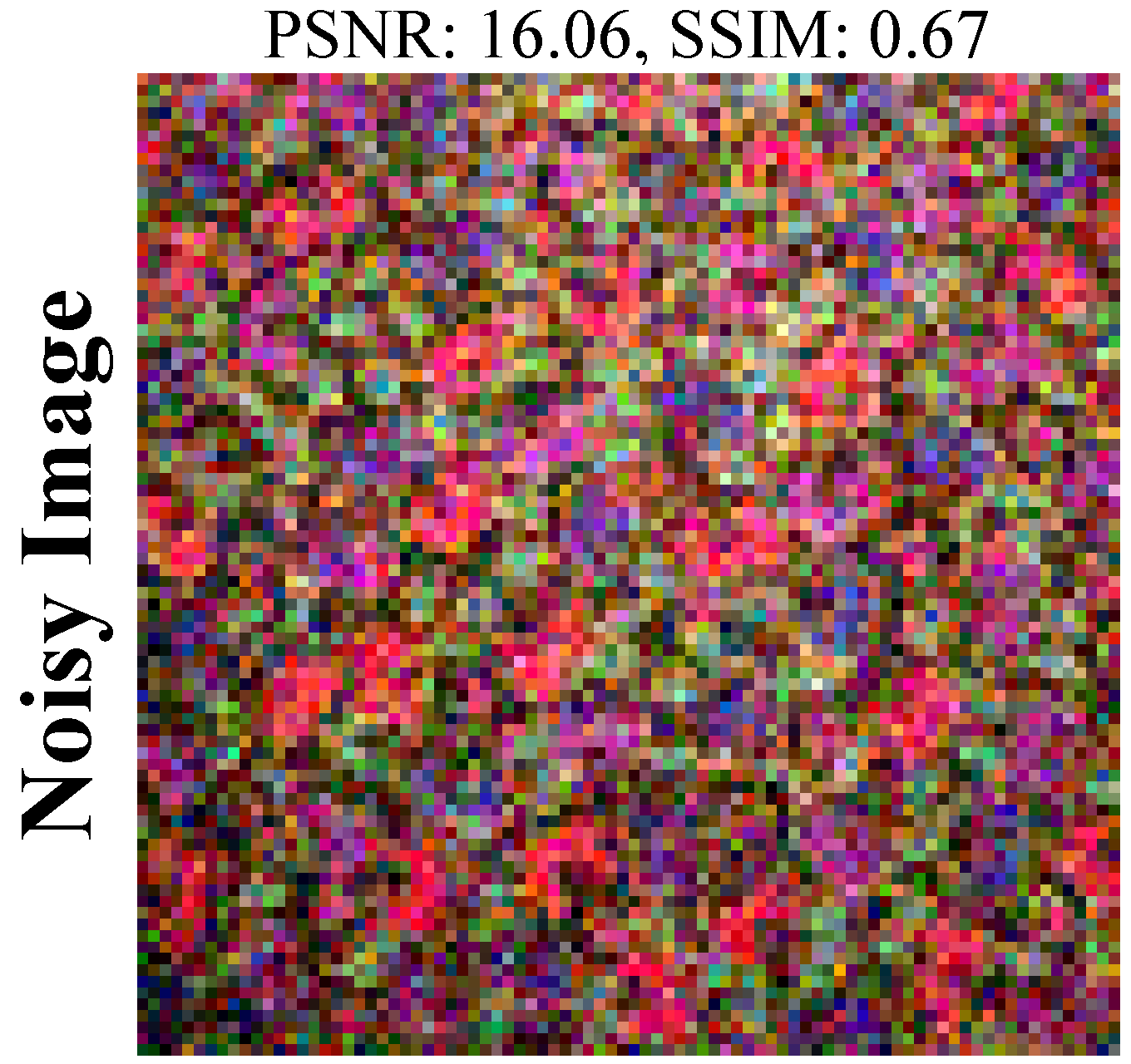}
		\end{subfigure}
		\begin{subfigure}{.13\textwidth}
			\centering
			\includegraphics[width=2.5cm, height=2.4cm]{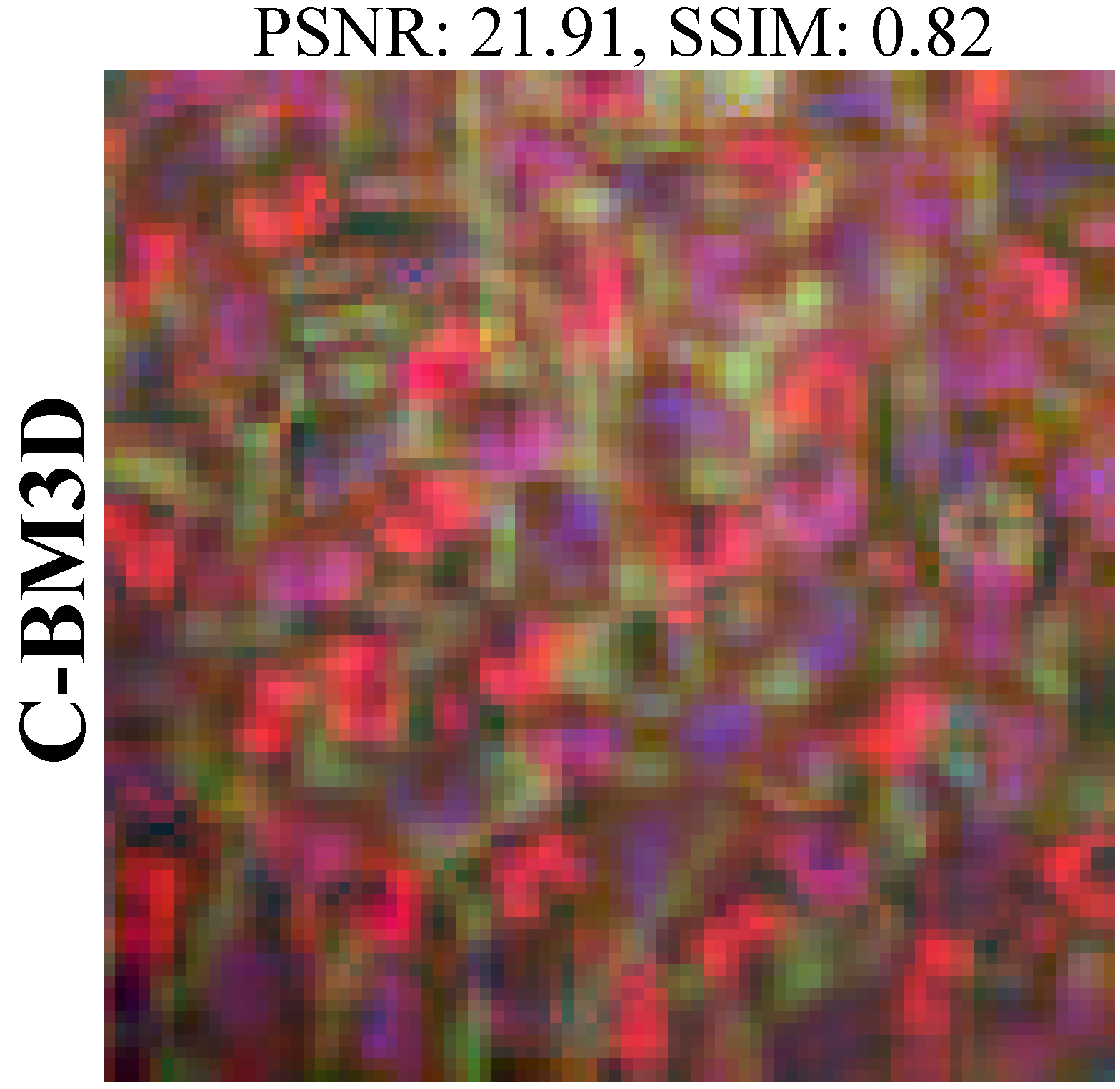}
		\end{subfigure}
		\begin{subfigure}{.13\textwidth}
			\centering
			\includegraphics[width=2.5cm, height=2.4cm]{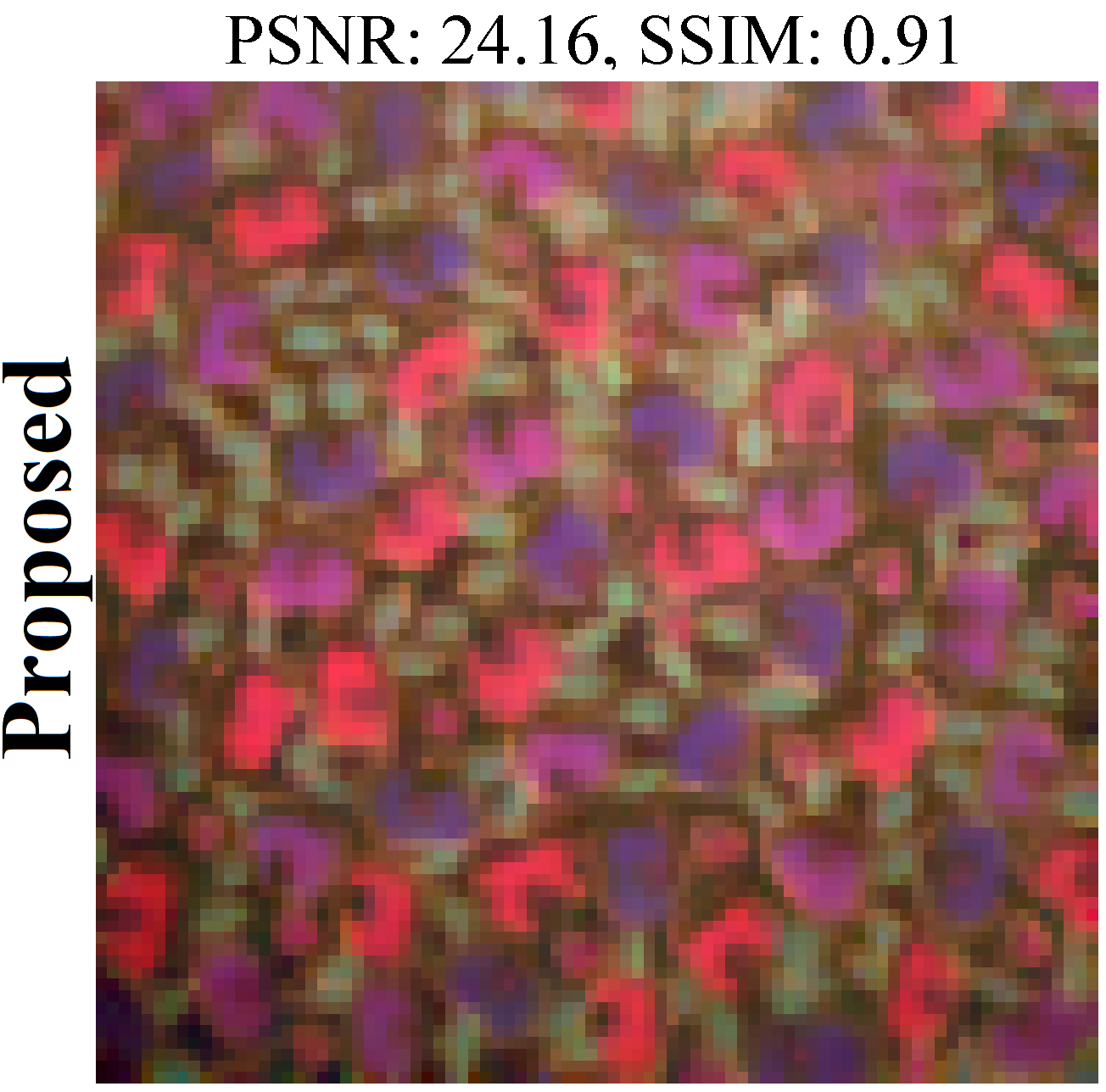}
		\end{subfigure}
	\end{subfigure} \\	
	\begin{subfigure}{\textwidth}
		\centering
		\begin{subfigure}{.13\textwidth}
			\centering
			\includegraphics[width=2.5cm, height=2.4cm]{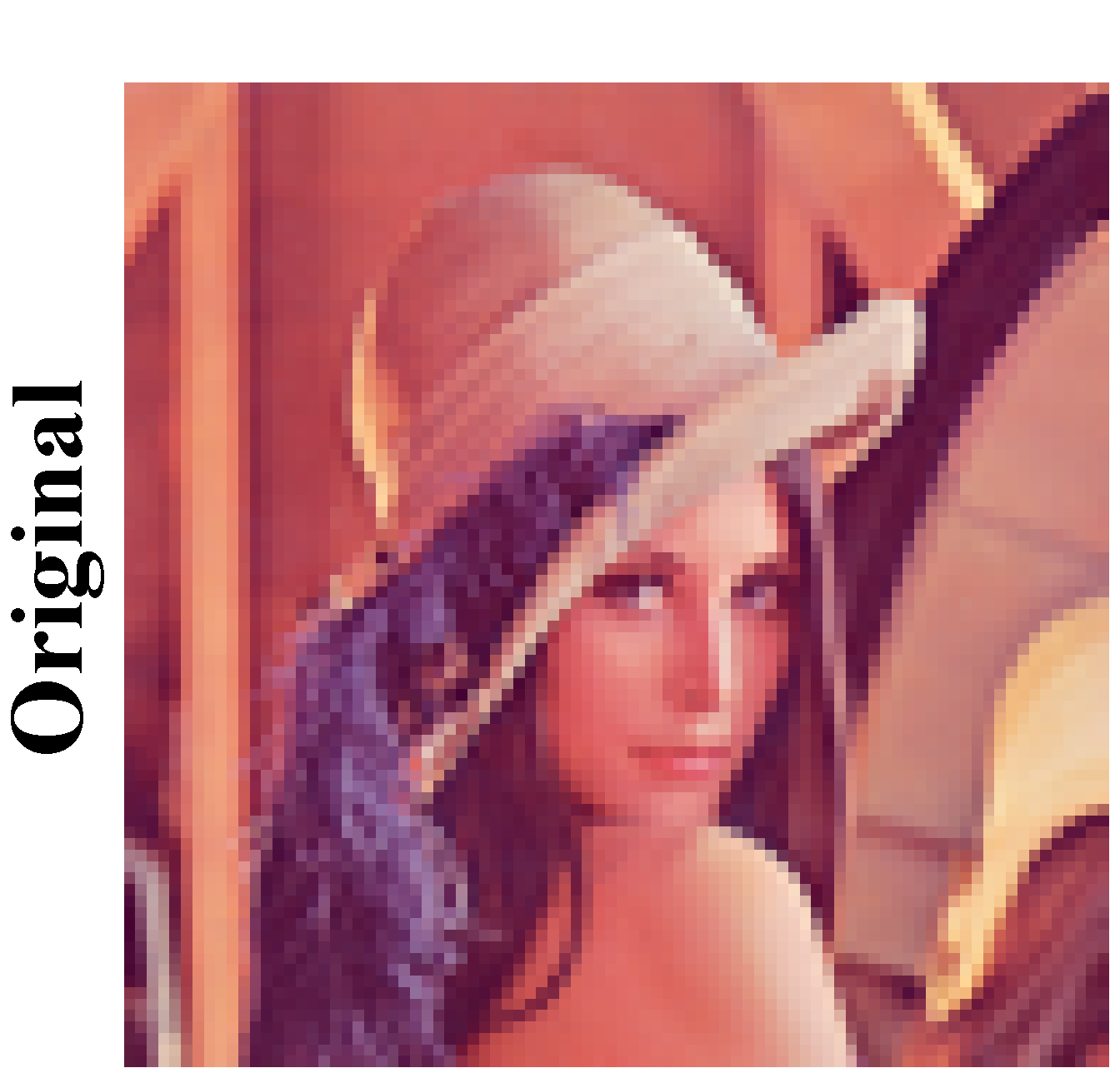}
		\end{subfigure}%
		\begin{subfigure}{.13\textwidth}
			\centering
			\includegraphics[width=2.5cm, height=2.4cm]{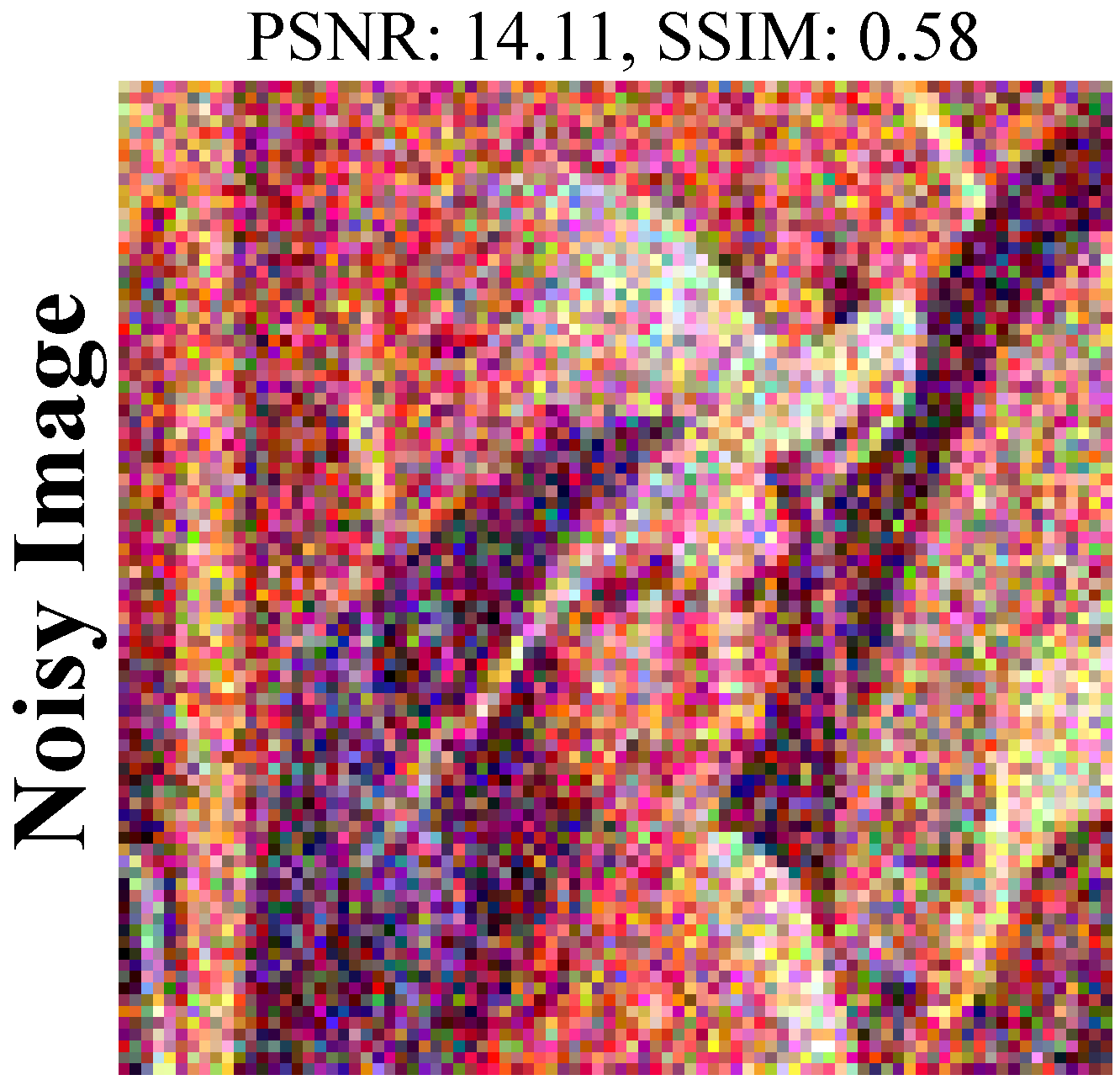}
		\end{subfigure}
		\begin{subfigure}{.13\textwidth}
			\centering
			\includegraphics[width=2.5cm, height=2.4cm]{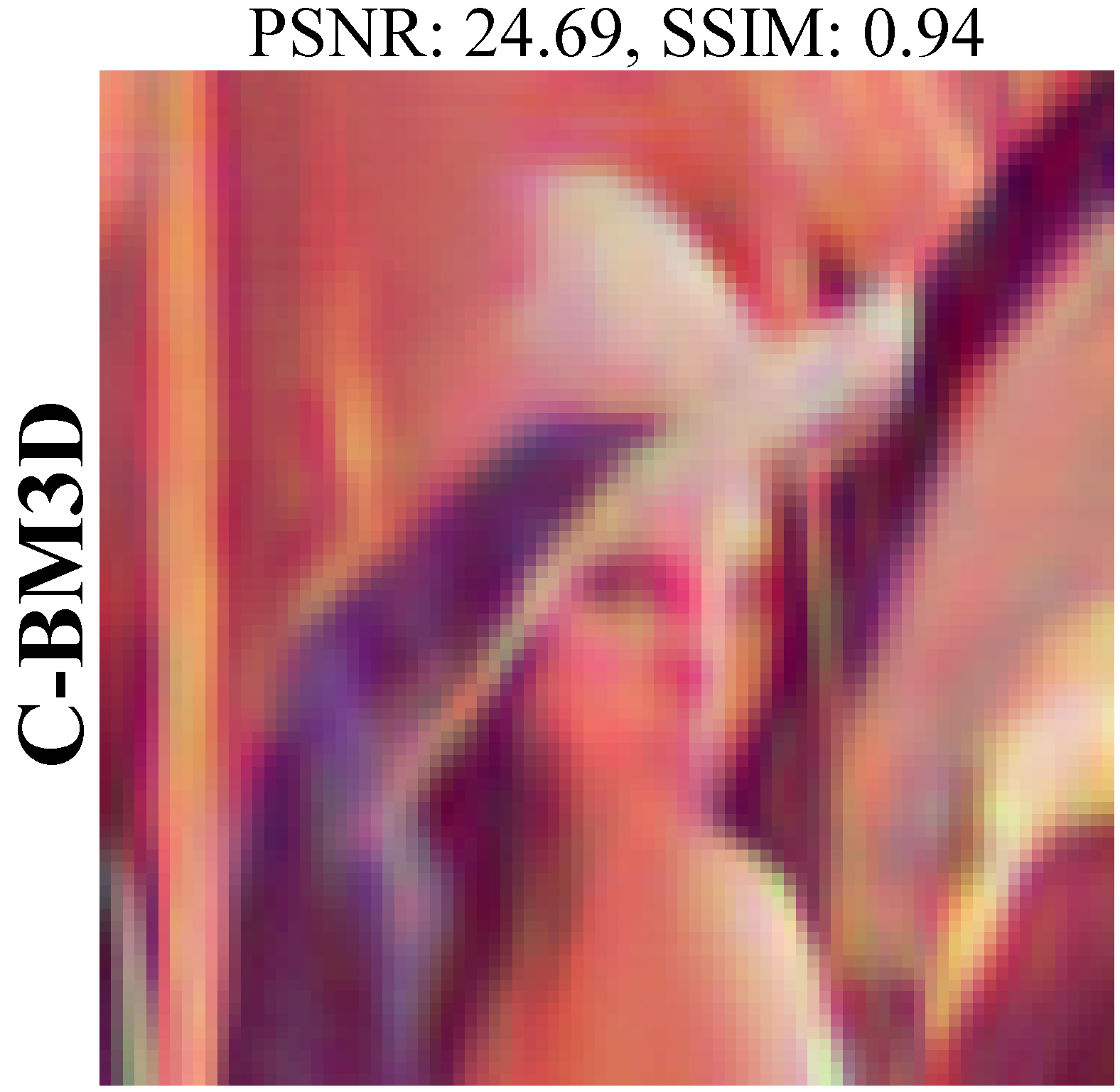}
		\end{subfigure}
		\begin{subfigure}{.13\textwidth}
			\centering
			\includegraphics[width=2.5cm, height=2.4cm]{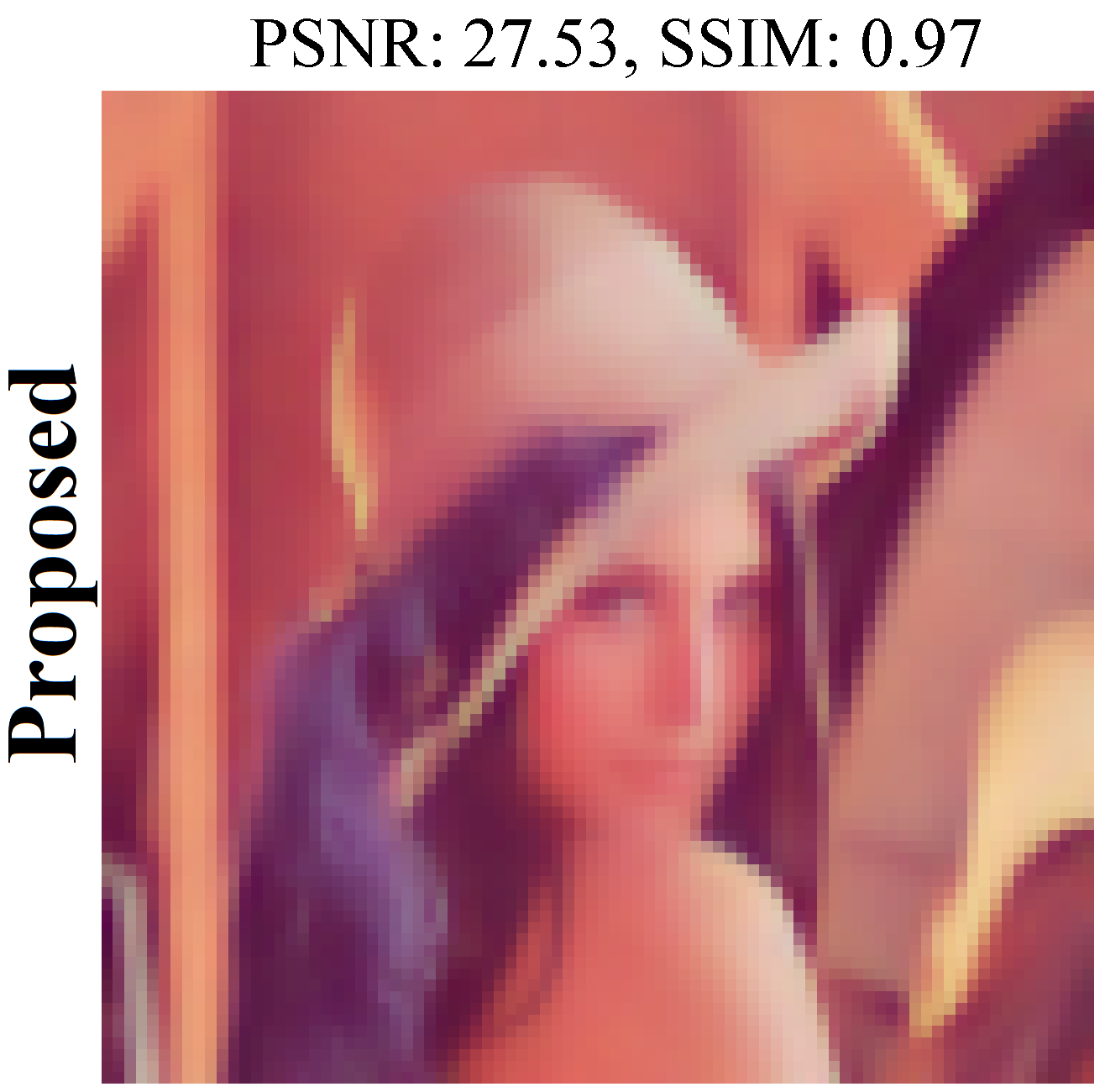}
		\end{subfigure}
		\begin{subfigure}{.13\textwidth}
			\centering
			\includegraphics[width=2.5cm, height=2.4cm]{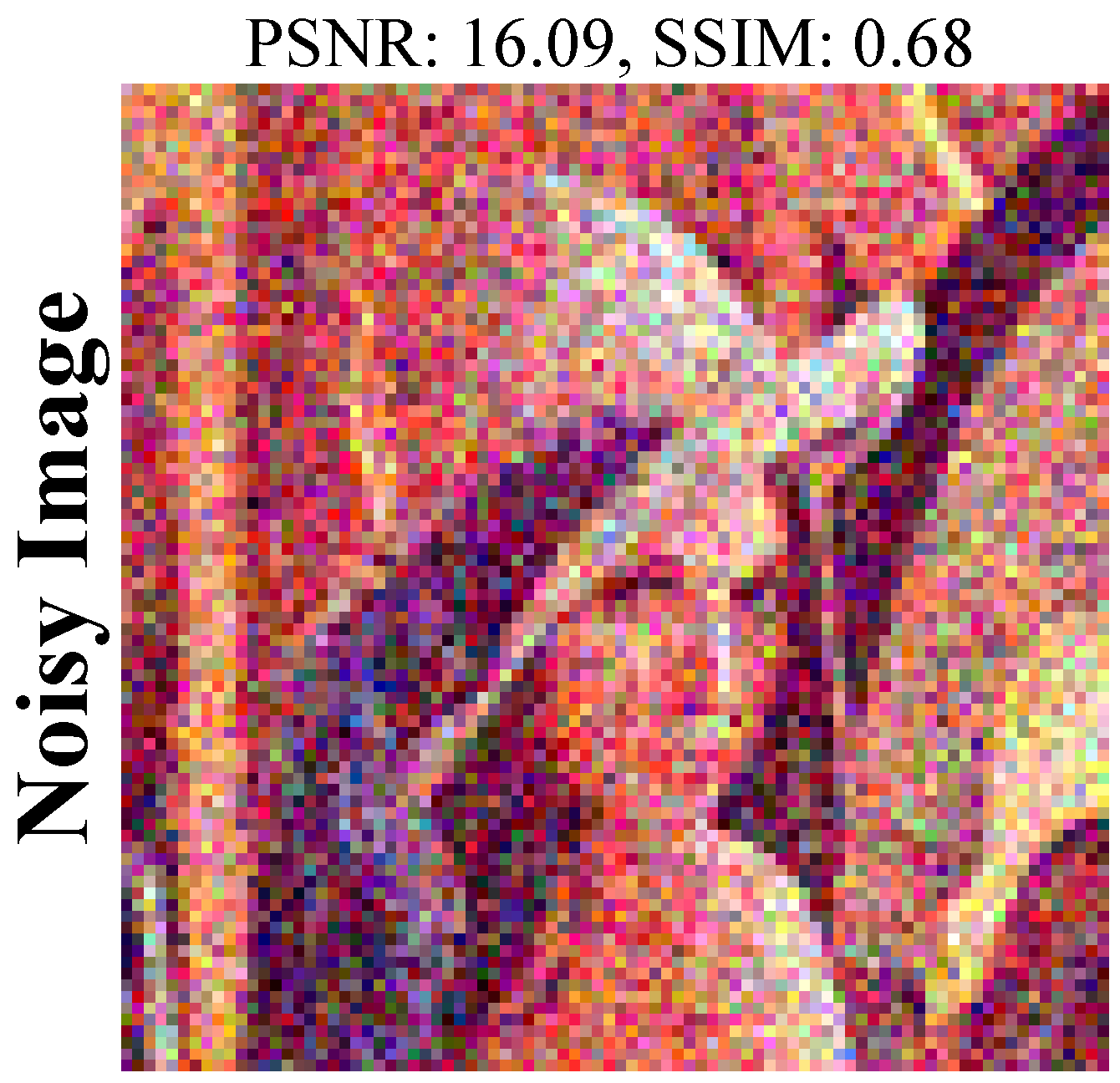}
		\end{subfigure}
		\begin{subfigure}{.13\textwidth}
			\centering
			\includegraphics[width=2.5cm, height=2.4cm]{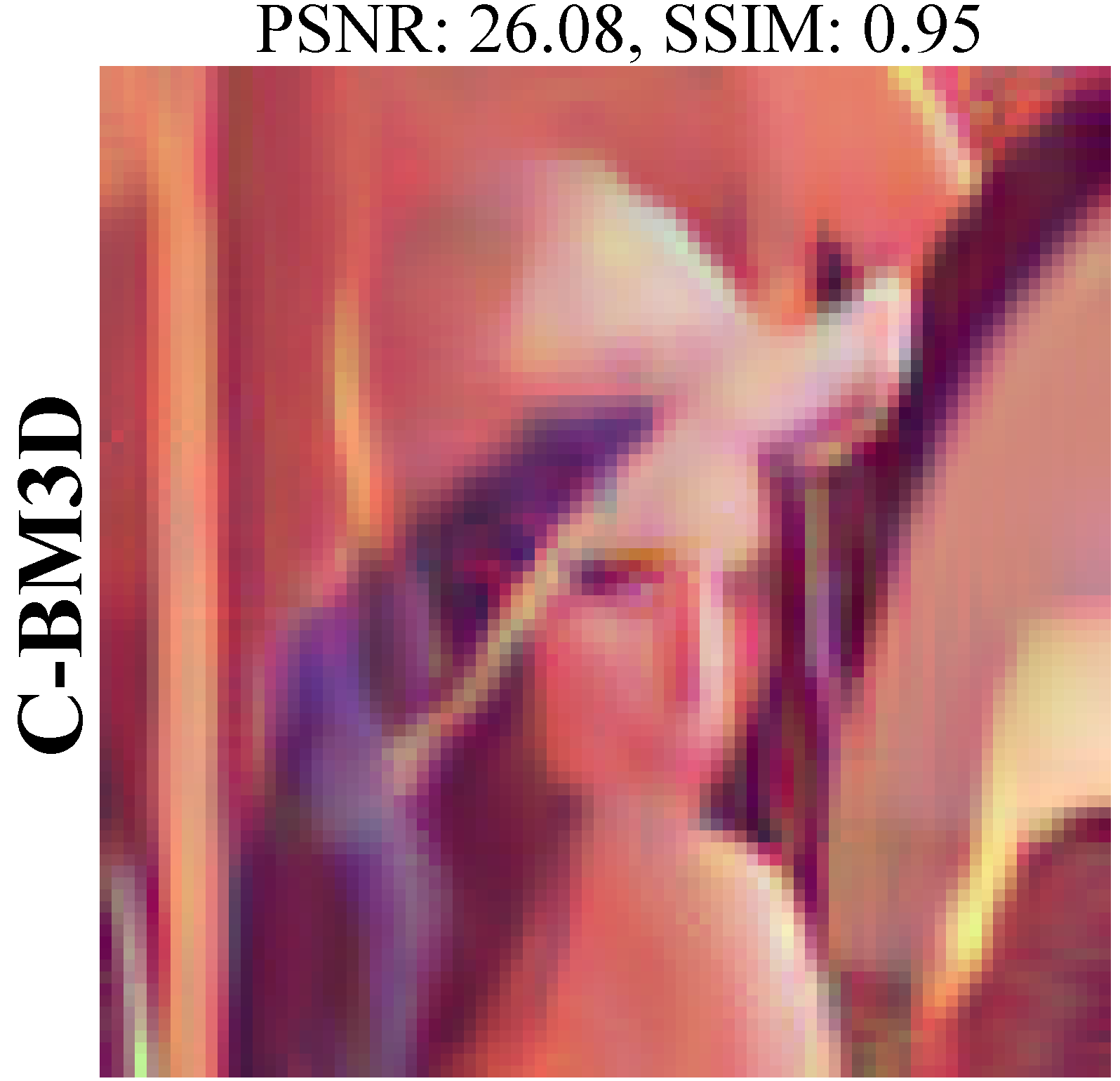}
		\end{subfigure}
		\begin{subfigure}{.13\textwidth}
			\centering
			\includegraphics[width=2.5cm, height=2.4cm]{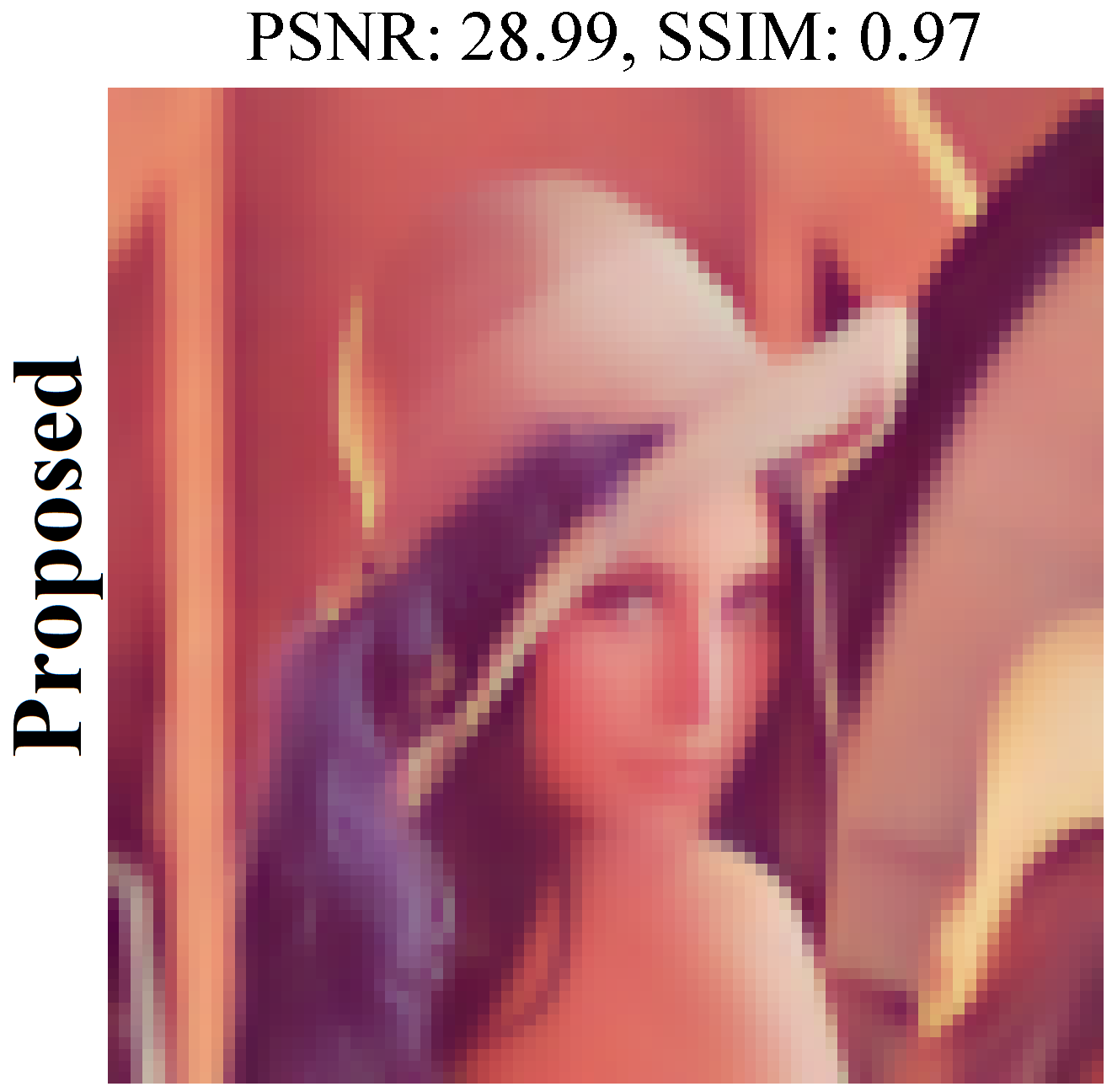}
		\end{subfigure}
	\end{subfigure}  
	\caption{Comparison of denoising color images by C-BM3D and C-C2DF. 1st column: original images, 2nd-4th columns: noisy images, denoised by C-BM3D and C-C2DF at $\mathcal{N} (0,50)$, 5th-7th columns: noisy images, denoised by C-BM3D and C-C2DF at $\mathcal{N} (0,40)$}
	\label{fig:Sim_Res_9}
\end{figure*}
\subsection{Color-Image Denoising}
\label{Color_Image_Denoising}
In this section, we present the extended version of our algorithm for color images (C-C2DF) over different scenarios. Since such images have R,G and B channels that together contribute to present the colors in an image, these channels provide a more efficient platform to perform collaboration. To take advantage of this channel correlation, we collaborate across the channels which provides us a much better sparse estimates. To do this, we find similar patches not only within the channel of reference but also across the other channels, and then performing the collaboration.

The results obtained from effectiveness of this improved collaboration step for different color images has been shown in Fig. \ref{fig:Sim_Res_9} under various high noise levels. As shown, the images even under severe noise contamination are recovered to a very good extent as opposed to the color version of the BM3D algorithm. We have also showed that our C-C2DF method is valid not only for natural images but is also equally efficient for synthetic images as shown in the third row of the figure under observation.

In addition to the images shown in Fig. \ref{fig:Sim_Res_9}, we summarize the denoising results of other color images in Table \ref{table4} to show the performance gain as compared to the Color-BM3D (C-BM3D). The stated results demonstrate significance of C2DF over a wide range of images, and proved that it can be used globally in any scenario outperforming existing state-of-the-art algorithms.

\section{Conclusion}
\label{Conclusion}
We proposed an image denoising algorithm by exploiting features from both spatial as well as transformed domain. Our method utilizes a patch-based collaborative approach using similar-structured patches using an intensity-independent approach. For each patch, the probabilities of taps being active are computed and then refined via collaboration among similar patches in transformed domain. This approach tremendously isolates the noisy components and thus improves the sparse estimates thereby producing high quality reconstructed image. For a further improvement in the denoised image, we deploy a region growing based spatially developed post-processor that refines the smooth parts further. An extension of proposed algorithm for color image denoising has also been presented performing cross-channel collaboration. The detailed results from extensive simulations under a number of wide range of scenarios dictate that our proposed method outperform the existing state-of-the-art methods by a very convenient margin.
%
%
%
%
\begin{table}[t!]
	\centering
	\begin{tabular}{|c|c|c|c|c|}\hline\hline
		\multicolumn{2}{|c|}{Image Name} & $\sigma_w = 50$ & $\sigma_w = 40$ & $\sigma_w = 30$  \\ \cline{1-4}
		\hline\hline
		\multirow{2}{*}{Jet}
		& C-BM3D & 23.47/0.75 & 24.68/0.79  & 26.40/0.86 \\ \cline{2-5}
		& C-C2DF & \textbf{26.26/0.79} & \textbf{27.63/0.83} & \textbf{29.80/0.87} \\ \hline
		
		\multirow{2}{*}{Lake}
		& C-BM3D & 21.52/0.83 & 22.83/0.87 & 24.37/0.90 \\ \cline{2-5}
		& C-C2DF & \textbf{26.01/0.92} & \textbf{27.71/0.94} & \textbf{30.18/0.96} \\ \hline
		
		\multirow{2}{*}{Mandrill}
		& C-BM3D & 22.48/0.77 & 23.36/0.81 & 24.69/0.85 \\ \cline{2-5}
		& C-C2DF & \textbf{24.37/0.82} & \textbf{25.83/0.86} & \textbf{28.13/0.90} \\ \hline
		
		\multirow{2}{*}{Peppers}
		& C-BM3D & 21.88/0.91 & 23.40/0.94 & 24.95/0.95 \\ \cline{2-5}
		& C-C2DF & \textbf{27.44/0.97} & \textbf{29.11/0.98} & \textbf{31.52/0.99} \\ \hline
		
		\multirow{2}{*}{Girl}
		& C-BM3D & 25.38/0.80 & 26.88/0.0.85 & 27.71/0.90 \\ \cline{2-5}
		& C-C2DF & \textbf{29.81/0.93} & \textbf{31.49/0.95} & \textbf{33.79/0.97} \\ \hline
		
		\multirow{2}{*}{Woman}
		& C-BM3D & 24.73/0.79 & 26.01/0.82 & 27.46/0.88 \\ \cline{2-5}
		& C-C2DF & \textbf{28.90/0.85} & \textbf{30.26/0.88} & \textbf{32.25/0.92} \\ \hline
	\end{tabular}
\caption{Results of denoising color images by C-BM3D and C-C2DF under $\Wm \sim \mathcal{N} (\textbf{0},\sigma_w\textbf{I})$}
\label{table4}
\end{table}

%
%
%



\ifCLASSOPTIONcaptionsoff
  \newpage
\fi

\bibliographystyle{ieeetr}
\bibliography{bare_jrnl_compsoc}

%

\begin{IEEEbiography}[{\includegraphics[width=1in,height=1.25in,clip,keepaspectratio]{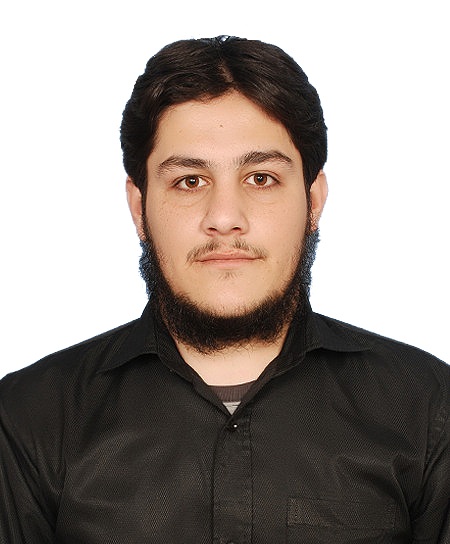}}]{Muzammil Behzad} received his partially-funded B.S. degree with distinctions (double medalist and valedictorian) from COMSATS Institute of Information Technology (CIIT), Pakistan, and his fully-funded M.S. degree from King Fahd University of Petroleum and Minerals (KFUPM), Saudi Arabia, both in Electrical Engineering. Presently, he is serving as Research Associate as well as Pioneer Member of Office of Hybrid Learning at CIIT. His research interests are oriented around signal and image processing, wireless sensor networks, and their applications. He is a lifetime member of Pakistan Engineering Council (PEC), student member of Institute of Electrical and Electronics Engineers (IEEE), and member of Society for Industrial and Applied Mathematics (SIAM). He currently holds more than 6 years of experience in teaching and research. He is also the recipient of employee of the year and the best supervised project award from CIIT in the fiscal year 2013-14.
\end{IEEEbiography}
%






\end{document}